\RequirePackage{fix-cm}
\documentclass[smallextended]{svjour3}       % onecolumn (second format)
\smartqed  % flush right qed marks, e.g. at end of proof
\usepackage{booktabs}
\usepackage{graphicx}
\usepackage{amsmath}
\usepackage{amssymb}
\usepackage{graphicx}
\usepackage{xspace}

\usepackage{url}
\usepackage{algorithmic}
\usepackage{algorithm} % For counting chapters
\algsetup{linenosize=\scriptsize}
\xspaceaddexceptions{=}
\xspaceaddexceptions{\}}
\xspaceaddexceptions{\in}
\algsetup{linenosize=\scriptsize}
\usepackage{algorithm} % For counting chapters

\usepackage[bf]{subfigure}
\usepackage[pdftex,table,fixpdftex]{xcolor}
\usepackage{longtable}
\usepackage{threeparttable}
\usepackage{multirow}
\usepackage{tabularx}
\usepackage{hhline}

\begin{document}

\title{Holarchic Structures for Decentralized Deep Learning -- A Performance Analysis
%\thanks{Grants or other notes
%about the article that should go on the front page should be
%placed here. General acknowledgments should be placed at the end of the article.}
}
%\subtitle{Do you have a subtitle?\\ If so, write it here}

%\titlerunning{Short form of title}        % if too long for running head

\author{Evangelos Pournaras \and Srivatsan Yadhunathan \and Ada Diaconescu}

%\authorrunning{Short form of author list} % if too long for running head

\institute{Evangelos Pournaras \and Srivatsan Yadhunathan \at
              Professorship of Computational Social Science \\
              ETH Zurich, Zurich, Switzerland\\
              Tel.: +41446320458\\
%              Fax: +123-45-678910\\
              \email{\{epournaras,ysrivatsan\}@ethz.ch}           %  \\
%             \emph{Present address:} of F. Author  %  if needed
           \and
          	  Ada Diaconescu \at
              Telecom ParisTech\\
              Paris-Saclay University, Paris, France\\
              Tel.: +33145818072\\
              \email{ada.diaconescu@telecom-paristech.fr}
}
\date{Received: date / Accepted: date}
% The correct dates will be entered by the editor

\maketitle

\begin{abstract}

Structure plays a key role in learning performance. In centralized computational systems, hyperparameter optimization and regularization techniques such as dropout are computational means to enhance learning performance by adjusting the deep hierarchical structure. However, in decentralized deep learning by the Internet of Things, the structure is an actual network of autonomous interconnected devices such as smart phones that interact via complex network protocols. Self-adaptation of the learning structure is a challenge. Uncertainties such as network latency, node and link failures or even bottlenecks by limited processing capacity and energy availability can significantly downgrade learning performance. Network self-organization and self-management is complex, while it requires additional computational and network resources that hinder the feasibility of decentralized deep learning. In contrast, this paper introduces a self-adaptive learning approach based on holarchic learning structures for exploring, mitigating and boosting learning performance in distributed environments with uncertainties. A large-scale performance analysis with 864000 experiments fed with synthetic and real-world data from smart grid and smart city pilot projects confirm the cost-effectiveness of holarchic structures for decentralized deep learning. 

\end{abstract}

\section{Introduction}\label{sec:intro}

Smart citizens' devices with increasing processing power and high energy autonomy are becoming pervasive and ubiquitous in everyday life. The Internet of Things empowers a high level of interconnectivity between smart phones, sensors and wearable devices. These technological developments provide unprecedented opportunities to rethink about the future of machine learning and artificial intelligence: Centralized computational intelligence can be often used for privacy-intrusive and discriminatory services that create `filter bubbles' and undermine citizens' autonomy by nudging~\cite{Friedman2015,Mik2016,Heidelberger2018}. In contrast, this paper envisions a more socially responsible design for digital society based on decentralized learning and collective intelligence formed by bottom-up planetary-scale networks run by citizens~\cite{Pournaras2018,Helbing2017}. 

In this context, the structural elements of decentralized deep learning processes play a key role. The effectiveness of several classification and prediction operations often relies heavily on hyperparameter optimization~\cite{Maclaurin2015,YAO2017} and on the learning structure, for instance, the number of layers in a neural network, the interconnectivity of the neurons, the activation or deactivation of certain pathways i.e. dropout regularization~\cite{Srivastava2014}, can enhance learning performance. Controlling and adjusting a deep hierarchical structure in a centralized computing system is straightforward in the sense that all meta information for model generation is locally available. However, in decentralized learning the challenge of optimizing the learning structure is not anymore exclusively a computational problem. Other challenges such as network latency, node and link failures as well as the overall complexity of building and maintaining an overlay network~\cite{Pournaras2013b} in a distributed environment perplex the feasibility of decentralized learning. 

This paper introduces the concept of \emph{holarchy} in deep hierarchical structures as the means to adapt to the aforementioned uncertainties of distributed environments. A learning process can be localized and performed over a holarchic structure in a recursive way without changing the core learning logic and without employing additional mechanisms to reconfigure the network. This is the proposed \emph{self-adaption approach} to decentralized learning that is by design highly reactive and cost-effective as it maximizes the utilization of the available communication and computational resources, in contrast to a complementary and more proactive \emph{self-organization approach} that requires additional interactions between agents and therefore can increase communication and computational cost. By using holarchies for learning, forward propagation and backpropagation become recursive in nested levels over the deep hierarchical structure as the means to (i) \emph{explore} improving solutions, (ii) \emph{mitigate} learning performance in case part of the network is disconnected or even (iii) \emph{boost} learning performance after the default learning process completes. These three scenarios are formalized by three holarchic schemes applied to a multi-agent system for decentralized deep learning in combinatorial optimization problems: I-EPOS, the \emph{Iterative Economic Planning and Optimized Selections}~\cite{Pournaras2018}.

A large-scale performance analysis with 864000 experiments is performed using synthetic and real-world data from pilot projects such as bike sharing, energy demand and electric vehicles. Several dimensions are studied, for instance, topological properties of the deep hierarchical structure, constraints by the agents' preferences and the scale of the holarchic structures, i.e. number of nested layers. Results confirm the cost-effectiveness of the holarchic learning schemes for exploration, mitigation and boosting of learning performance in dynamic distributed environments. Nevertheless, in stable environments the localization of the learning process within holarchic structures may result in lower performance compared to a learning applied system-wide. 

In summary, the contributions of this paper are the following:

\begin{itemize}

\item A novel self-adaptation approach to decentralized deep learning as the means to decrease or avoid the communication and computational cost of self-organization in distributed environments with uncertainties. 
\item The concept of holarchy as a self-adaptation approach for decentralized deep learning.
\item The introduction of three holarchic schemes as self-adaptation means to explore, mitigate and boost deep learning performance. 
\item The applicability and extension of I-EPOS with the three holarchic schemes to perform collective decision-making in decentralized combinatorial optimization problems.
\item An empirical performance analysis of 864000 benchmark experiments generated with synthetic and real-world data from Smart Grid and Smart City pilot projects.
\end{itemize}

This paper is organized as follows: Section~\ref{sec:holarchy} introduces the concept of holarchy in decentralized learning as well as three holarchic schemes to explore, mitigate and boost deep learning performance under uncertainties of distributed environments. Section~\ref{sec:i-epos} illustrates a case study of a decentralized deep learning system to which the three holarchic schemes are applied: I-EPOS. Section~\ref{sec:experimental-methodology} shows the experimental methodology followed to conduct a large-scale performance analysis of the three holarchic schemes. Section~\ref{sec:experimental-evaluation} illustrates the results of the experimental evaluation. Section~\ref{sec:discussion} summarizes and discusses the main findings. Section~\ref{sec:related-work} positions and compares this paper with related work. Finally, Section~\ref{sec:future-work} concludes this paper and outlines future work.

%In future work we will aim to capitalise on the reusability of existing optimisation (sub-)solutions in holarchic sub-trees for further speeding-up the convergence of the global optimisation process. 

\section{Holarchic Structures for Decentralized Deep Learning}\label{sec:holarchy}

This paper studies \emph{decentralized deep learning} processes in which the deep hierarchical structure is fully distributed and self-organized by remote autonomous (software) agents that interact over a communication network. In other words, the learning process is crowd-sourced to citizens, who participate by contributing their computational resources, for instance their personal computers or smart phones. The agents reside on these devices and collectively run the decentralized learning process. For example, in contrast to a conventional model of a centralized neural network, which performs training by locally accessing all citizens' data, a collective neural network consists of neurons that are remote citizens' devices interacting over the Internet to form an overlay peer-to-peer network~\cite{Pournaras2013b}. It is this overlay network that represents the hierarchical neural network structure. 

Methods for hyperparameter optimization of the hierarchical structure to improve the learning performance of a model are usually designed for centralized systems in which all information, including input data, the network structure and the learning model itself are locally available. This provides a large spectrum of flexibility to change the deep hierarchical structure offline or even online~\cite{Grande2013,Reymann2018} and determine via hyperparameter optimization the settings that maximize the learning performance. In contrast to learning based on centralized computational systems, in decentralized deep learning the structure cannot arbitrary change without paying for some computational and communication cost. Network uncertainties such as node and link failures, latency as well as limited resources in terms of bandwidth, computational capacity or even energy in case of sensors and smart phones can limit performance, interrupt the learning process and increase the design complexity of decentralized hyperparameter optimization. 

The aforementioned uncertainties of distributed environments introduce endogenous constraints in parts of the deep hierarchical structure: the learning process is interrupted and becomes \emph{localized} within branches of the hierarchical structure. For instance, node and link failures or suspension of the learning processes due to conservation of resources~\cite{Aziz2013,Sterbenz2010} in nodes are scenarios under which learning can be localized. This paper poses here the following question: How to preserve the learning capacity of a decentralized system, whose learning process is interrupted and localized by aforementioned uncertianties in distributed environments? On the one hand, it is known that localization and greedy optimization can underperform with search becoming trapped to locally optimum solutions that have a significant divergence from global optimality~\cite{Bang2004}. On the other hand, limiting the learning units of hierarchical structures can also increase learning performance by preventing overfitting as known by the dropout concept in neural networks~\cite{Srivastava2014}. This paper sheds light on the role of localization in decentralized learning. 

In this context, the management of the learning performance of an algorithm is no longer entirely a computational challenge but rather a multifaceted self-adaptation process: Other aspects such as performance \emph{exploration}, \emph{mitigation} and \emph{boosting} come to the analysis foreground. Exploration adapts the learning process and the search space to improve learning performance under localization. Mitigation is the maintenance of a high learning performance under a localization of the learning process. Finally, the feasibility of boosting the learning performance under localization is subject of this study as well. 

\begin{figure}[!htb]
\centering
\includegraphics[width=0.65\textwidth]{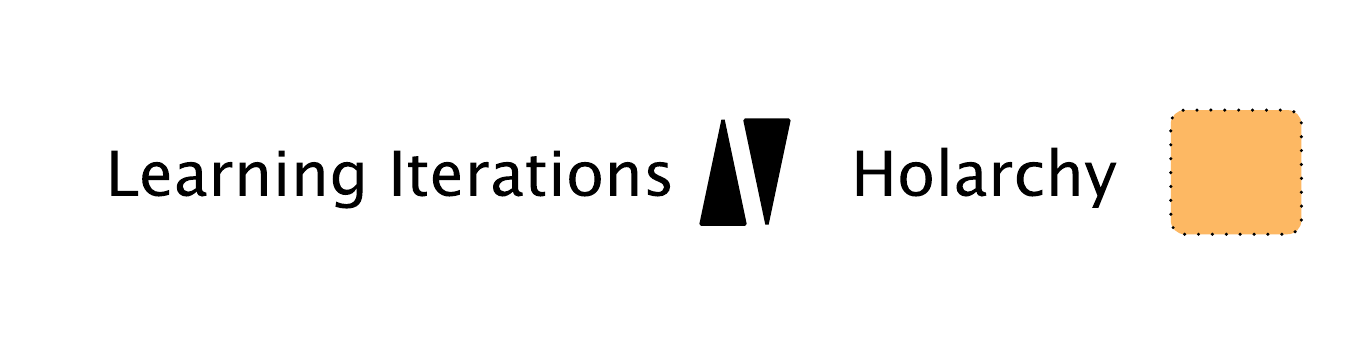}\\
\subfigure[Learning iterations between the leaves and their parents.]{\includegraphics[width=0.32\textwidth]{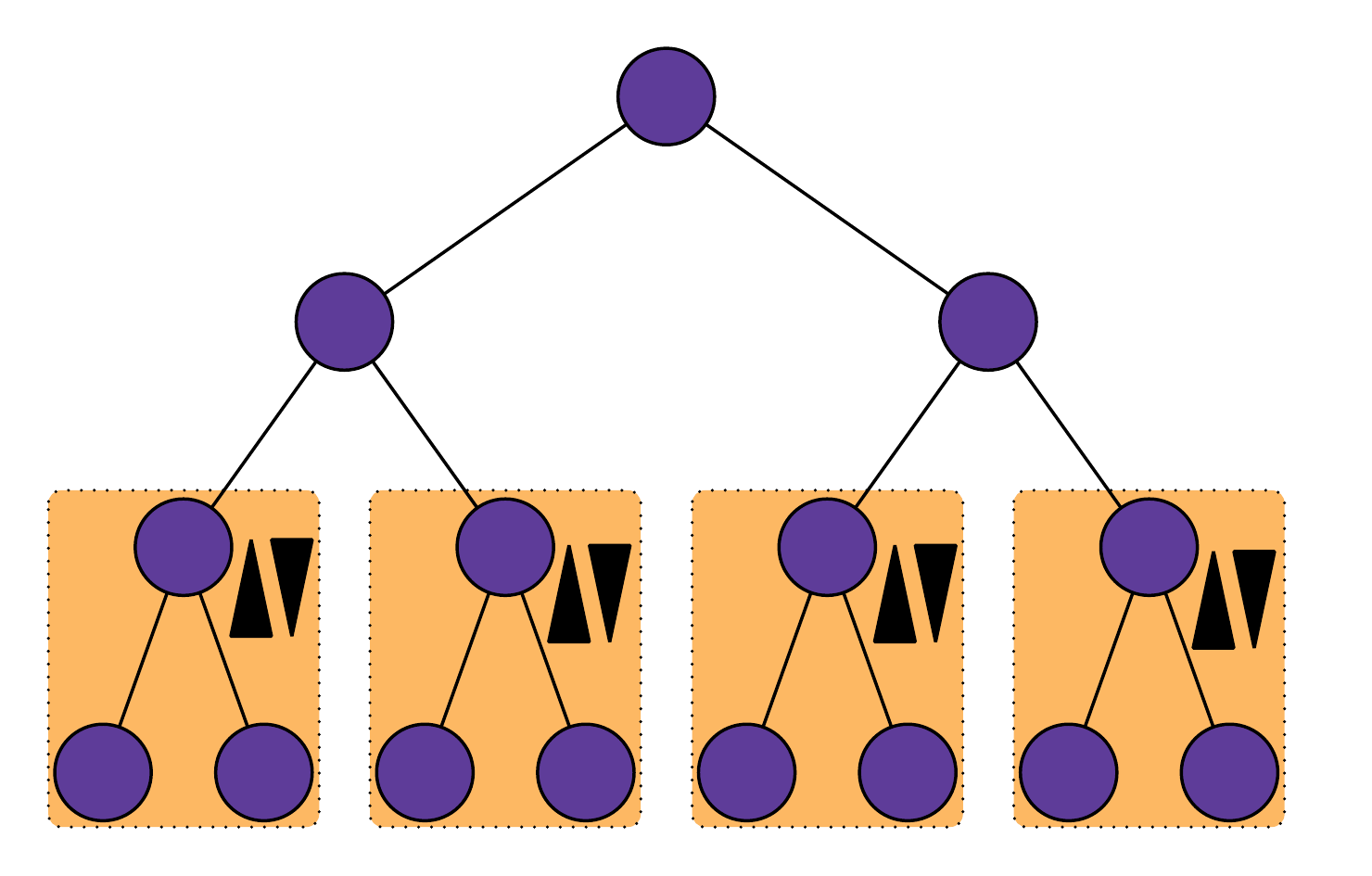}}
\subfigure[Progress to the next level. Learning iterations are performed within tree branches.]{\includegraphics[width=0.32\textwidth]{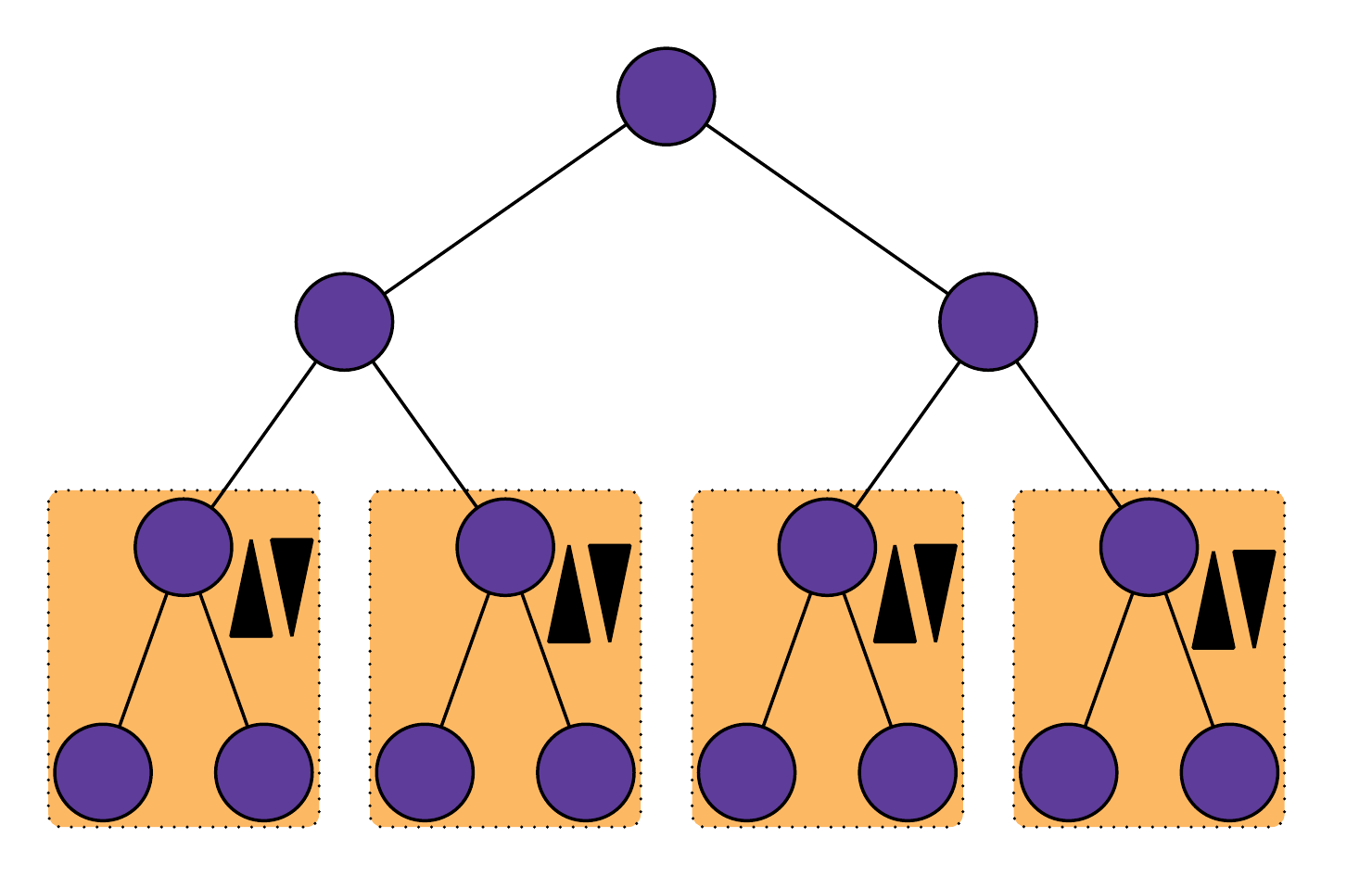}}
\subfigure[Holarchic learning completes with learning iterations performed over the whole tree structure.]{\includegraphics[width=0.32\textwidth]{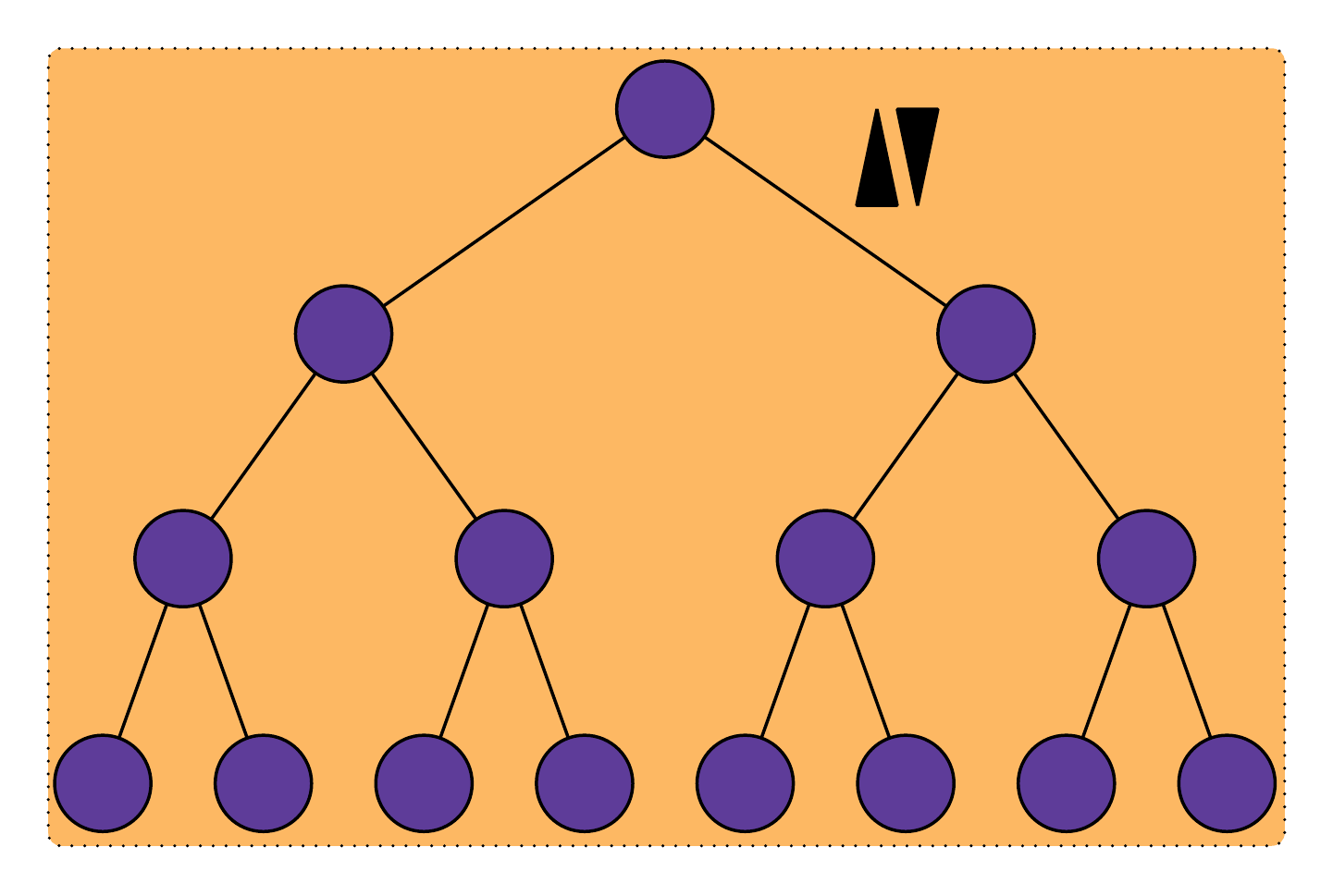}}
\caption{The concept of holarchic learning. Learning iterations are performed in nested branches, the holons. Figure~\ref{fig:holarchy}c actually depicts the default baseline learning strategy, while Figure~\ref{fig:holarchy}a and~\ref{fig:holarchy}b show the earlier learning iterations performed within the holons.}\label{fig:holarchy}
\end{figure}

This paper introduces the concept of \emph{holarchy} in deep learning hierarchical structures to study the performance exploration, mitigation and boosting potential under the aforementioned uncertainties of distributed environments. A holarchy is a recursive hierarchical network of \emph{holons} that represent part of the deep hierarchical structure as well as the whole structure. In the case of a tree topology, every possible branch (part) in the whole tree topology is also a tree topology (whole). When an agent (parent) connects two branches, it forms another nested holon that is the next level of the holarchic structure. This recursive process starts from the parents of the leaves in the tree and progresses up to the root as shown in Figure~\ref{fig:holarchy}. Learning iterations can be independently executed in every holonic branch before the process progresses to the next level of the holarchic structure, in which a new series of learning iterations are executed. The top holonic branch is actually the whole tree topology and therefore the execution of learning iterations at this top level corresponds to the learning iterations without a holarchic structure. In other words, the concept of holarchy introduces multiple localized, nested and incremental learning processes. 

The nested learning processes of a holarchic system enable the management of each level independently from the higher and lower levels, which further allows the management separation between the agents at each level. This is a divide-and-conquer approach to complex system management and hence improves localization, parallelism, reusablity and diversity with a potential in improving system performance and robustness while decreasing costs~\cite{gott-oc2017}. From a software engineering perspective, the following key properties of a holonic design are identified~\cite{alife2018,gott-oc2017}: (i) \textit{Bottom-up abstraction} that represents the aggregation of information from lower levels. (ii) \textit{Partial isolation} within and among levels -- ensuring an agent's decision results from the aggregation of agents' decisions at a lower level and results in the agent's decision at a higher level, hence other agents do not influence. (iii) \textit{Inter-level time tuning} -- ensuring that the relative execution times at different holarchic levels are set in order to avoid cross-level oscillations and divergence.

This paper studies whether learning processes structured within holarchies are cost-effective countermeasures to adapt to the uncertainties of distributed environments. Holarchies can provide the following operational flexibility and adaptation: (i) The learning process can be limited to a targeted part of the network to prevent a network-wide use of computational and communication resources, i.e. agents can participate in the learning process on-demand. Any failure to serve the participation does not disrupt the learning process that can continue within part of the network. (ii) The mapping and deployment of holarchies on the network can be designed according to the network heterogeneity, i.e. varying latency, computational and battery capacities. For instance, higher performing nodes can be placed at the bottom of a holarchy to serve their more frequent recurrent use in bottom-up learning interactions.

\begin{figure}[!htb]
\centering
\subfigure[\textbf{Baseline}: Throughout runtime a fixed number of learning iterations is performed during which convergence is potentially achieved. The bottom-up phase starts from the leaves and progresses level-by-level up to the root, while the reverse process (backpropagation) is performed in the top-down phase. In both phases, the parent-children interactions always progress to the next level, in contrast to the holarchic strategies in which learning iterations are performed in nested tree branches as shown below.]{\includegraphics[width=1.0\textwidth]{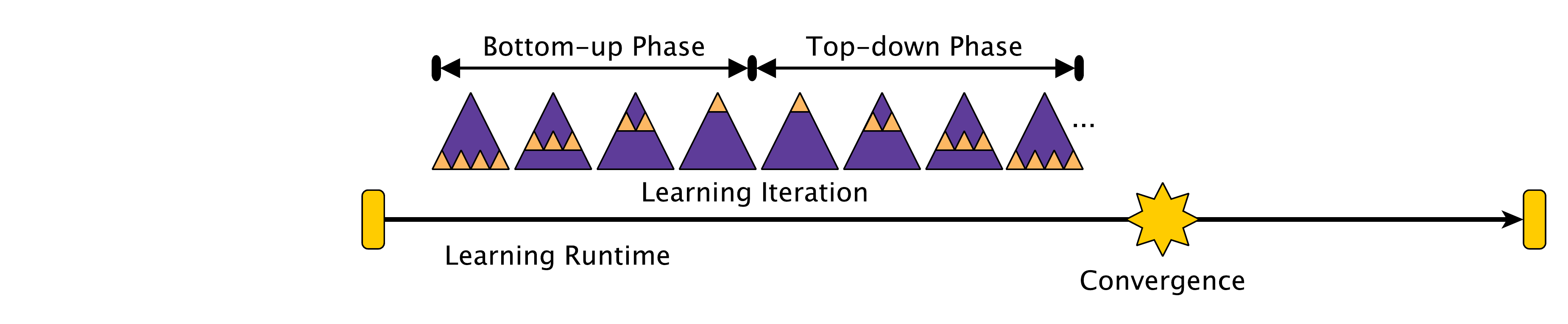}}
\subfigure[\textbf{Holarchic initialization}: Before the baseline execution, holarchic learning is performed.]{\includegraphics[width=1.0\textwidth]{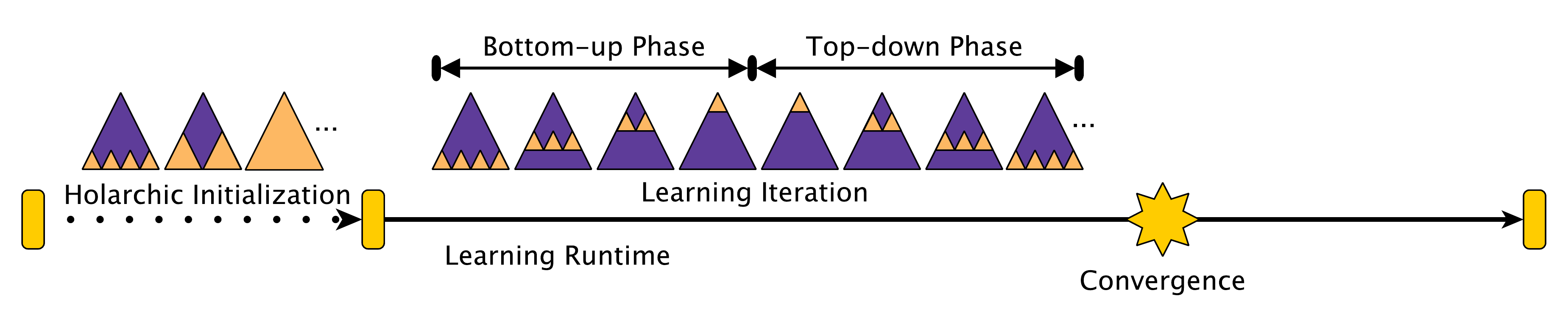}}
\subfigure[\textbf{Holarchic runtime}: Holarchic learning is performed throughout the runtime. Note that one learning iteration corresponds to multiple holarchic iterations performed in each holon.]{\includegraphics[width=1.0\textwidth]{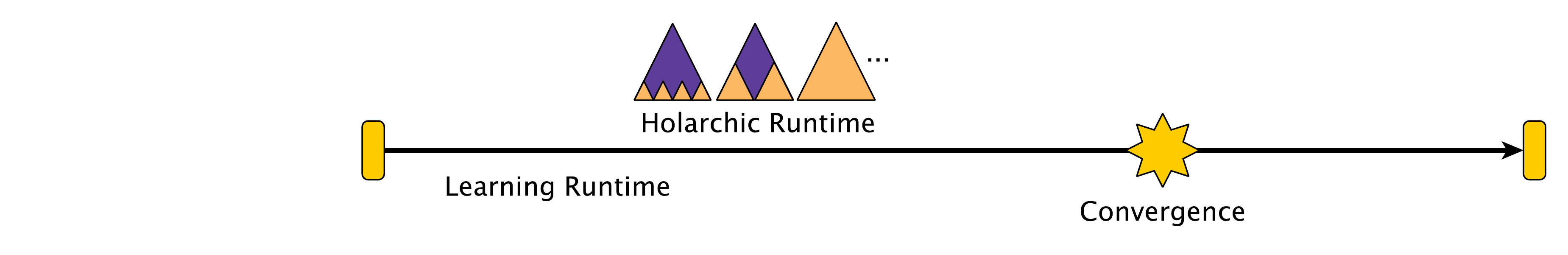}}
\subfigure[\textbf{Holarchic termination}: Holarchic learning is activated after convergence is reached with baseline to potentially discover improved solutions.]{\includegraphics[width=1.0\textwidth]{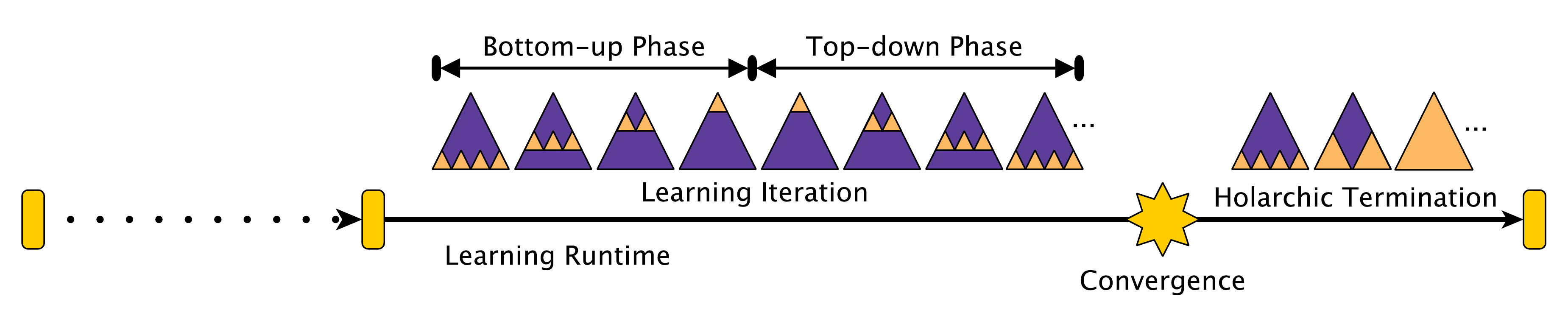}}
\caption{The four learning schemes studied and compared in this paper.}\label{fig:holarchic-schemes}
\end{figure}

This paper studies three self-adaptation scenarios of the holarchic concept in decentralized deep learning each designed for performance exploration, mitigation and boosting respectively: (i) \emph{holarchic initialization}, (ii) \emph{holarchic runtime} and (iii) \emph{holarchic termination}. Assume a \emph{baseline} scheme that involves a tree structure with agents interacting in a (i) \emph{bottom-up phase} and a (ii) \emph{top-down phase} that both complete a \emph{learning iteration}. The former phase may represent a fit forward learning process starting from the leaves and completing to the root while the latter phase a backpropagation starting from the root and reaching back the leaves. Without loss of generality, an exact decentralized learning algorithm realizing these concepts is presented in Section~\ref{sec:i-epos}. Learning iterations repeat to decrease a \emph{cost function}. Learning \emph{converges} when a certain number of iterations is performed or when the cost function cannot be decreased further. Figure~\ref{fig:holarchic-schemes}a illustrates the baseline scheme. 

Figure~\ref{fig:holarchy}c depicts one baseline learning iteration, while within each nested holon formed during the bottom-up phase several learning iterations are performed. This process is common in all holarchic schemes. Holarchic initialization is applied before baseline to perform an exploration of the search space. Several learning iterations can be performed before switching to baseline as illustrated in Figure~\ref{fig:holarchic-schemes}b. In contrast, holarchic runtime applies holarchic learning throughout runtime without switching to baseline as shown in Figure~\ref{fig:holarchic-schemes}c. This scheme is applicable in self-adaptation scenarios of failures or conservation of resources to mitigate losses of learning performance. Finally, holarchic termination is applied after the baseline convergence as the means to further boost the baseline performance. This self-adaptation scheme is shown in Figure~\ref{fig:holarchic-schemes}d.

\section{Applicability of Holarchic Schemes}\label{sec:i-epos}

This section illustrates a case study for the applicability of holarchic structures in decentralized deep learning for combinatorial optimization problems: I-EPOS\footnote{Available at \url{epos-net.org} (last accessed: September 2018)}, the \emph{Iterative Economic Planning and Optimized Selections}~\cite{Pournaras2018,Pilgerstorfer2017}. I-EPOS consists of agents that autonomously plan resources they consume and produce. Planning is a process of \emph{resource scheduling} or \emph{resource allocation}. For instance agents may represent smart phone apps (personal assistants), cyber-physical controllers or smart home information systems with the capability to plan the energy consumption of residential appliances, the charging of electric vehicles or the choices of bike sharing stations. Planning serves the local resource requirements of users as well as system-wide objectives, for instance, the decrease of demand oscillations in the whole power grid to prevent blackouts~\cite{Pournaras2017b}, the synchronization of power demand with the availability of renewables~\cite{Aghaei2013} or the load-balancing of bike sharing stations to decrease the operational costs of manual bike relocations~\cite{Singla2015,Pournaras2018}. 

I-EPOS introduces the computational model of locally generated \emph{possible plans} that represent users' operational flexibility on how resources can be consumed or produced. For instance, a user may turn on a laundry machine earlier or later in time, can choose among two or more stations to return a shared bike, etc. Computationally, plans are vectors of real values and agents need to collectively choose one and only one of these plans to execute so that the summation of all these selected plans satisfies a system-wide objective measured by a \emph{global cost} function. This paper focuses on the computational problem of the variance minimization that is a quadrtic cost function, which cannot be locally minimized: coordination and collective decision-making is required~\cite{Rockafellar2000}. Choosing the optimum combination of plans, whose summation minimizes a quadratic cost function is a non-convex combinatorial optimization problem known to be NP-hard~\cite{Puchinger2005}. 

Among the global cost that captures system-wide objectives, agents can also assess a \emph{local cost} of their possible plans that may represent a notion of \emph{discomfort or inconvenience}~\cite{epos_fairness}. For instance, the longer the time period a user shifts the energy consumption earlier or later in time to prevent a power peak (global cost reduction) the higher the level of disruption is in the regular residential activities of a user (local cost increase). The agents' choices can be autonomously parameterized to increase or decrease the priority of minimizing the local cost over the global cost. This trade-off is regulated by the $\lambda$ parameter for each agent. A $\lambda=0$ results in choices that exclusively minimize the global cost, i.e. variance, and ignores the local cost. In contrast, a $\lambda>0$ biases the agents' choices to favor plans with a lower local cost. 

Learning is performed as follows: agents self-organize~\cite{Pournaras2013b} in a tree network topology over which collective decision-making is performed -- a plan is chosen by taking into account the following aggregate information (element-wise summation of vectors): (i) the aggregate plan choices of the agents in the branch underneath made available during the bottom-up phase and (ii) the aggregate plan choices of all agents at the previous learning iteration made available during the top-down phase. Note that decision-making remains highly localized and decentralized as the planning information of the other agents is always at an aggregate level, i.e. the possible plans of other agents are not explicitly required. 

Several of the following factors influence the learning performance defined by the level of the global cost reduction and the number of learning iterations required to minimize variance: (i) the positioning of the agents in the tree that determines the order of the collective choices made, (ii) the $\lambda$ parameter that regulates the trade-off of global versus local cost and (iii) the overall topological structure and specifically in this paper the number of children $c$ in balanced trees is studied. 

Improving the learning performance by repositioning the agents in the tree or adapting the topology is complex and costly in distributed environments as the aforementioned self-organization methodologies are based on supplementary distributed protocols that consume resources, i.e. exchange of messages, computational power, energy, etc. Moreover, any node or link failure limits the computation of the aggregate plans at a branch level and therefore the collective decision-making cannot be anymore performed over all participating agents. The applicability of the holarchic schemes does not require any change in the logic of I-EPOS. The algorithm is localized and applied within multiple nested and connected branches even when the network becomes disconnected due to node and link failures: A disconnected agent triggers adaptation by children that turn to roots of holons and initiate the top-down phase of I-EPOS. The fact that a number of agents is isolated and does not participate in the learning process is the self-adaptation means to traverse the optimization space via alternative pathways. This has the potential to explore improving solutions, mitigate performance loss compared to a total interruption of I-EPOS, or even boost the reduction of the global cost that cannot be decreased anymore with I-EPOS. 

Note that I-EPOS is selected to evaluate the applicability of holarchic schemes as it is a fully decentralized and hierarchical learning algorithm. The goal of learning in terms of whether it is designed for classification, prediction, pattern recognition, etc. or whether textual or image data are used, do not influence the applicability of holarchic schemes. The rest of this paper illustrates an empirical performance analysis of the three holarchic schemes and their applicability to I-EPOS. 

\section{Experimental Methodology}\label{sec:experimental-methodology}

This section illustrates the system prototyping, the varying dimensions in the experiments and the experimental settings. It also illustrates the evaluation metrics used to assess the holarchic learning schemes.

\subsection{Prototyping and test runs}\label{subsec:prototyping}

An open-source implementation of I-EPOS\footnote{Available at \url{https://github.com/epournaras/EPOS} (last accessed: September 2018).} is used for the prototyping of the holarchic schemes\footnote{Available at \url{https://github.com/ysrivatsan/EPOS/tree/Srivatsan} (last accessed: September 2018).}. The software is implemented using the Protopeer distributed prototyping toolkit~\cite{Galuba2009} and is designed to run in a simulation and live mode. A software artifact\footnote{Available at \url{http://epos-net.org/software/exemplar/} (last accessed: September 2018).} of EPOS for the broader scientific community is available for further research and evaluations~\cite{Pournaras2018,Pilgerstorfer2017}. The actual experiments are deployed and parallelized in the Euler\footnote{Available at \url{https://scicomp.ethz.ch/wiki/Euler} (last accessed: September 2018).} cluster computing infrastructure of ETH Zurich.

\subsection{Varying dimensions and performed experiments}\label{subsec:varying dimensions}

The following system dimensions are studied: (i) \emph{application scenarios}, (ii) \emph{holarchic schemes}, (iii) \emph{scale of holarchy}, (iv) \emph{number of children $c$ in the tree network}, (v) \emph{different agent preferences $\lambda$}. 

Synthetic and empirical plans are generated for 1000 agents using data from real-world pilot projects. These four application scenarios are referred to as follows: (i) \emph{synthetic}, (ii) \emph{bike sharing}, (iii) \emph{energy demand} and (iv) \emph{electric vehicles}. 

The synthetic dataset consists of 16 possible plans of size 100 generated from a standard normal distribution with a mean of 0 and a standard deviation of 1. A random local cost is assigned to the plans, i.e. the index of the plan represents its cost.

The bike sharing dataset\footnote{Available at \url{http://epos-net.org/shared/datasets/EPOS-BICYCLES.zip} (last accessed: September 2018).} of the Hubway bike sharing system\footnote{The plans are generated using the dataset made available in the context of the Hubway Data Visualization Challenge: \url{http://hubwaydatachallenge.org/} (last accessed: September 2018). Although this dataset does not contain personalized records, user trips are extracted from user information: zip-code, year of birth and gender. All trips that have common values in these fields are assumed to be made by the same user. The timeslot is chosen from 8:00 am to 10:00 am. All historic unique trips a user did in the defined timeslot of a week day are considered as the possible plans for that day.} in Paris is used to generate a varying number of plans of size 98 for each agent based on the unique historic trips performed by each user. Therefore, the plans represent the trip profiles of the users and they contain the number of incoming/outgoing bike changes made by each user in every station~\cite{Pournaras2017c}. The local cost of each plan is defined by the likelihood of a user to not perform a trip instructed in the plan~\cite{Pournaras2018}. For instance, if three plans are chosen $4$, $5$ and $1$ days during the measured time period respectively, the local cost for these plans is $0.6$, $0.5$ and $0.9$ respectively. 

%The SHUFFLE generation scheme randomly shuffles the values in the first plan, and as this scheme perturbates all 144 values of the original plan, SHUFFLE plans have the lowest preference score equal to $\frac{1}{144}$. The SWAP-15 generation scheme randomly selects 15 pairs of values of the original plan to swap, and the preference of the plans is adjusted to $\frac{1}{15}$. Similarly, SWAP-30 scheme randomly selects 30 pairs to swap and their preference score is $\frac{1}{30}$. Consequently, mean and standard deviation of possible plans are constant for each agent.\\
%
%Preference scores are used for local cost optimization which implies selecting possible plans with high preference scores with respect to the other global objectives. As \iepos is a minimization algorithm, preference scores are converted to their respective \textit{complements} called \textit{costs} as follows: if preference score is $h$, then its cost is $1-h$. Note that each cost value is between 0 and 1. \\

The energy demand dataset\footnote{Available at \url{http://epos-net.org/shared/datasets/EPOS-ENERGY-SUBSET.zip} (last accessed: September 2018).} is generated via disaggregation of aggregate load obtained from the Pacific Northwest Smart Grid Demonstration Project (PNW) by Battelle\footnote{Available upon request at \url{http://www.pnwsmartgrid.org/participants.asp} (last accessed: September 2018)}. The disaggregation algorithm and the raw data are illustrated in earlier work~\cite{Pournaras2017b}. The generated dataset contains 5600 agents representing residential consumers. Every agent has 10 possible plans, each with length 144 containing electricity consumption records for every 5 minutes. The first plan corresponds to the raw data. The next three possible plans are obtained via the \textsc{\scriptsize SHUFFLE} generation scheme~\cite{Pournaras2017b} that randomly permutes the values of the first plan. The next three plans are generated with the \textsc{\scriptsize SWAP-15} generation scheme~\cite{Pournaras2017b} that randomly picks a pair of values from the first plan and swaps their values. The process repeats 15 times. Respectively, the last three plans are generated by \textsc{\scriptsize SWAP-30}~\cite{Pournaras2017b} that applies the same process 30 times. The local cost of each plan represents the level of perturbation introduced on the first plan, i.e. on the disaggregated data, by the plan generation scheme. It is measured by the standard deviation of the difference between the raw and the perturbed plan values, element-wise: $\sigma(x_{1}-y_{1},...,x_{144}-y_{144})$.

The plans\footnote{Available at \url{http://epos-net.org/shared/datasets/EPOS-ELECTRIC-VEHICLES.zip} (last accessed: September 2018).} for the electric vehicles are generated using data\footnote{Available at \url{www.nrel.gov/tsdc} (last accessed: September 2018). Electric vehicles equipped with GPS are selected.} from the Household Travel Survey of the California Department of Transportation during 2010--2012. The plans concern the energy consumption of the electric vehicles by charging from the power grid. Four plans per agent with size 1440 are generated by extracting the vehicle utilization using the historical data and then computing the state of charge by redistributing the charging times over different time slots. The methodology is outlined in detail in earlier work~\cite{Pournaras2017c}. The local cost of each plan is measured by the likelihood of the vehicle utilization during the selected charging times.

The learning schemes studied are the \emph{baseline} that is default I-EPOS and the three holarchic schemes: \emph{Holarchic initialization} is used as an \emph{exploration strategy} to evaluate its likelihood to improve the learning capacity. \emph{Holarchic runtime} is used as a \emph{mitigation strategy} to evaluate the maintenance of learning capacity in distributed environments under uncertainties. Finally, \emph{holarchic termination} is used as a \emph{boosting strategy} to evaluate the likelihood of improving the learning capacity. In each holon of the holarchic schemes within one main learning iteration, $\tau=5$ holarchic iterations are executed. 

Two holarchic scales are evaluated: (i) \emph{full} and (ii) \emph{partial}. The full scale uses all levels of the baseline tree network to apply a holarchic scheme as also shown in Figure~\ref{fig:holarchy}. In contrast, partial scale is applied in one branch under the root.

The influence of the topological tree structure and agent preferences on the learning capacity is studied by varying respectively the number of children as $c=2,..,5$ and the $\lambda$ parameter as $\lambda=0,0.25,0.5,0.75$. Table~\ref{table:experiments} summarizes the dimensions and their variations in the performed experiments. 

\begin{table}[!htb]
\centering
\caption{Dimensions and their variations in the total of 864000 experiments.}\label{table:experiments}
\footnotesize{
\begin{tabular}{lp{1.7cm}p{1.7cm}p{1.7cm}p{1.7cm}}  
\toprule
%\multicolumn{2}{c}{Item} \\
%\cmidrule(r){1-2}
\textbf{Dimension} & \textbf{Variation 1} & \textbf{Variation 2} & \textbf{Variation 3} & \textbf{Variation 4}\\
\midrule
Application scenario & Synthetic & Bike sharing & Energy demand & Electric vehicles \\\addlinespace
\midrule
Learning scheme & Baseline & Holarchic initialization & Holarchic runtime & Holarchic termination \\\addlinespace
\midrule
Holarchic scale & Full & Partial & - & - \\\addlinespace
\midrule
Number of children ($c$) & $c=2$ & $c=3$ & $c=4$ & $c=5$ \\\addlinespace
\midrule
Agent preferences ($\lambda$) & $\lambda=0$ & $\lambda=0.25$ & $\lambda=0.5$ & $\lambda=0.75$ \\
\bottomrule\addlinespace
\textbf{Total experiments}: & \textbf{864000} &  &  &  \\
\bottomrule
\end{tabular}
}
\end{table}

The total experiments performed are calculated as follows: For the dimension of application scenarios, 4 variations are counted and 3 variations for the learning schemes given that the baseline and the holarchic termination can be generated within one experiment. The partial scale uses all possible branch combinations: $2+3+4+5=14$ variations plus 4 variations for the full scale result in 18 total variations for the two dimensions of holarchic scale and number of children. Finally, 4 variations are counted for the dimension of agent preferences. The total number of $4\ast 3\ast 18\ast 4=864$ variation combinations is the total number of experiments performed. Each experimental combination is repeated 1000 times by (i) 1000 samples of possible plans in the synthetic scenario and (ii) 1000 random assignments of the agents in the tree network in the other three application scenarios. Therefore, the total number of experiments performed is $864\ast 1000=864000$.

\subsection{Evaluation metrics}\label{subsec:evaluation-metrics}

Performance is evaluated with the following metrics: (i) \emph{standardized global cost}, (ii) \emph{improvement index} and (iii) \emph{communication cost}.

The standardized global cost is the variance of the global plan at convergence time. The minimization of the variance is the optimization objective and therefore the variance is used as the criterion of the learning capacity. Standardization\footnote{Standardization transforms the global cost values to have zero mean and units of variance as follows: $\frac{x-\mu}{\sigma}$} is applied on the variance so that the learning capacity among different datasets can be compared. 

The improvement index $I$ measures the reduction or increase of the global cost at convergence time for the holarchic schemes compared to the baseline. Positive values indicate an improvement of the baseline, while negative values show a deterioration. The improvement index is measured as follows: 

\begin{align}\label{eq:improvement-index}
I=\frac{C_{\mathsf{b}}-C_{\mathsf{h}}}{C_{\mathsf{b}}+C_{\mathsf{h}}}
\end{align}

\noindent where $C_{\mathsf{b}}$ is the global cost for the baseline and $C_{\mathsf{h}}$ is the global cost for a holarchic scheme, both at convergence time. The improvement index is calculated based on the principle of the symmetric mean absolute error, which compared to the mean absolute error can handle single zero values, it is bound to $[-1,1]$ and eliminates very large values originated by low denominators~\cite{Hyndman2006}.

The communication cost measures the number of messages exchanged to complete a learning iteration and can be distinguished to \emph{total} and \emph{synchronized}. The total communication cost counts all exchanged messages between the agents during a learning iteration and it is calculated for the baseline $M_{\mathsf{b}}$ as follows:

\begin{align}\label{eq:total-communication-baseline}
M_{\mathsf{b}}=2(c^{0}+...+c^{l}-1)
\end{align}

\noindent where $c$ is the number of children in a balanced tree. Equation~\ref{eq:total-communication-baseline} sums up the number of agents in each level of the tree and substracts 1 to count the number of links. Multiplication by 2 counts both bottom-up and top-down phases. The total communication cost of a holarchic scheme can be measured as follows:

\begin{align}\label{eq:total-communication-holarchy}
M_{\mathsf{t}}=2\tau \sum_{j=0}^{l} c^{l-j}(c^{0}+...+c^{j}-1) 
\end{align}

\noindent where $c$ is here again the number of children in a balanced tree/holarchy and $\tau$ is the number of holarchic iterations. The summation starts from leaves (level $j=0$), and progresses to the root of the holarchy (at level $j=l$). Equation~\ref{eq:total-communication-holarchy} multiplies the number of agents $c^{l-j}$ at each level $l-j$ with $2\tau$ times ($\tau$ bottom-up and $\tau$ top-down holarchic iterations) the number of agents in the branches underneath: $c^{0}+...+c^{j}-1$. For example, $j=0$ corresponds to Figure~\ref{fig:holarchy}a with a communication cost of $2\tau 2^{2}(2^{0}+2^{1}-1)=16\tau$. $j=1$ corresponds to Figure~\ref{fig:holarchy}b with a communication cost of $2\tau 2^{1}(2^{0}+2^{1}+2^{2}-1)=24\tau$ and respectively, for $j=2$ and Figure~\ref{fig:holarchy}c communication cost is calculated as $2\tau 2^{0}(2^{0}+2^{1}+2^{2}+2^{3}-1)=28\tau$. These nested calculations for the full holarchy sum up to $M_{\mathsf{t}}=(16+24+28)\tau=68\tau$ messages. 

The synchronized communication cost counts the number of messages exchanged within holons, while counting this number only once for holons at the same level which can exchange messages in parallel. For instance, Figure~\ref{fig:holarchy}a illustrates four parallel holons with a total communication cost of $2\tau 2^{2}(2^{0}+2^{1}-1)=16\tau$ messages. Instead, the synchronized communication cost counts for $2\tau (2^{0}+2^{1}-1)=4\tau$ messages. The synchronized communication cost of a full holarchy can be measured as follows:

\begin{align}\label{eq:synchronized-communication-holarchy}
M_{\mathsf{s}}=2\tau\sum_{j=0}^{l} (c^{0}+...+c^{j}-1)
\end{align}

\noindent where $c$ is the number of children and $\tau$ the number of holarchic iterations. Note that the synchronized communication cost considers holons performing in parallel and not individual agents, that is why within each holon all messages exchanged are counted. For the same reason, the synchronized communication cost for baseline is not defined for a fairer comparison with holarchic schemes.

\subsection{Computational challenge}\label{subsec:computational-challenge}

\begin{figure}[!htb]
\centering
\includegraphics[width=0.244\textwidth]{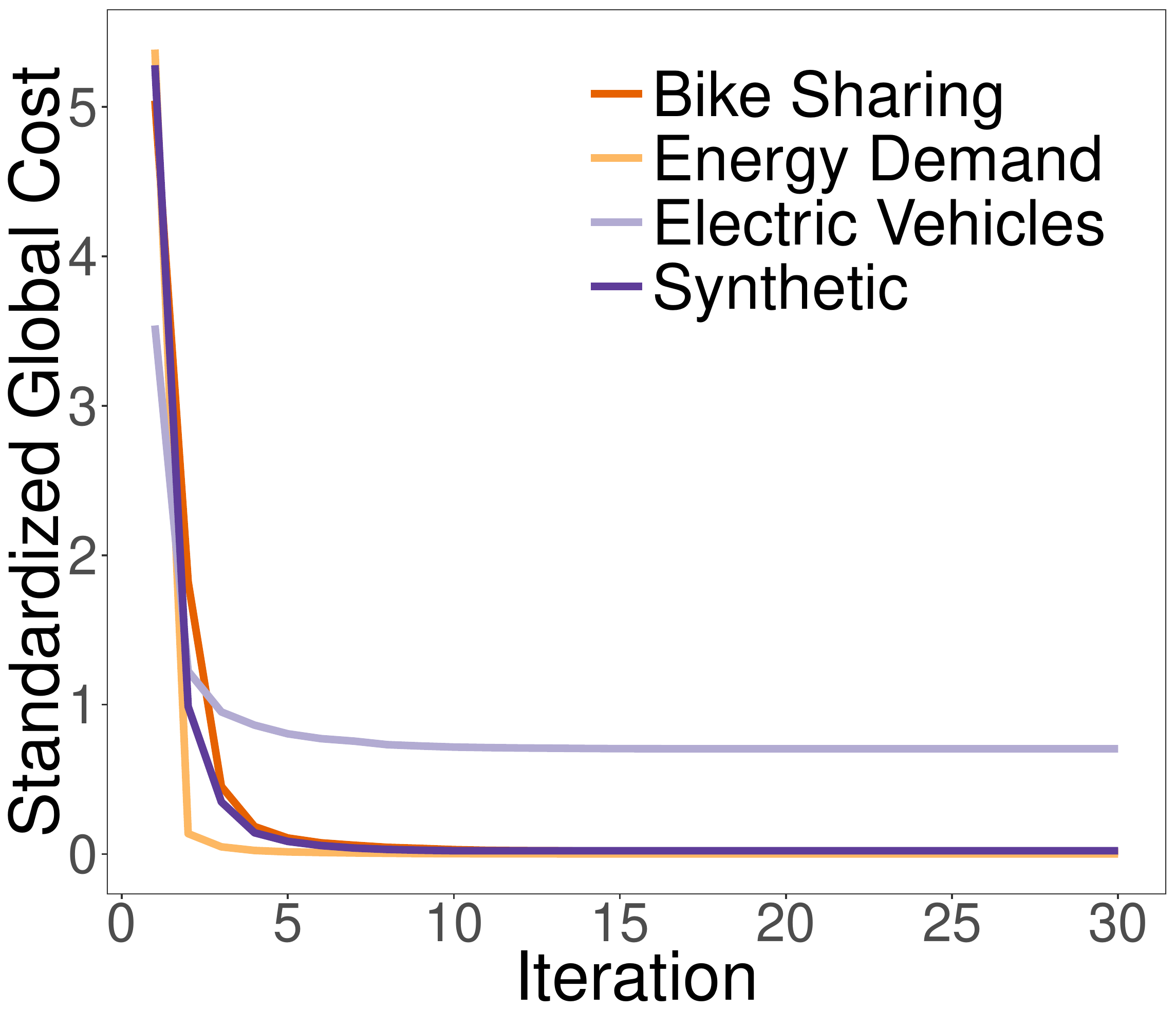}
\caption{Learning curves of I-EPOS for the four benchmark application scenarios with 1000 agents generating 4 plans and choosing a plan with $\lambda=0$.}\label{fig:learning-baseline}
\end{figure}

To better understand the computational challenge and context in which the performance of the holarchic schemes is studied, the performance characteristics of the baseline are illustrated in this section. Figure~\ref{fig:learning-baseline} illustrates the learning curves of the baseline in the four application scenarios. 

\begin{figure}[!htb]
\centering
\subfigure[Synthetic]{\includegraphics[width=0.244\textwidth]{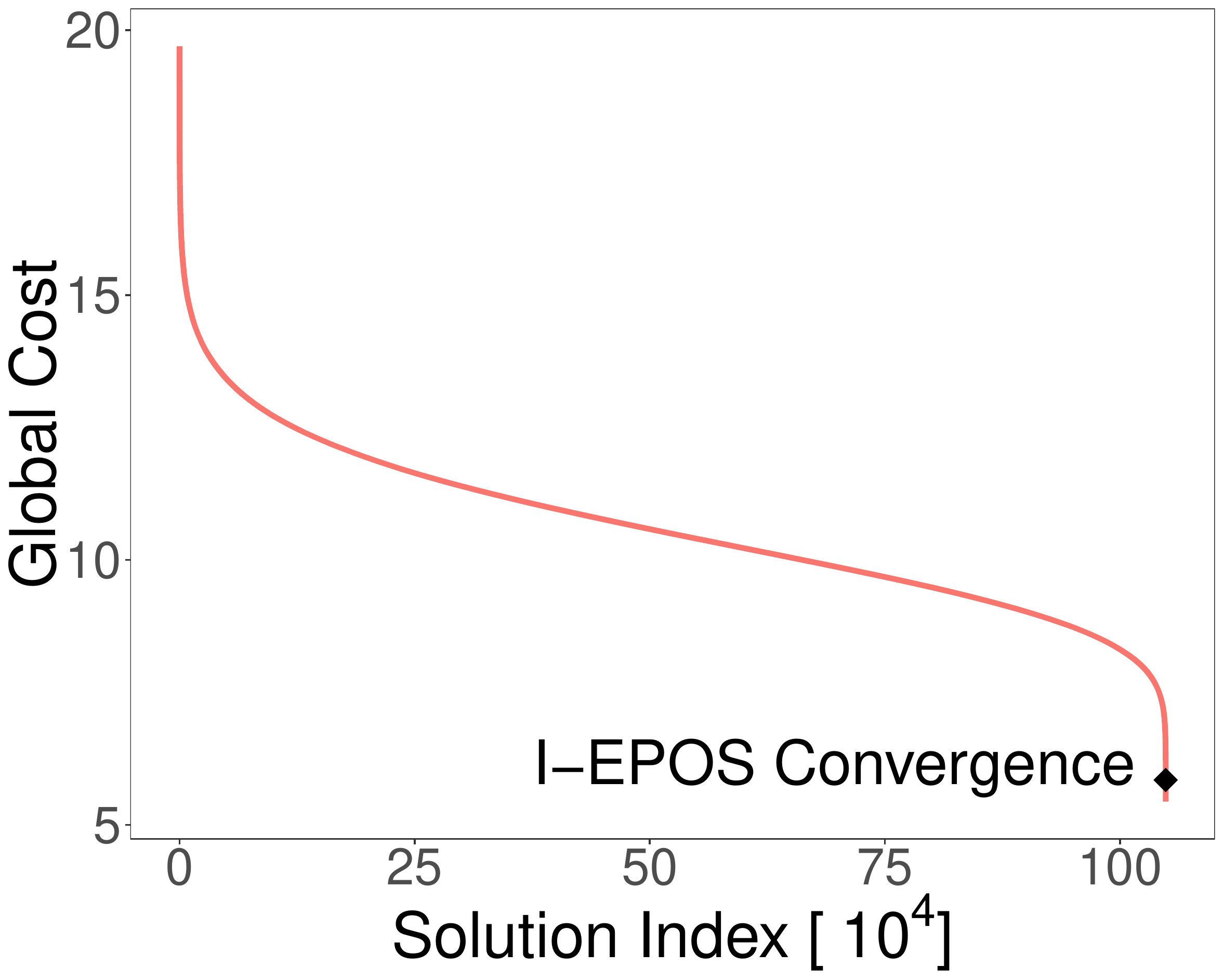}}
\subfigure[Bike sharing]{\includegraphics[width=0.244\textwidth]{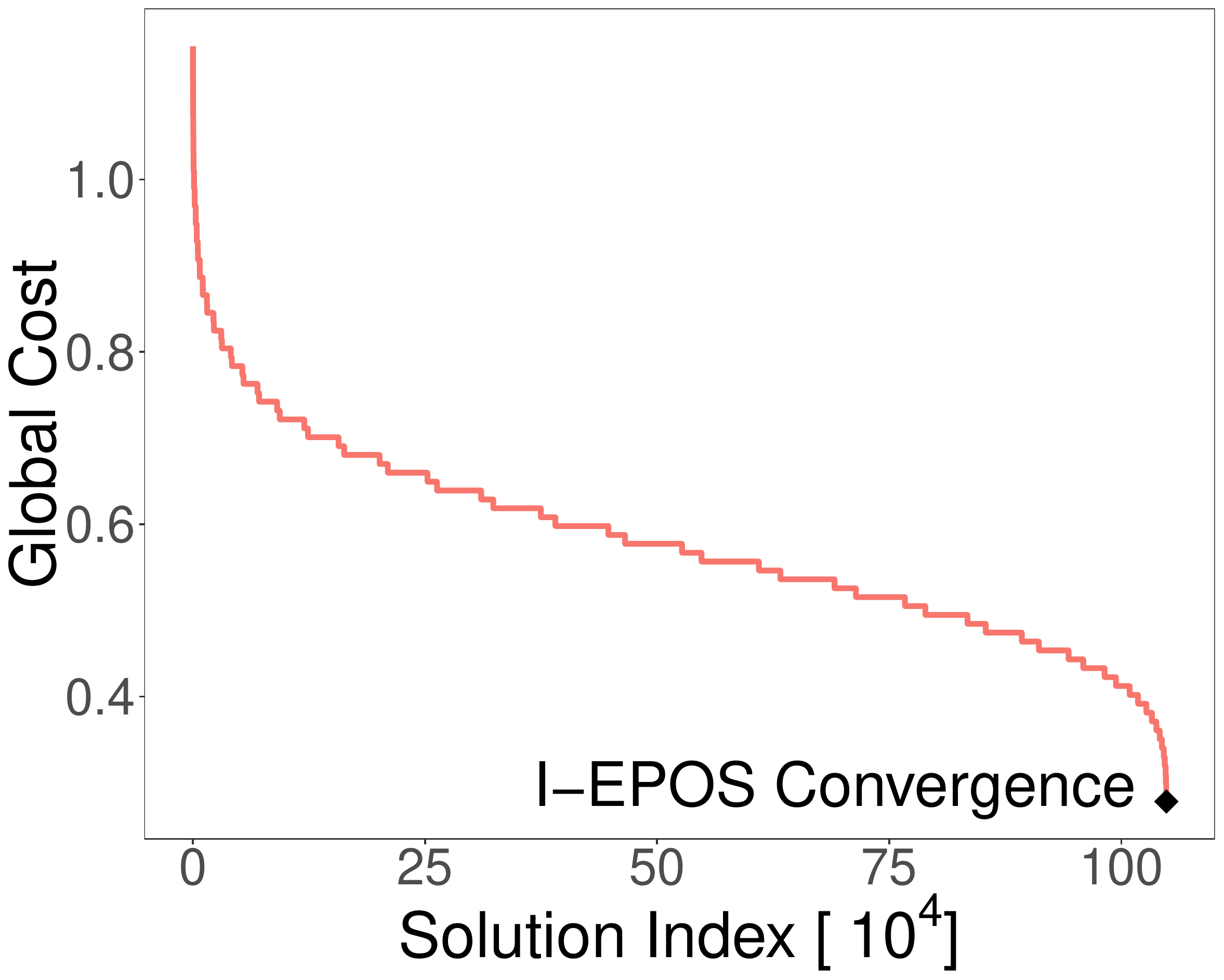}}
\subfigure[Energy demand]{\includegraphics[width=0.244\textwidth]{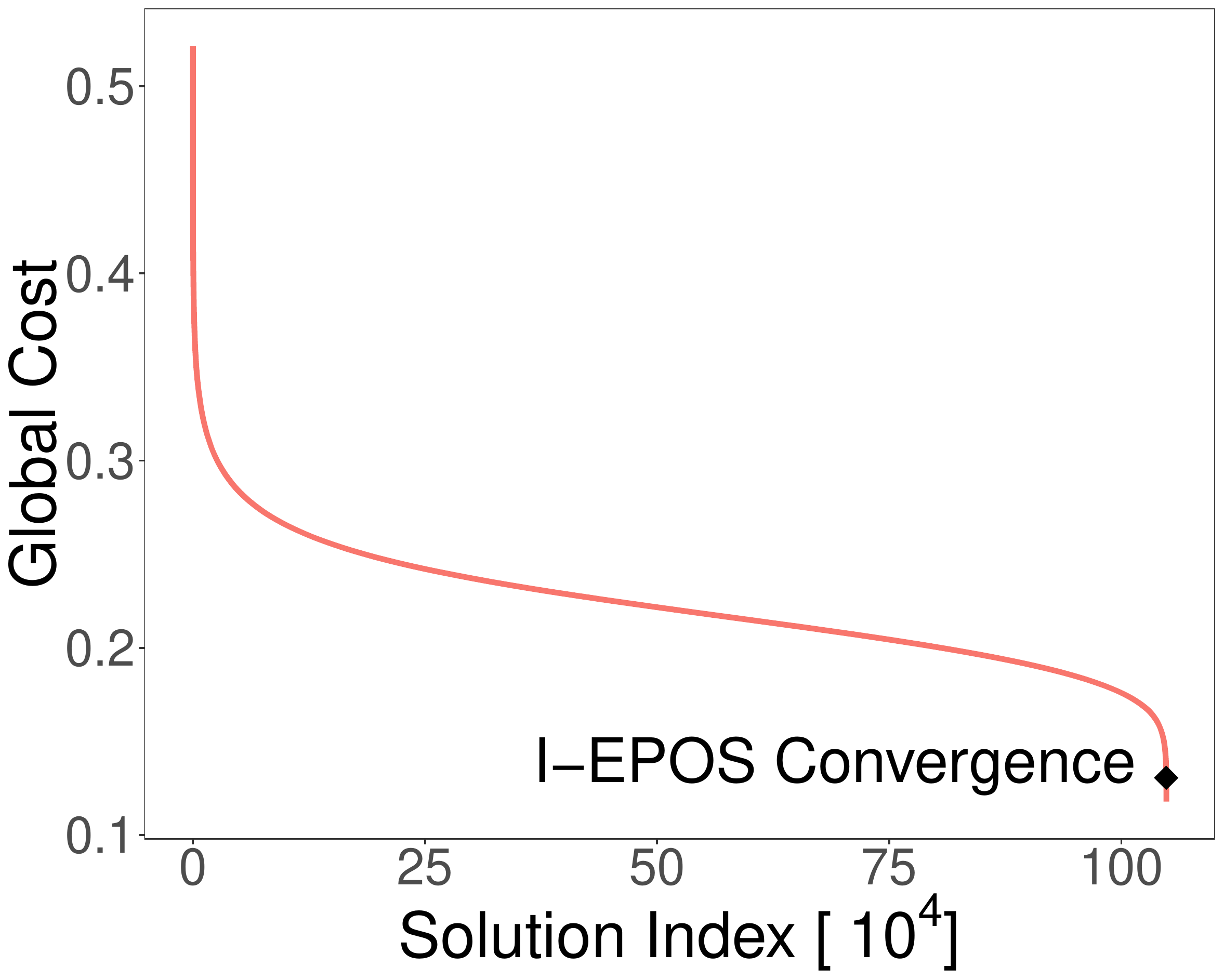}}
\subfigure[Electric vehicles]{\includegraphics[width=0.244\textwidth]{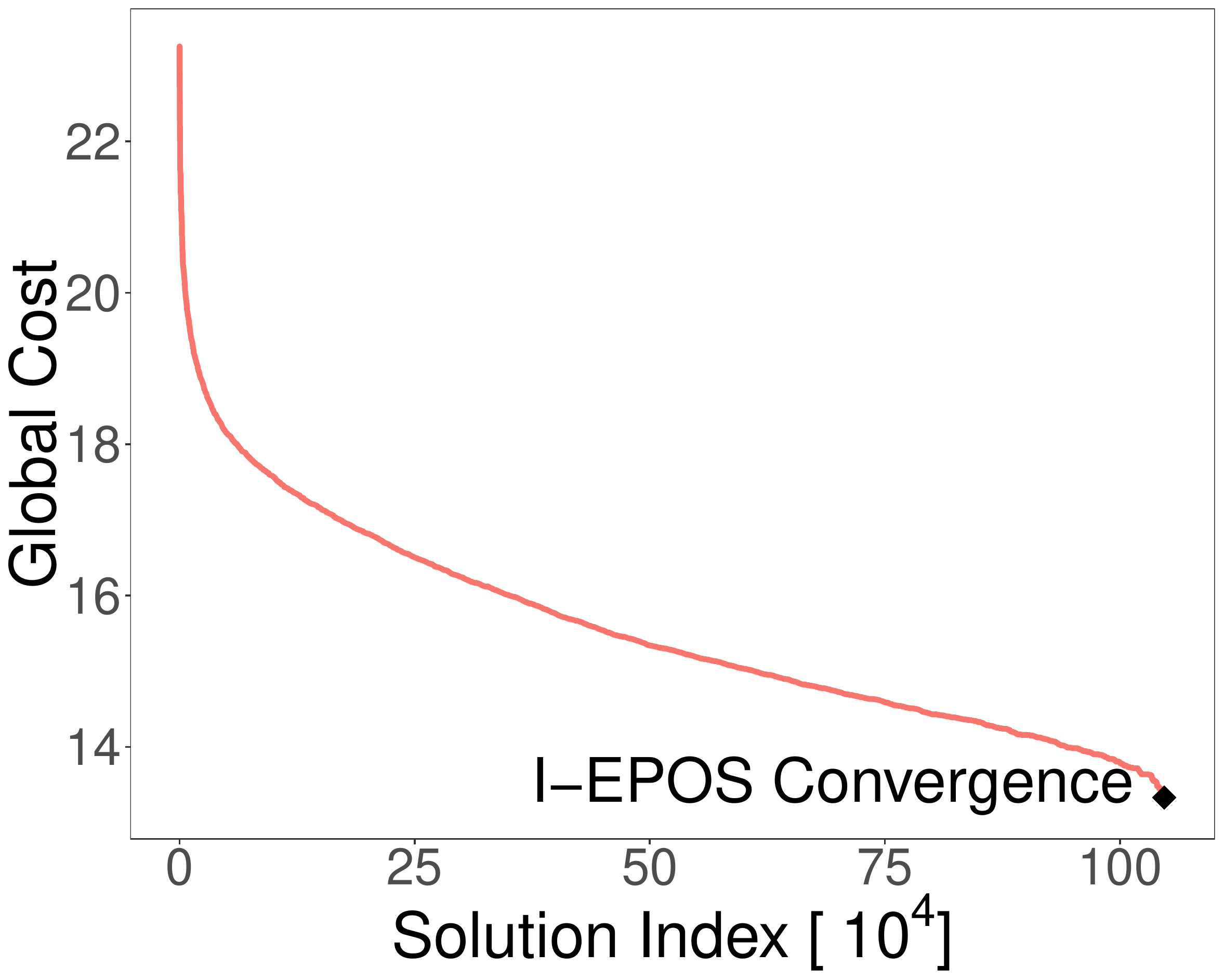}}
\caption{Global optimality of I-EPOS for the four benchmark applicaiton scenarios with 10 agents generating 4 plans and choosing a plan with $\lambda=0$.}\label{fig:optimality}
\end{figure}

The learning performance of I-EPOS shows the following behavior: Global cost decreases dramatically in  very few iterations in all application scenarios, while the decrease is monotonous. Therefore, I-EPOS has a superior efficiency to learn fast combinations of plans that minimize the variance by executing 10-15 iterations. Convergence does not necessarily mean that the globally optimum solution is found. The evaluation of the global optimality in a system with 1000 agents and more than 4 plans per agents is computationally infeasible given the exponential complexity of the combinatorial space: $4^{1000}$. For this reason, the global optimality is evaluated using brute force search in a small-scale system of 10 agents with 4 plans per agent and $\lambda=0$. Therefore the total number of solutions is $4^{10}$. Each experiment is repeated 10 times by shuffling the agents over a binary tree. Figure~\ref{fig:optimality} illustrates the global optimality results for each application scenario. 

I-EPOS finds the 0.007\%, 0\%, 0.017\% and 0.153\% top solution in each of the application scenarios of Figure~\ref{fig:optimality}. Note that such a significant optimality may not be achieved for systems with thousands of agents and several plans. Nevertheless, designing a new learning scheme to overpass this performance level, without introducing additional complexity and resources is a challenge and potentially not an endeavor worth pursuing. Instead this paper studies holarchic structures as learning strategies to explore, mitigate and boost the cost-effectiveness of I-EPOS in distributed environments and in this sense the notion of holarchy is the means for decentralized learning to tolerate their uncertainties.

\section{Experimental Evaluation}\label{sec:experimental-evaluation}

This section illustrates the learning capacity of the three holarchic schemes followed by the trade-offs and cost-effectiveness of the holarchic runtime. All main experimental findings are covered in this section and an appendix provides supplementary results that cover the broader range of the varying parameters. The results in the appendix are included for the sake of completeness of the paper and future reference.

\subsection{Learning capacity}\label{subsec:learning-capacity}

\begin{figure}[!htb]
\centering
\includegraphics[width=1.0\textwidth]{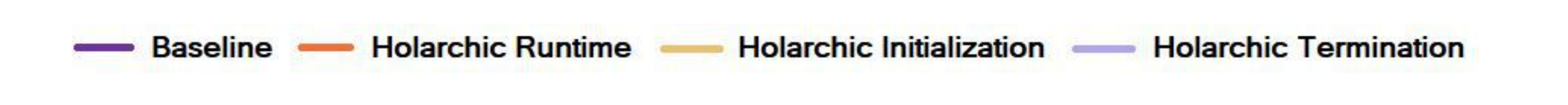}\\
\subfigure[Synthetic]{\includegraphics[width=0.244\textwidth]{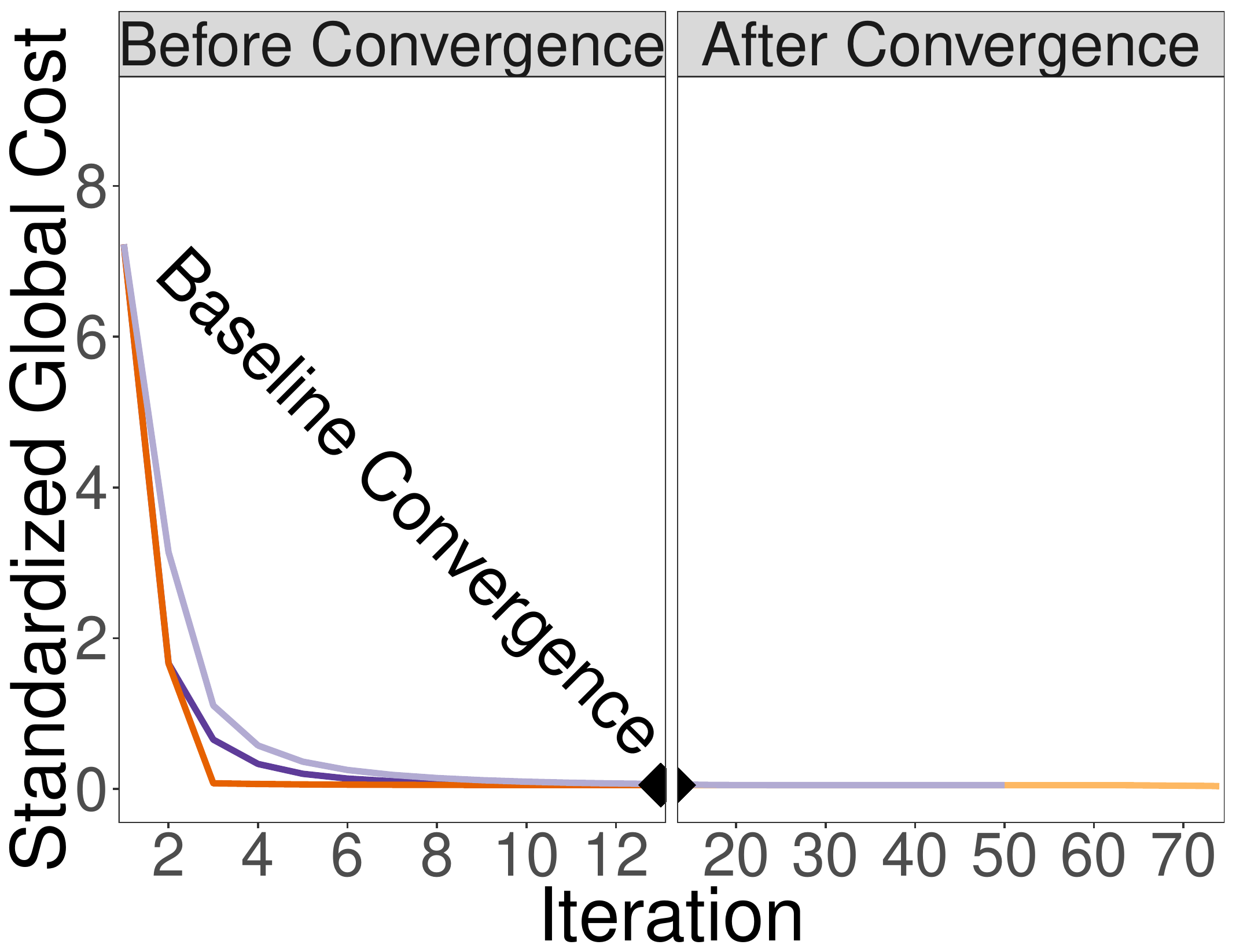}}
\subfigure[Bike sharing]{\includegraphics[width=0.244\textwidth]{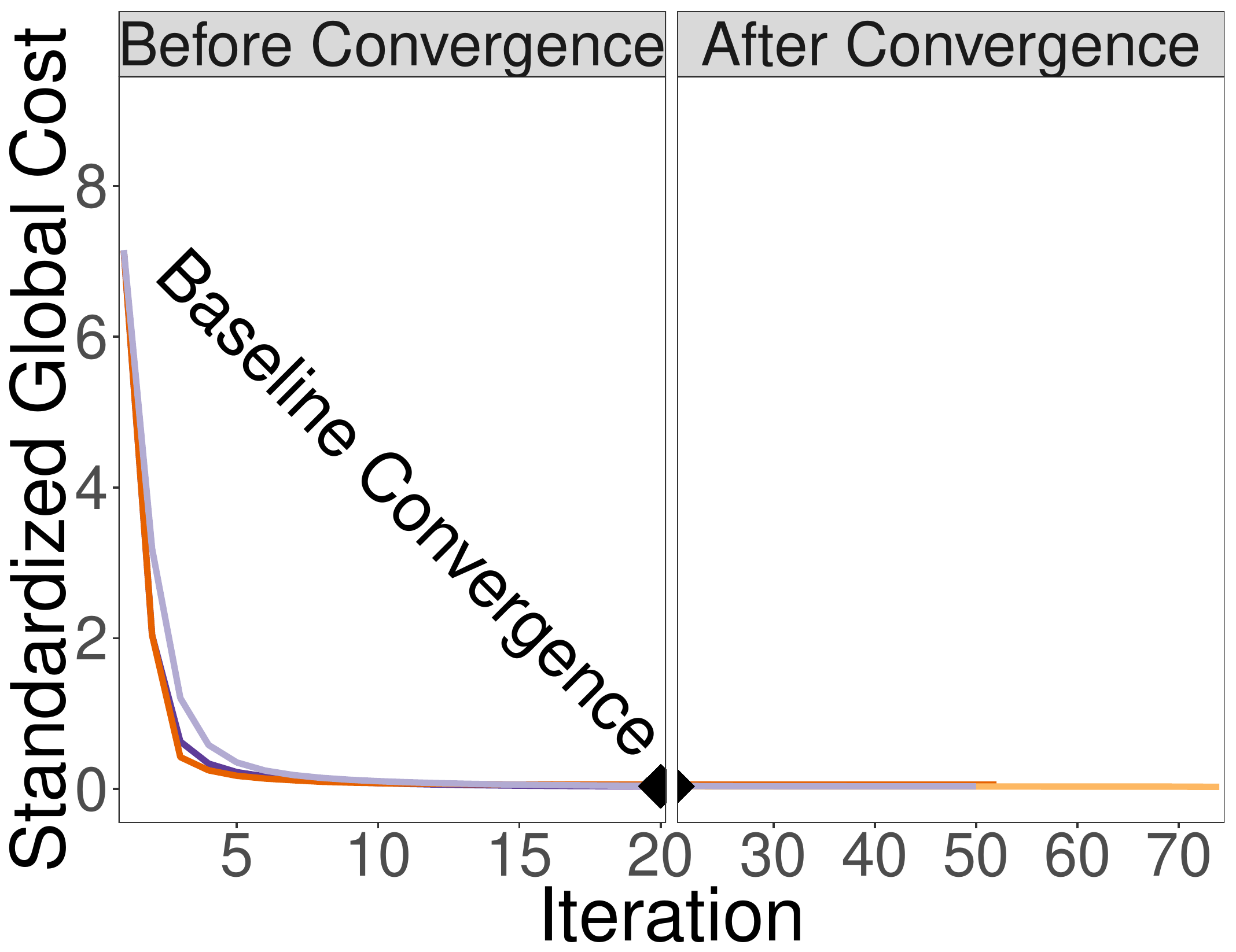}}
\subfigure[Energy demand]{\includegraphics[width=0.244\textwidth]{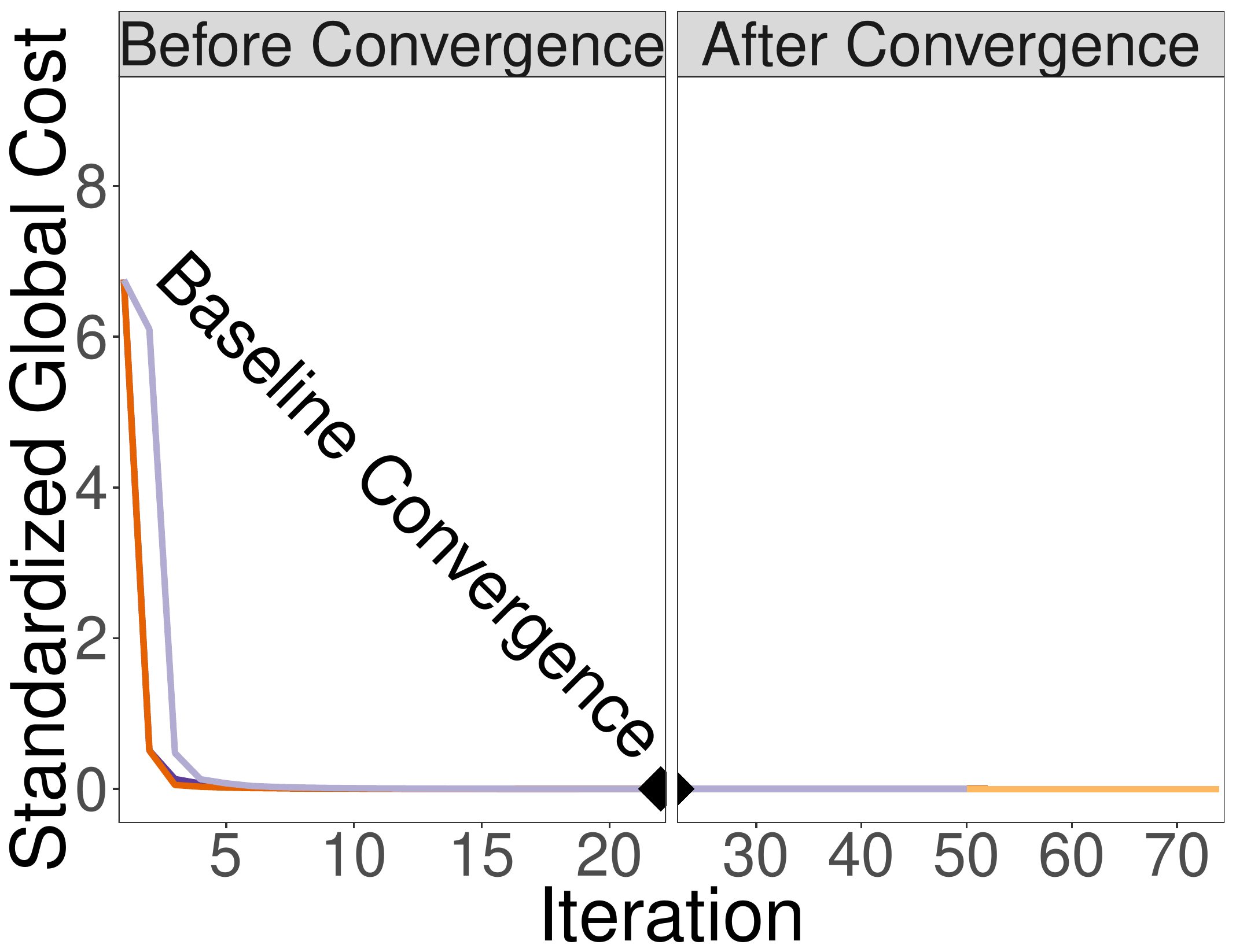}}
\subfigure[Electric vehicles]{\includegraphics[width=0.244\textwidth]{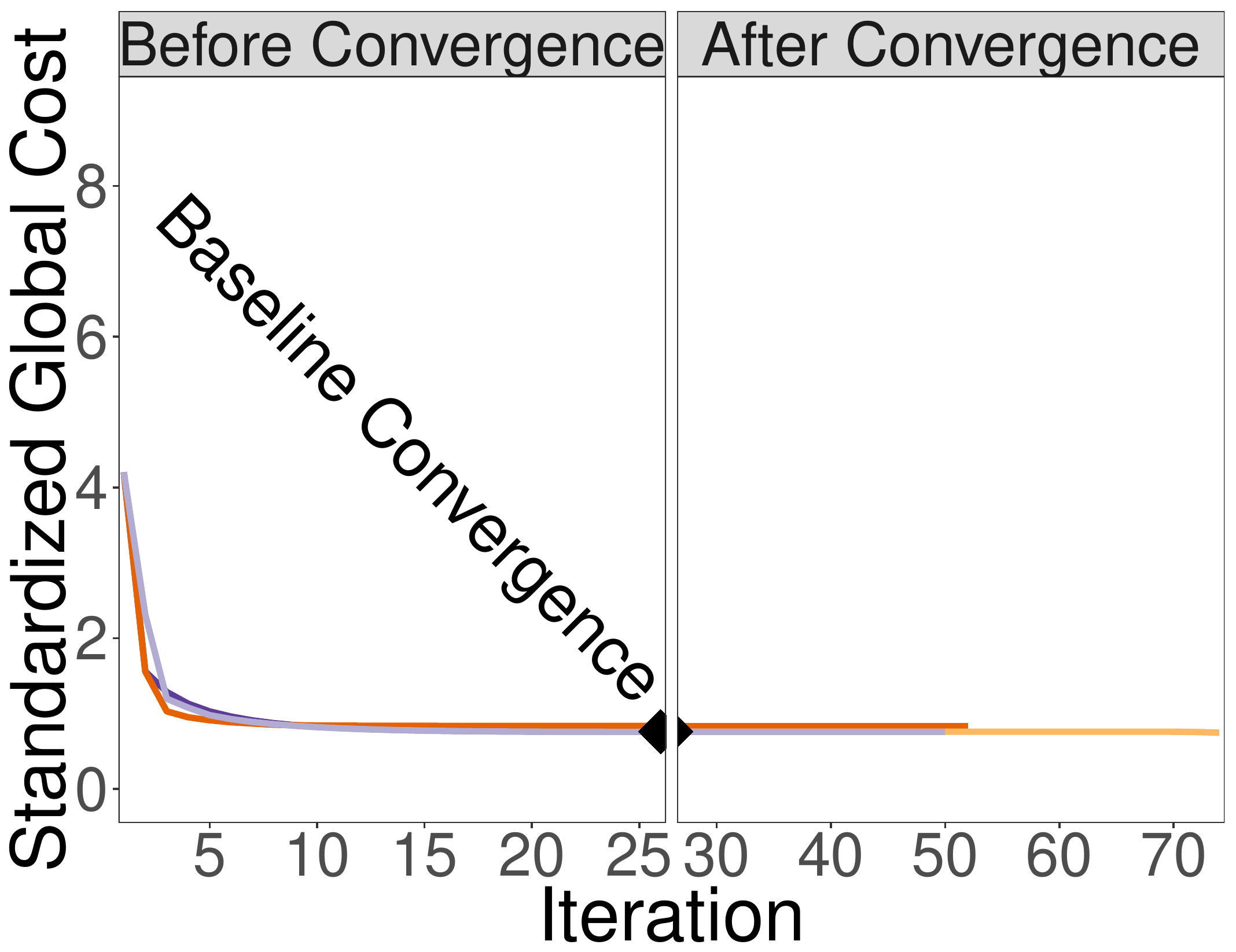}}
\caption{Learning curves. \emph{Dimensions}: learning schemes, application scenarios. \emph{Settings}: partial scale, $\lambda=0$, $c=2$.}\label{fig:learning-curves-partial-lambda-zero-c-two}
\end{figure}

Figure~\ref{fig:learning-curves-partial-lambda-zero-c-two} illustrates the learning curves for the four application scenarios and learning schemes. The partial scale with $\lambda=0$ and $c=2$ is illustrated. The learning curves for the full scale, $\lambda=0.5$ and $c=5$ are illustrated in Figure~\ref{fig:learning-curves-comparisons} of Appendix~\ref{sec:details}.   

The following observations can be made in Figure~\ref{fig:learning-curves-partial-lambda-zero-c-two}: Holarchic runtime achieves the fastest convergence speed given the several multi-level holarchic iterations performed within a main learning iteration. However, a performance sacrifice in global cost is observed, which is though low and observable for the scenario of electric vehicles in which the global cost is 10\% higher than the baseline. Moreover, within 7-8 iterations all holarchic schemes achieve the global cost reduction of the baseline. The convergence speed of the holarchic initialization is 1-2 iterations slower than the baseline, though this insignificant difference is not anymore observable in the scenario of electric vehicles.

\begin{figure}[!htb]
\centering
\includegraphics[width=1.0\textwidth]{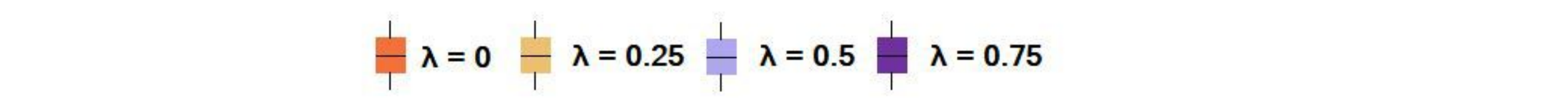}\\
\subfigure[Synthetic]{\includegraphics[width=0.244\textwidth]{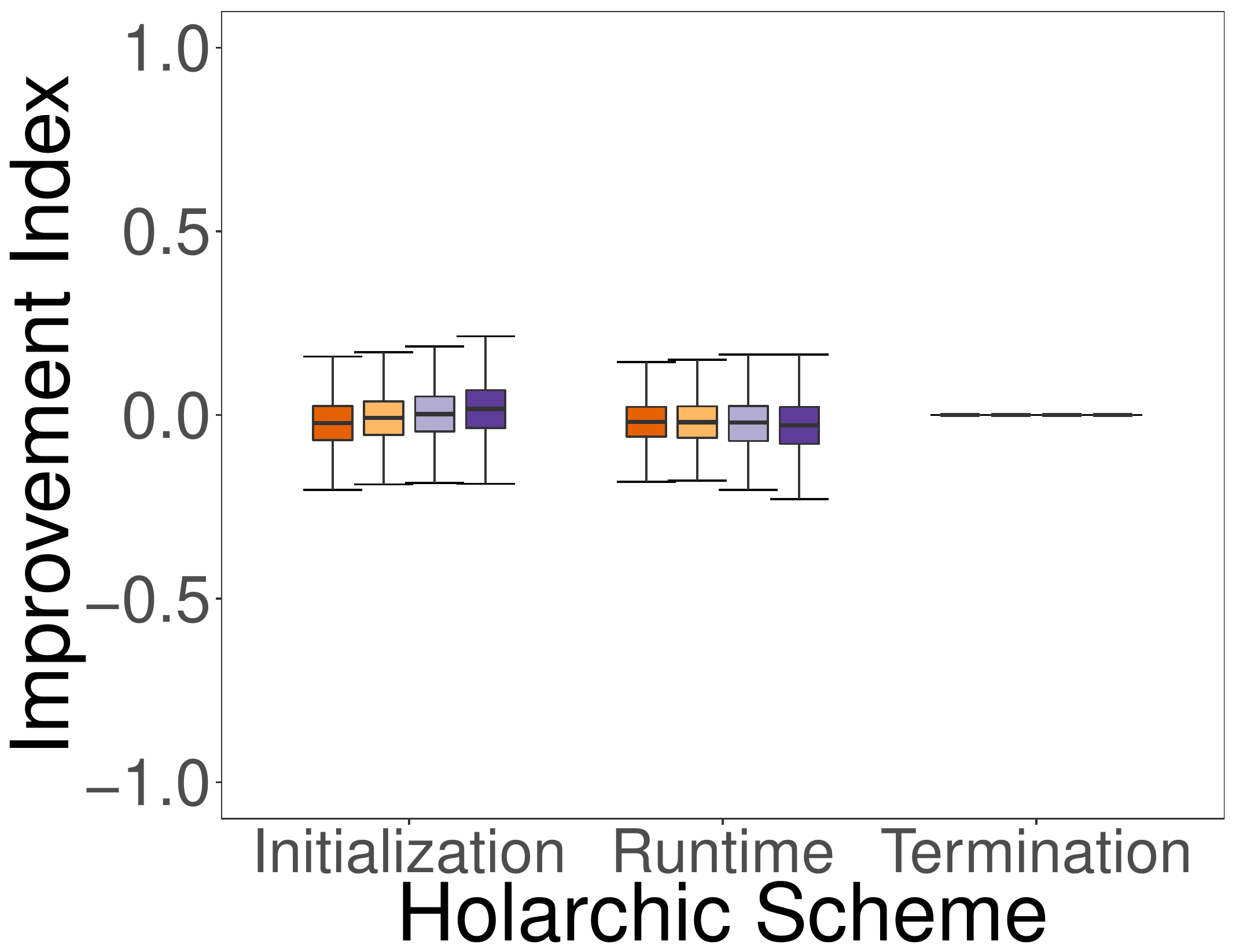}}
\subfigure[Bike sharing]{\includegraphics[width=0.244\textwidth]{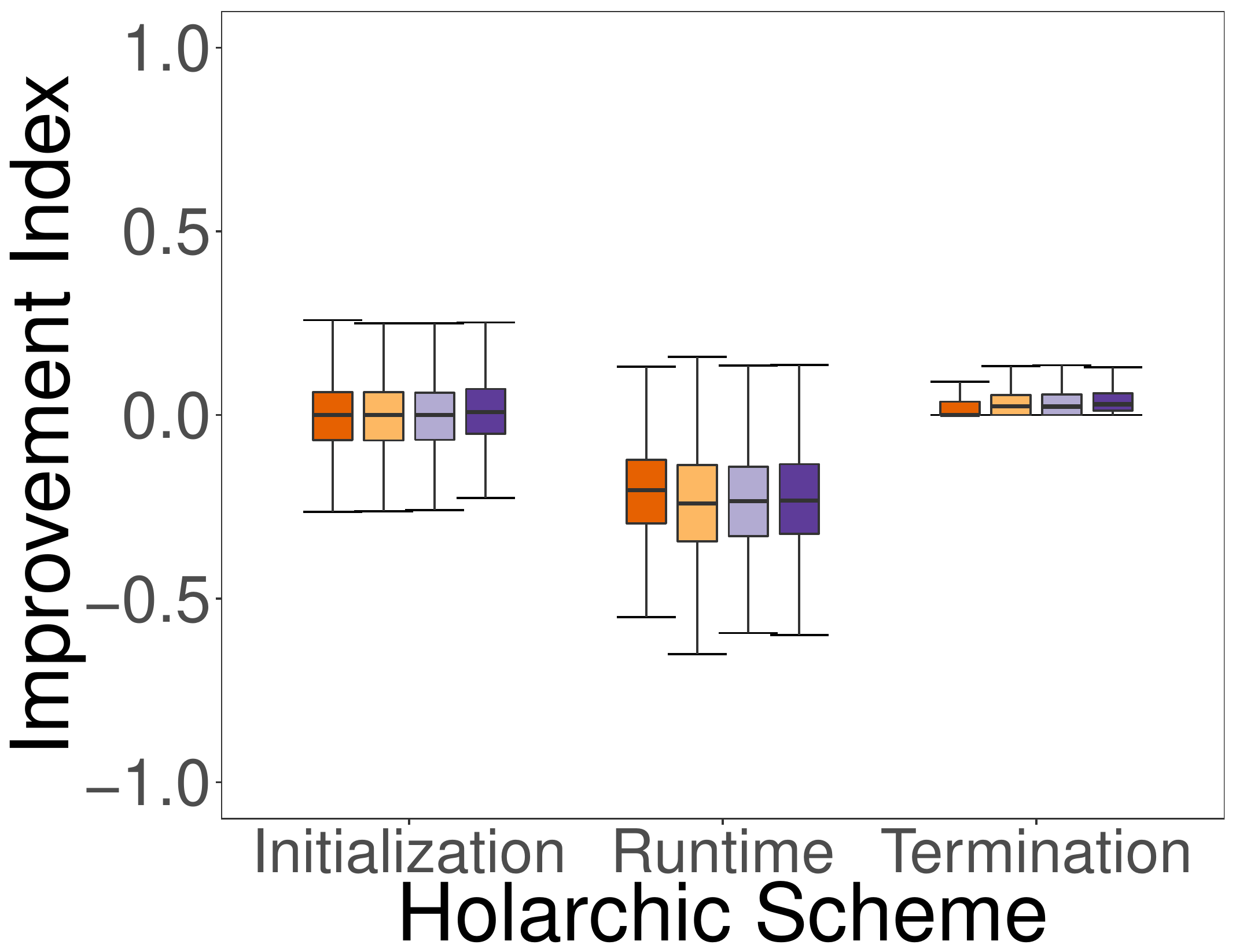}}
\subfigure[Energy demand]{\includegraphics[width=0.244\textwidth]{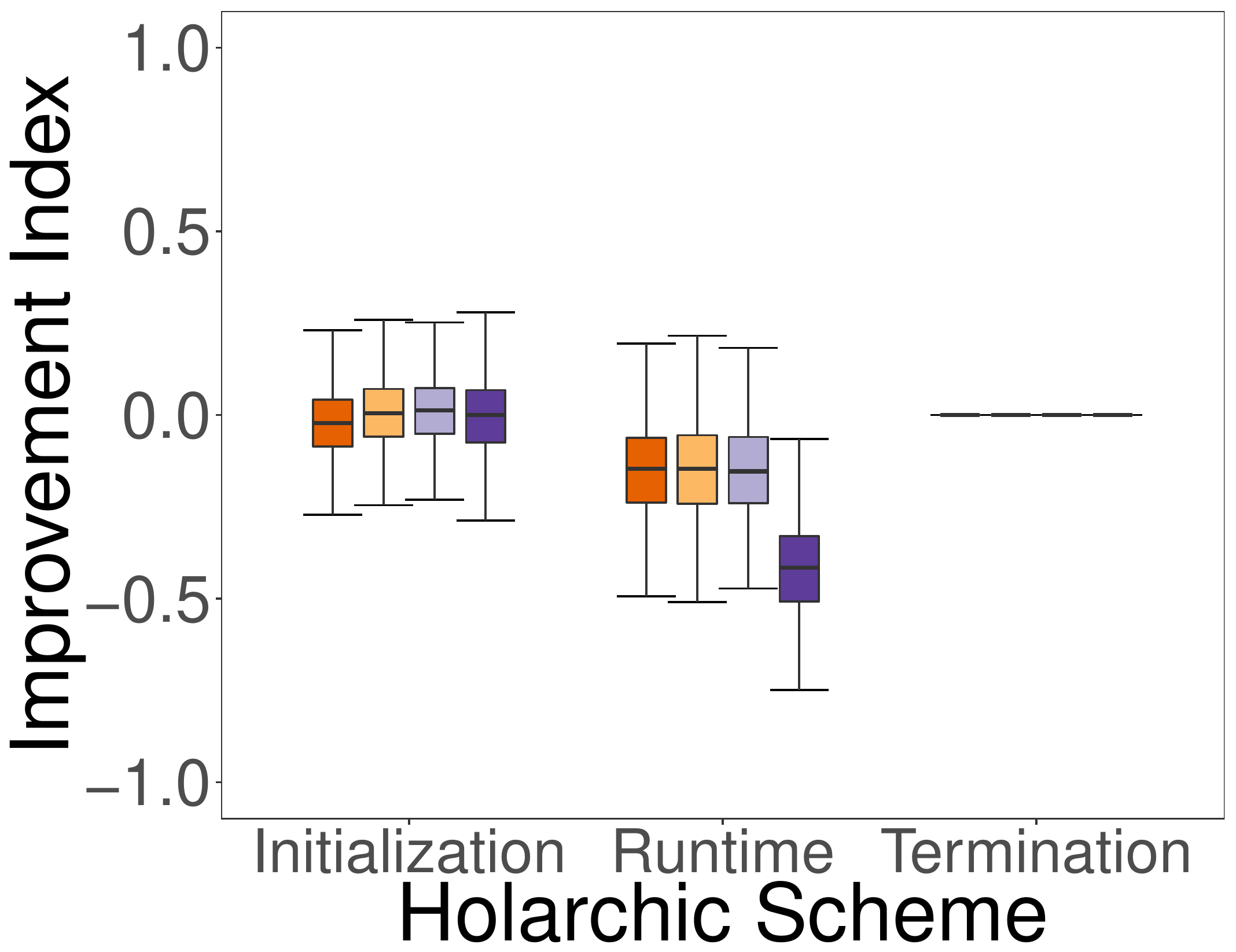}}
\subfigure[Electric vehicles]{\includegraphics[width=0.244\textwidth]{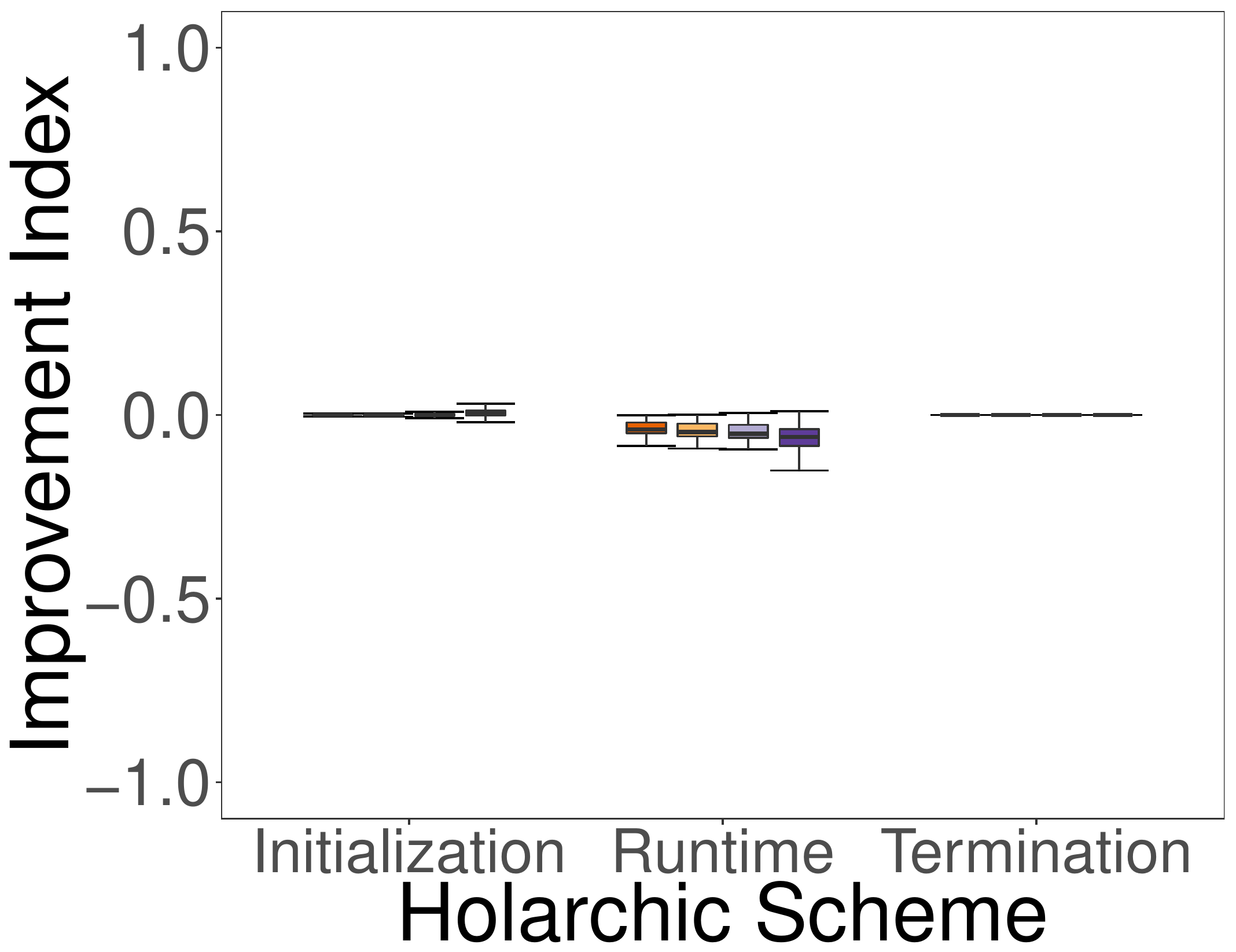}}
\caption{Improvement index. \emph{Dimensions}: holarchic schemes, application scenarios, different $\lambda$ values. \emph{Settings}: partial scale, $c=2$.}\label{fig:improvement-index-lambda-partial}
\end{figure}

Figure~\ref{fig:improvement-index-lambda-partial} illustrates the improvement index of the holarchic schemes for different application scenarios and different $\lambda$ values. The exploration potential of the holarchic initialization is indicated by the error bars above the mean. Moreover, holarchic initialization does not influence significantly the global cost reduction, however, for higher $\lambda$ values, i.e. $\lambda=0.75$, an average increase of 0.5\% is observed in the improvement index. This means that in more constrained optimization settings, the exploration strategy of the holarchic initialization contributes a low performance improvement. In contrast, the mitigation strategy of the holarchic runtime manages to preserve the baseline performance with an improvement index of values close to 0. An average performance boosting of 1.65\% is observed via the holarchic termination in the bike sharing scenario that has sparse data, while in the other scenarios no significant improvement is observed.

\begin{figure}[!htb]
\centering
\includegraphics[width=1.0\textwidth]{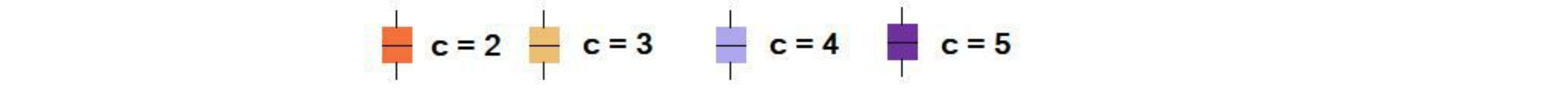}\\
\subfigure[Synthetic]{\includegraphics[width=0.244\textwidth]{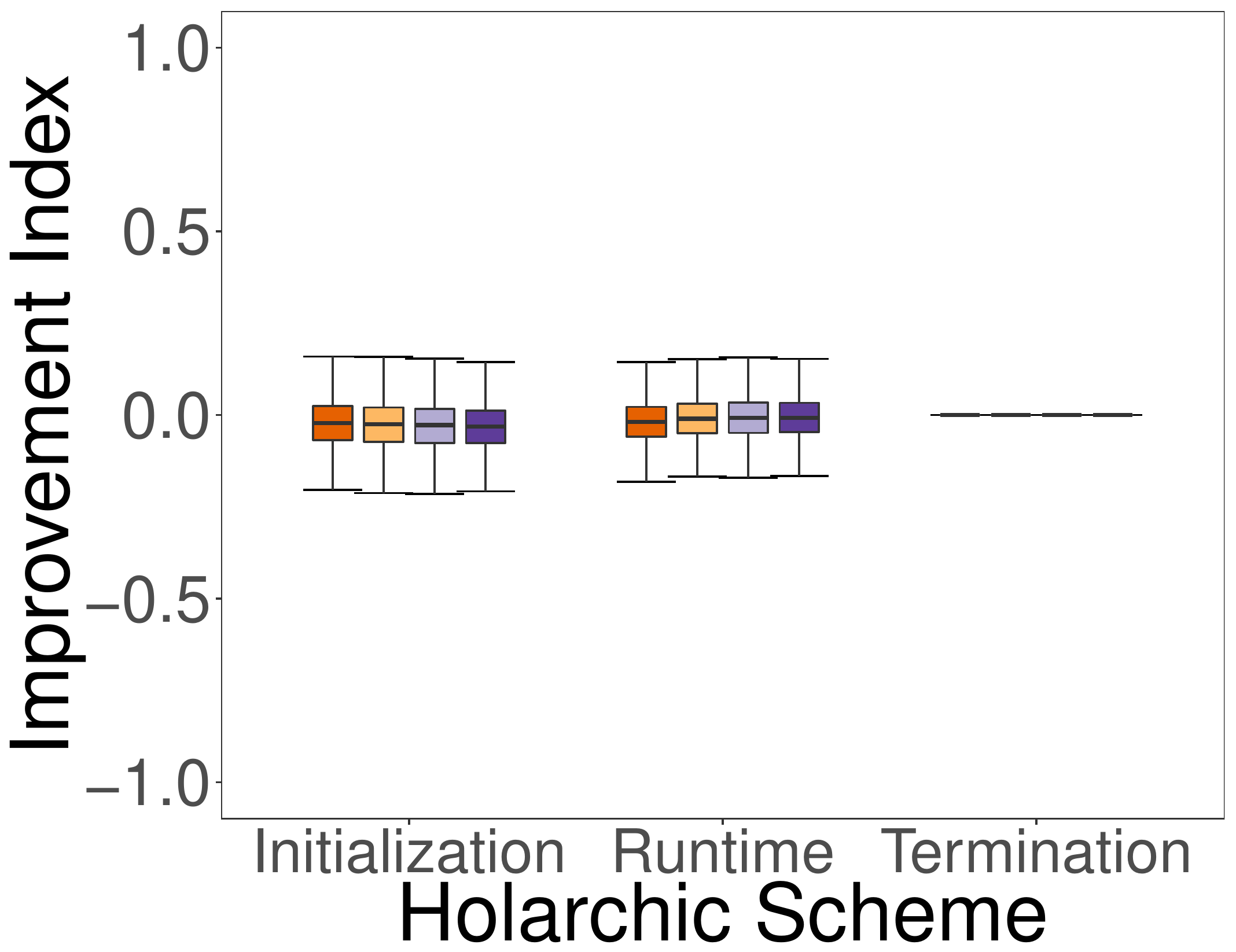}}
\subfigure[Bike sharing]{\includegraphics[width=0.244\textwidth]{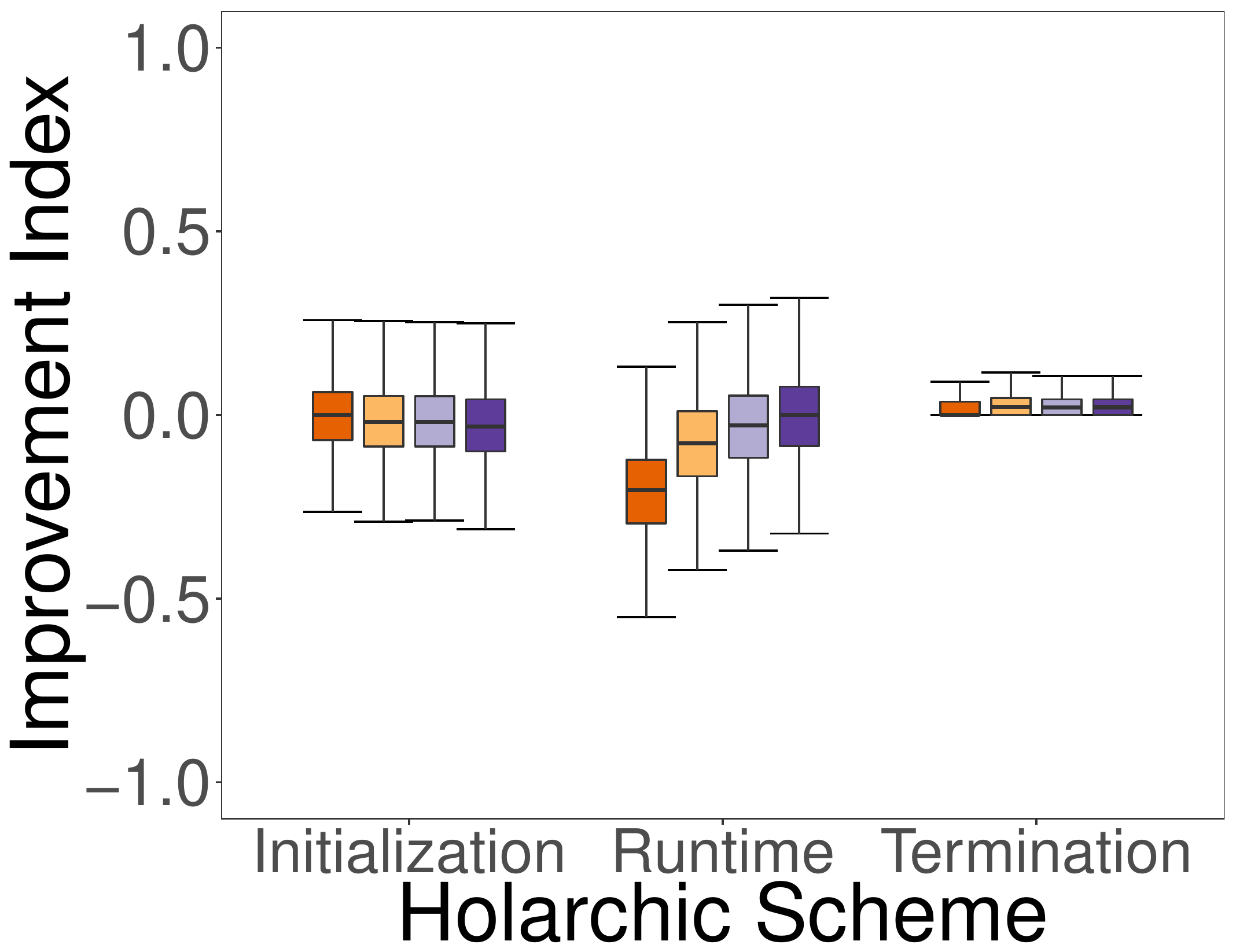}}
\subfigure[Energy demand]{\includegraphics[width=0.244\textwidth]{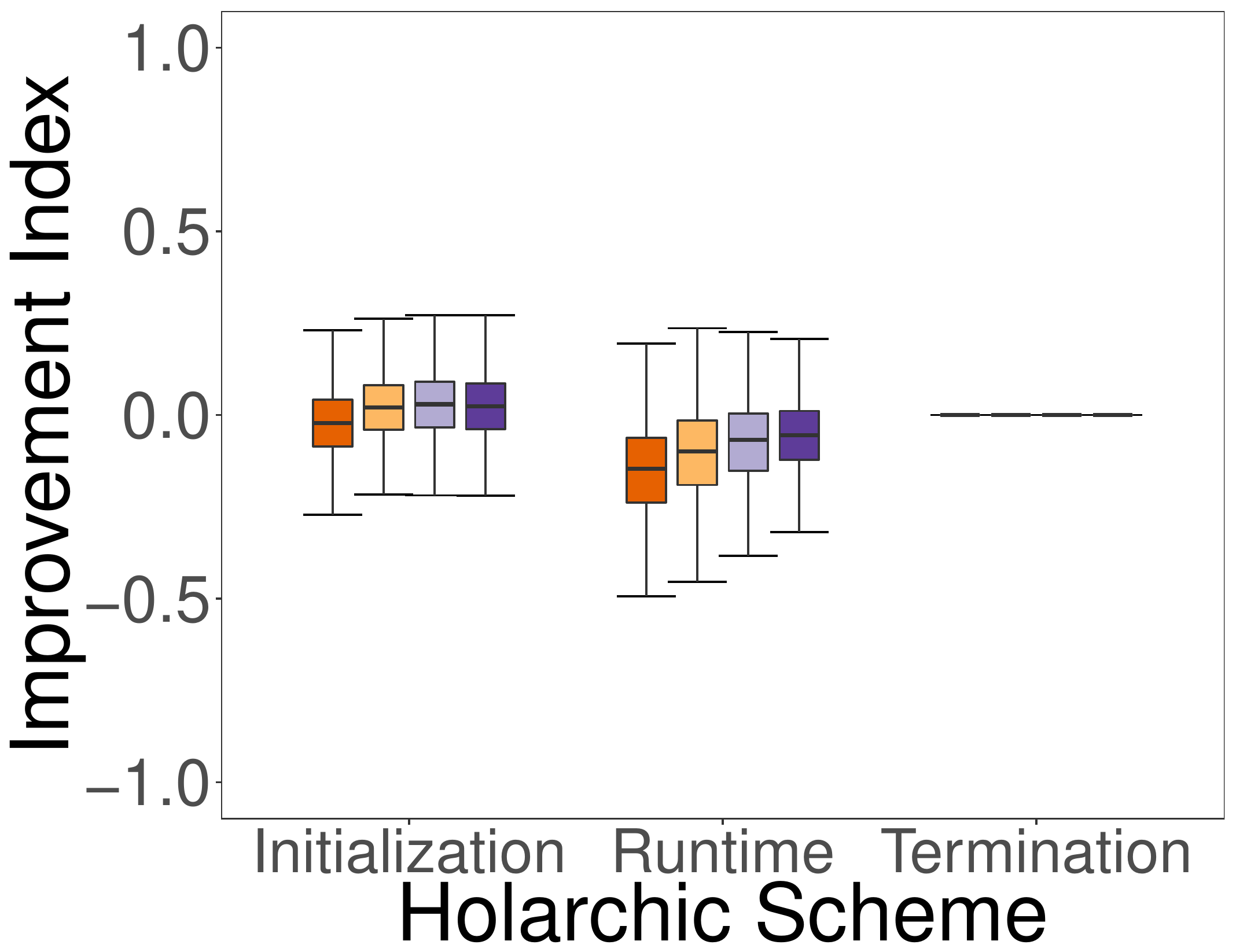}}
\subfigure[Electric vehicles]{\includegraphics[width=0.244\textwidth]{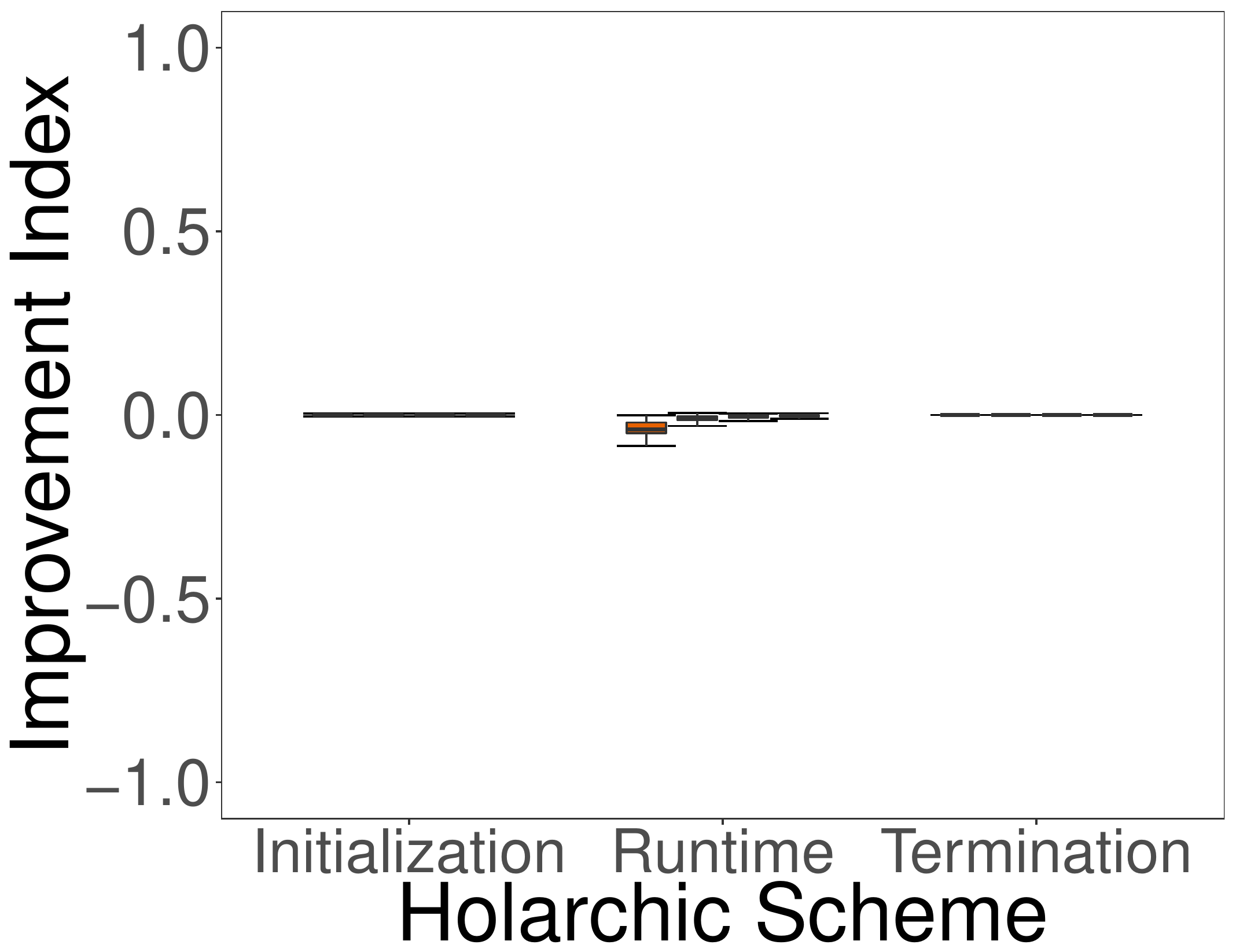}}
\caption{Improvement index. \emph{Dimensions}: holarchic schemes, application scenarios, varying number of children. \emph{Settings}: partial scale, $\lambda=0$.}\label{fig:improvement-index-children-partial}
\end{figure}

Figure~\ref{fig:improvement-index-children-partial} illustrates the improvement index of the holarchic schemes for different application scenarios and different $c$ values. Holarchic initialization retains an average improvement index of -0.008 while it can scale up the improvement index to values of 0.225 on average. The number of children does not influence the performance of this holarchic scheme. In contrast, the holarchic runtime shows an average increase of 4.1\% in the improvement index by increasing $c$ from 2 to 5, while this holarchic scheme serves well its performance mitigation role: an average improvement index of -0.041. Finally, holarchic termination boosts performance by 1.1\% in the bike sharing scenario. 

\begin{figure}[!htb]
\centering
\includegraphics[width=1.0\textwidth]{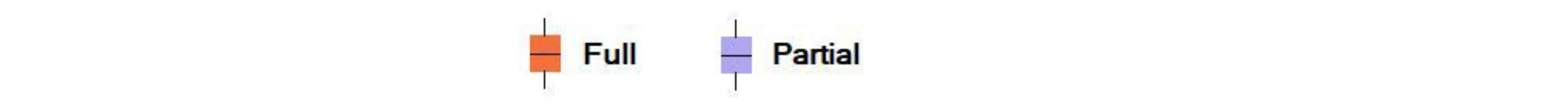}\\
\subfigure[Synthetic]{\includegraphics[width=0.244\textwidth]{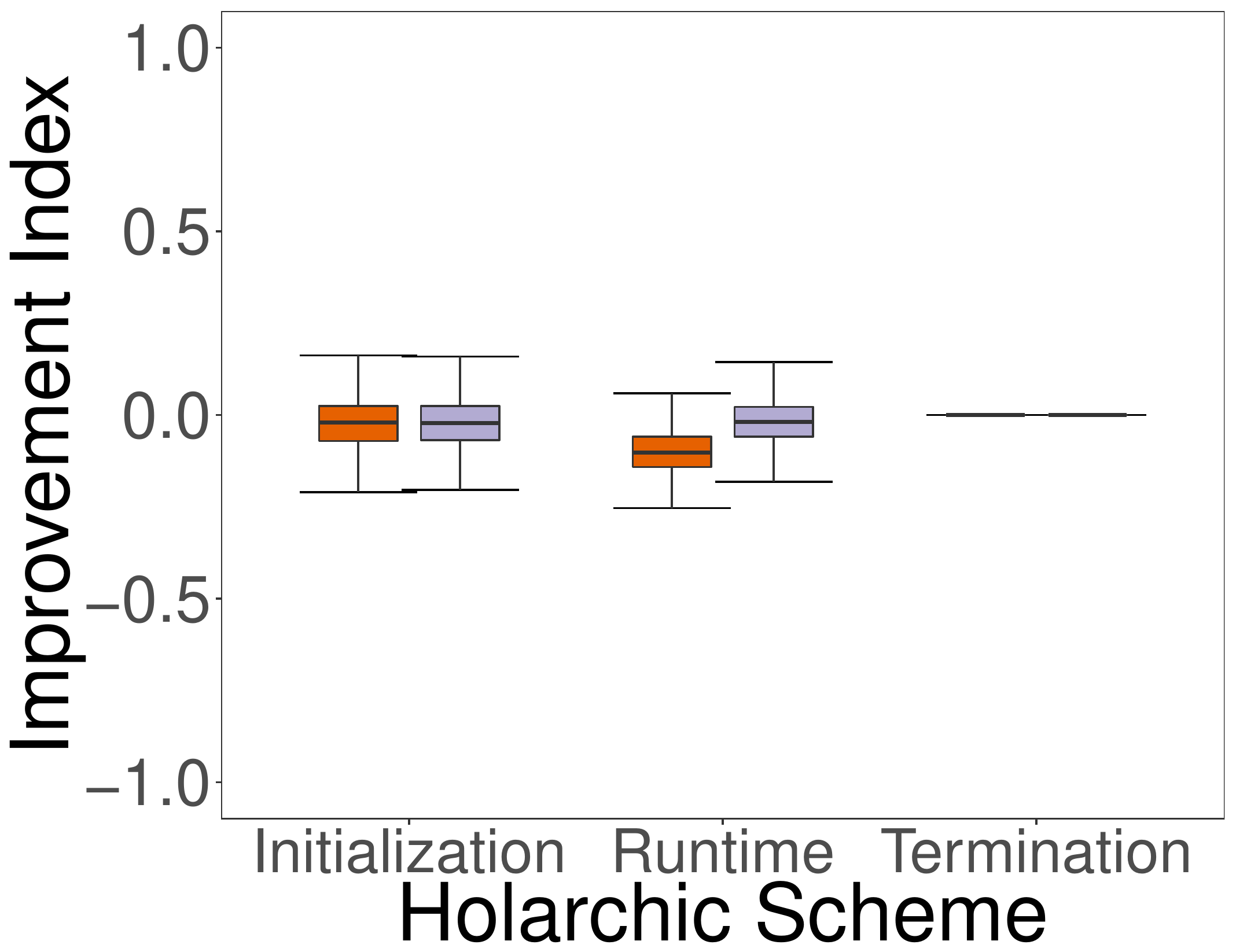}}
\subfigure[Bike sharing]{\includegraphics[width=0.244\textwidth]{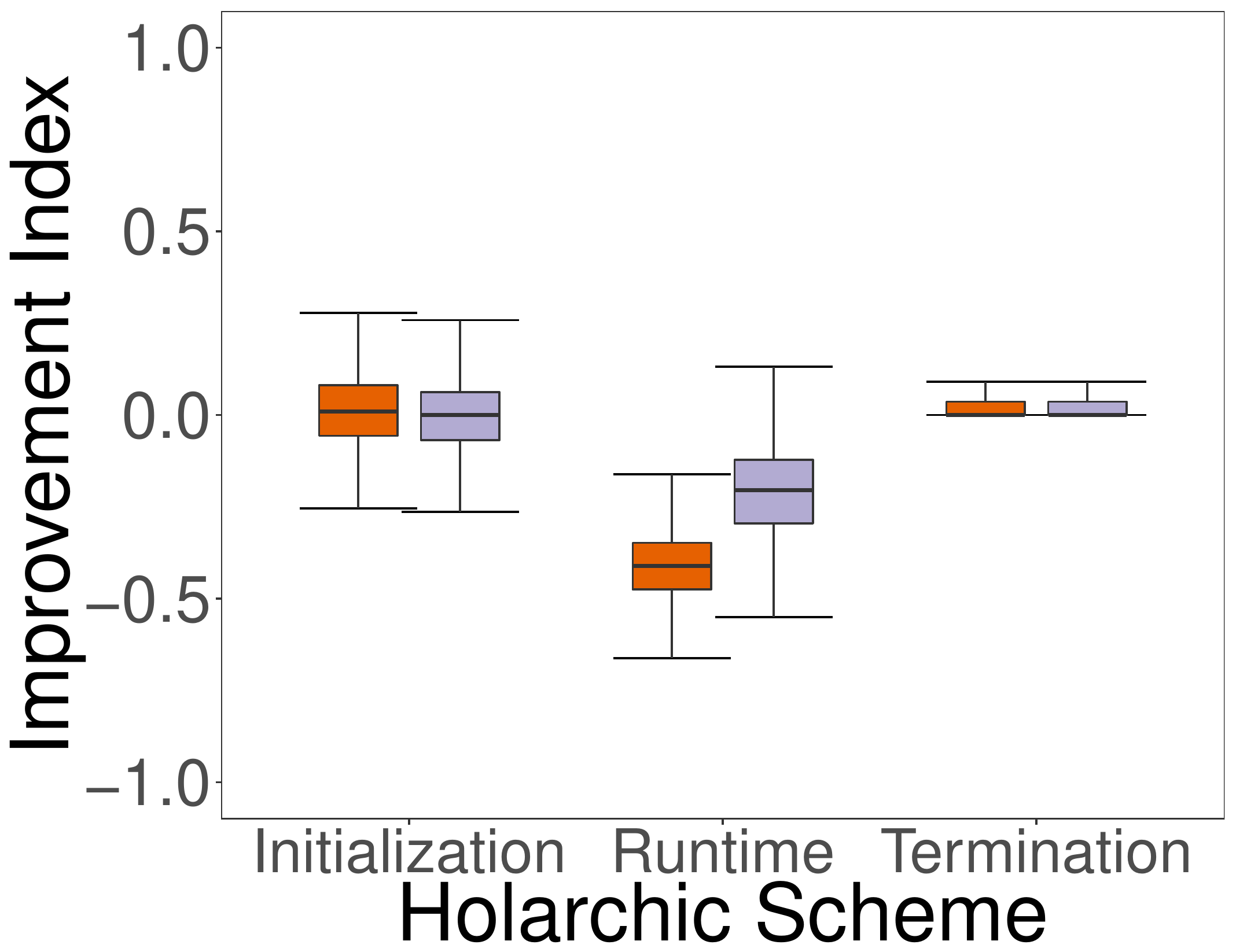}}
\subfigure[Energy demand]{\includegraphics[width=0.244\textwidth]{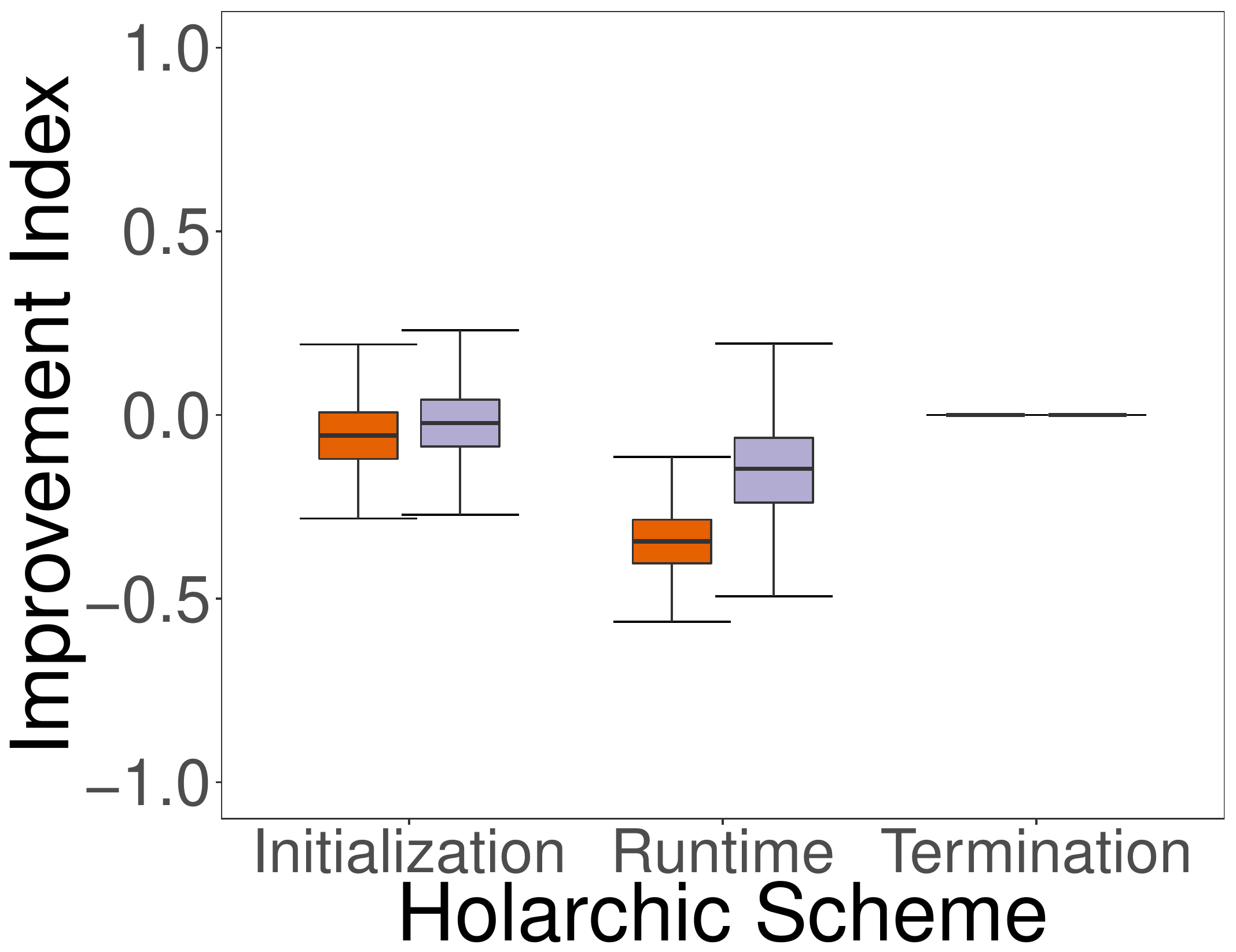}}
\subfigure[Electric vehicles]{\includegraphics[width=0.244\textwidth]{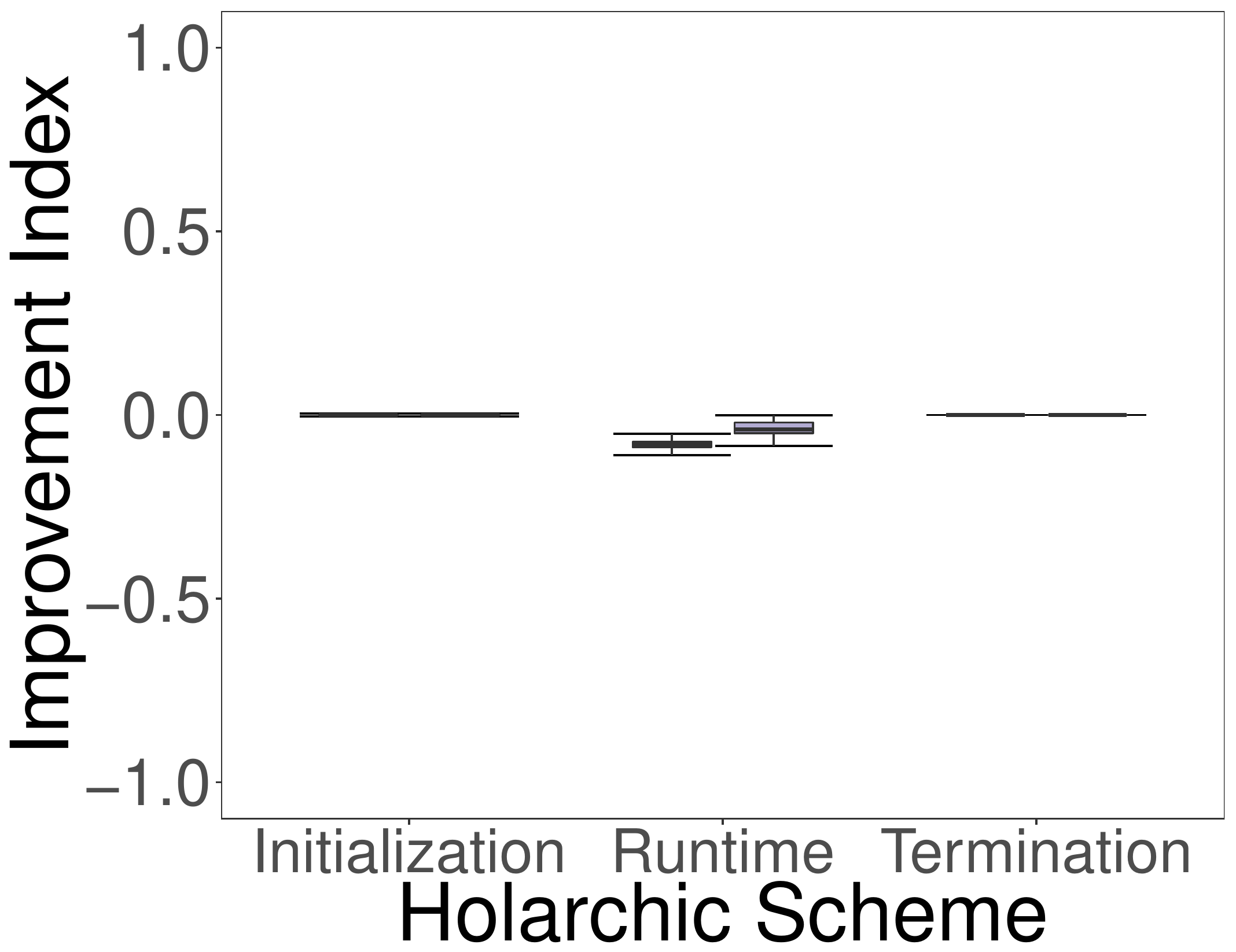}}
\caption{Improvement index. \emph{Dimensions}: holarchic schemes, application scenarios, partial versus full scale. \emph{Settings}: $\lambda=0$, $c=2$.}\label{fig:improvement-index-scale}
\end{figure}

Figure~\ref{fig:improvement-index-scale} demonstrates the higher performance that the partial scale shows compared to full scale: 0.25\% higher improvement index for holarchic initialization and 6.5\% for holarchic runtime.

Figure~\ref{fig:density-improvement-index-lambda-partial},~\ref{fig:density-improvement-index-children-partial} and~\ref{fig:density-improvement-index-scale} of Appendix~\ref{sec:details} show the probability density of the improvement index for different $\lambda$, $c$ values and holarchic scales respectively. In these figures one can study in more detail the density of the improvement index values behind the error bars of the respective Figure~\ref{fig:improvement-index-lambda-partial},~\ref{fig:improvement-index-children-partial} and~\ref{fig:improvement-index-scale}.

The rest of this section studies the trade-offs and the cost-effectiveness of the holarchic runtime designed for the mitigation of the learning performance.

\subsection{Trade-offs and cost-effectiveness}\label{subsec:cost-effectiveness}

\begin{figure}[!htb]
\centering
\subfigure[Total versus synchronized communication for different $c$.]{\includegraphics[width=0.244\textwidth]{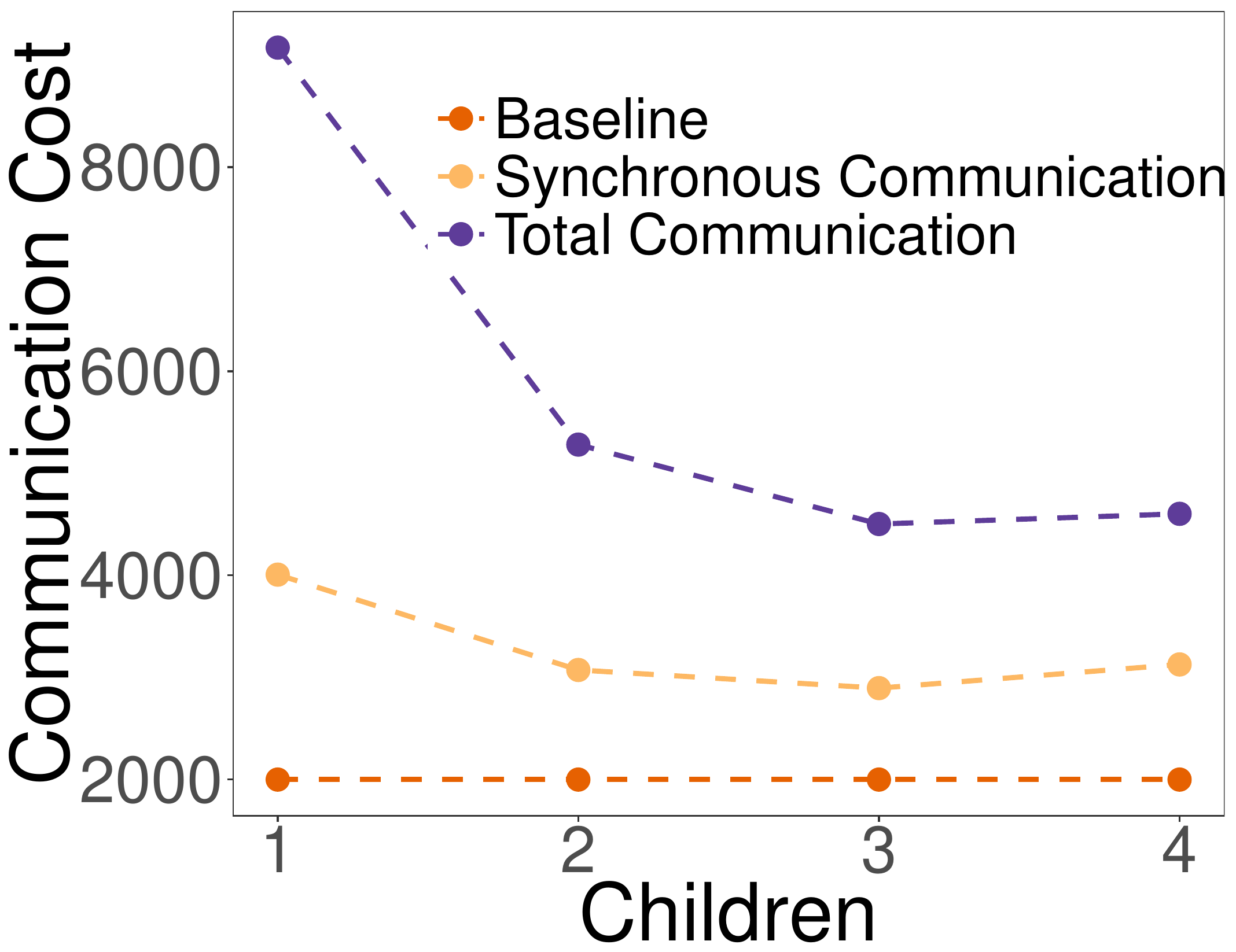}}
\subfigure[Total communication]{\includegraphics[width=0.244\textwidth]{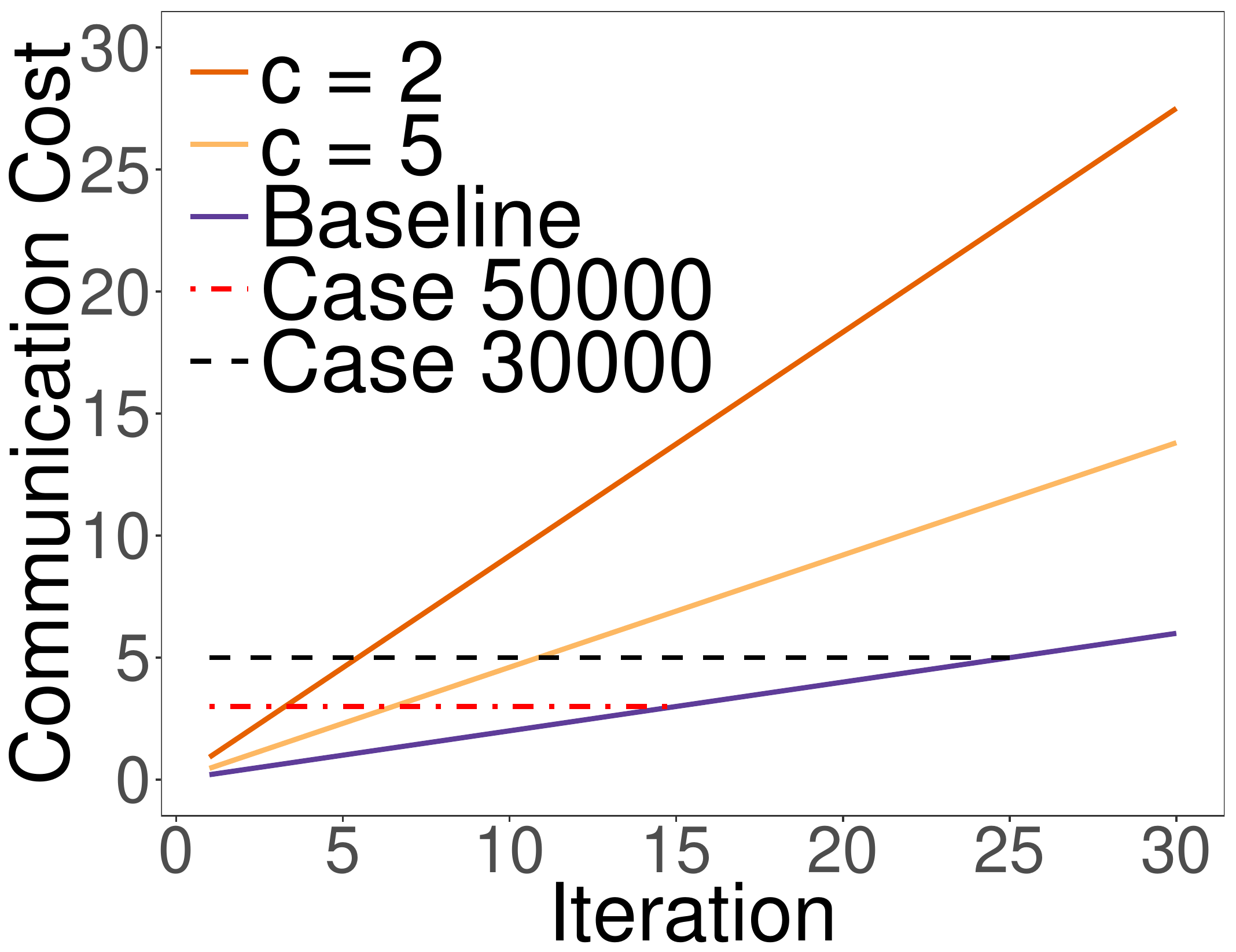}}
\subfigure[Synchronized communication]{\includegraphics[width=0.244\textwidth]{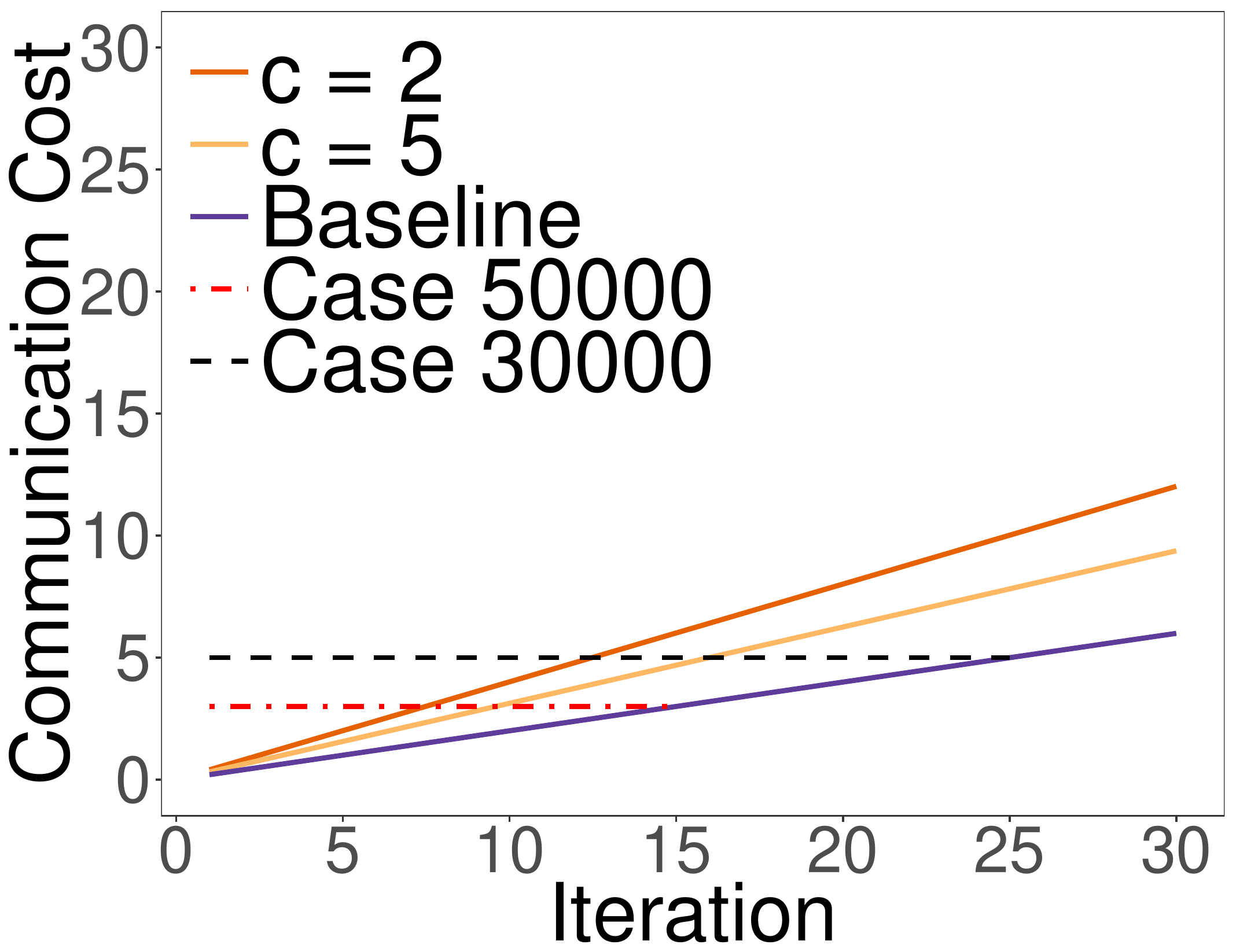}}
\caption{Communication cost. \emph{Settings}: partial scale, $\lambda=0$ (a) \emph{Dimensions}: varying number of children, total versus synchronized communication cost, baseline versus holarchic runtime. (b) and (c) Sampling case 30000 and case 50000 under total versus synchronized communication cost. \emph{Dimensions}: $c=2$ versus $c=5$, number of iterations, baseline versus holarchic runtime.}\label{fig:communication-cost-partial-lambda-zero}
\end{figure}

Figure~\ref{fig:communication-cost-partial-lambda-zero}a illustrates the communication cost per iteration of the baseline versus the total and synchronized communication cost of the holarchic runtime. This is a worse case scenario as applying the holarchy to a smaller branch or for a fewer than 5 holarchic iterations can make the communication cost equivalent\footnote{The communication cost of the holarchic runtime can even become lower than baseline assuming a holarchy at partial scale with the agents that do not belong to the holarchy being disconnected.} to the one of the baseline. In Figure~\ref{fig:communication-cost-partial-lambda-zero}a, the communication cost of the holarchic runtime decreases as the number of children increases given that the recursion of the holarchy is limited to a lower number of levels in the tree, i.e. fewer holons are formed. The synchronized communication cost is on average 45\% lower than the total communication cost. 

\begin{figure}[!htb]
\centering
\includegraphics[width=1.0\textwidth]{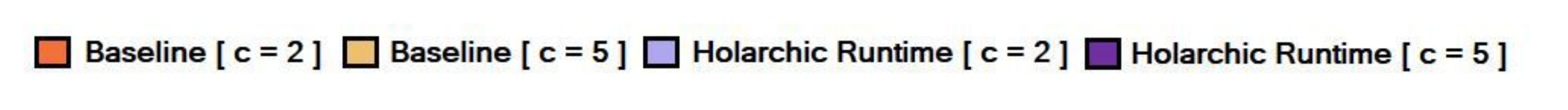}\\
\subfigure[Synthetic, total communication]{\includegraphics[width=0.244\textwidth]{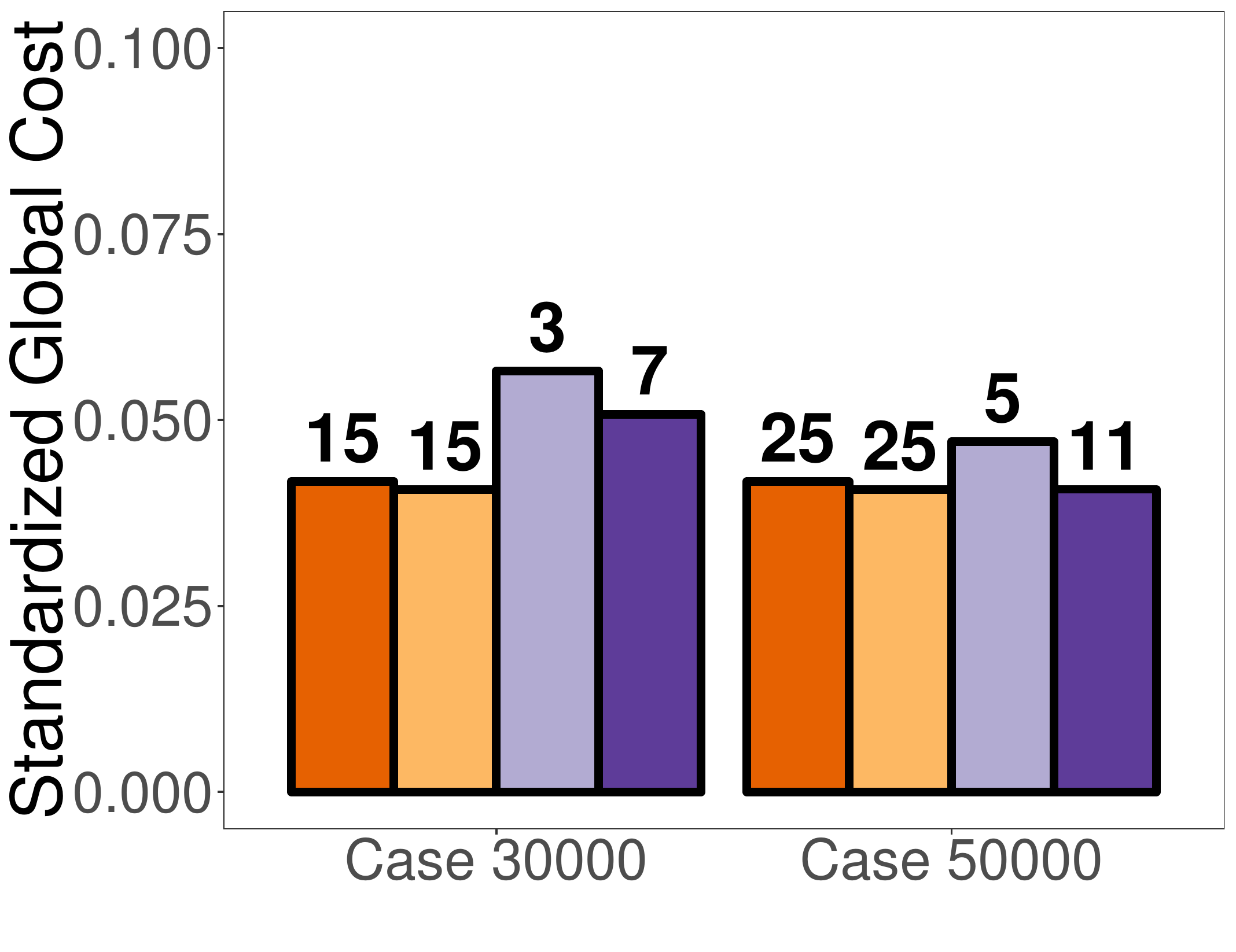}}
\subfigure[Bike sharing, total communication]{\includegraphics[width=0.244\textwidth]{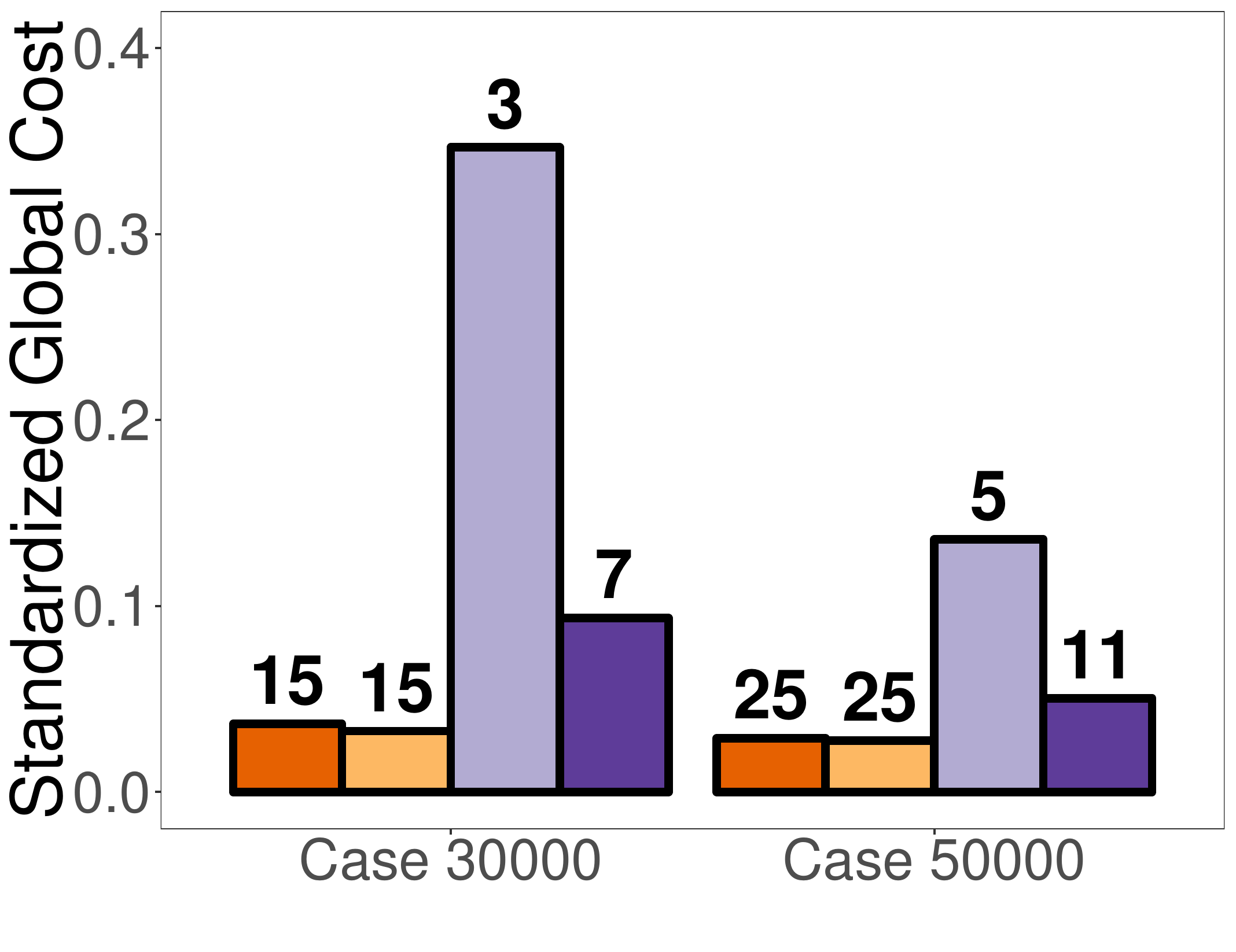}}
\subfigure[Energy demand, total communication]{\includegraphics[width=0.244\textwidth]{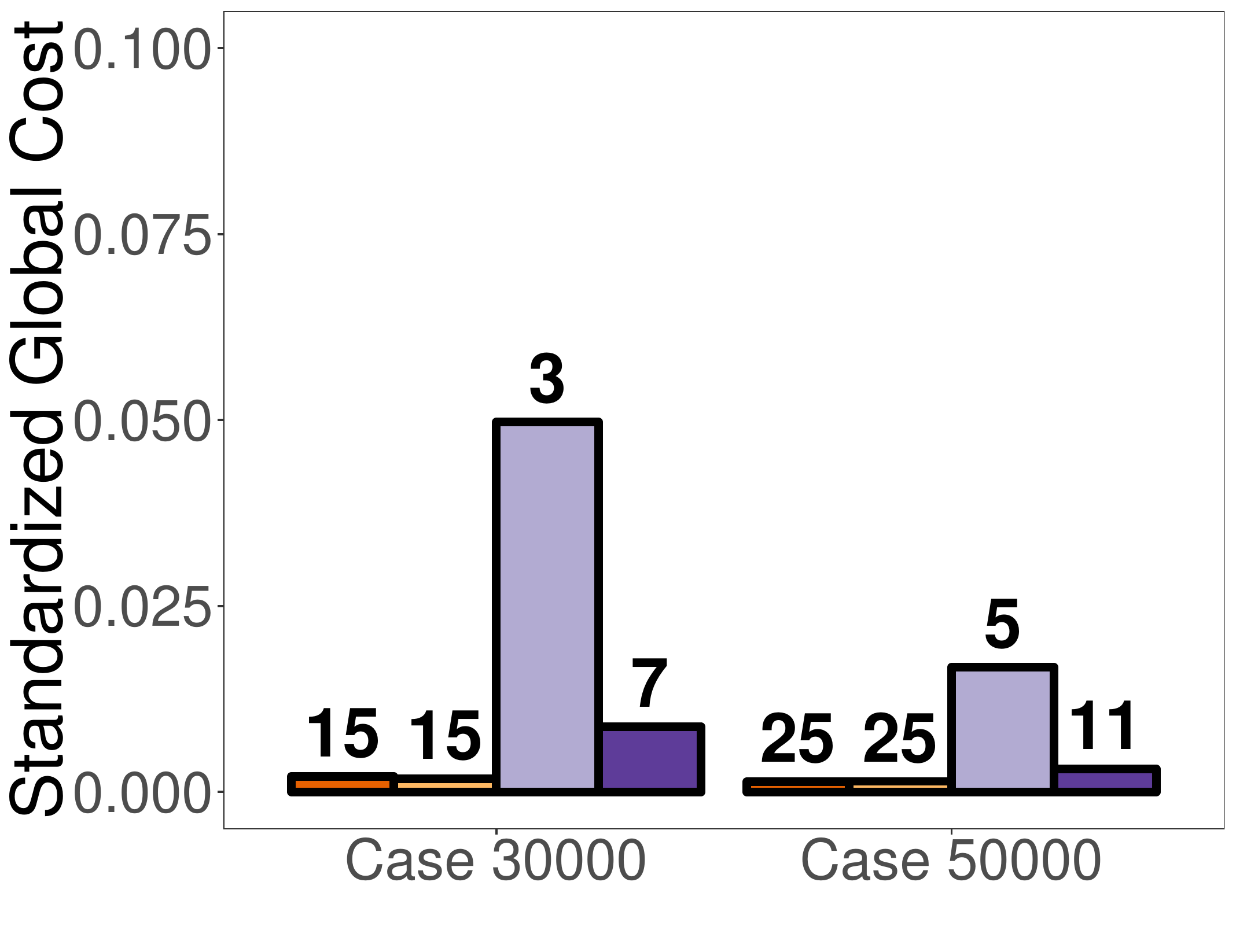}}
\subfigure[Electric vehicles total communication]{\includegraphics[width=0.244\textwidth]{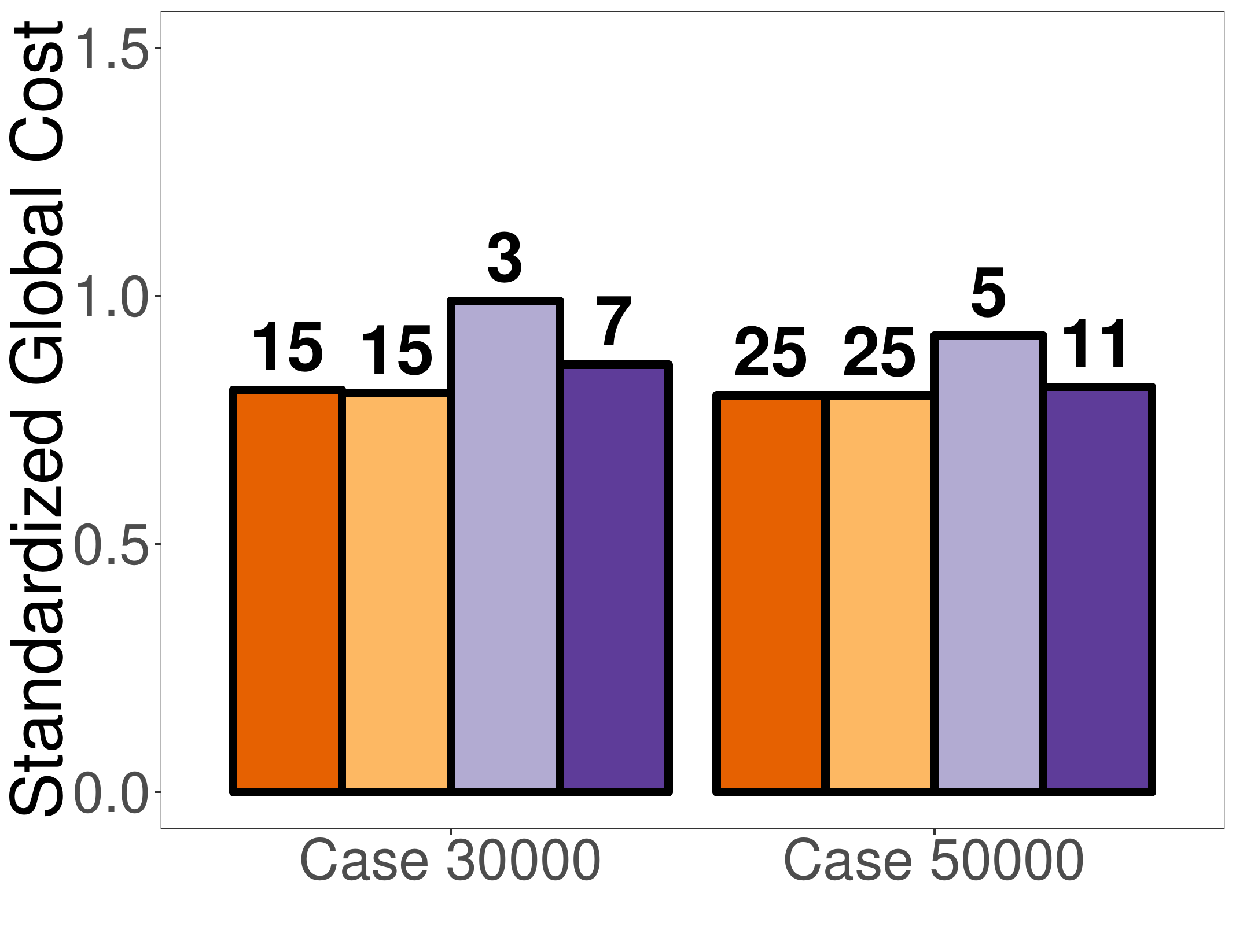}}
\subfigure[Synthetic, synchronized communication]{\includegraphics[width=0.244\textwidth]{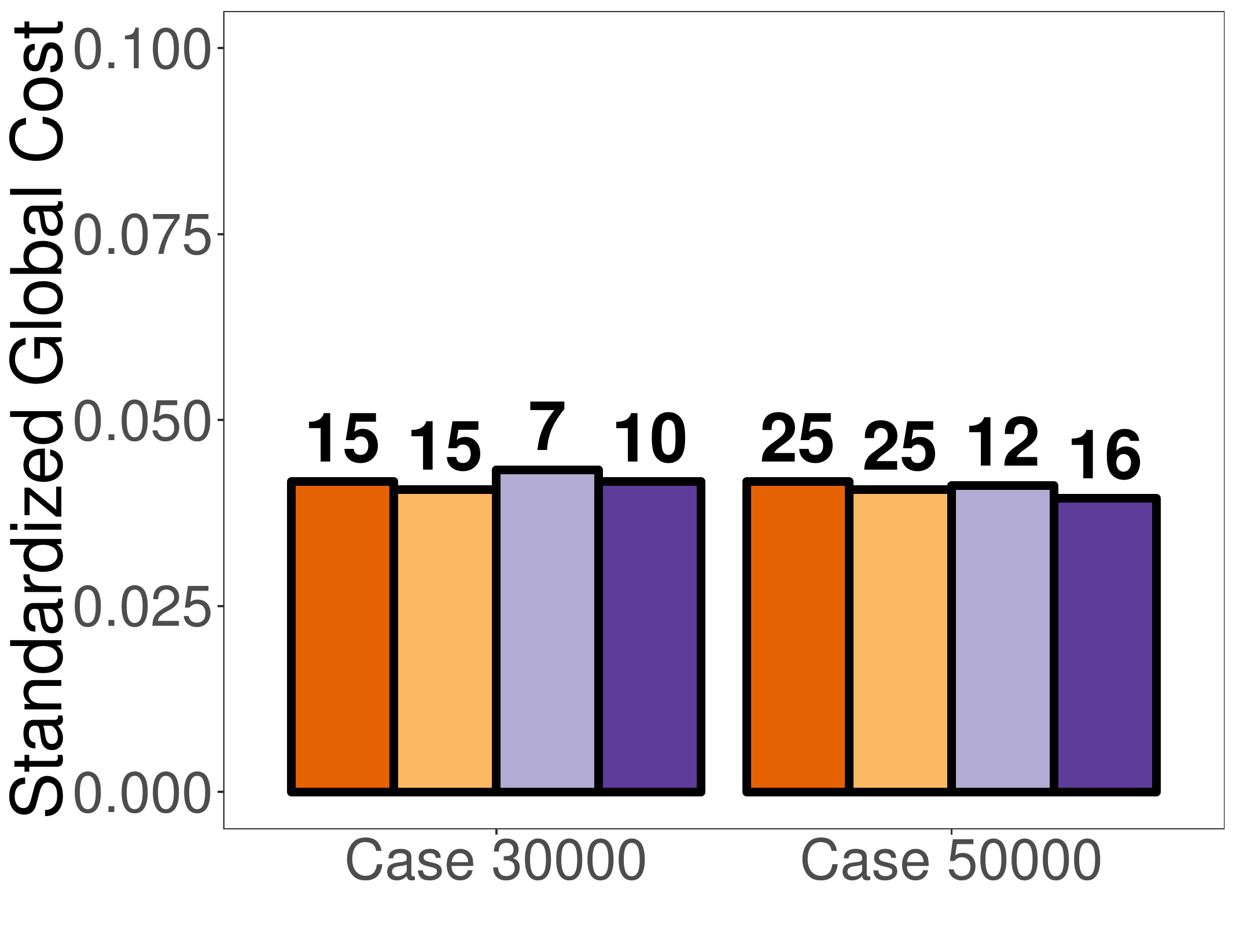}}
\subfigure[Bike sharing, synchronized communication]{\includegraphics[width=0.244\textwidth]{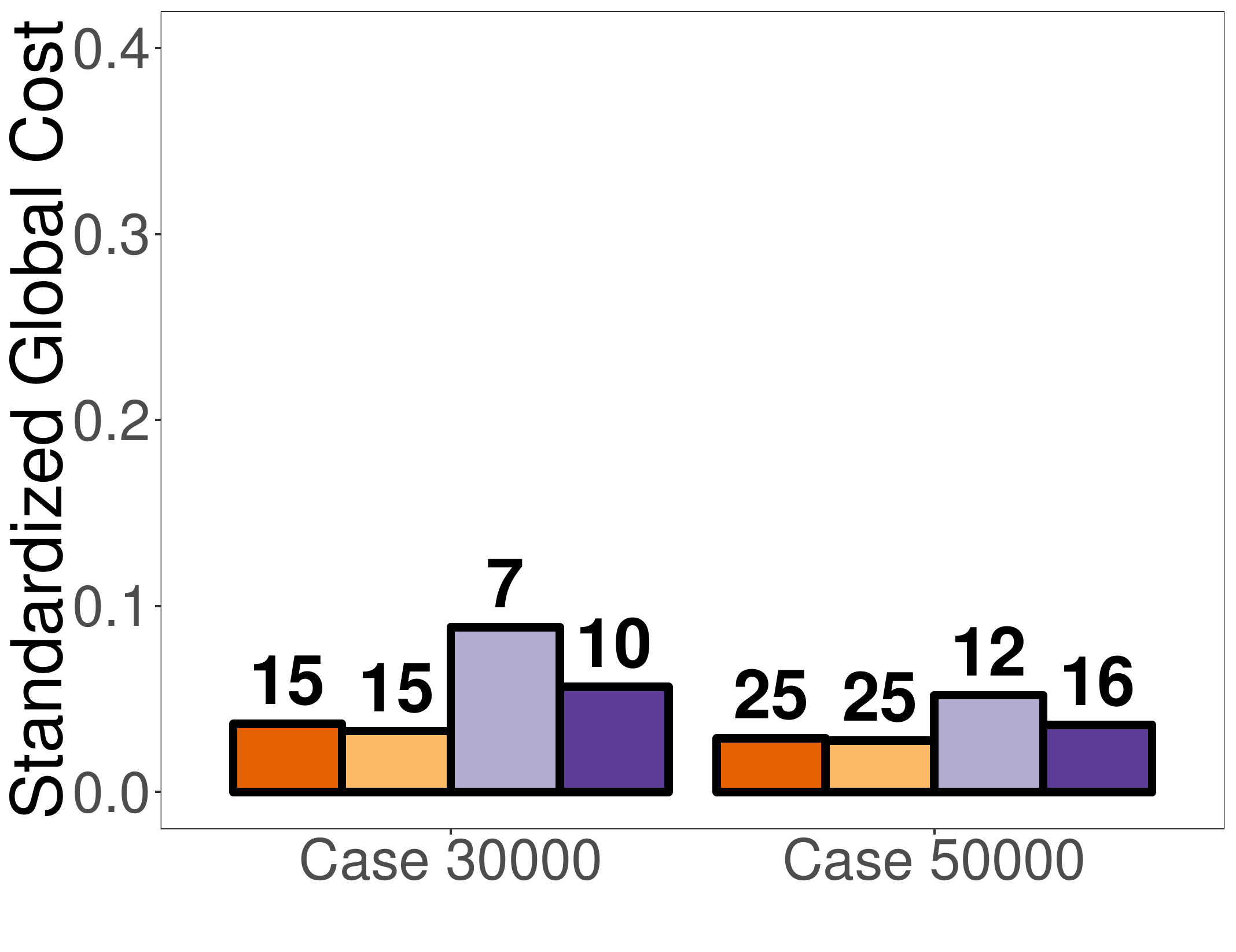}}
\subfigure[Energy demand, synchronized communication]{\includegraphics[width=0.244\textwidth]{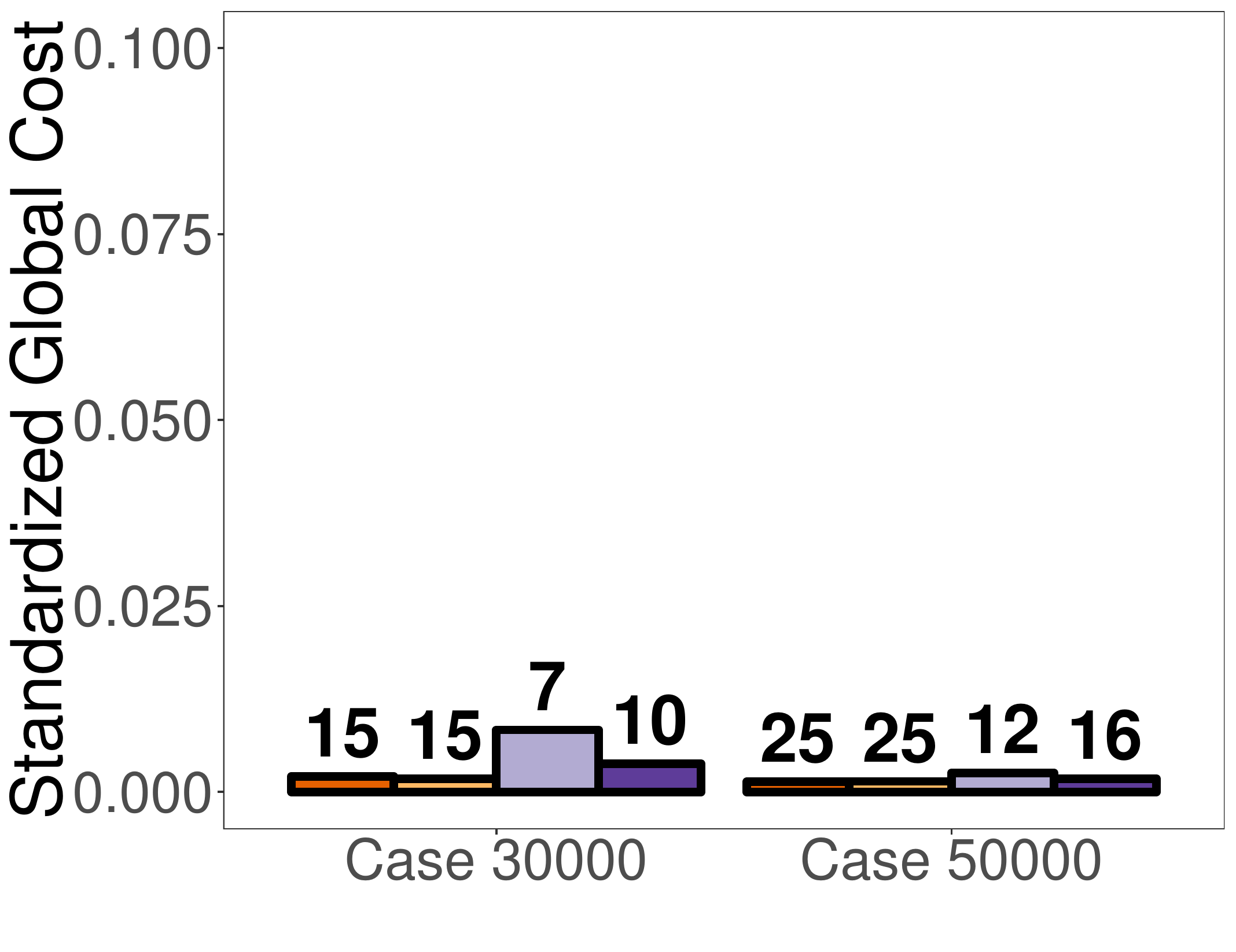}}
\subfigure[Electric vehicles, synchronized communication]{\includegraphics[width=0.244\textwidth]{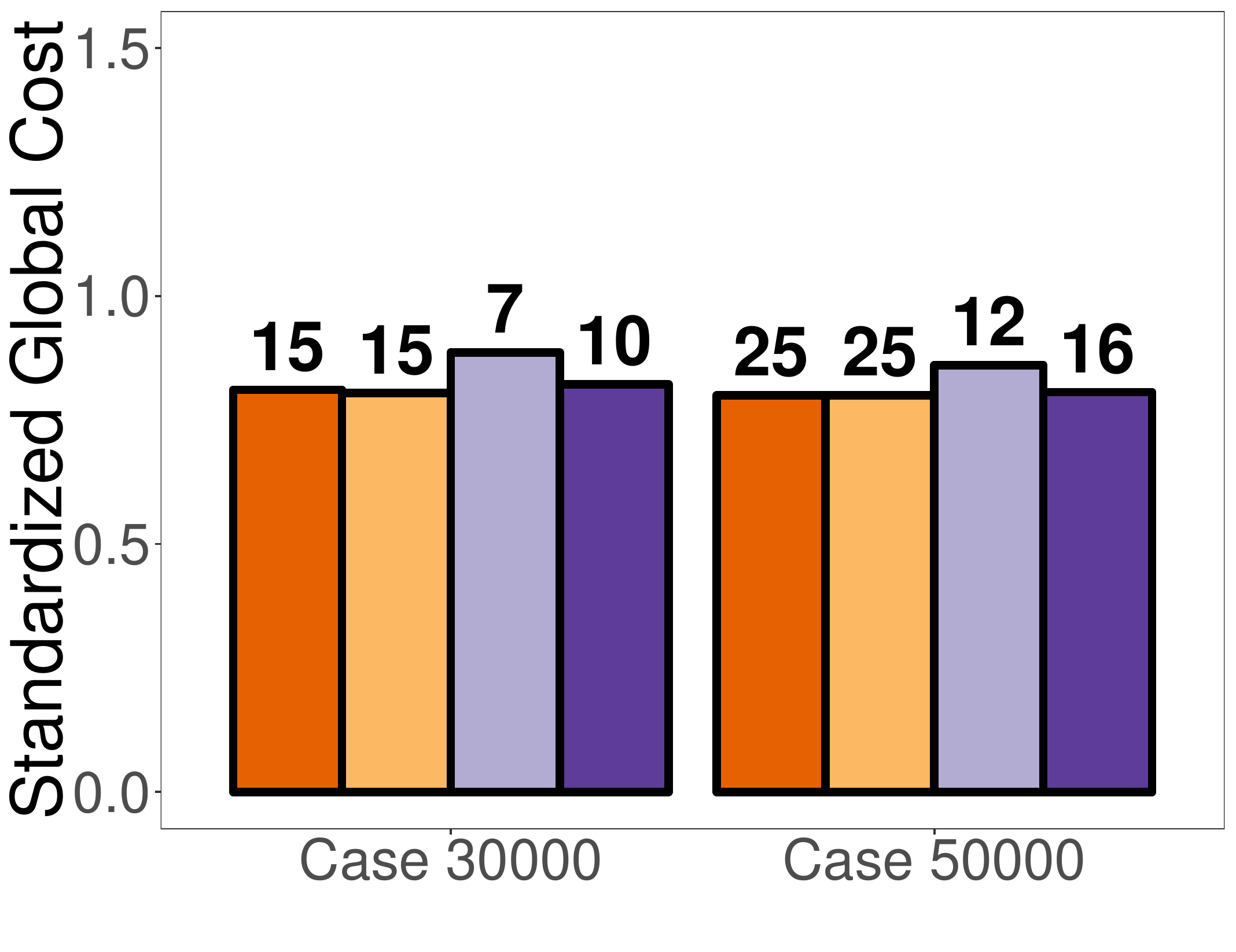}}
\caption{Global cost. \emph{Dimensions}: baseline versus holarchic runtime, total versus synchronized communication cost, application scenarios, $c=2$ versus $c=5$, number of iterations given a communication cost, case 30000 versus case 50000. \emph{Settings}: partial scale, $\lambda=0$.}
\label{fig:global cost-cases-partial}
\end{figure}

The cost-effectiveness of the holarchic runtime is studied by fixing the communication cost for both baseline and holarching runtime and looking into the global cost reduction achieved for the same number of messages exchanged. Figure~\ref{fig:communication-cost-partial-lambda-zero}b and~\ref{fig:communication-cost-partial-lambda-zero}c illustrate this process for total and synchronous communication cost. Two cases are determined: (i) \emph{case 30000} and (ii) \emph{case 50000}. Each case runs for a given number of iterations that is determined by the intersection with the horizontal dashed lines for each of the $c=2$ and $c=5$. Then, the global costs can be compared for the same number of exchanged messages as shown in Figure~\ref{fig:global cost-cases-partial}. 

The following observations can be made in Figure~\ref{fig:global cost-cases-partial}a to~\ref{fig:global cost-cases-partial}d for the total communication cost: When $c=5$, the holarchic runtime achieves a highly equivalent performance with the baseline. This also holds for $c=2$ in the scenarios of synthetic and electric vehicles. The performance mitigation becomes 13.64\% more significant when the synchronized communication cost is counted in Figure~\ref{fig:global cost-cases-partial}e to~\ref{fig:global cost-cases-partial}h. This is also shown by the shifted probability densities of the improvement index in Figure~\ref{fig:density-improvement-index-communication-children-partial} of Appendix~\ref{sec:details}. 

\begin{figure}[!htb]
\centering
\includegraphics[width=1.0\textwidth]{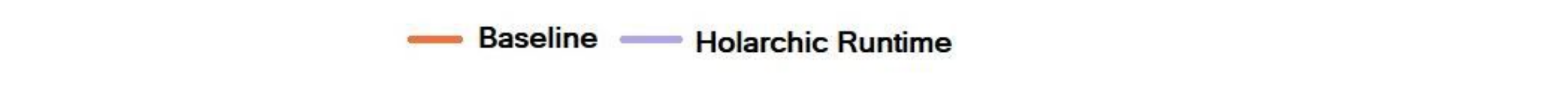}\\
\subfigure[Synthetic, total communication]{\includegraphics[width=0.244\textwidth]{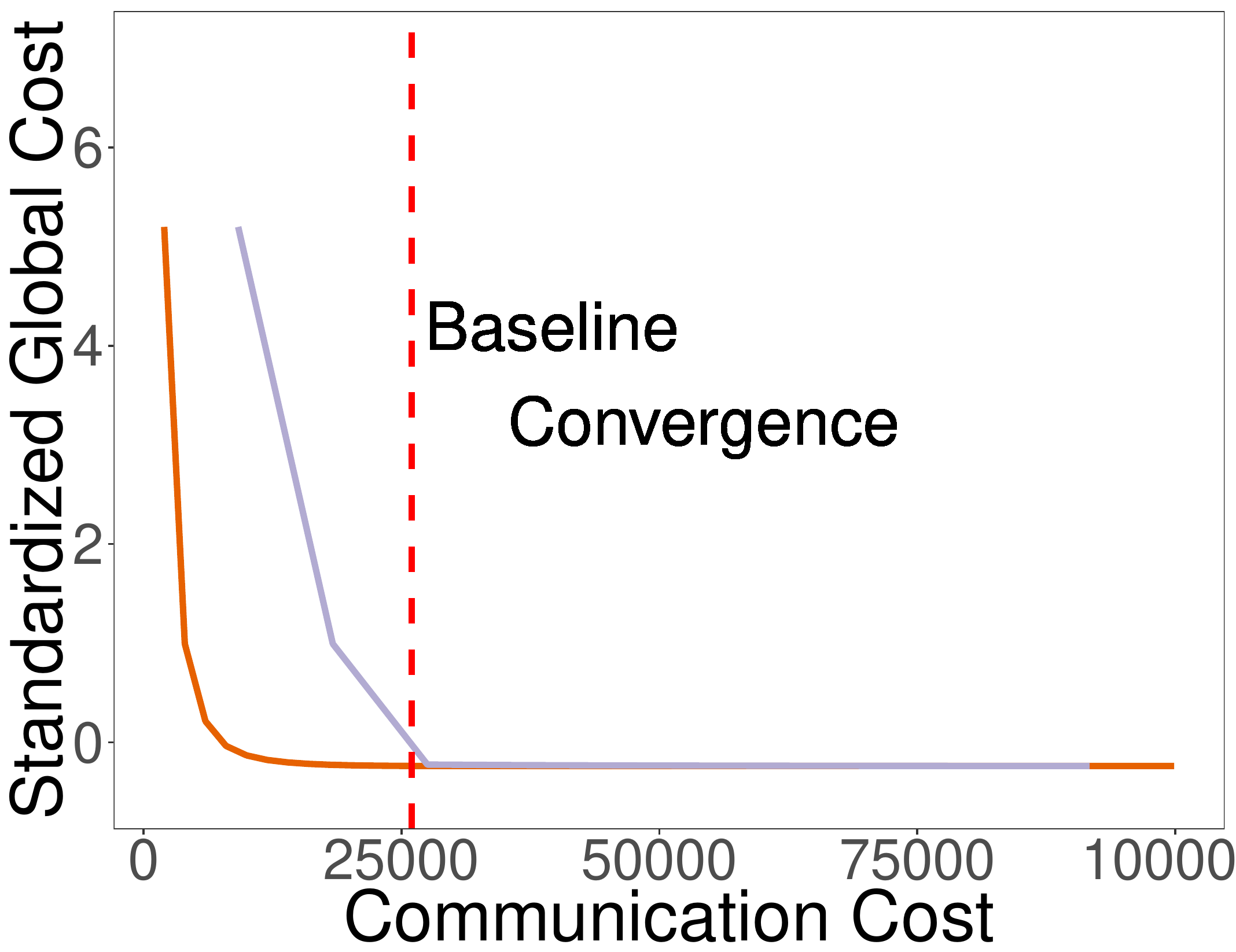}}
\subfigure[Bike sharing, total communication]{\includegraphics[width=0.244\textwidth]{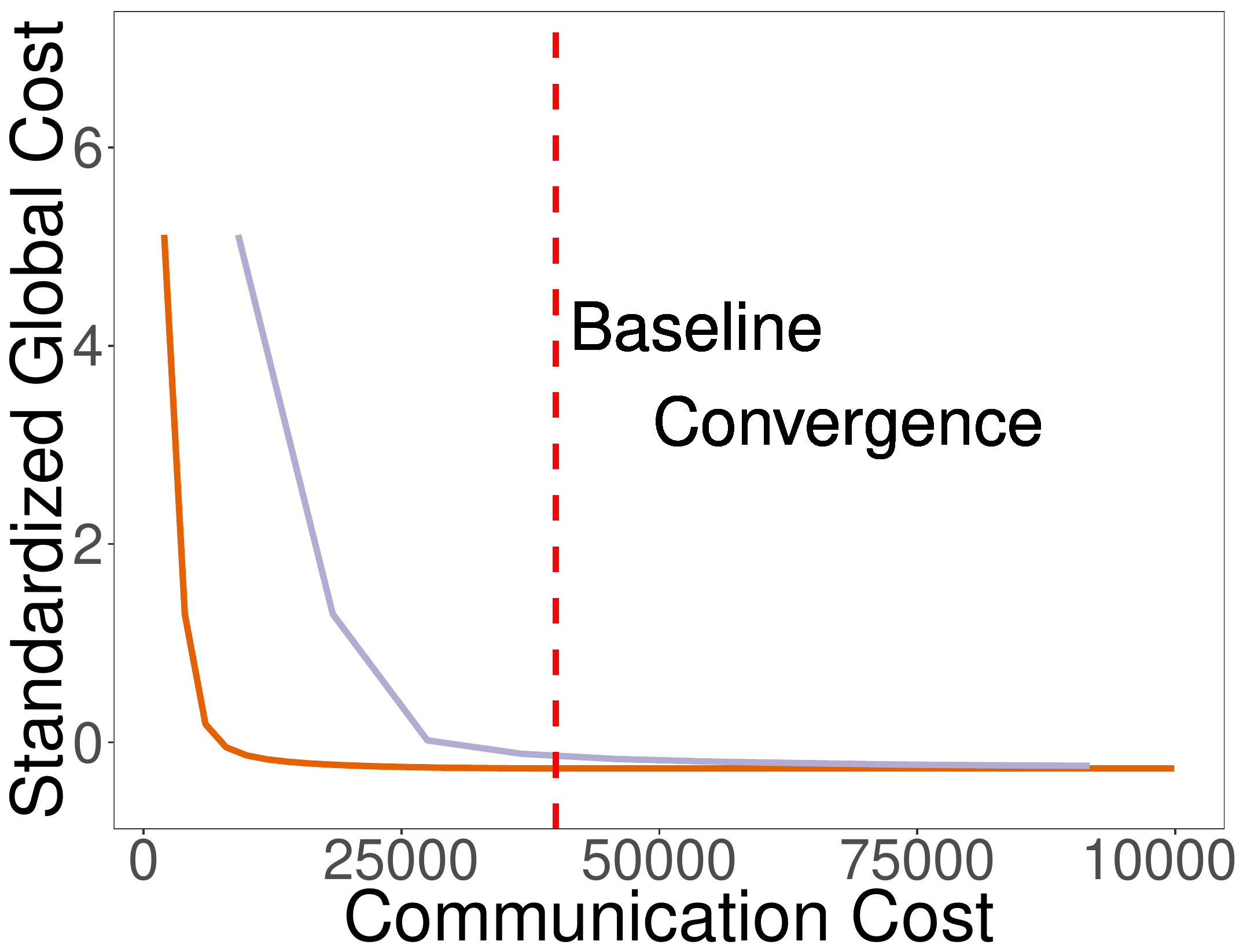}}
\subfigure[Energy demand, total communication]{\includegraphics[width=0.244\textwidth]{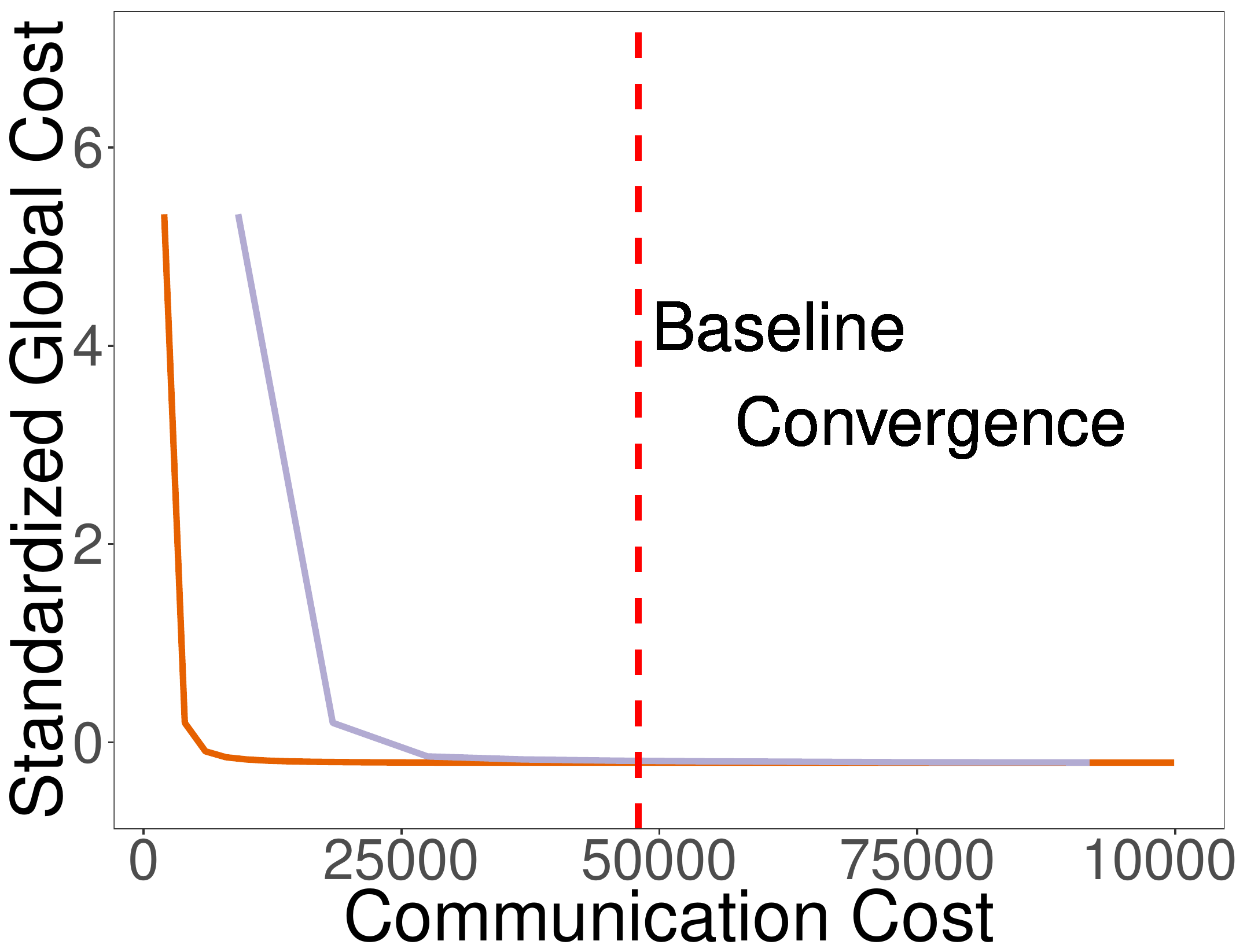}}
\subfigure[Electric vehicles, total communication]{\includegraphics[width=0.244\textwidth]{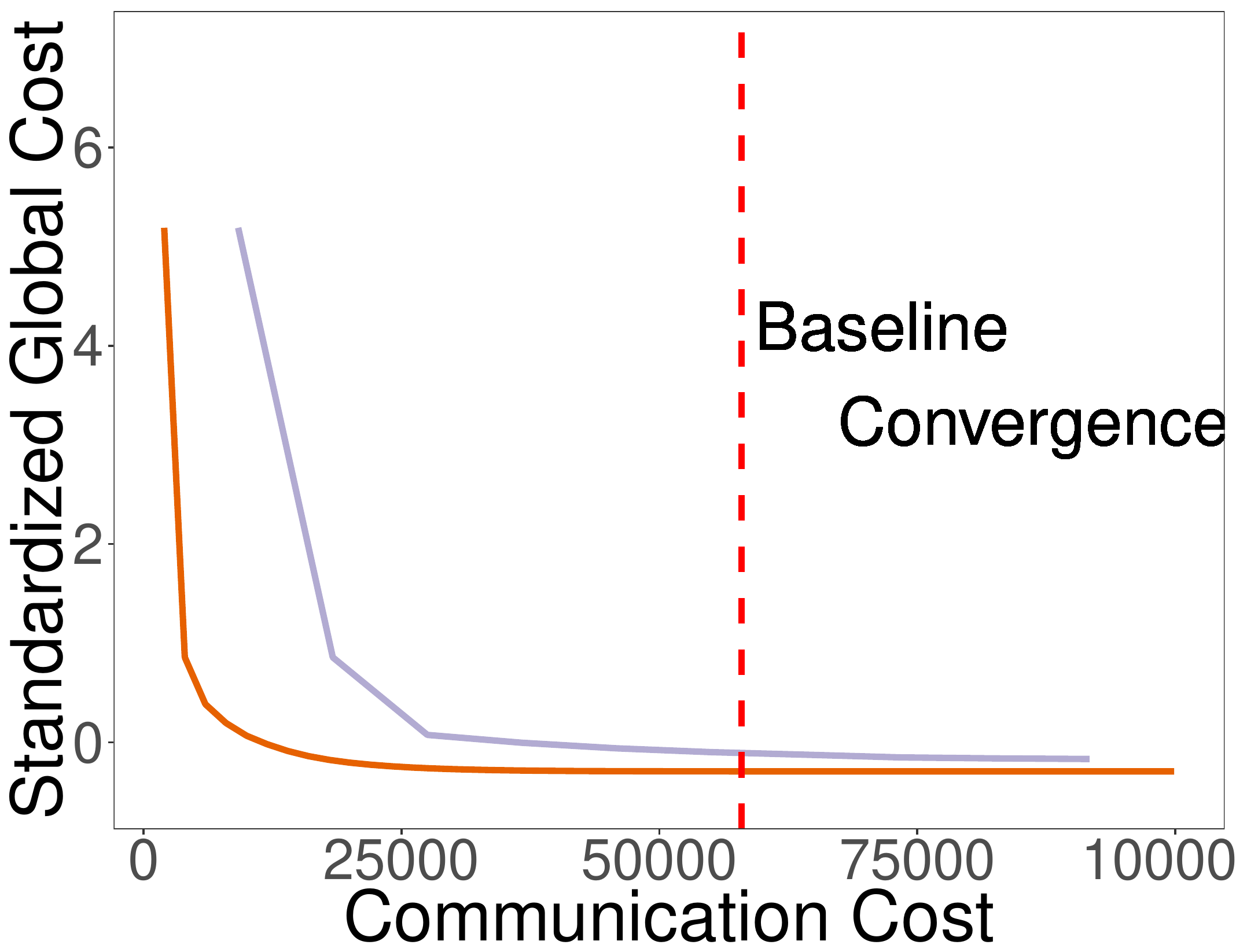}}
\subfigure[Synthetic, synchronized communication]{\includegraphics[width=0.244\textwidth]{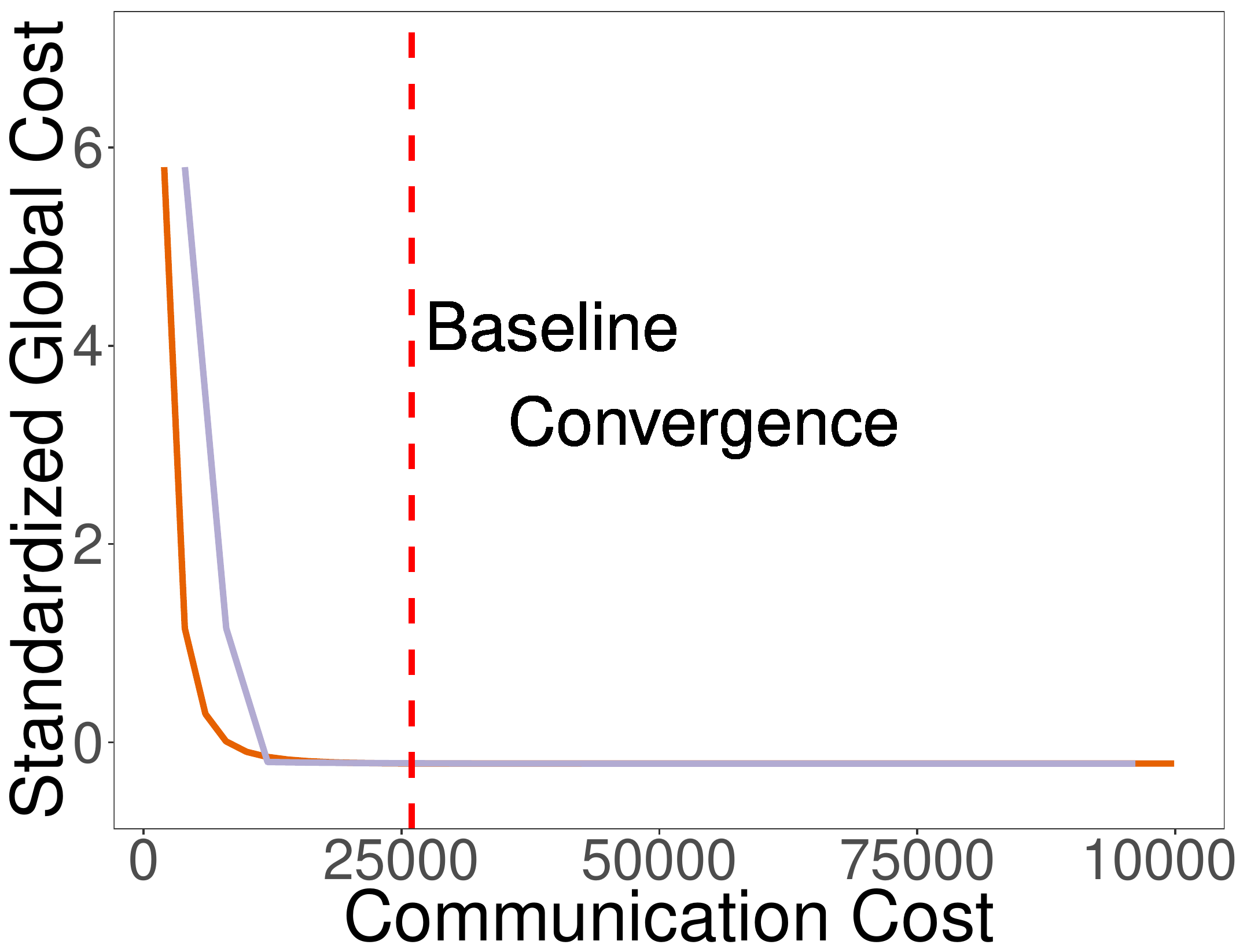}}
\subfigure[Bike sharing, synchronized communication]{\includegraphics[width=0.244\textwidth]{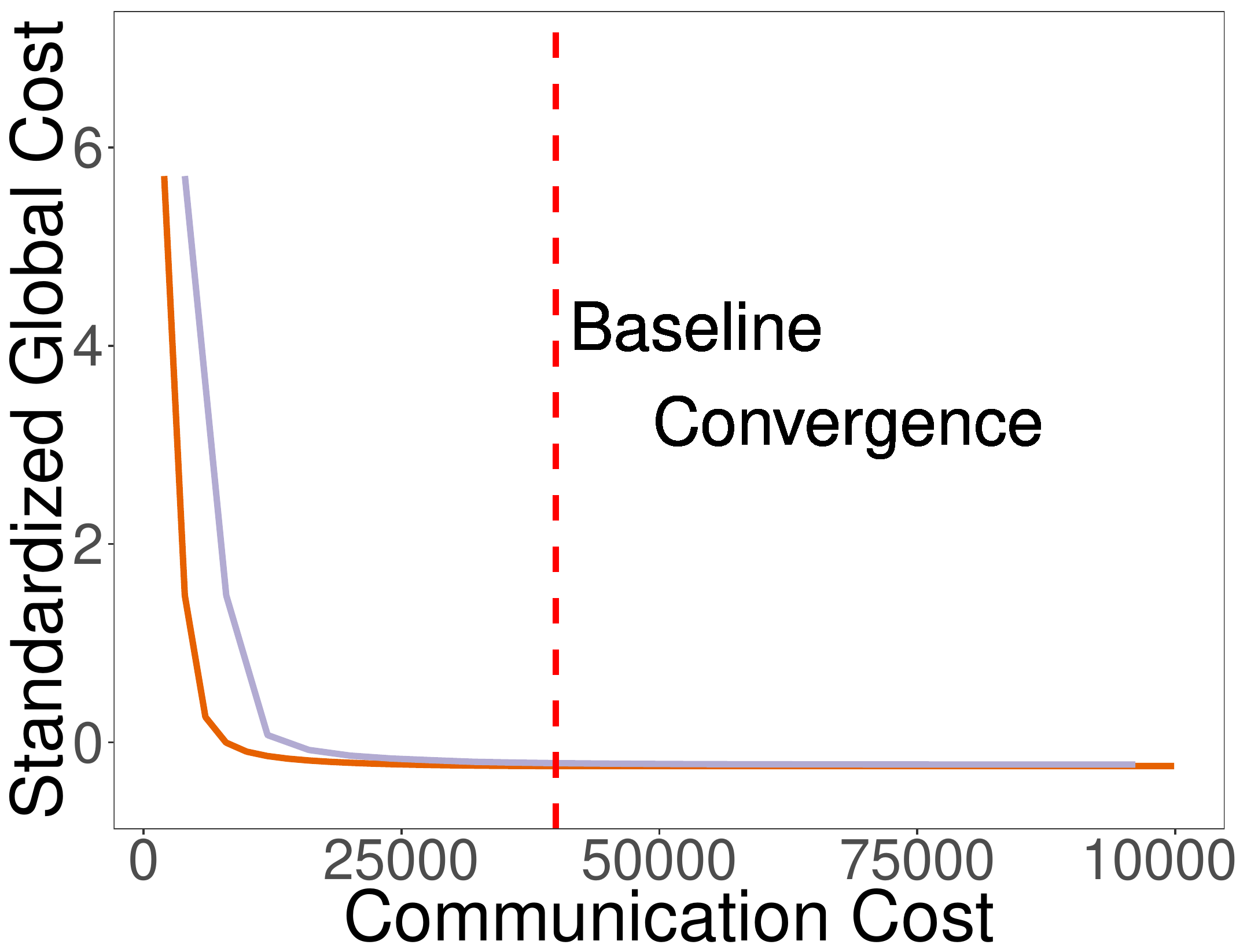}}
\subfigure[Energy demand, synchronized communication]{\includegraphics[width=0.244\textwidth]{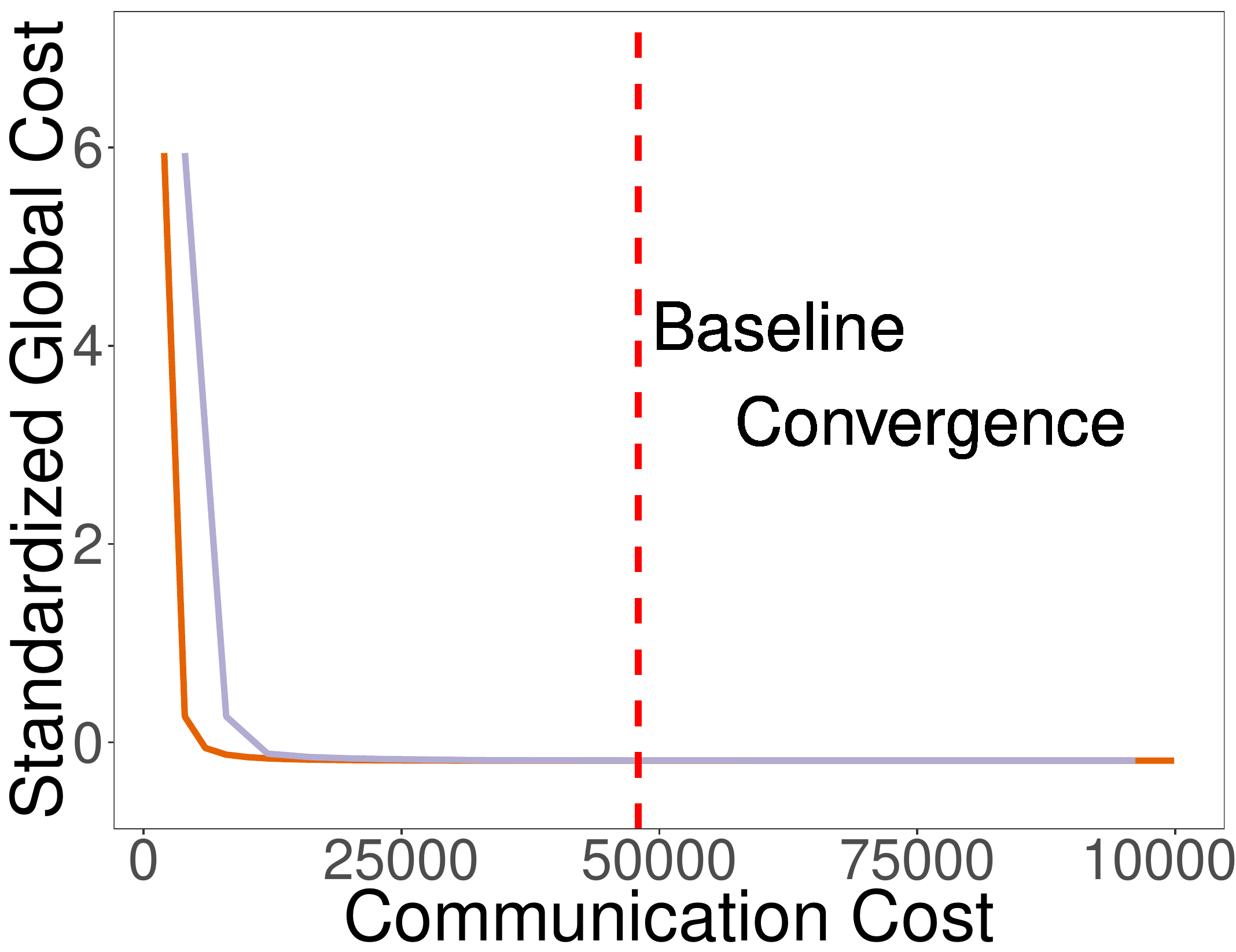}}
\subfigure[Electric vehicles, synchronized communication]{\includegraphics[width=0.244\textwidth]{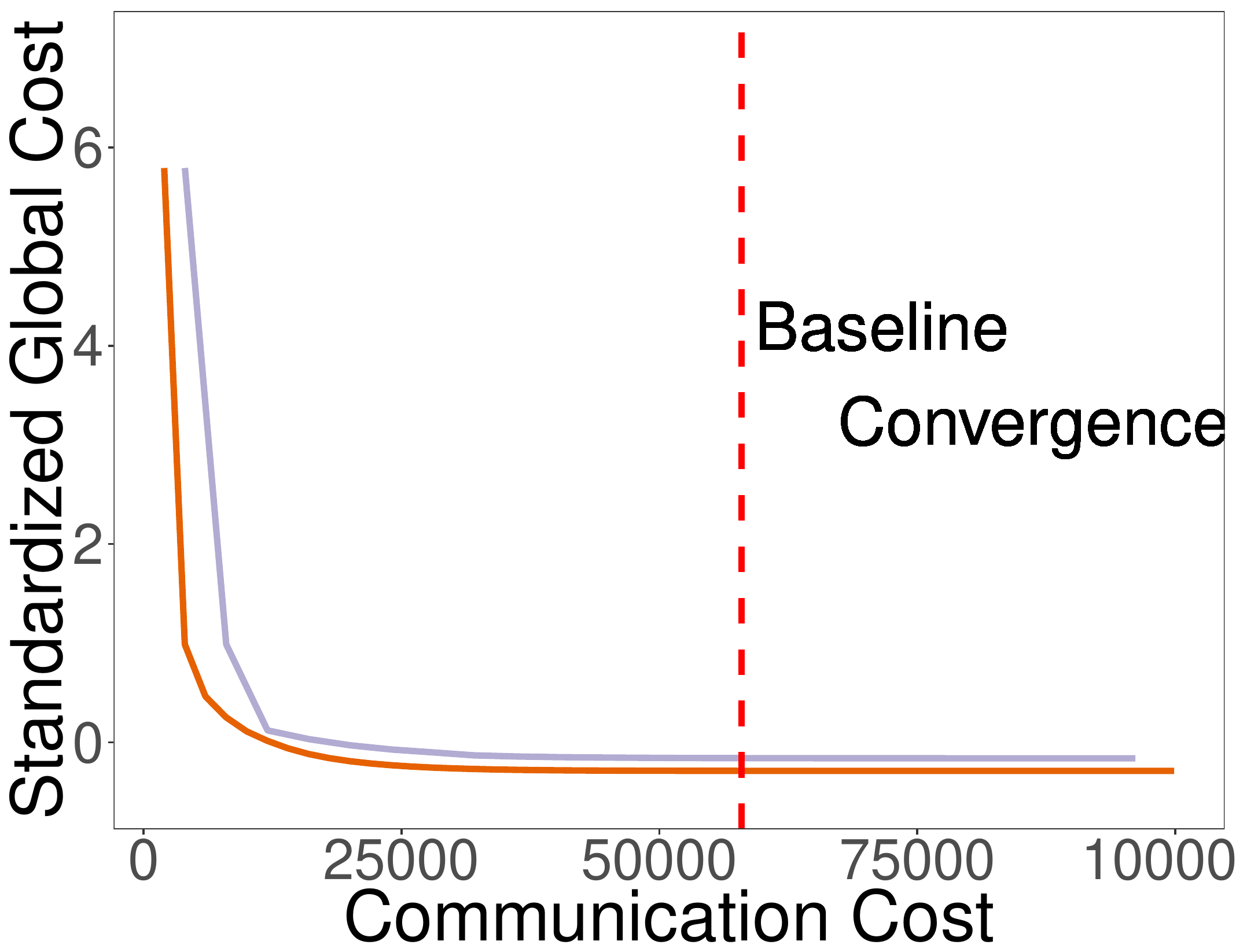}}
\caption{Cost-effectiveness. \emph{Dimensions}: baseline versus holarchic runtime, application scenarios, total versus synchronized communication cost. \emph{Settings}: partial scale, $\lambda=0$, $c=2$.}\label{fig:cost-effectiveness-two-children}
\end{figure}

These findings can be generalized further for the broader range of communication cost as shown in Figure~\ref{fig:cost-effectiveness-two-children} for $c=2$. The key observation that confirms the mitigation capability of the holarchic runtime is the comparable global cost achieved using the communication cost required for baseline to converge. This mitigation potential is also demonstrated by the probability density of the relative global cost between baseline and holarchic runtime in Figure~\ref{fig:density-relative-global-cost-reduction-communication-children-partial} of Appendix~\ref{sec:details}. 

%On average X and X total and synchronized messages are required to reach a lower than 10\% and 20\% divergence in global cost, which are usually below the required messages for the convergence of the baseline. 

Figure~\ref{fig:cost-effectiveness-two-children} can also be compared with Figure~\ref{fig:cost-effectiveness-five-children} of Appendix~\ref{sec:details} that shows the cost-effe\-cti\-ve\-ness under $c=5$. The performance mitigation is even higher with this setting.

\section{Summary of Results and Discussion}\label{sec:discussion}

The following key observations can be made about the learning capacity of holarchic structures: (i) The limitation to contribute improvements in terms of global cost as motivated in Figure~\ref{fig:learning-baseline} and~\ref{fig:optimality} is confirmed in the performed experiments. (ii) The performance exploration, mitigation and boosting for which the holarchic schemes are designed are confirmed: A low and sporadic performance improvement is observed by holarchic initialization, especially in constrained environments of high $\lambda$ values that justifies the exploration potential of this scheme. Similarly, the mitigation potential of the holarchic runtime is also confirmed by the fast convergence and preservation of the performance, especially for trees with a higher number of children $c$. Performance boosting by holarchic termination is scarce but observable in the bike sharing scenario that has sparser data. (ii) Strikingly, holarchies applied at a partial scale demonstrate higher cost-effectiveness than full scale in all holarchic schemes, i.e. lower cost due to the higher localization that limits communication cost and higher effectiveness in terms of higher improvement index. 

In the trade-offs of cost-effectiveness, the synchronized communication cost is almost half of the total communication cost via a parallel execution of the learning process using holarchic structures. Moreover, for the same communication cost in baseline and holarchic runtime, the global costs become equivalent, especially in terms of the synchronized communication cost and for trees with a higher number of children $c$. 

Overall the results suggest that several parallel and small-scale holarchic structures decrease the probability of trapping to local optima, require a lower parallelizable communication cost, while they better serve the purpose they are designed for: make decentralized deep learning more resilient to the uncertainties of distributed environments. Because of the high efficiency of I-EPOS to which holarchic structures are applied, conclusions cannot be reached whether the learning capacity of other mechanisms can be enhanced using holarchic structures. Nevertheless, another promising use case of holarchies is the multi-level optimization of complex techno-socio-economic systems formed by agents with different goals. In other words, the localization of the learning process in part of the deep hierarchical structure can represent the collective effort of a community to meet its goal that can differ or even oppose the goal of another community, i.e. another part of the deep hierarchical structure.

\section{Positioning and Comparison with Related Work}\label{sec:related-work}

Earlier work~\cite{simon62archCmplx,simon1996sciences} identifies the key role that hierarchical design  plays in dealing with the complexity~\cite{cmplx-oc2017} of large-scale and highly dynamic systems. More specifically, it is studied how such hierarchies can address the challenge of \textit{limited rationality}, which stems from the combinatorial explosion of alternative system compositions. In the context of this paper, the combinatorial explosion results from the multitude of resource planning alternatives. Such hierarchies are later on introduced as `holarchies' by emphasizing their recursive and self-encapsulated nature~\cite{Koestler67} that represent both an entire system (whole) and a mere subsystem (part). 

General-purpose holonic designs that integrate the above principles are proposed via holonic multi-agent system platforms~\cite{rodriguez2011,sarl-iat2014}; multi-level modeling and simulation approaches~\cite{Gaud2008}; and hierarchical problem solving~\cite{archCmplxSys-Landauer2001}. Customised holonic designs are been applied to various domains, including hierarchical planning~\cite{Nolle2002}, traffic control~\cite{Ficher99}, manufacturing~\cite{Christensen2003,Giret2009} and smart grids~\cite{lassig2011,schiendorfer14,FreyTAAS2015}. The success of these applications motivate the adoption of similar holonic principles for designing deep learning systems for large-scale distributed environments.

The design of I-EPOS~\cite{Pournaras2018,Pilgerstorfer2017}, used in this paper as case study and performance baseline, adopts a hierarchical approach, featuring: (i) bottom-up abstraction -- via the aggregation of plans from the lower level; (ii) partial isolation -- as different tree branches operate in parallel; and (iii) time tuning -- by synchronizing the execution of hierarchical levels. However, the learning process in I-EPOS executes over the entire hierarchy -- sequentially, level-by-level -- rather than being nested within multiple holarchic levels. 

Various deep learning techniques employ a hierarchical structure for different purposes. This depends on the nature of the learning problem, e.g. classification or prediction, and on the context within which the learning process executes, i.e. amount of input data and availability of computational resources. Hierarchical models are employed to deal with complex processing of data, by performing learning tasks progressively within incremental modelling levels, e.g. image classification~\cite{ruslan2013,huang2012} and text categorisation~\cite{kowsari2017}. Hierarchical structures are employed to process online large amounts of distributed input data, hence scaling up machine learning techniques~\cite{surveyDML2012,mlnet2015}. Such learning approaches rely on data partitions -- pre-existing or artificially created -- over which they distribute the learning process. The partial results from each partition are then collected and aggregated into an overall learning outcome. The work of this paper is positioned and compared below with some of these approaches. 

An earlier survey~\cite{surveyDML2012} compares distributed learning approaches with respect to their capability to (i) combine learning outcomes among heterogeneous representations of data partitions and (ii) deal with privacy constraints. The studied approaches feature a two-layer hierarchy: one layer for distributed learning across data partitions and a second layer for collecting and aggregating the results. In contrast, the holarchic learning design proposed in this paper introduces multiple levels to further to enhance the learning capacity in distributed environments under uncertainties. The localized holons operating in parallel reuse their learning outcomes even when the topology is clustered by node or link failures. 

MLNet~\cite{mlnet2015} introduces a special-purpose communication layer for distributed machine learning. It uses tree-based overlay networks to aggregate progressively partial learning results in order to reduce network traffic. This approach draws parallels with the design proposed here, while it is more applicable at the lower communication layers.

In dynamic environments, rescheduling, or reoptimization~\cite{meignan2014} becomes a critical function for dealing with unpredictable disturbances. It usually raises the additional constraints of minimizing reoptimization time as well as the distance between the initial optimization solution and the one of reoptimization, e.g. dynamic rescheduling in manufacturing systems~\cite{Leitao2008} or shift rescheduling~\cite{meignan2014}. For instance, a two-level holarchy is earlier adopted to combine global optimization scheduling with fast rescheduling when dynamic system disturbances occur~\cite{meignan2014}. In contrast, the holarchic schemes of this paper do not undermine the learning performance, while adding resilience by localizing the learning process without reinitiating it. 

Another line of relevant related work concerns the initial selection and maintenance of the topology over which the distributed learning process operates. Multi-agent approaches often rely on self-organization by changing their interactions and system structure at runtime. A holonic multi-agent approach for optimizing facility location problems is earlier introduced~\cite{hmas-gaud2007}. Facilities include distribution of bus stops, hospitals or schools within a geographical area. The agents react to mutual attraction and repulsion forces to self-organize into a holarchy. Stable solutions represent optimal facility localization distributions. This self-organization process has a considerable computational and communication cost, in case of remote agents' interactions. In contrast, the holarchic schemes studied in this paper preserve the agents' organizational structure, while the decentralized learning process used for system optimization is self-adapted by localizing the learning span within part of the tree network. 

The experimental work of this paper shows how different topological configurations, i.e. agents' positioning and number of children, influence learning performance. The key role that such hyperparameters play in ensuring the effectiveness of learning approaches is also confirmed by related work on the optimization of the hierarchical structure and its configuration variables, for instance grid search, i.e. exhaustive search of all possibilities, (which, however, suffers from exponential combinatorics), random search, Bayesian optimization, e.g. in neural networks and deep belief systems~\cite{Bergstra2011,Bergstra2012} as well as gradient-based optimization~\cite{Maclaurin2015}. This is also relevant for deep learning applications via unsupervised pre-training~\cite{YAO2017} and evolutionary algorithms, e.g. in deep learning neural networks~\cite{MiikkulainenLMR17,Young2017}. In the context of this work, such approaches can be used to determine the most effective holarchic structures and hyperparameter configurations.

\section{Conclusion and Future Work}\label{sec:future-work}

This paper concludes that holarchic structures for decentralized deep learning can be a highly cost-effective organizational artifact for managing learning performance under uncertainties of distributed environments. The communication cost of self-organization can be eliminated by self-adapting the span of the learning process at a more localized level within the hierarchical structure as the means to cope with failures, latency and constrained computational resources. An extensive experimental evaluation with more than 864000 experiments fed with synthetic and real-world data from pilot projects confirm the potential to explore, mitigate and boost the learning performance using three respective holarchic schemes applied to the I-EPOS decentralized deep learning system for solving combinatorial optimization problems. 

Results show that the exploration of improving solutions is feasible and more likely to happen under stricter agents' constraints, while performance mitigation is more effective in balanced tree topologies with higher number of children. Boosting the learning performance via holarchic structures is challenging yet consistently observed under computational problems with sparse data, i.e. the bike sharing application scenario. The partial scale of the holarchic structures is more cost-effective than the full scale. Nevertheless, when the uncertainties of distributed environments are not anymore a constraint, holarchic schemes cannot outperform the cost-effectiveness of learning systems that make use of the whole hierarchical structure, which is a finding consistent with earlier work on greedy optimization and suboptimum heuristics trapped in local optima~\cite{Bang2004}. 

Mechanisms for an automated activation and deactivation of holarchic schemes as well as the applicability of these schemes in other more complex hierarchical structures than tree topologies are subject of future work. Applying the concept of holarchy in non-hierarchical structures such as the unstructured learning network of COHDA~\cite{cohdaOrig} can provide new means to control high communication costs, while preserving a high learning performance.

\begin{acknowledgements}
This work is supported by the European Community’s H2020 Program under the scheme `INFRAIA-1-2014-2015: Research Infrastructures’, grant agreement \#654024 `SoBigData: Social Mining \& Big Data Ecosystem’ (http://www.sobigdata.eu) and the European Community’s H2020 Program under the scheme `ICT-10-2015 RIA', grant agreement \#688364 `ASSET: Instant Gratification for Collective Awareness and Sustainable Consumerism'.
\end{acknowledgements}

% BibTeX users please use one of
%\bibliographystyle{spbasic}      % basic style, author-year citations
\bibliographystyle{spmpsci}      % mathematics and physical sciences
\bibliography{I-EPOS}   % name your BibTeX data base

\appendix

\section{Detailed Experimental Results}\label{sec:details}

\begin{figure}[!htb]
\centering
\includegraphics[width=1.0\textwidth]{legend-holarchic-schemes.pdf}\\
\subfigure[Synthetic, full scale, $\lambda=0$, $c=2$]{\includegraphics[width=0.244\textwidth]{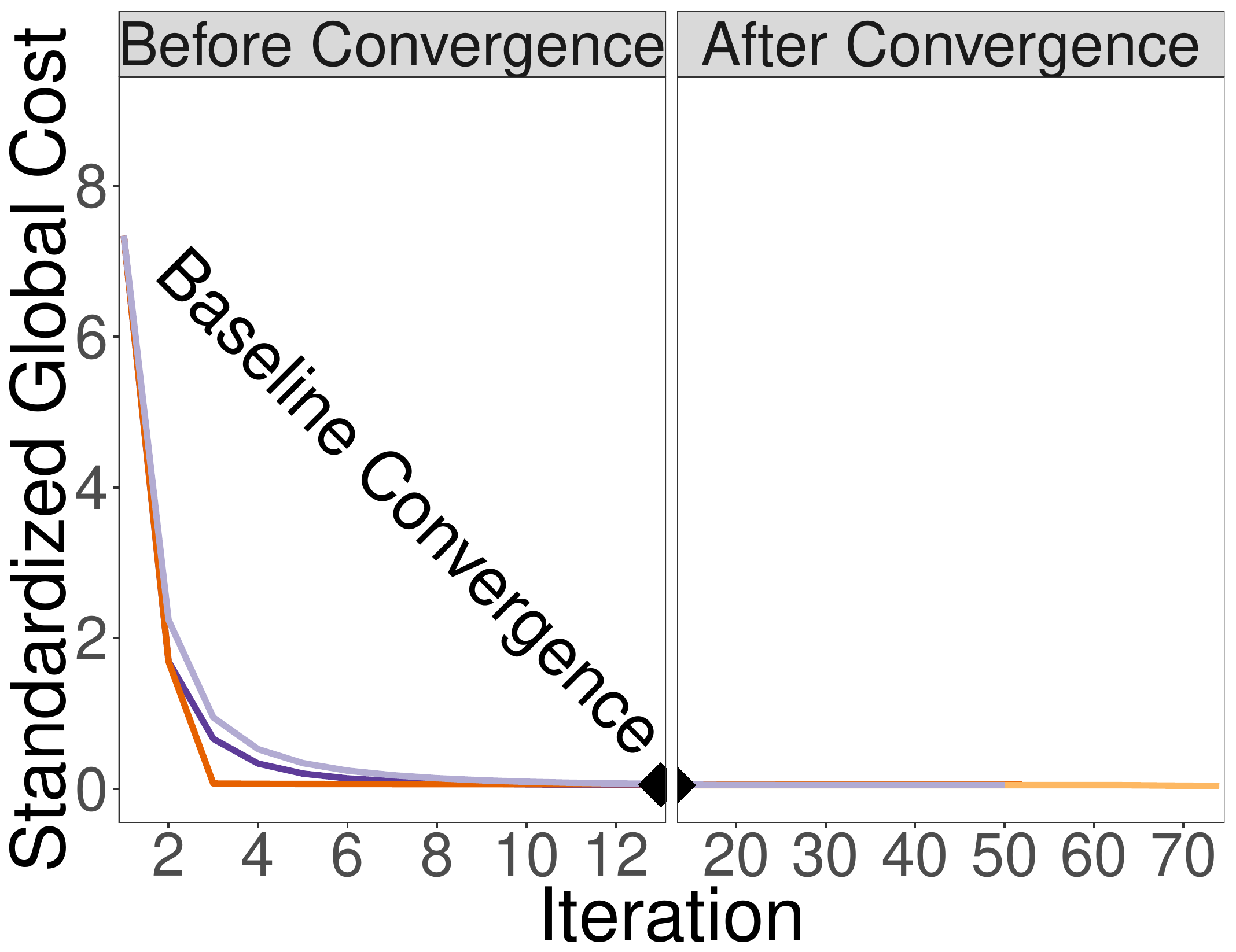}}
\subfigure[Bike sharing, full scale, $\lambda=0$, $c=2$]{\includegraphics[width=0.244\textwidth]{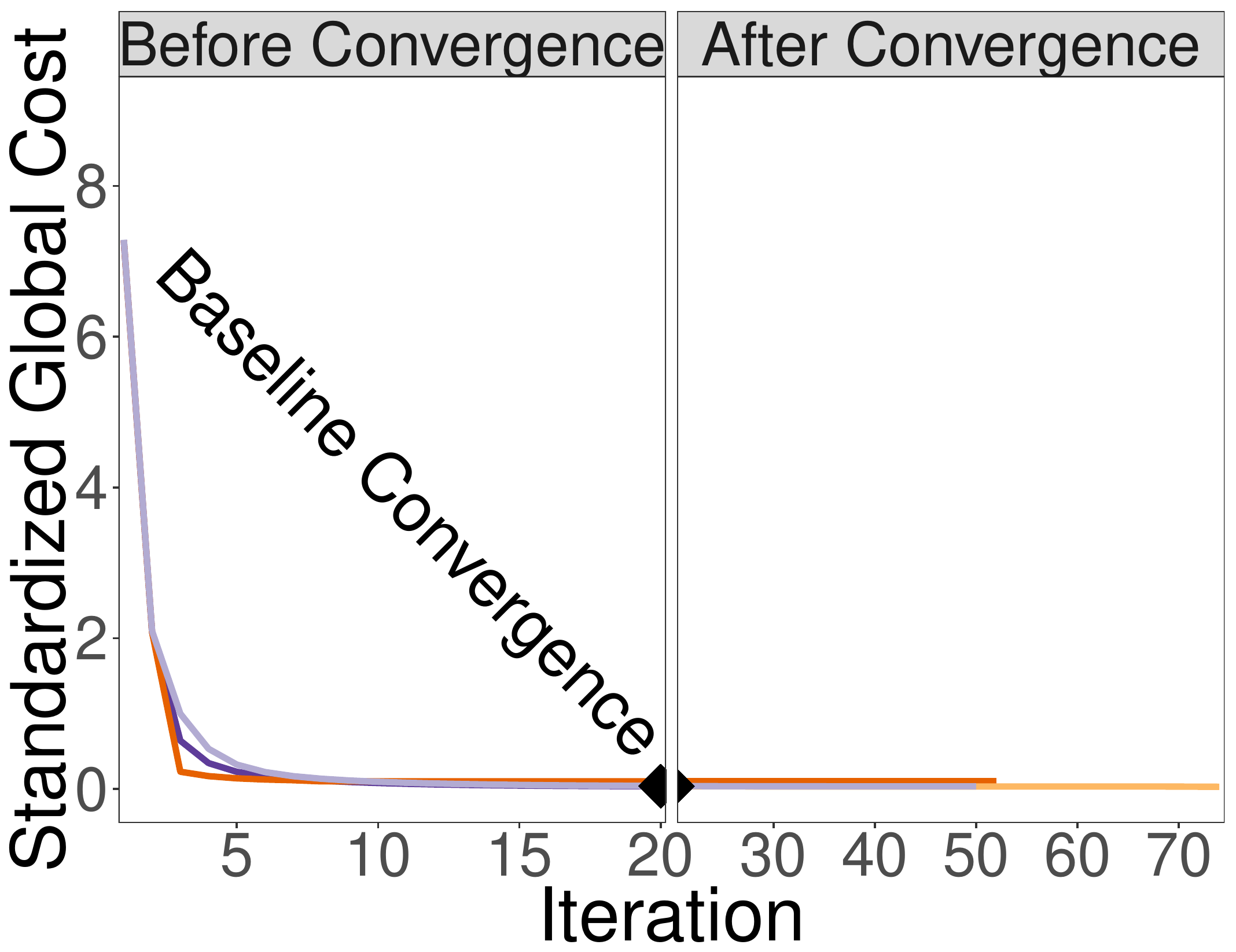}}
\subfigure[Energy demand, full scale, $\lambda=0$, $c=2$]{\includegraphics[width=0.244\textwidth]{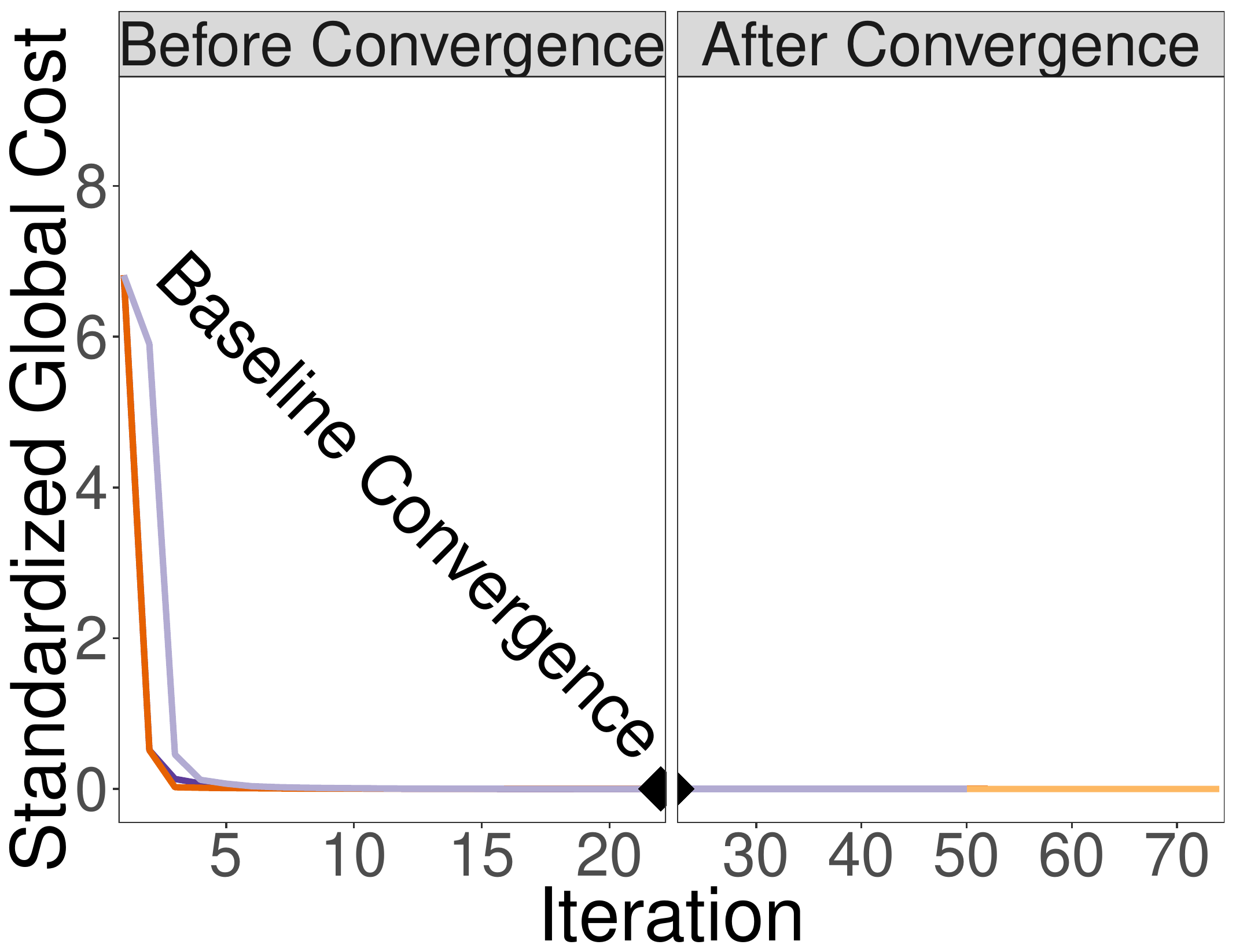}}
\subfigure[Electric vehicles, full scale, $\lambda=0$, $c=2$]{\includegraphics[width=0.244\textwidth]{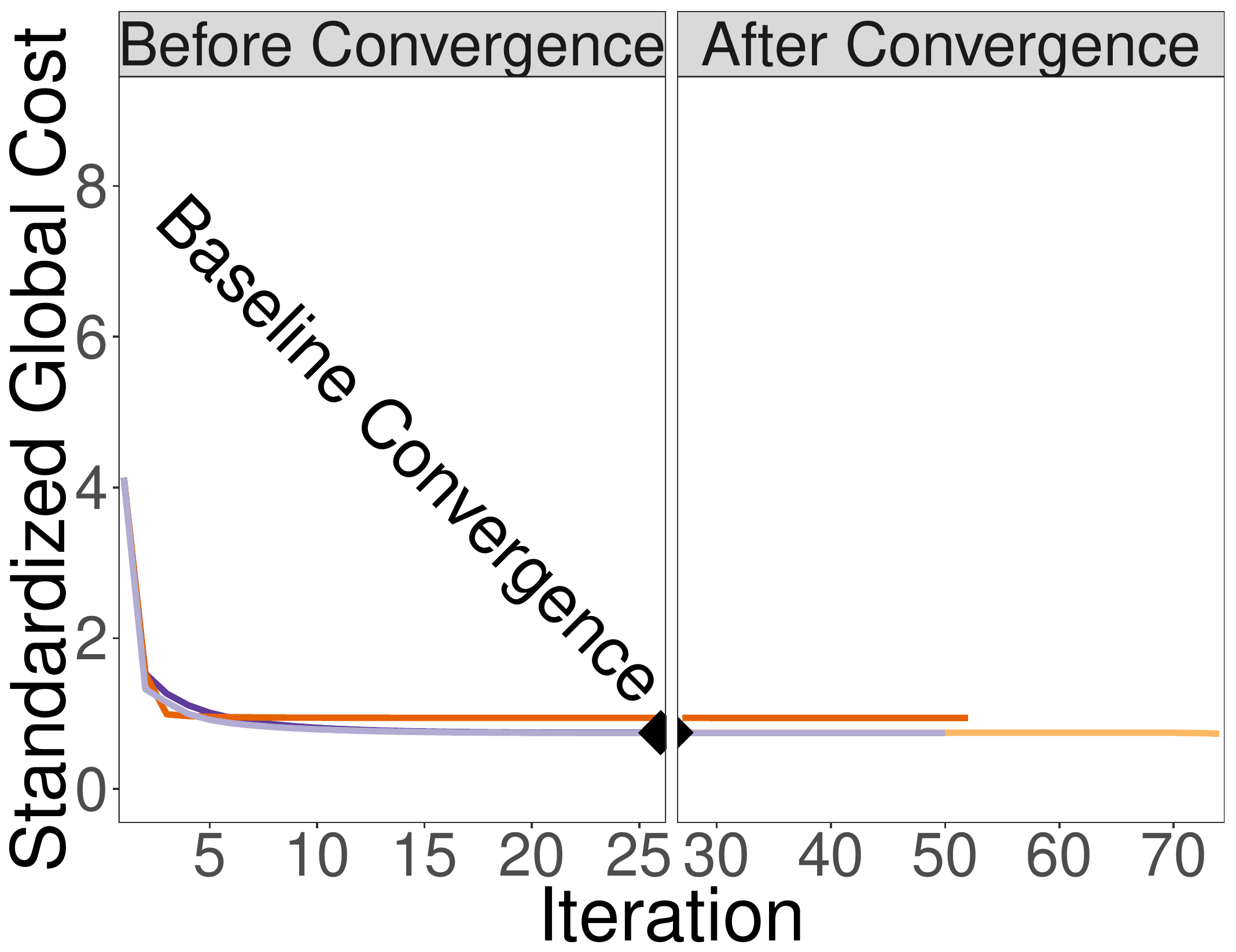}}
\subfigure[Synthetic, partial scale, $\lambda=0.5$, $c=2$]{\includegraphics[width=0.244\textwidth]{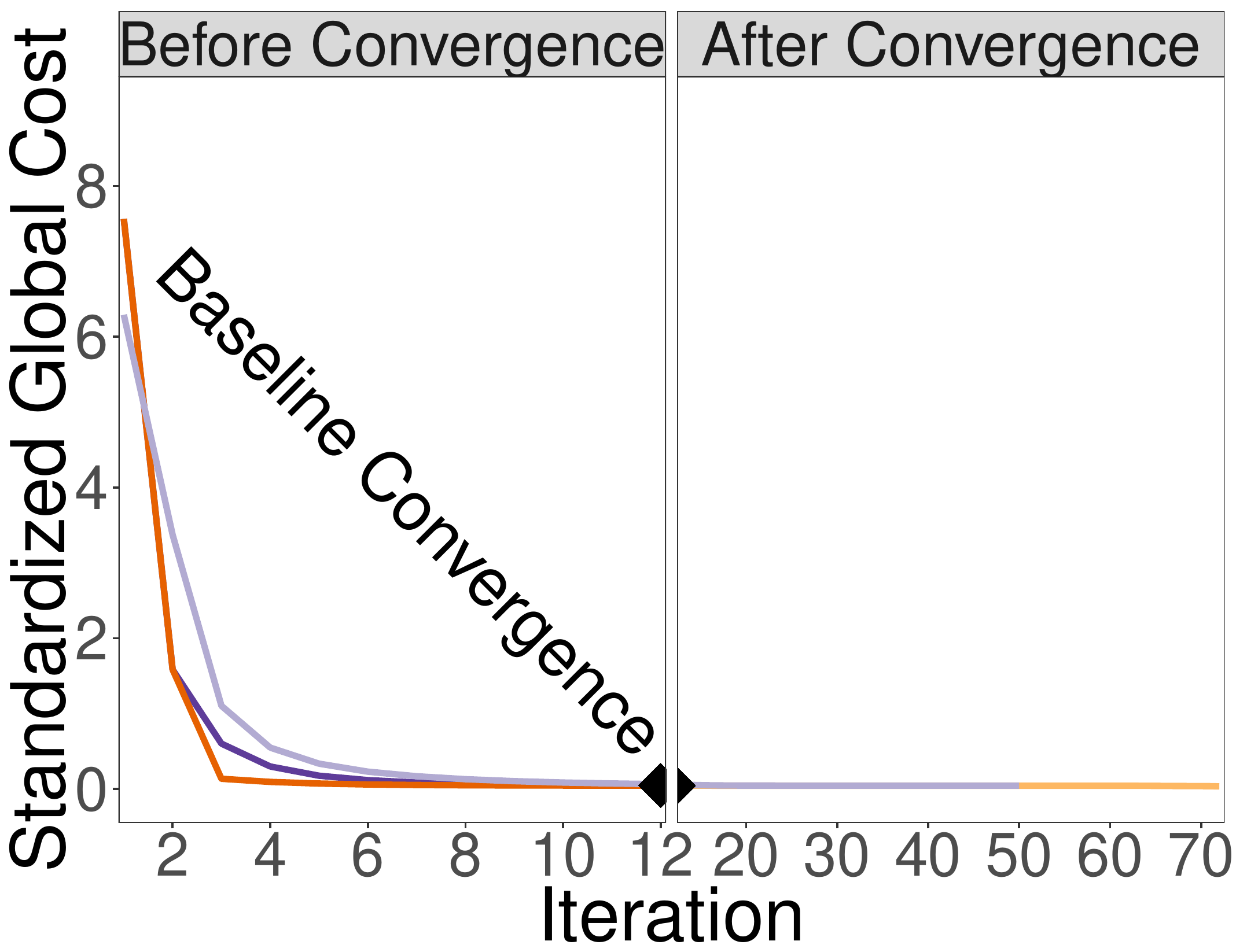}}
\subfigure[Bike sharing, partial scale, $\lambda=0.5$, $c=2$]{\includegraphics[width=0.244\textwidth]{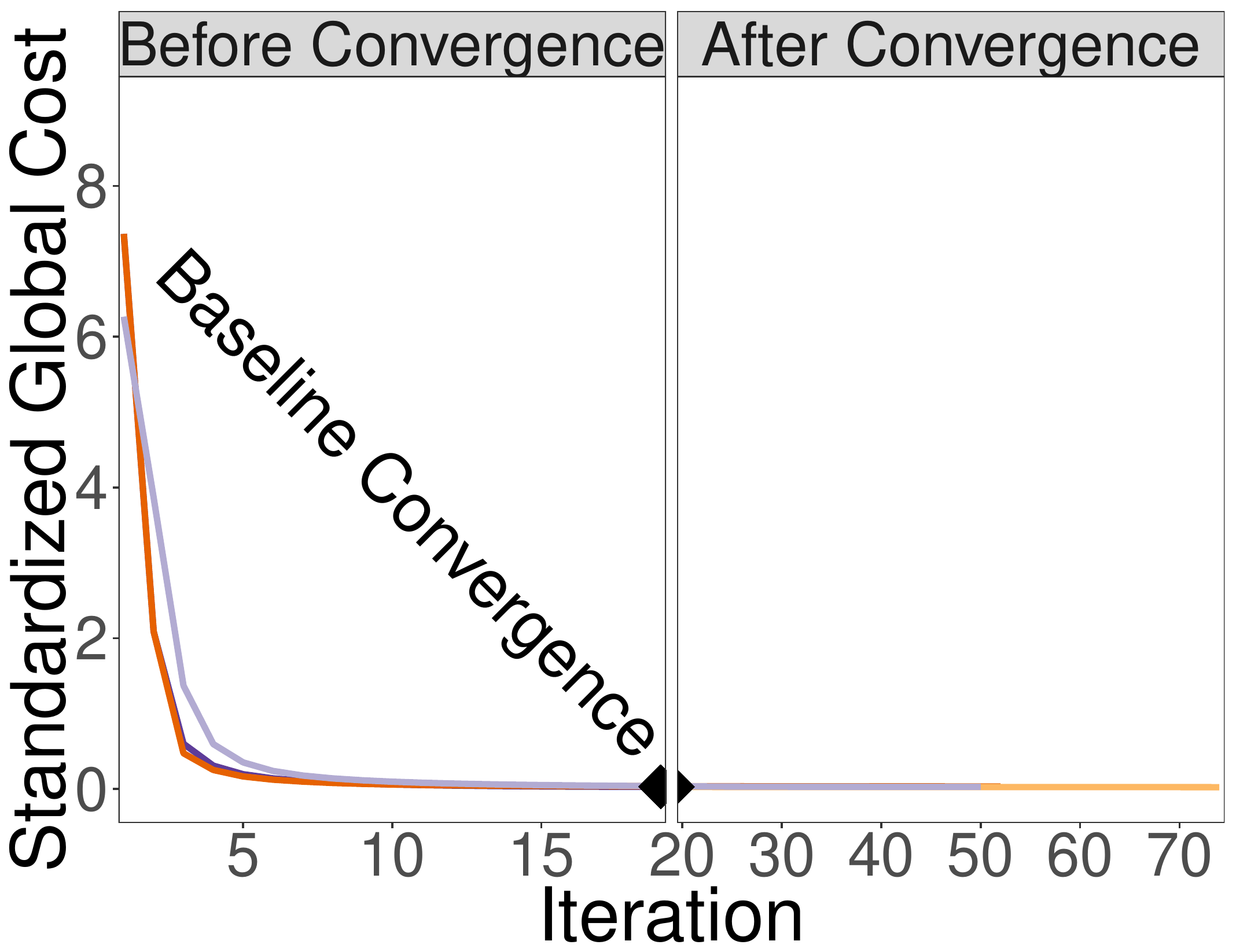}}
\subfigure[Energy demand, partial scale, $\lambda=0.5$, $c=2$]{\includegraphics[width=0.244\textwidth]{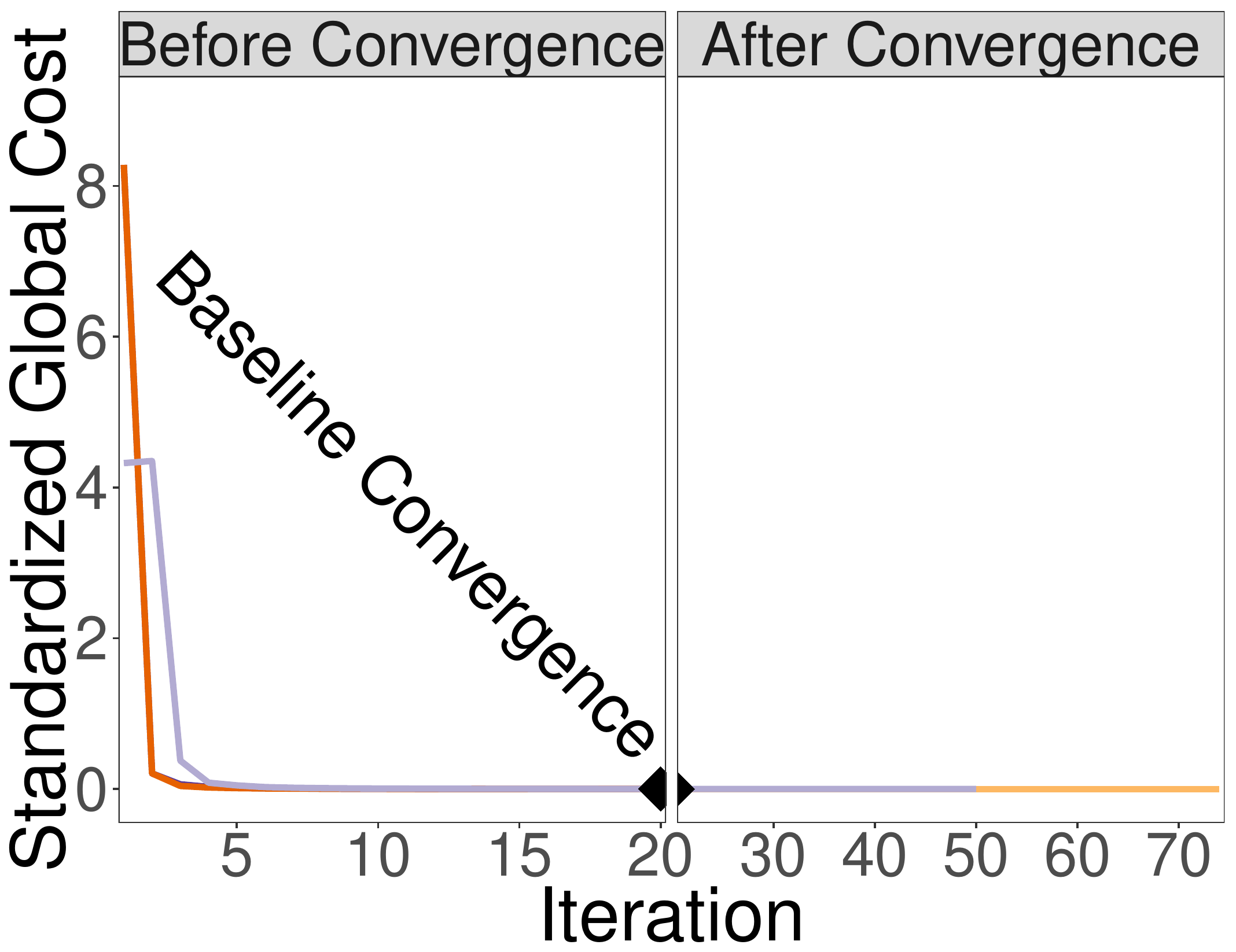}}
\subfigure[Electric vehicles, partial scale, $\lambda=0.5$, $c=2$]{\includegraphics[width=0.244\textwidth]{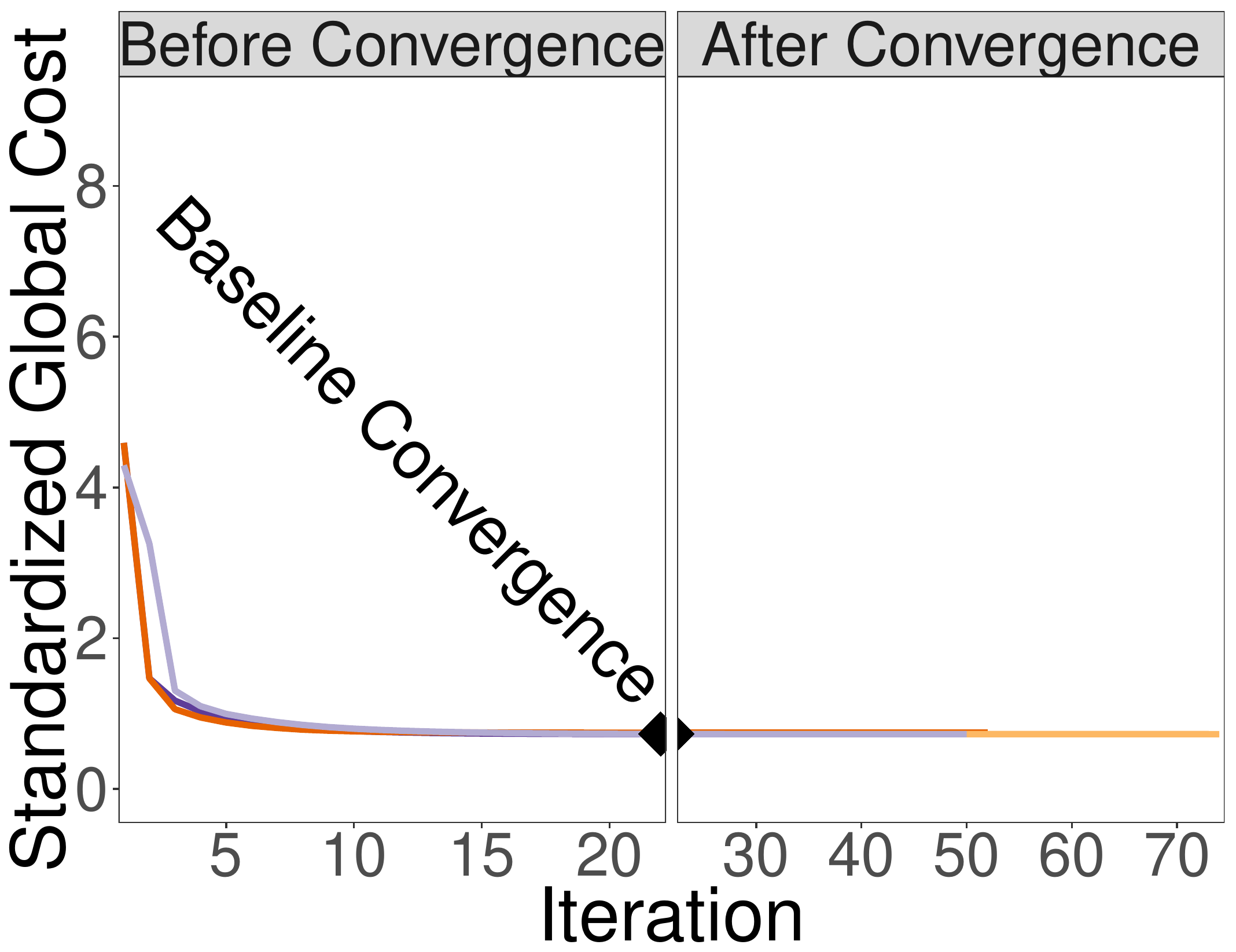}}
\subfigure[Synthetic, partial scale, $\lambda=0$, $c=5$]{\includegraphics[width=0.244\textwidth]{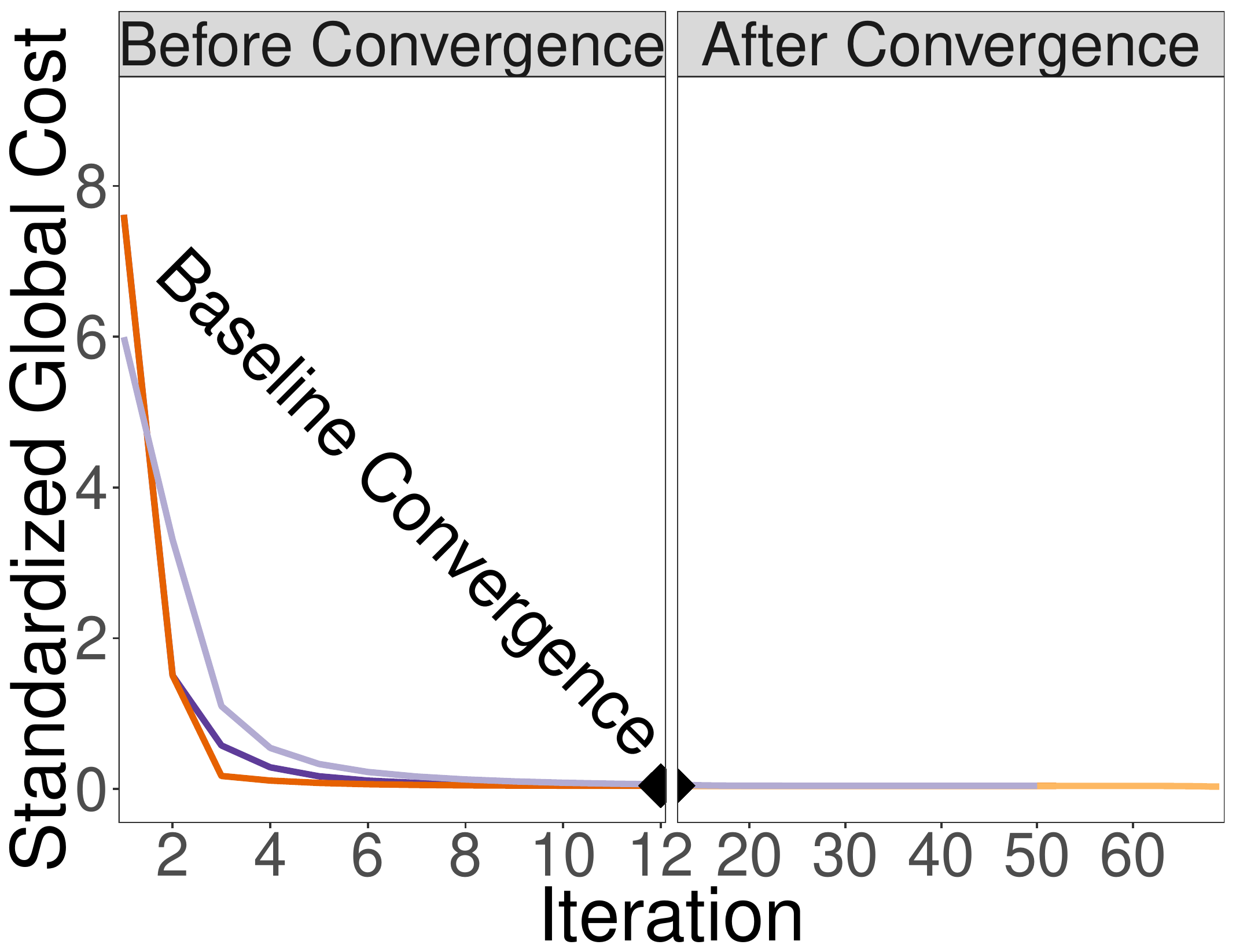}}
\subfigure[Bike sharing, partial scale, $\lambda=0$, $c=5$]{\includegraphics[width=0.244\textwidth]{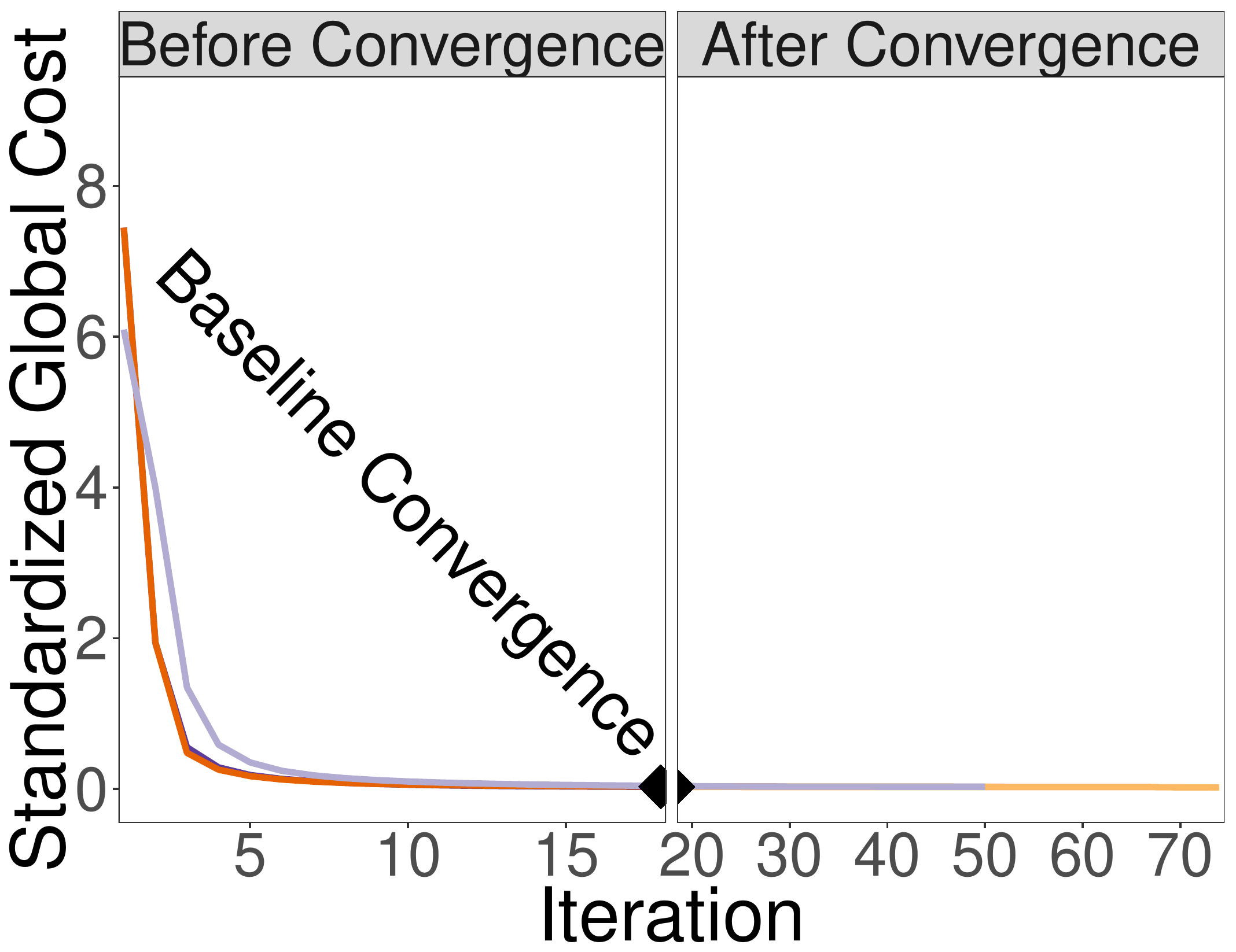}}
\subfigure[Energy demand, partial scale, $\lambda=0$, $c=5$]{\includegraphics[width=0.244\textwidth]{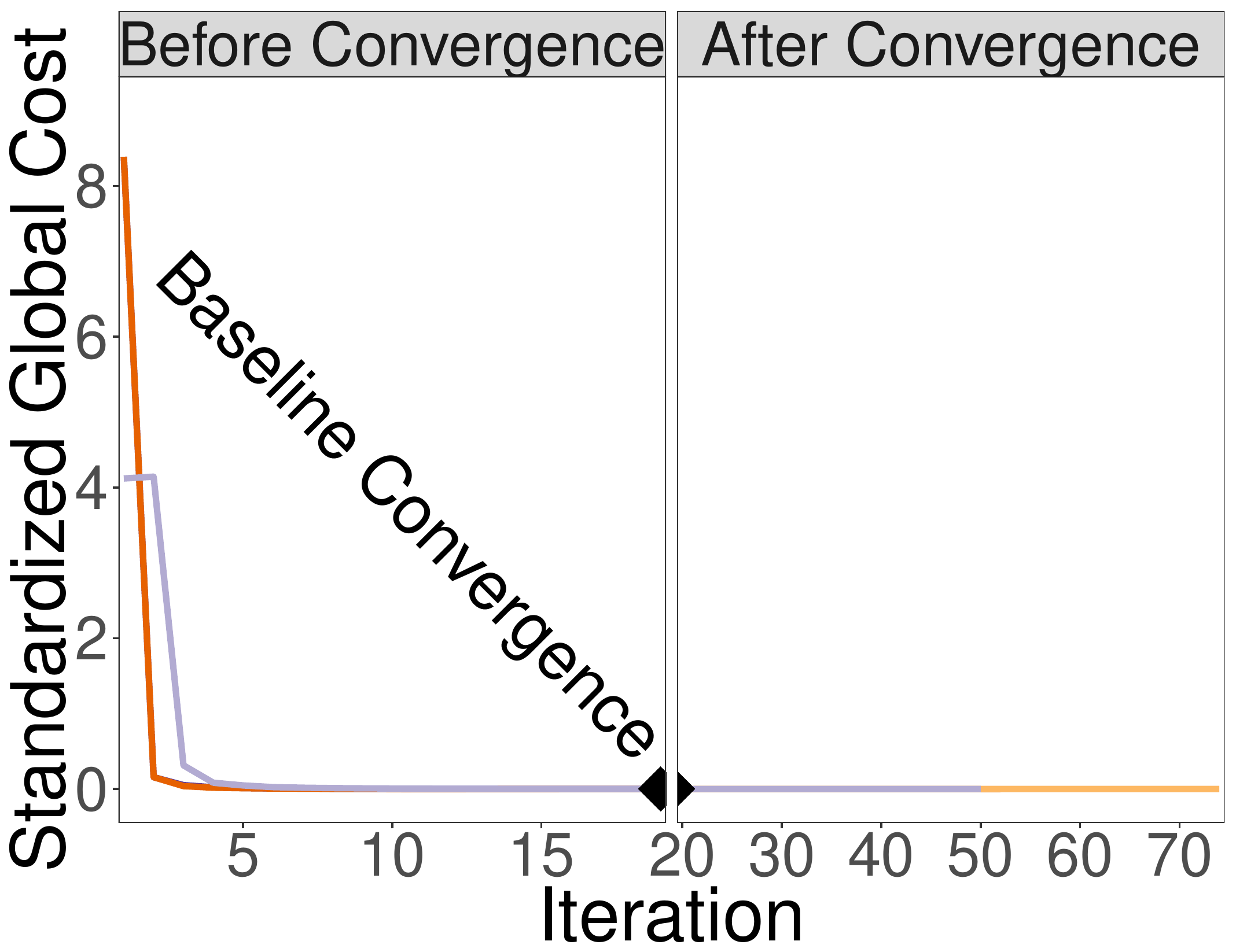}}
\subfigure[Electric vehicles, partial scale, $\lambda=0$, $c=5$]{\includegraphics[width=0.244\textwidth]{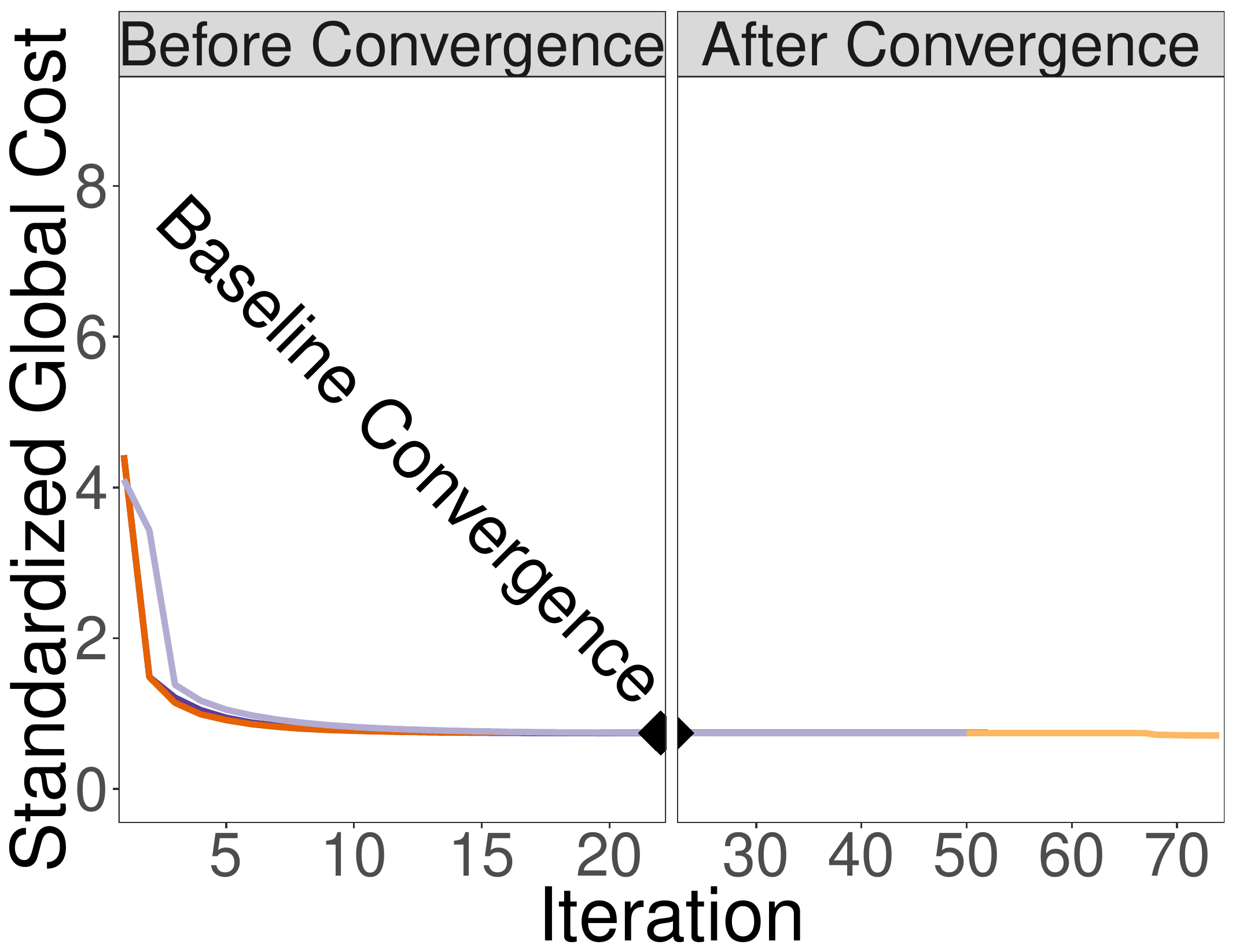}}
\caption{Learning curves for comparison with Figure~\ref{fig:learning-curves-partial-lambda-zero-c-two}. \emph{Dimensions}: holarchic schemes, application scenarios, full versus partial scale, different $\lambda$ values, varying number of children.}\label{fig:learning-curves-comparisons}
\end{figure}

Figure~\ref{fig:learning-curves-comparisons} compares with Figure~\ref{fig:learning-curves-partial-lambda-zero-c-two} by varying partial scale to full (Figure~\ref{fig:learning-curves-comparisons}a-\ref{fig:learning-curves-comparisons}d), $\lambda=0$ to $\lambda=0.5$ (Figure~\ref{fig:learning-curves-comparisons}e-\ref{fig:learning-curves-comparisons}h) and $c=0$ to $c=5$ (Figure~\ref{fig:learning-curves-comparisons}i-\ref{fig:learning-curves-comparisons}l). 

%\begin{figure}[!htb]
%\centering
%\subfigure[Synthetic]{\includegraphics[width=0.244\textwidth]{{IMPROVEMENT-INDEX-GAUSSIAN-CHILDREN-2-FULL}.pdf}}
%\subfigure[Bike sharing]{\includegraphics[width=0.244\textwidth]{{IMPROVEMENT-INDEX-BICYCLE-CHILDREN-2-FULL}.pdf}}
%\subfigure[Energy demand]{\includegraphics[width=0.244\textwidth]{{IMPROVEMENT-INDEX-ENERGY-CHILDREN-2-FULL}.pdf}}
%\subfigure[Electric vehicles]{\includegraphics[width=0.244\textwidth]{{IMPROVEMENT-INDEX-EV-CHILDREN-2-FULL}.pdf}}
%\caption{Improvement index. \emph{Dimensions}: holarchic schemes, application scenarios, different $\lambda$ values. \emph{Settings}: full scale, $c=2$.}\label{fig:improvement-index-lambda-full}
%\end{figure}

\begin{figure}[!htb]
\centering
\subfigure[Synthetic, holarchic initialization]{\includegraphics[width=0.244\textwidth]{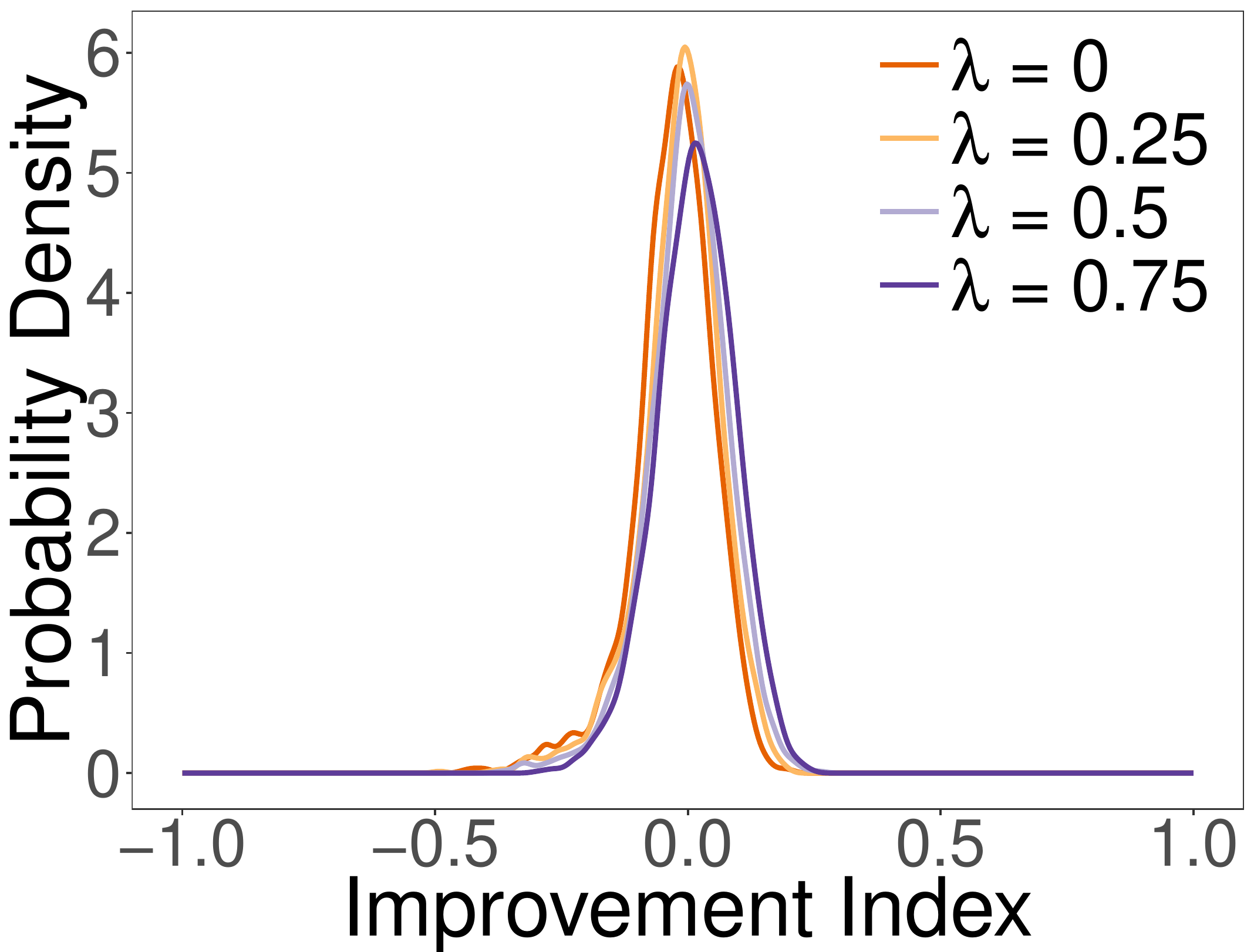}}
\subfigure[Bike sharing, holarchic initialization]{\includegraphics[width=0.244\textwidth]{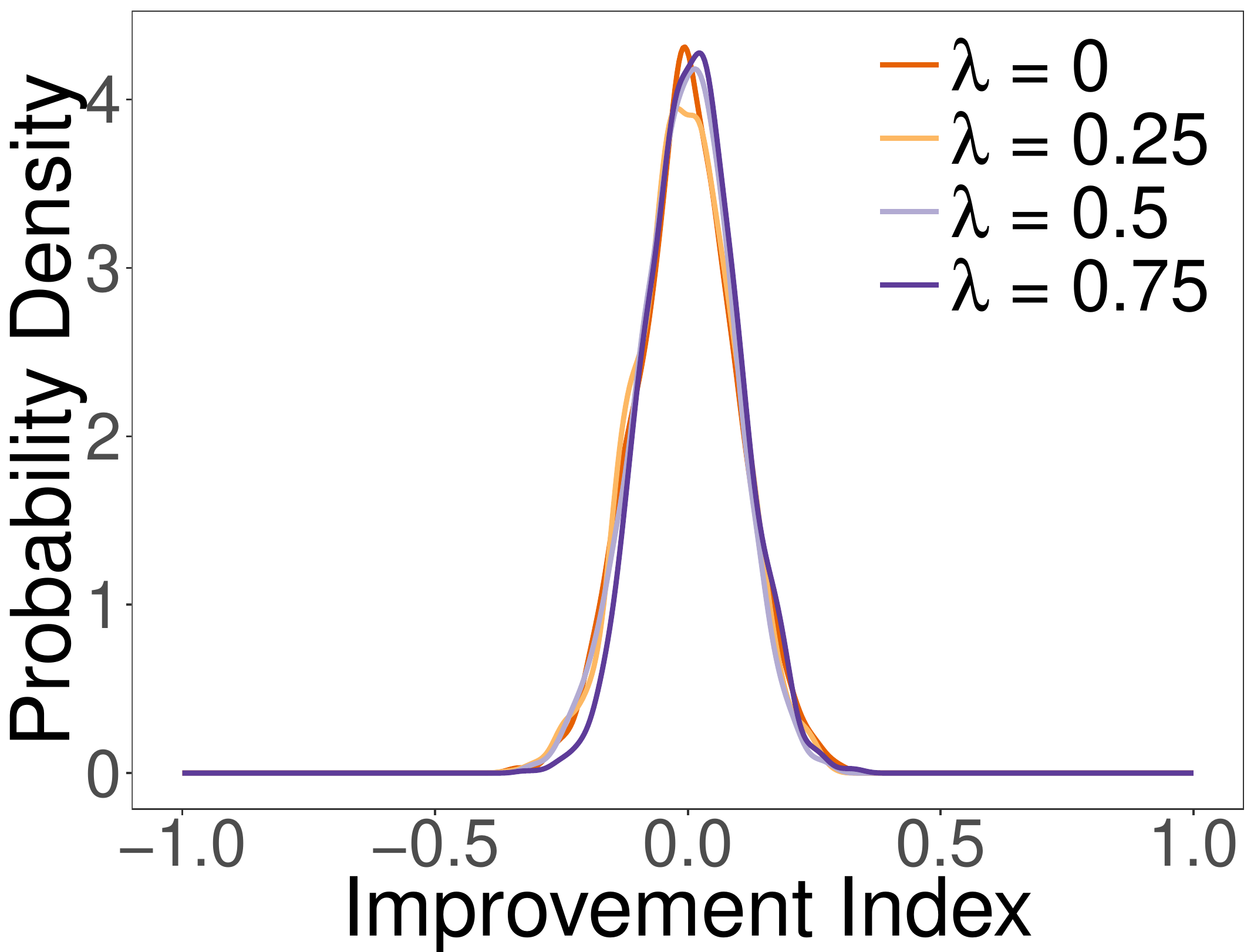}}
\subfigure[Energy demand, holarchic initialization]{\includegraphics[width=0.244\textwidth]{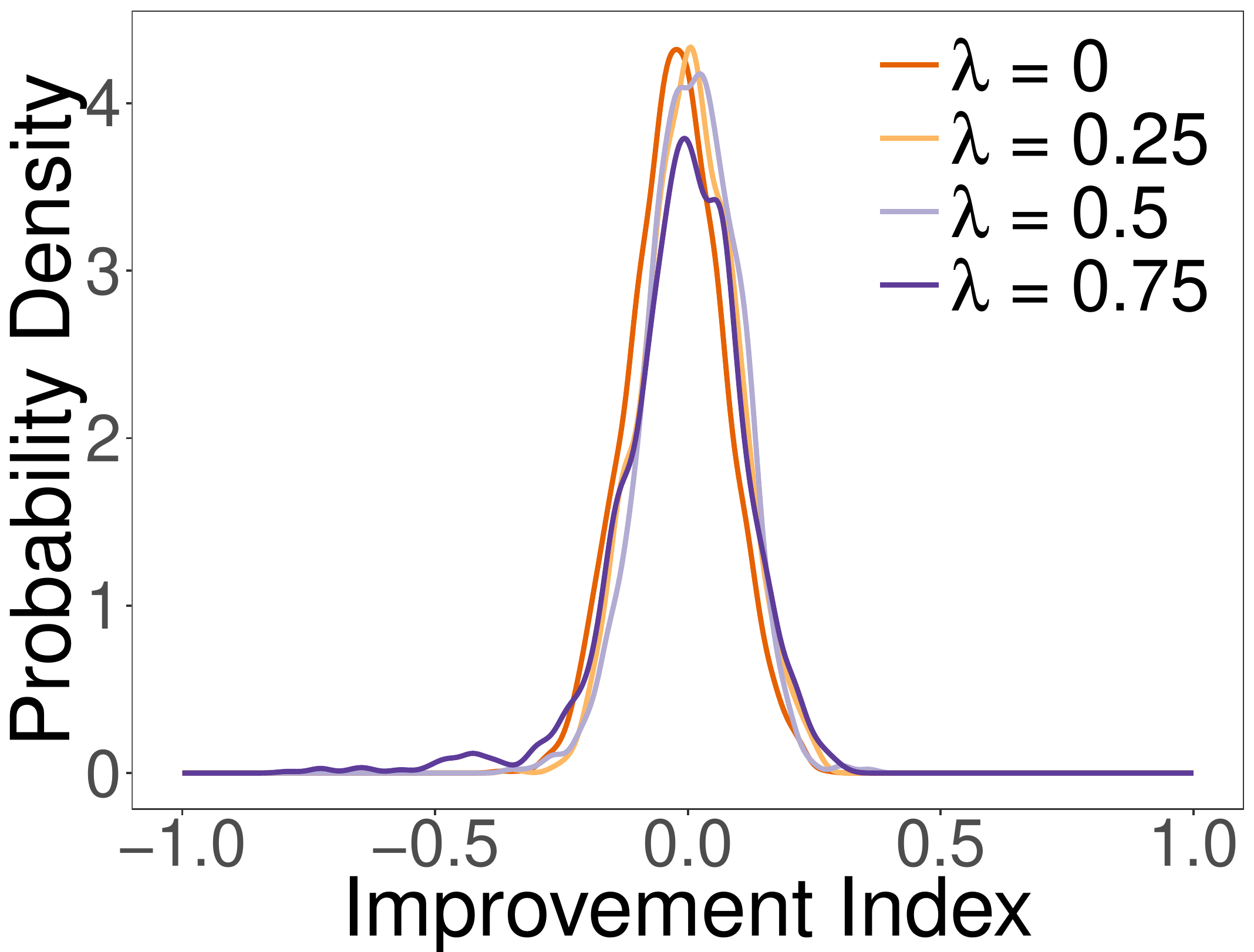}}
\subfigure[Electric vehicles, holarchic initialization]{\includegraphics[width=0.244\textwidth]{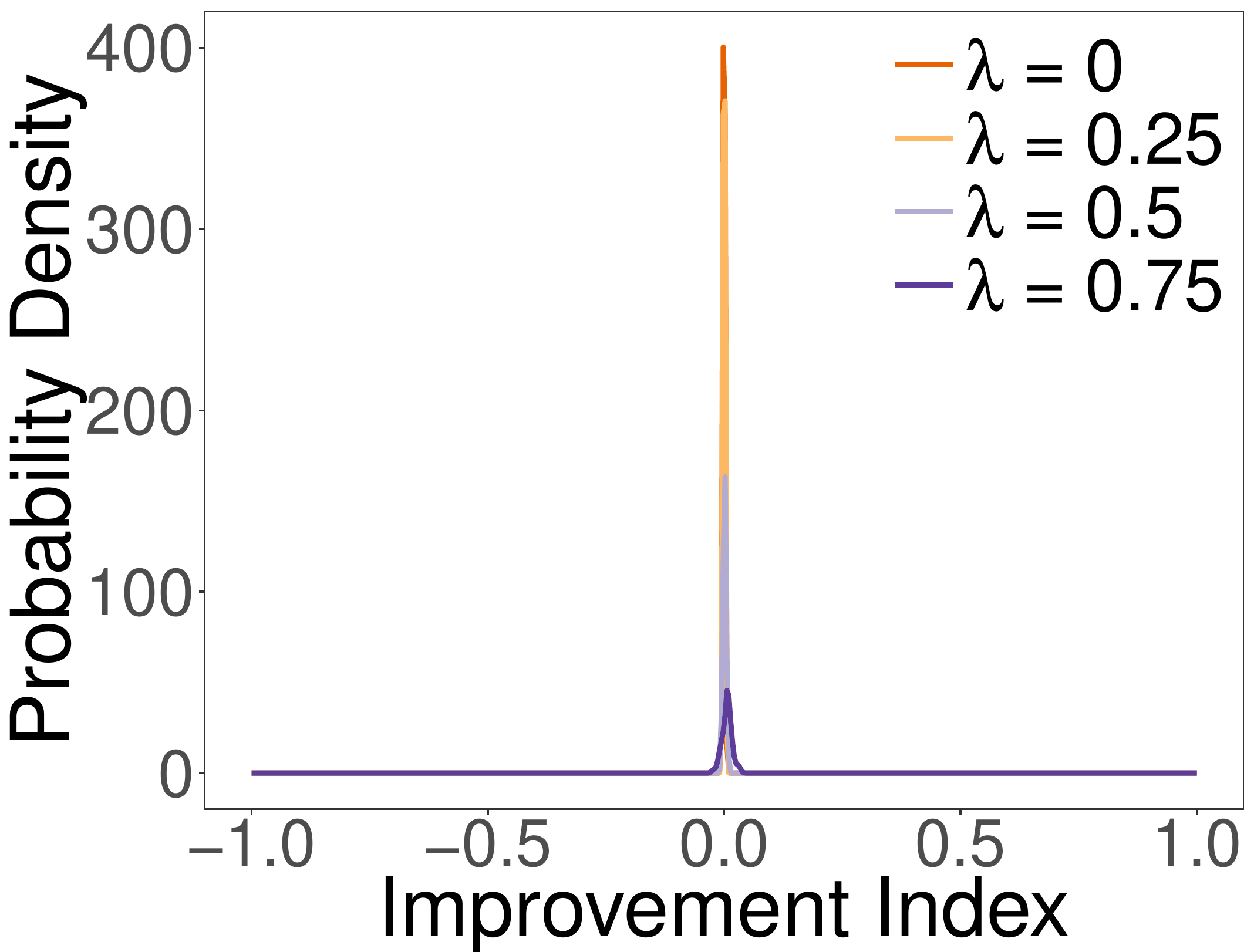}}
\subfigure[Synthetic, holarchic runtime]{\includegraphics[width=0.244\textwidth]{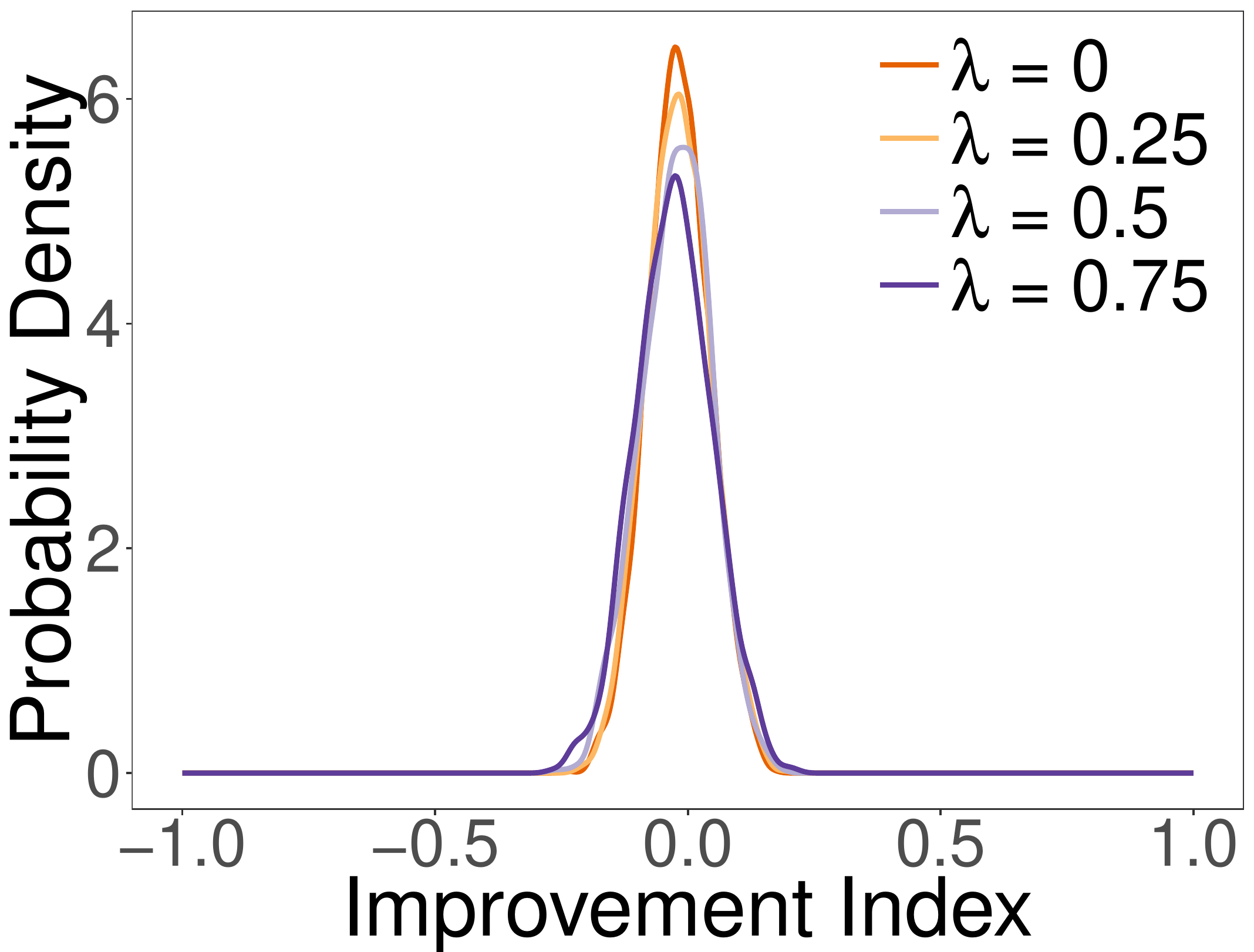}}
\subfigure[Bike sharing, holarchic runtime]{\includegraphics[width=0.244\textwidth]{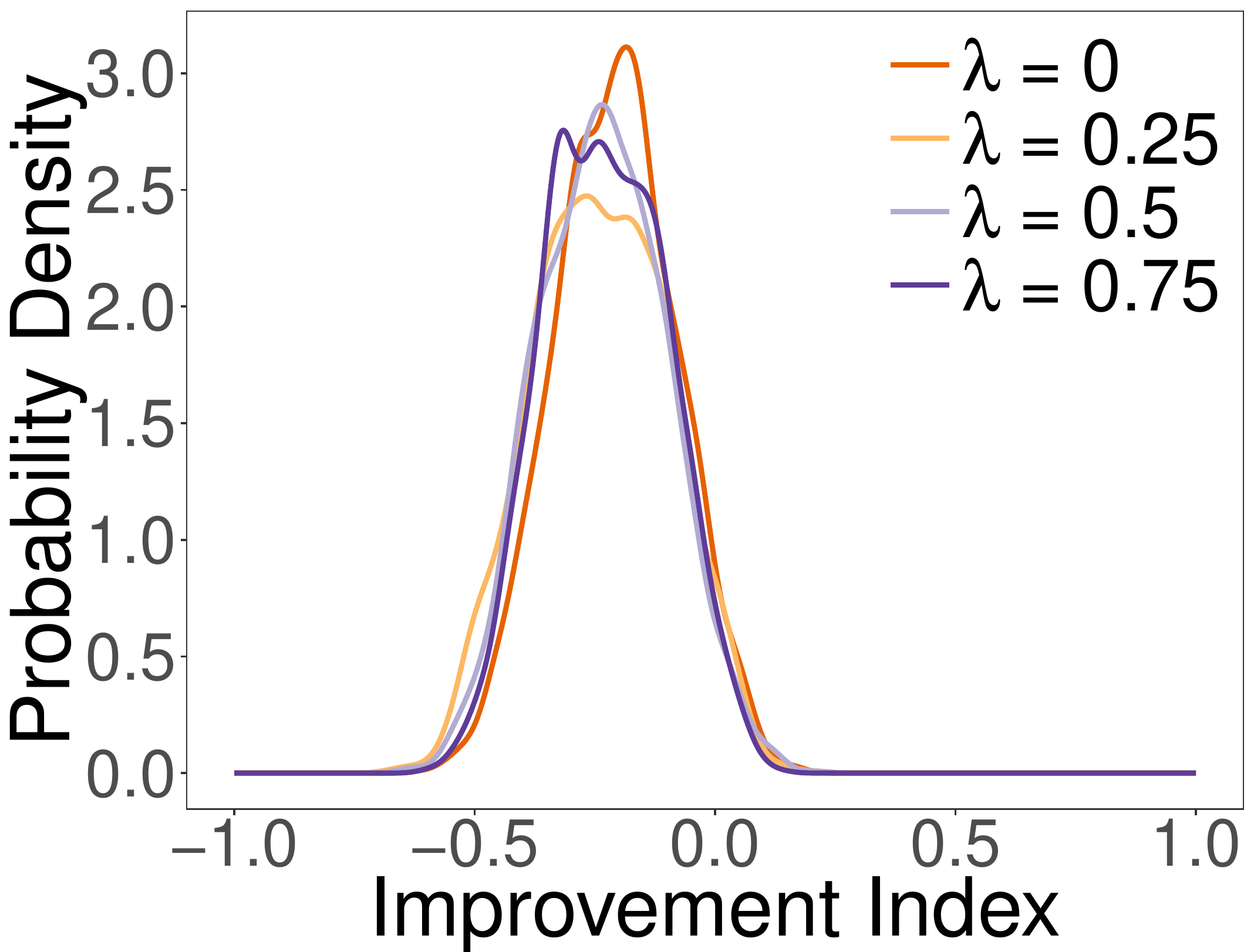}}
\subfigure[Energy demand, holarchic runtime]{\includegraphics[width=0.244\textwidth]{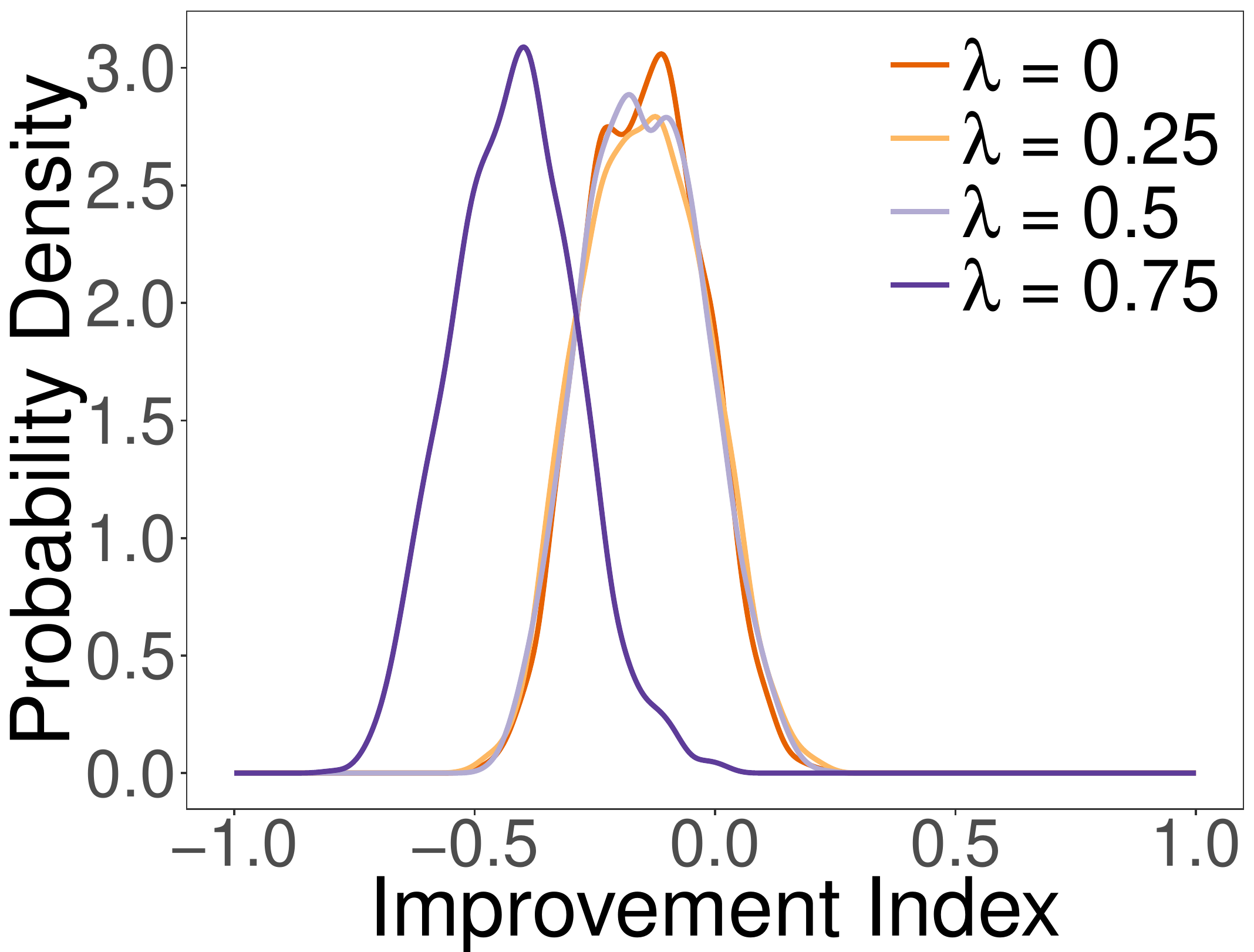}}
\subfigure[Electric vehicles, holarchic runtime]{\includegraphics[width=0.244\textwidth]{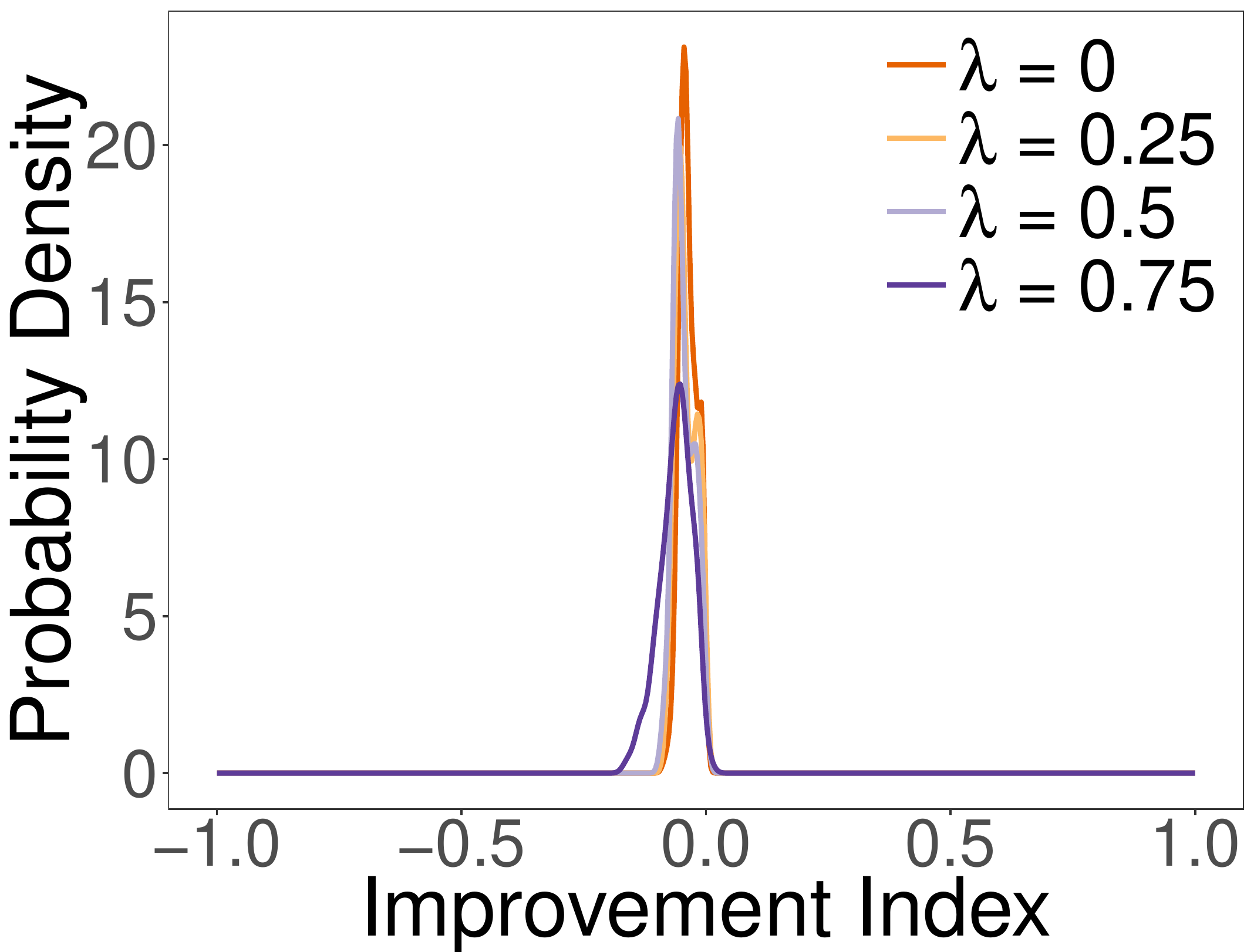}}
\subfigure[Synthetic, holarchic termination]{\includegraphics[width=0.244\textwidth]{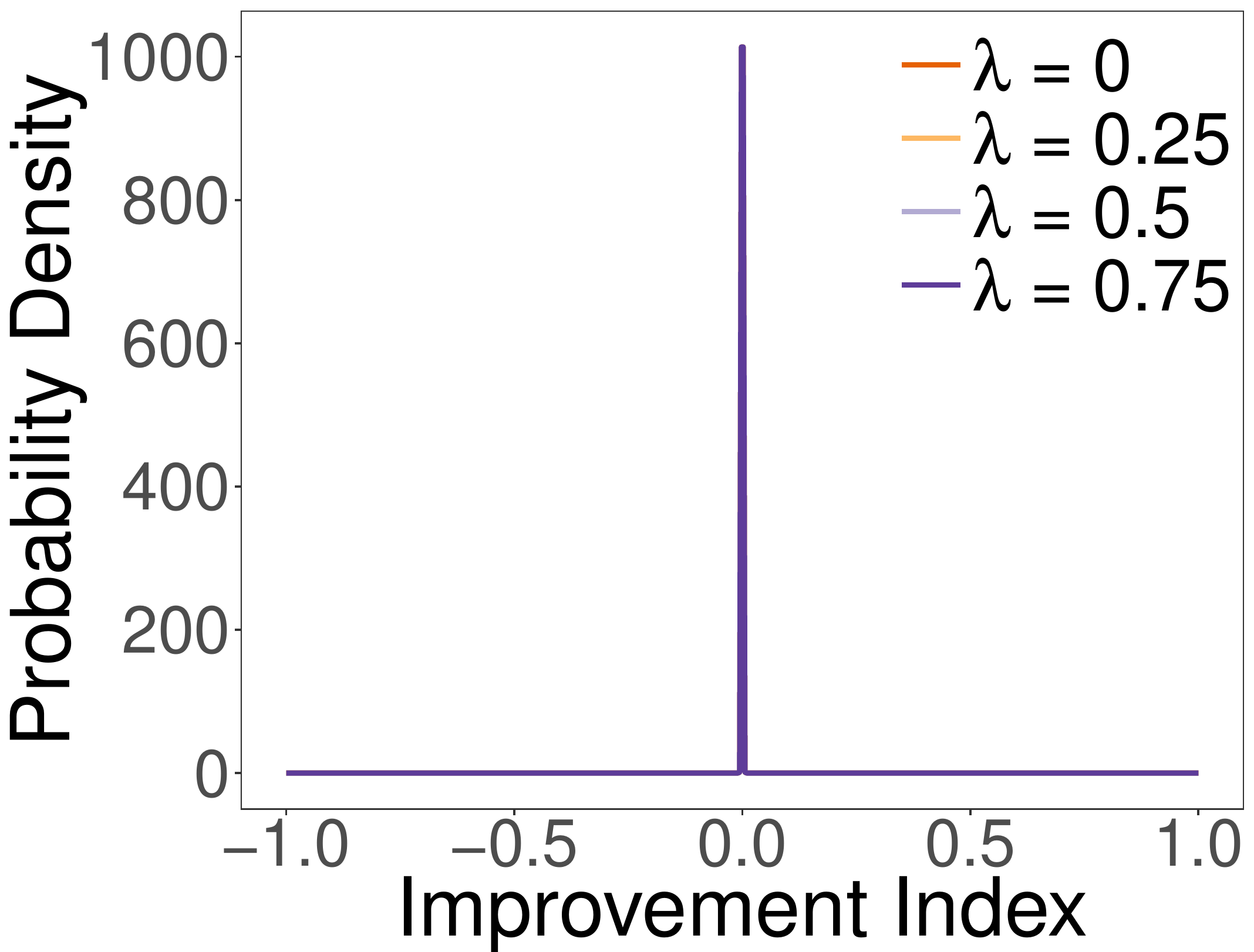}}
\subfigure[Bike sharing, holarchic termination]{\includegraphics[width=0.244\textwidth]{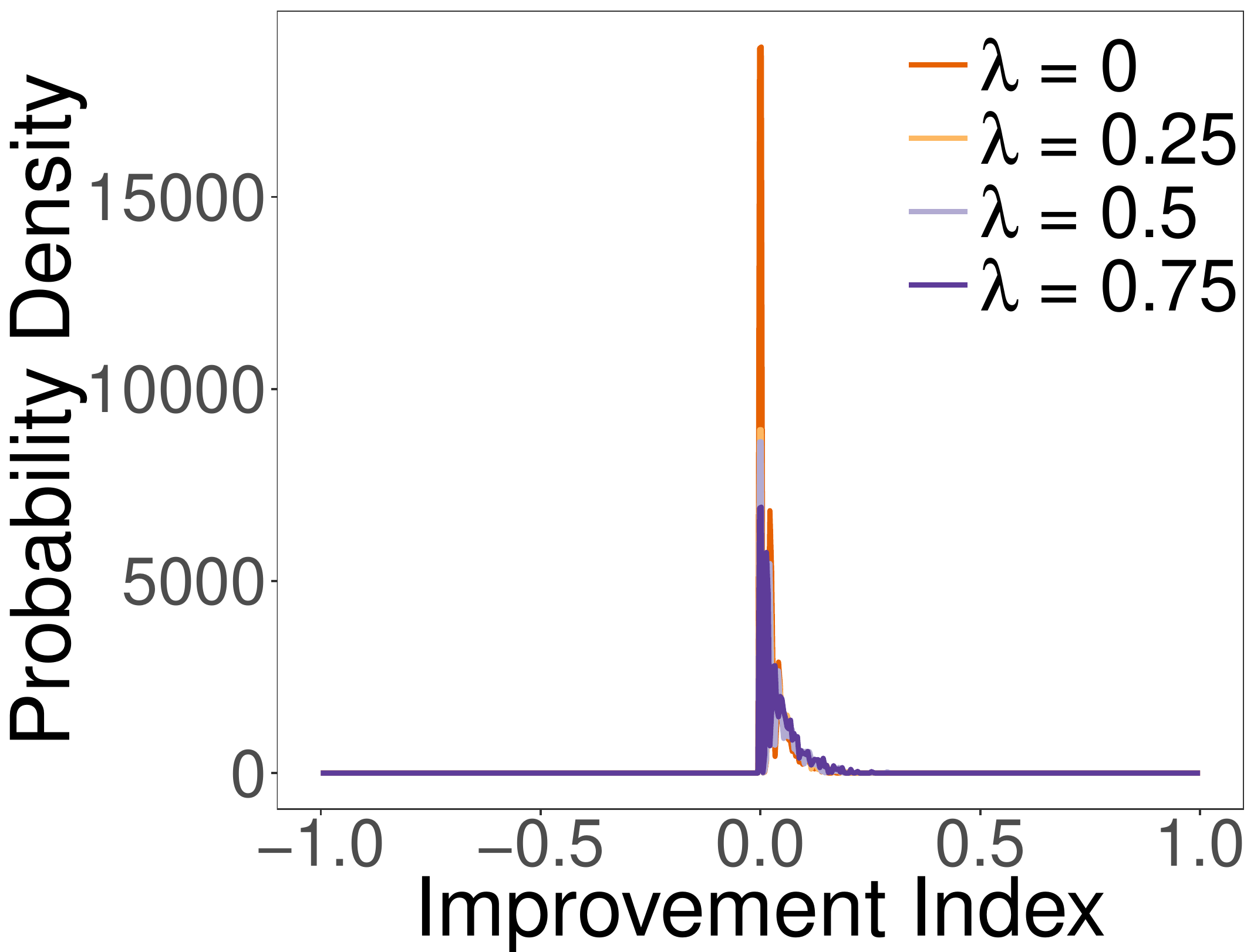}}
\subfigure[Energy demand, holarchic termination]{\includegraphics[width=0.244\textwidth]{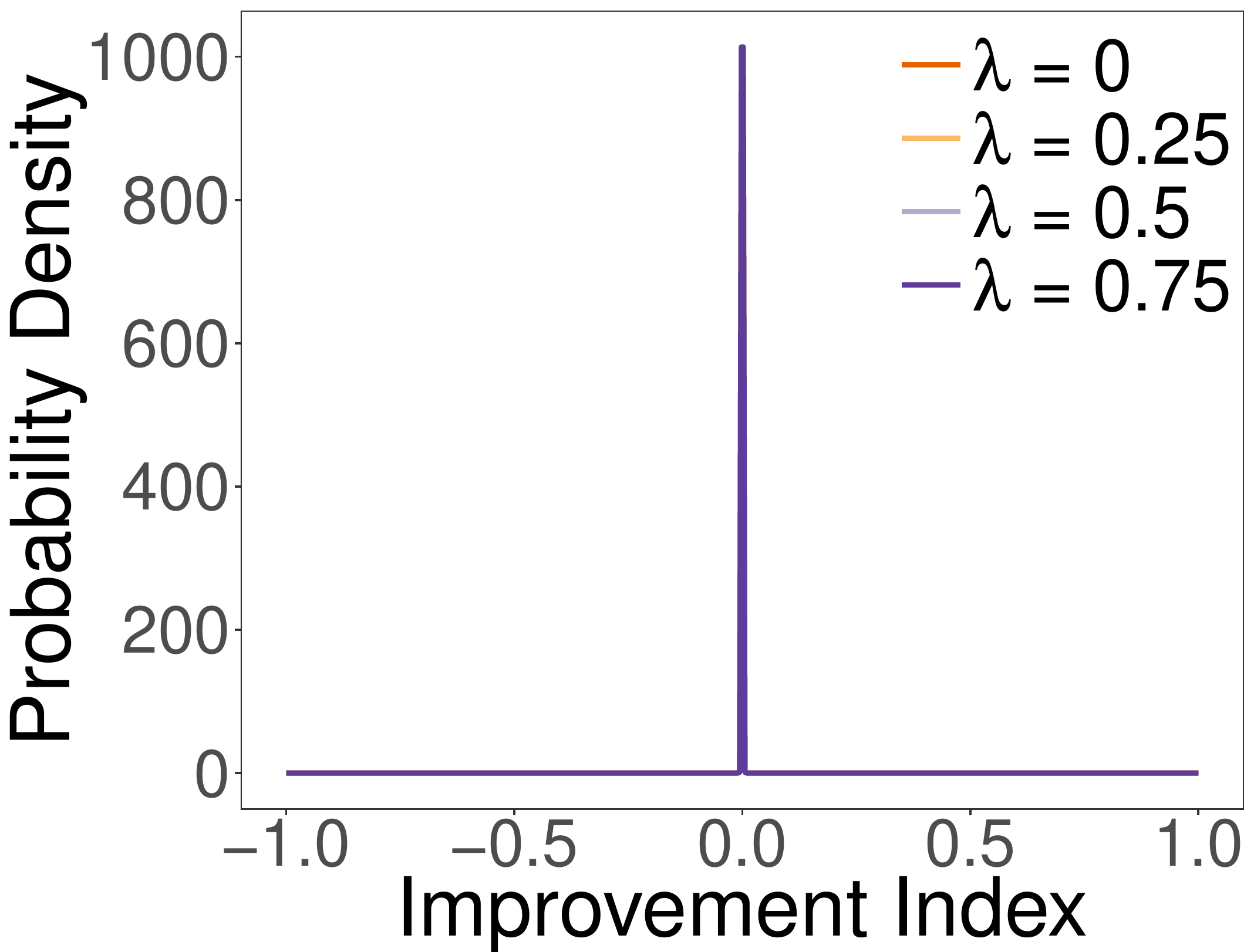}}
\subfigure[Electric vehicles, holarchic termination]{\includegraphics[width=0.244\textwidth]{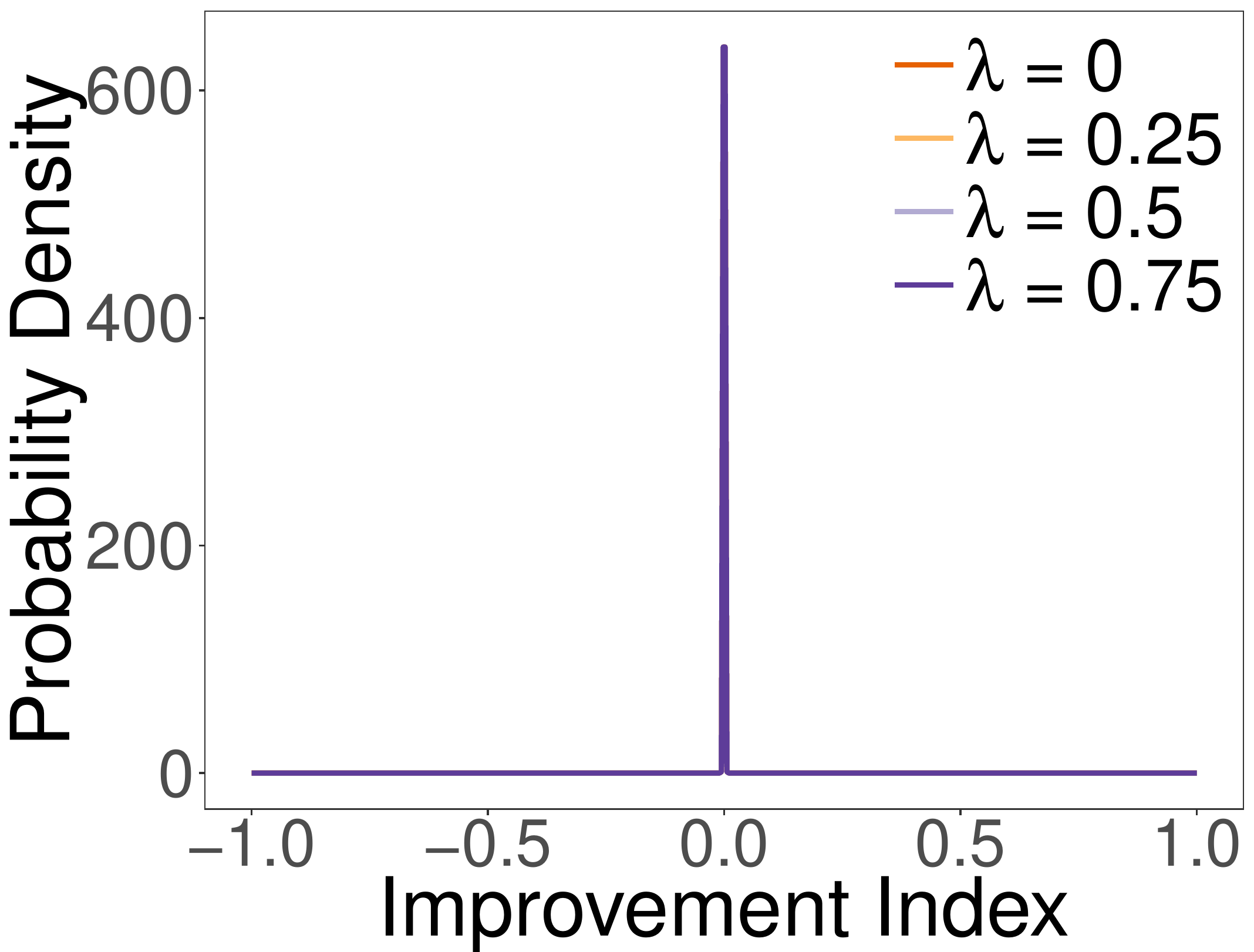}}
\caption{Probability density of the improvement index that elaborates on Figure~\ref{fig:improvement-index-lambda-partial}. \emph{Dimensions}: holarchic schemes, application scenarios, different $\lambda$ values. \emph{Settings}: partial scale, $c=2$.}\label{fig:density-improvement-index-lambda-partial}
\end{figure}

Figure~\ref{fig:density-improvement-index-lambda-partial} elaborates on the Figure~\ref{fig:improvement-index-lambda-partial}. It illustrates the probability density function of the improvement index by fixing the holarchic scale to partial, $c=2$ and varying all other dimensions. 

\begin{figure}[!htb]
\centering
\subfigure[Synthetic, holarchic initialization]{\includegraphics[width=0.244\textwidth]{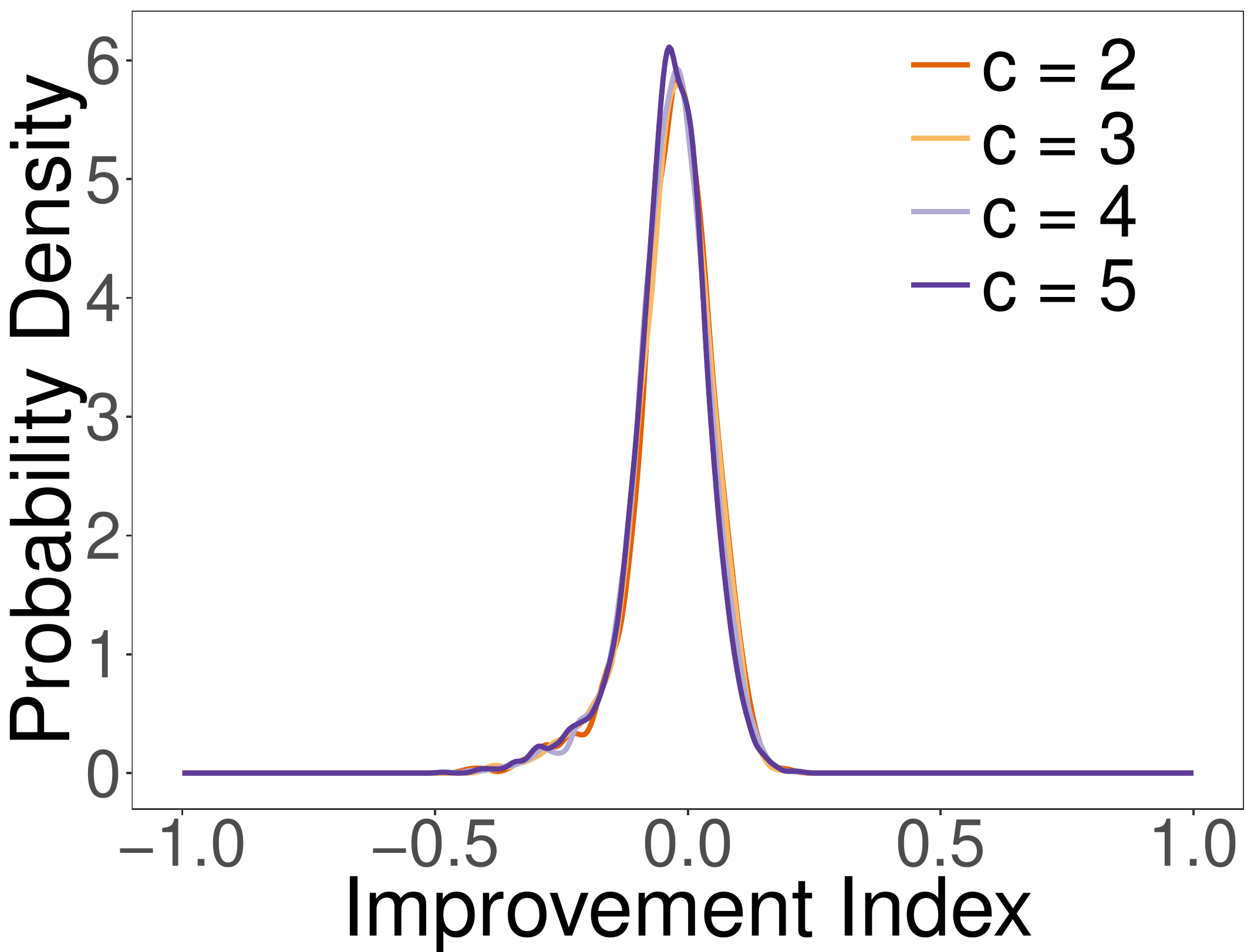}}
\subfigure[Bike sharing, holarchic initialization]{\includegraphics[width=0.244\textwidth]{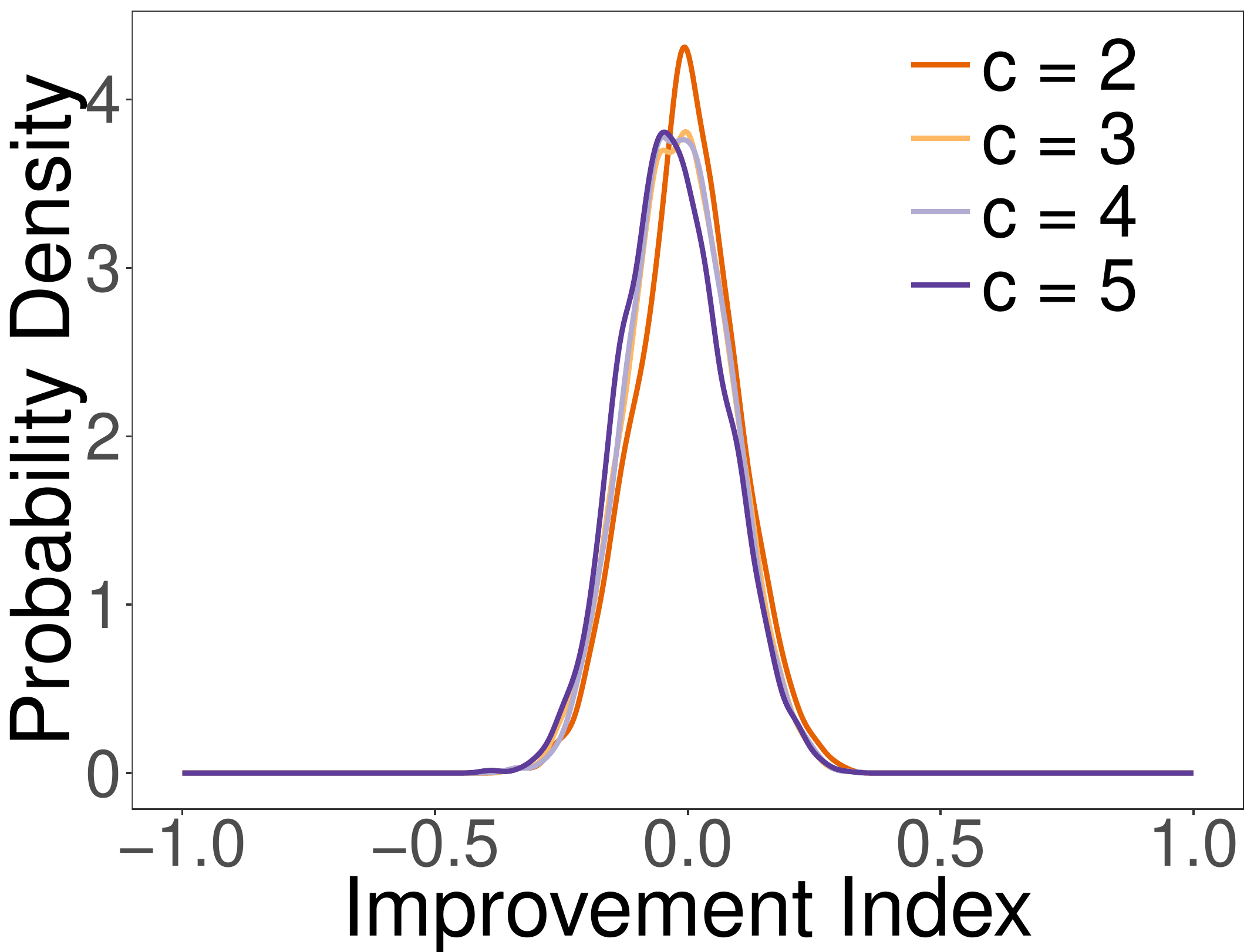}}
\subfigure[Energy demand, holarchic initialization]{\includegraphics[width=0.244\textwidth]{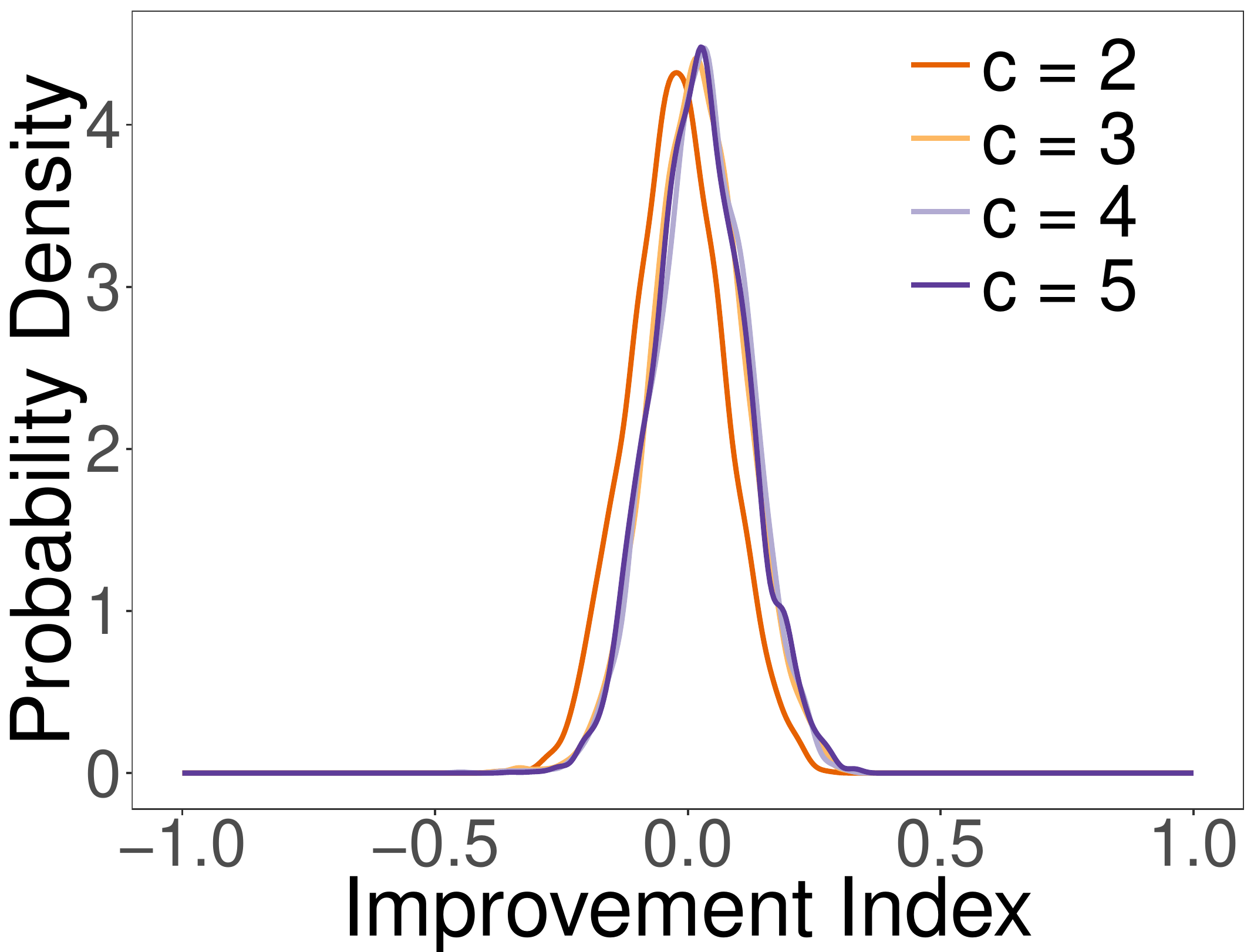}}
\subfigure[Electric vehicles, holarchic initialization]{\includegraphics[width=0.244\textwidth]{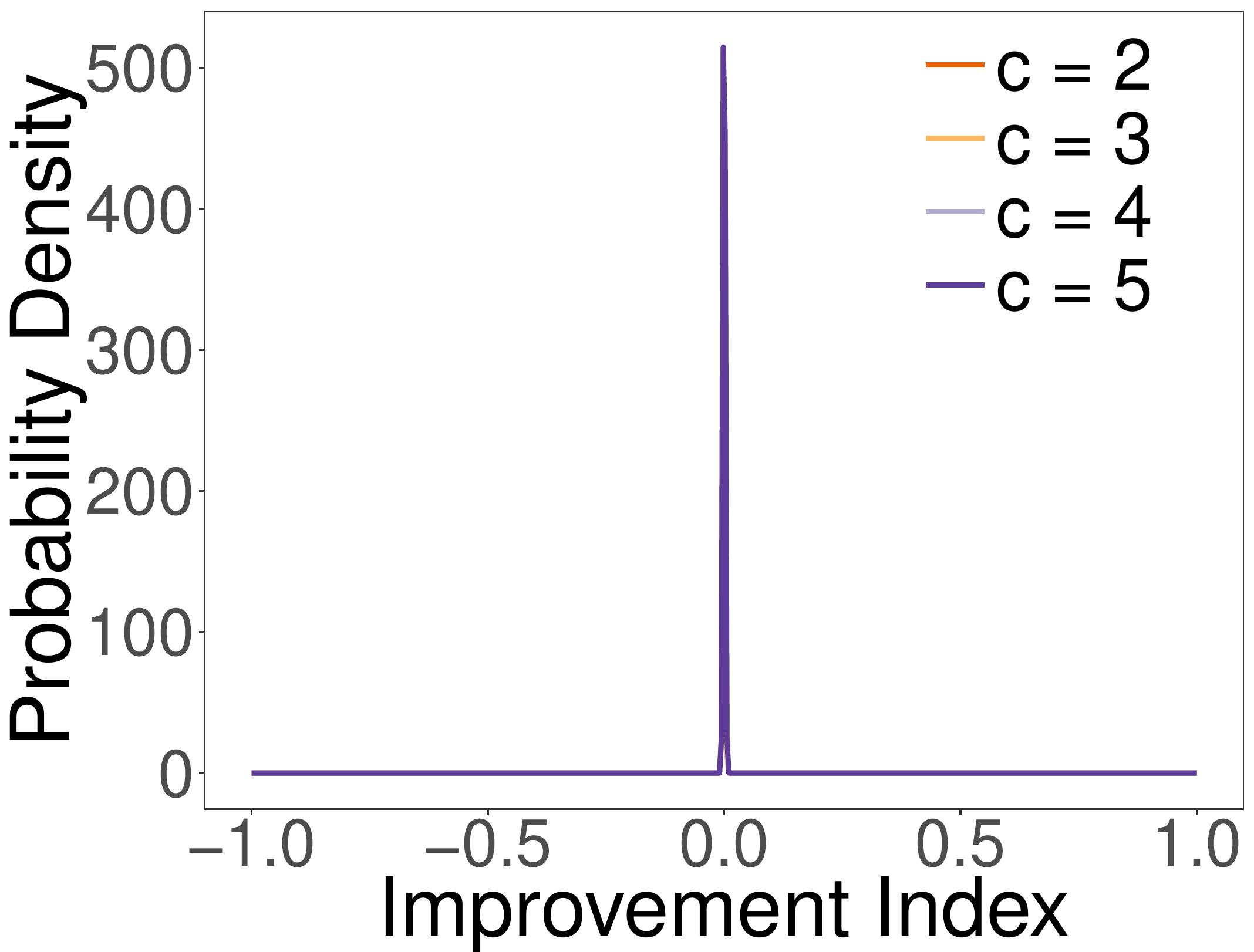}}
\subfigure[Synthetic, holarchic runtime]{\includegraphics[width=0.244\textwidth]{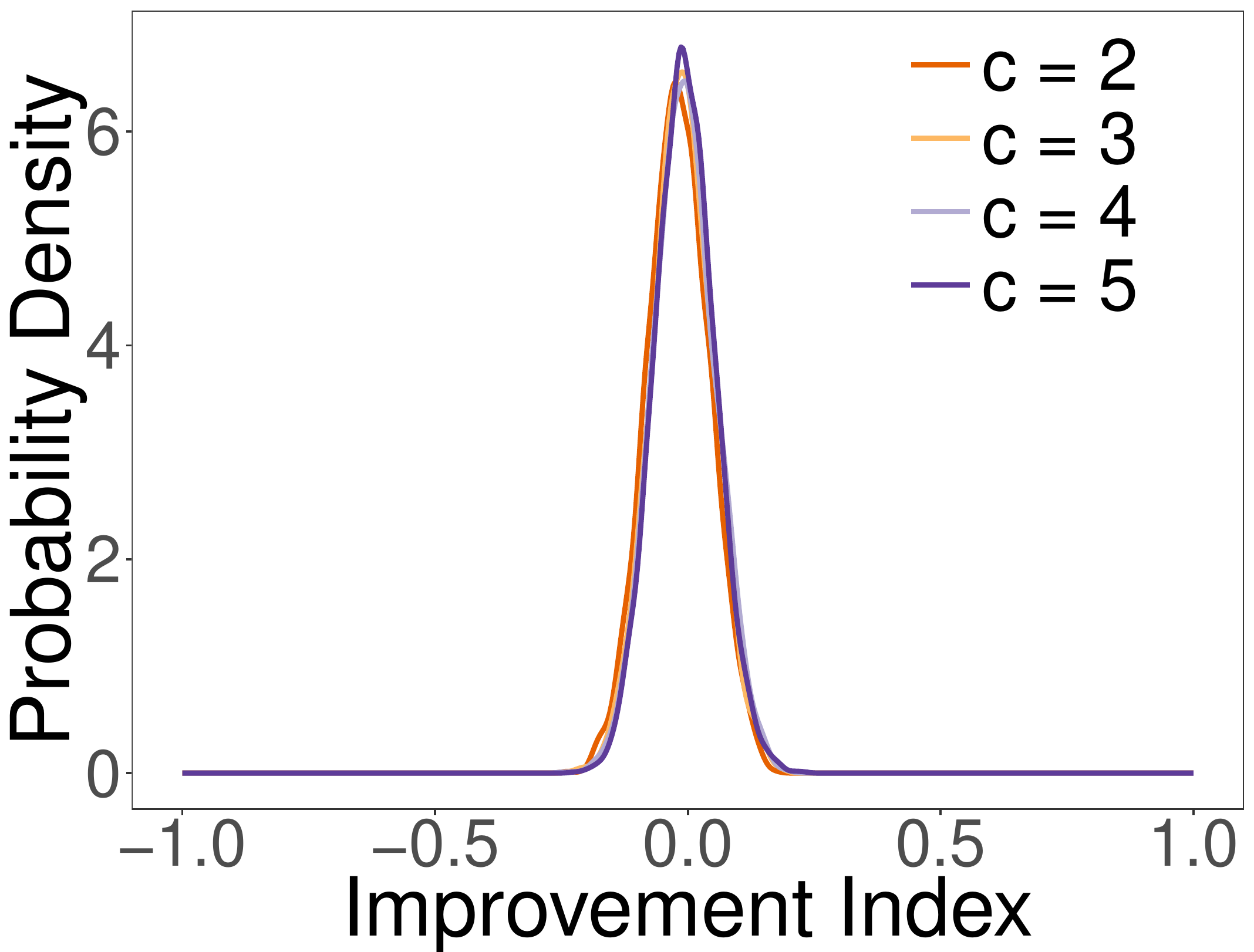}}
\subfigure[Bike sharing, holarchic runtime]{\includegraphics[width=0.244\textwidth]{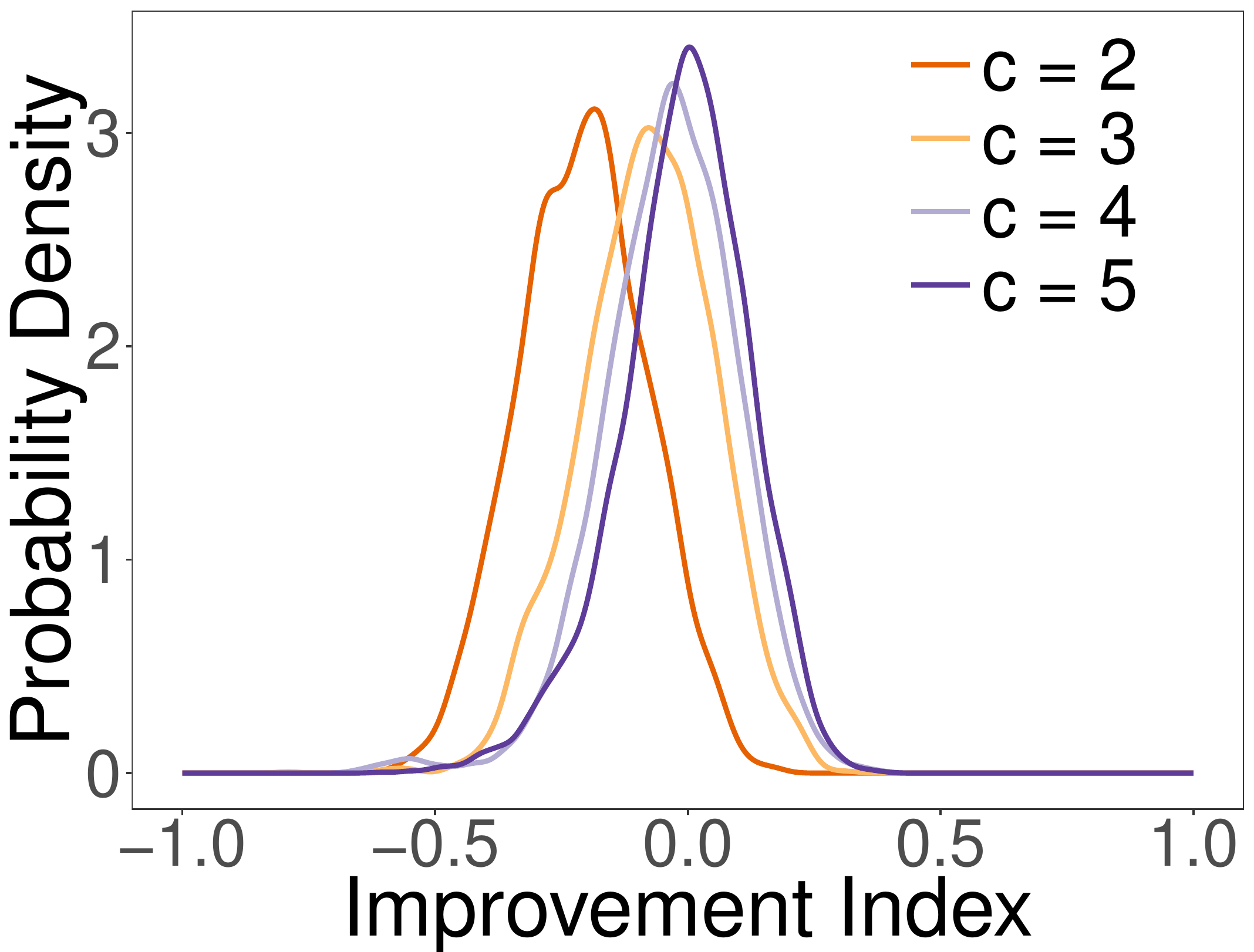}}
\subfigure[Energy demand, holarchic runtime]{\includegraphics[width=0.244\textwidth]{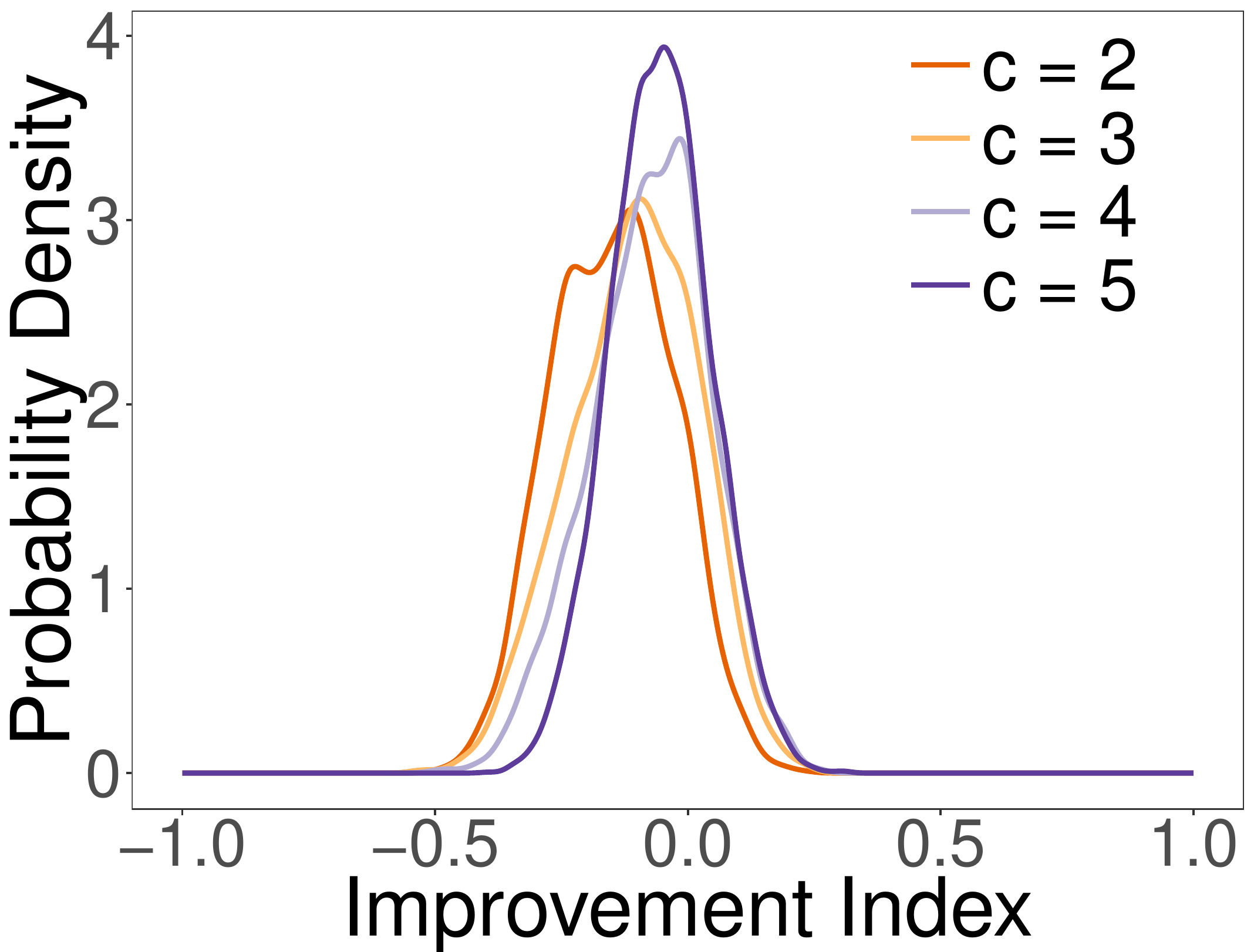}}
\subfigure[Electric vehicles, holarchic runtime]{\includegraphics[width=0.244\textwidth]{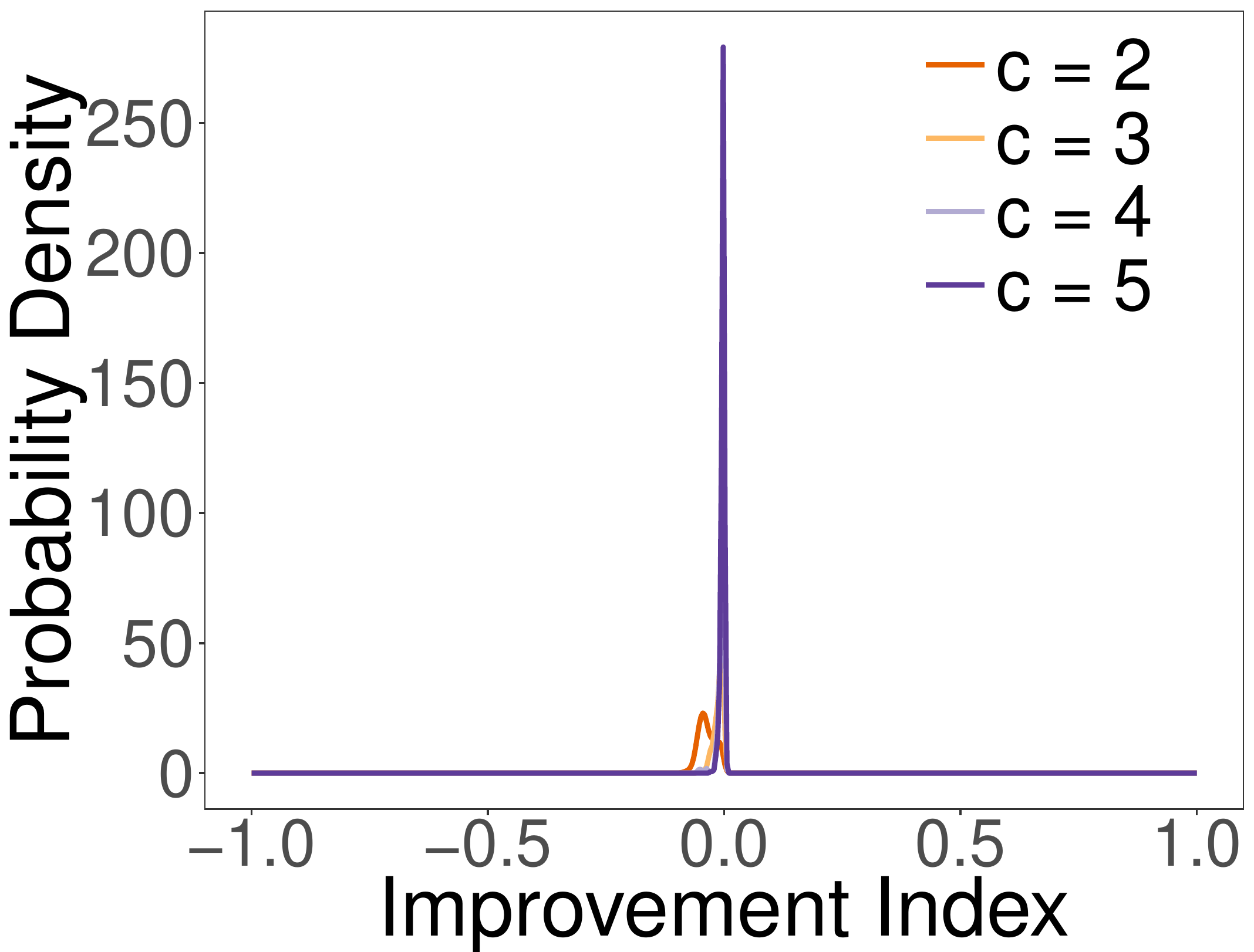}}
\subfigure[Synthetic, holarchic termination]{\includegraphics[width=0.244\textwidth]{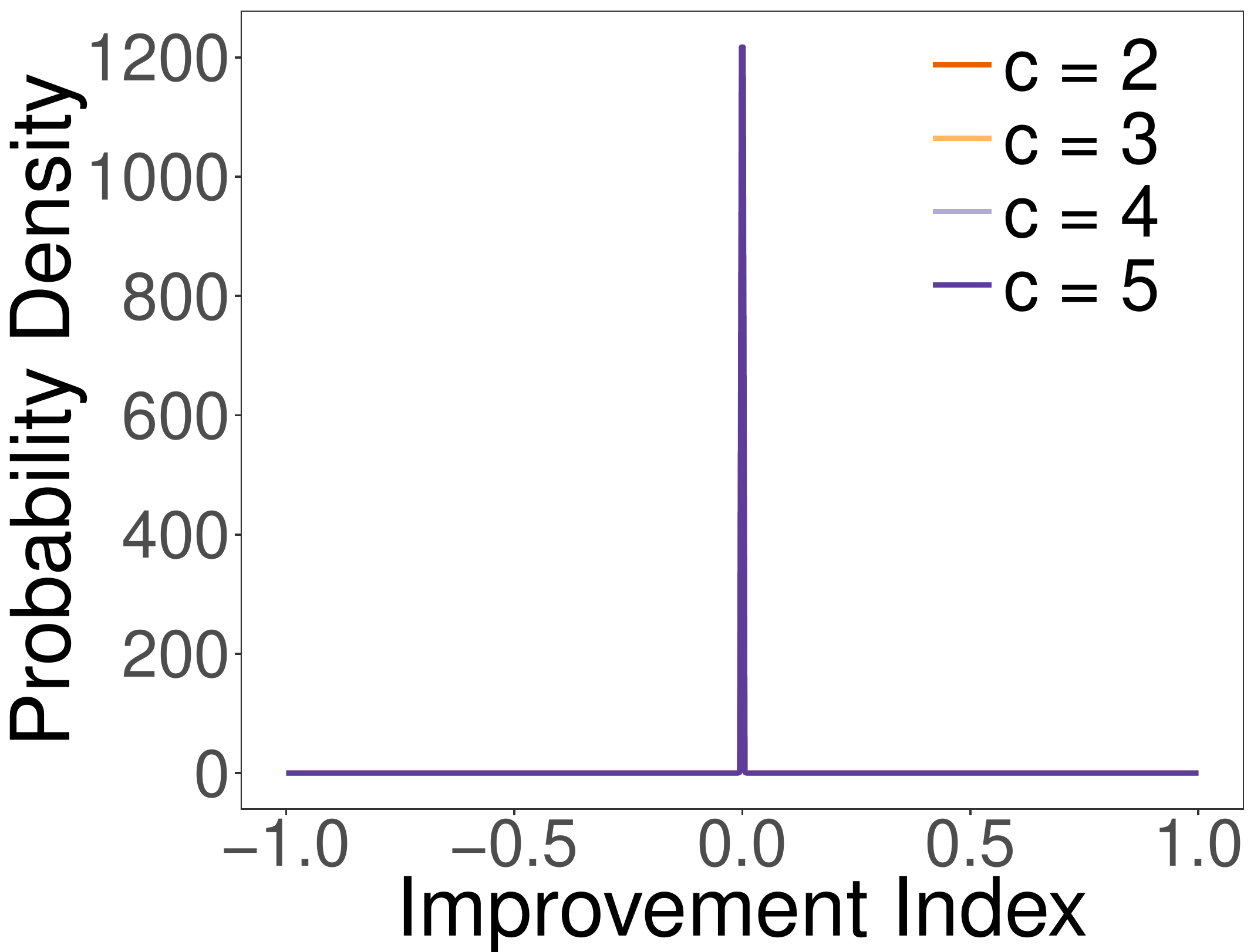}}
\subfigure[Bike sharing, holarchic termination]{\includegraphics[width=0.244\textwidth]{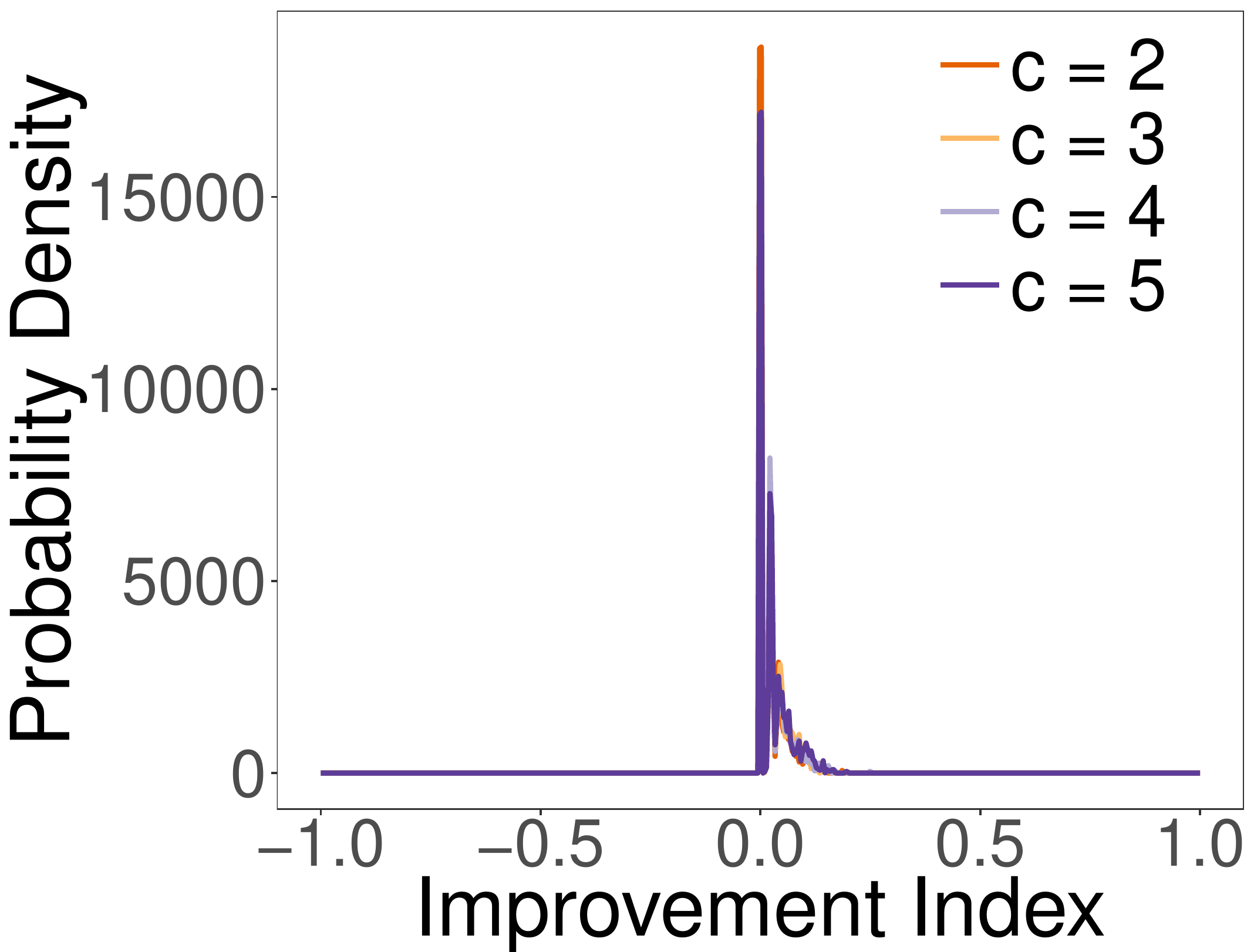}}
\subfigure[Energy demand, holarchic termination]{\includegraphics[width=0.244\textwidth]{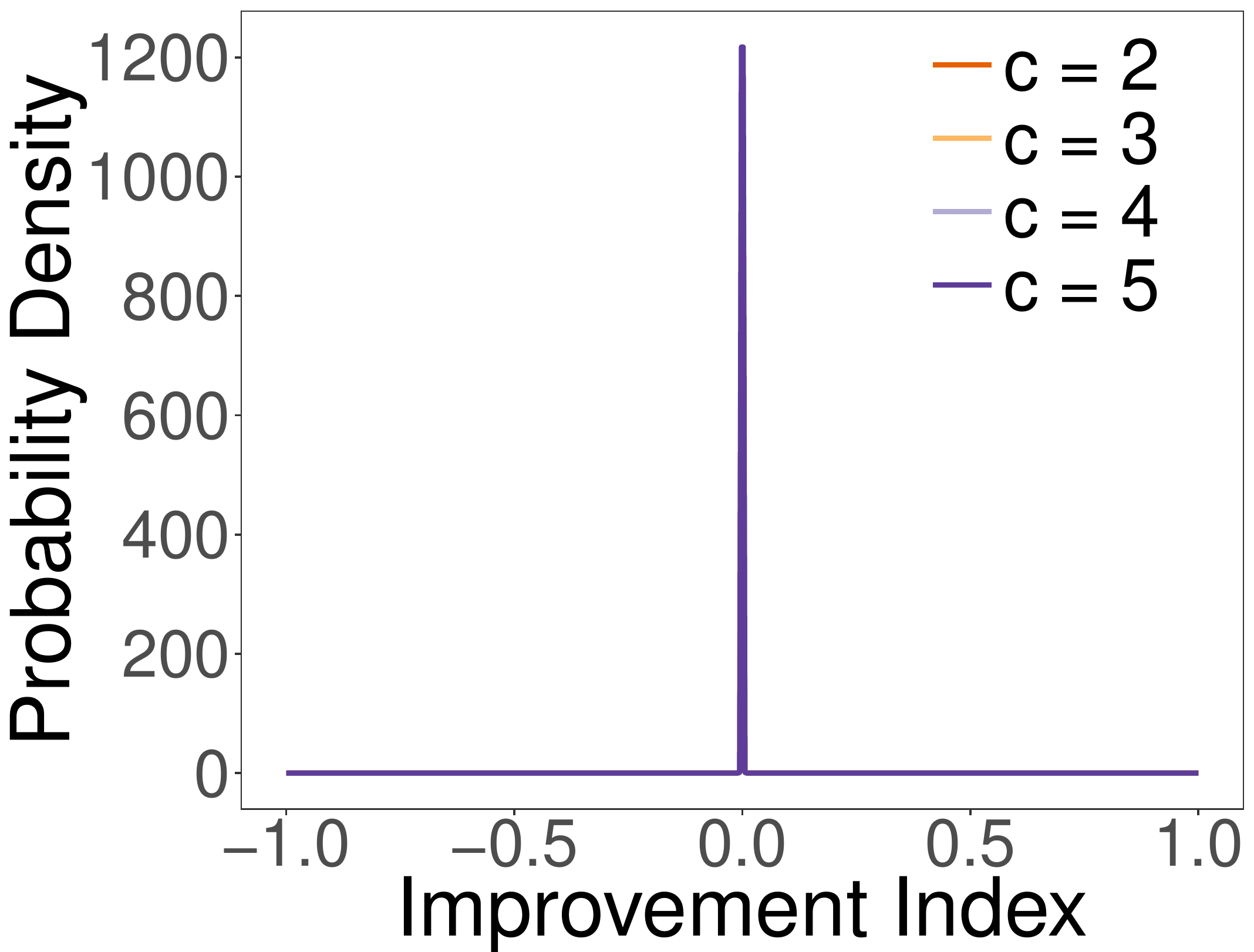}}
\subfigure[Electric vehicles, holarchic termination]{\includegraphics[width=0.244\textwidth]{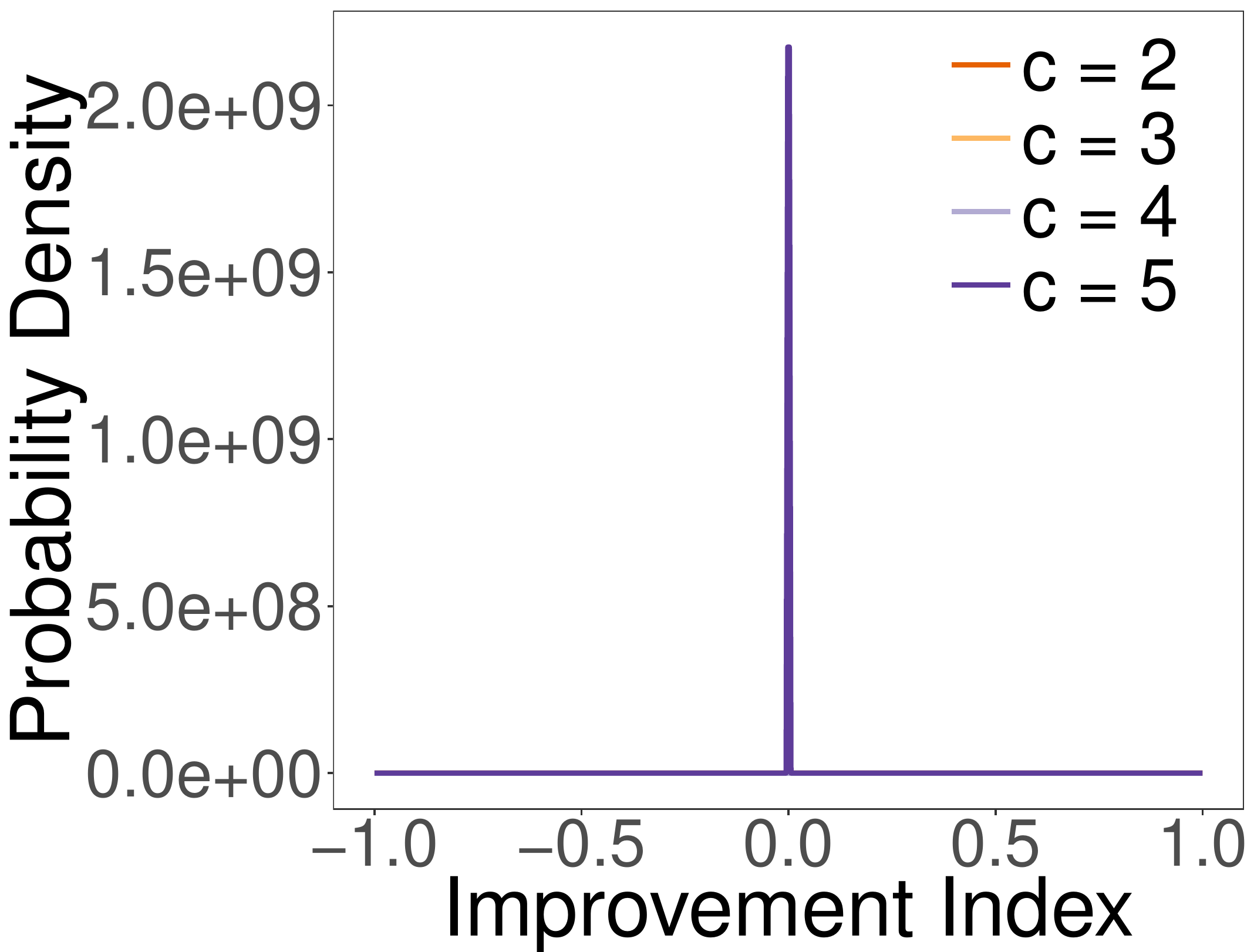}}
\caption{Probability density of the improvement index that elaborates on Figure~\ref{fig:improvement-index-children-partial}. \emph{Dimensions}: holarchic schemes, application scenarios, varying number of children. \emph{Settings}: partial scale, $\lambda=0$.}\label{fig:density-improvement-index-children-partial}
\end{figure}

Figure~\ref{fig:density-improvement-index-children-partial} elaborates on the Figure~\ref{fig:improvement-index-children-partial}. It illustrates the probability density function of the improvement index by fixing the holarchic scale to partial, $\lambda=0$ and varying all other dimensions. 

\begin{figure}[!htb]
\centering
\subfigure[Synthetic, holarchic initialization]{\includegraphics[width=0.244\textwidth]{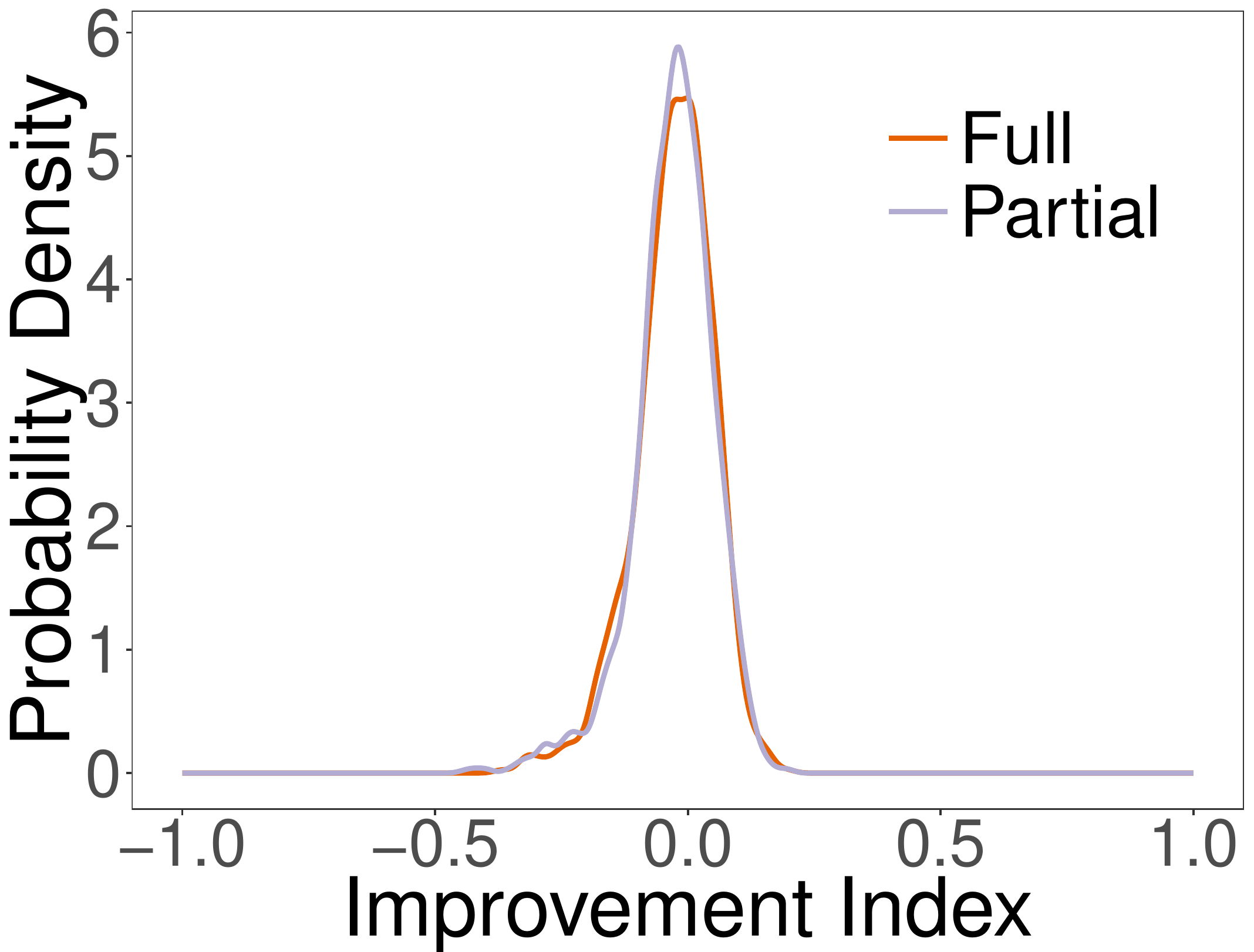}}
\subfigure[Bike sharing, holarchic initialization]{\includegraphics[width=0.244\textwidth]{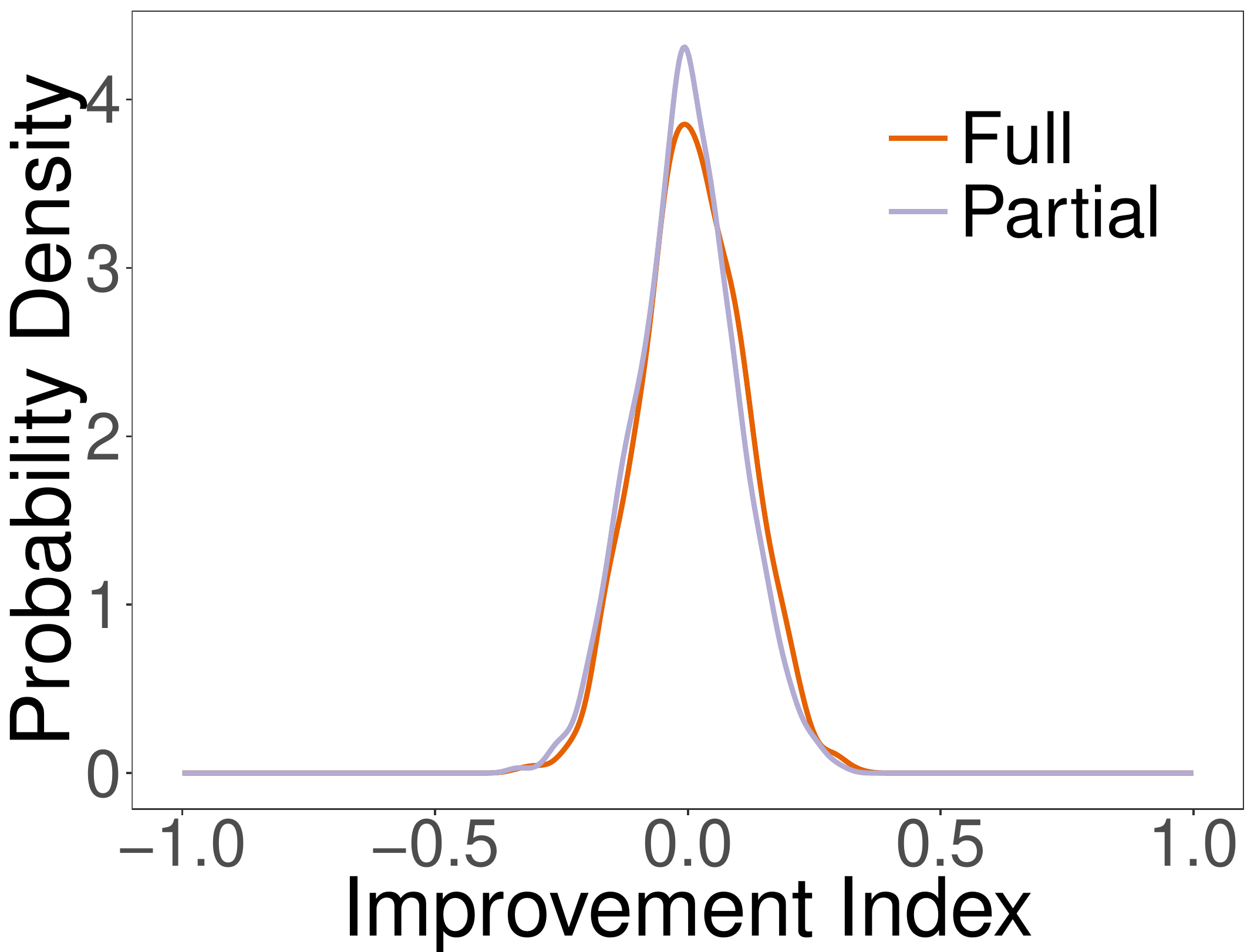}}
\subfigure[Energy demand, holarchic initialization]{\includegraphics[width=0.244\textwidth]{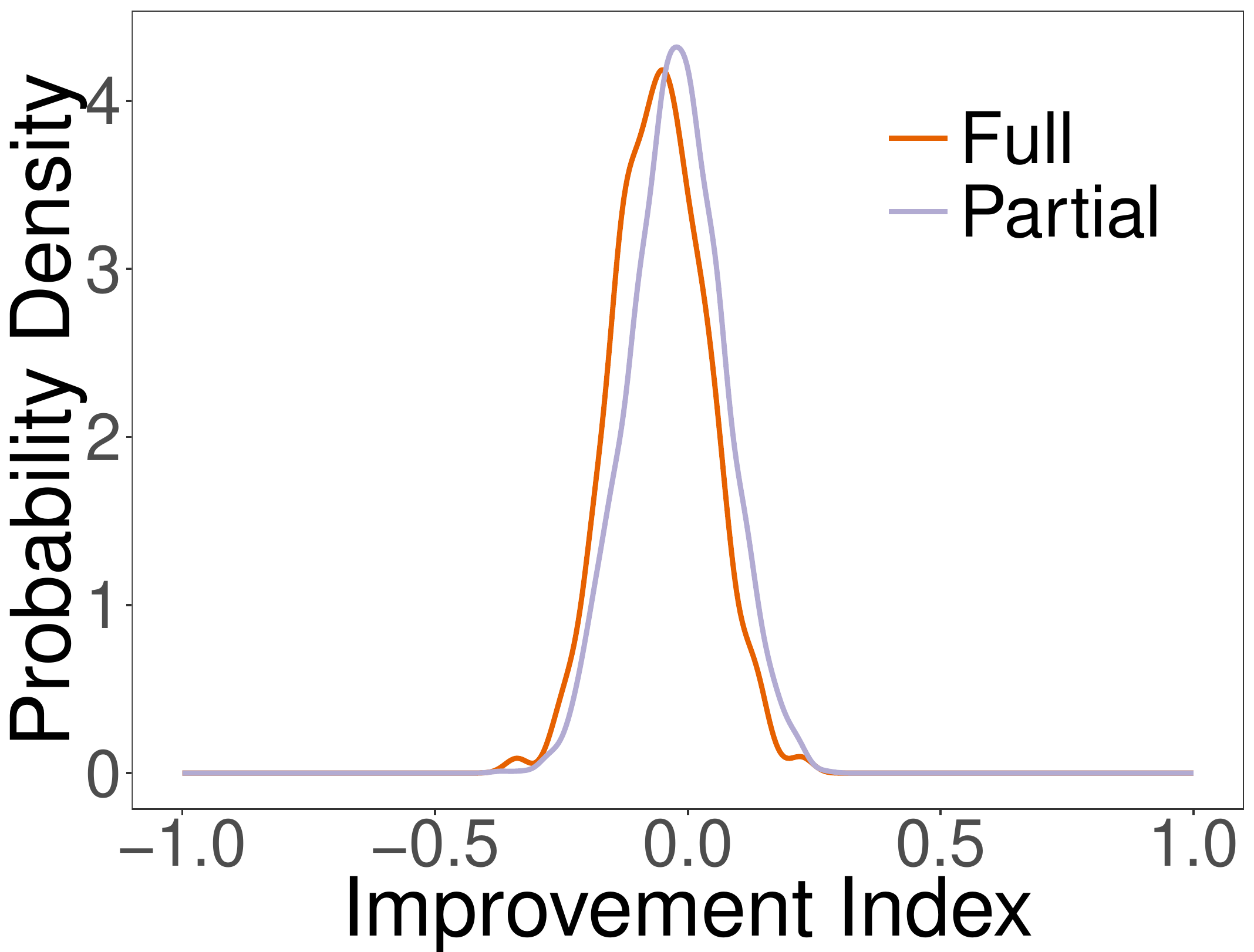}}
\subfigure[Electric vehicles, holarchic initialization]{\includegraphics[width=0.244\textwidth]{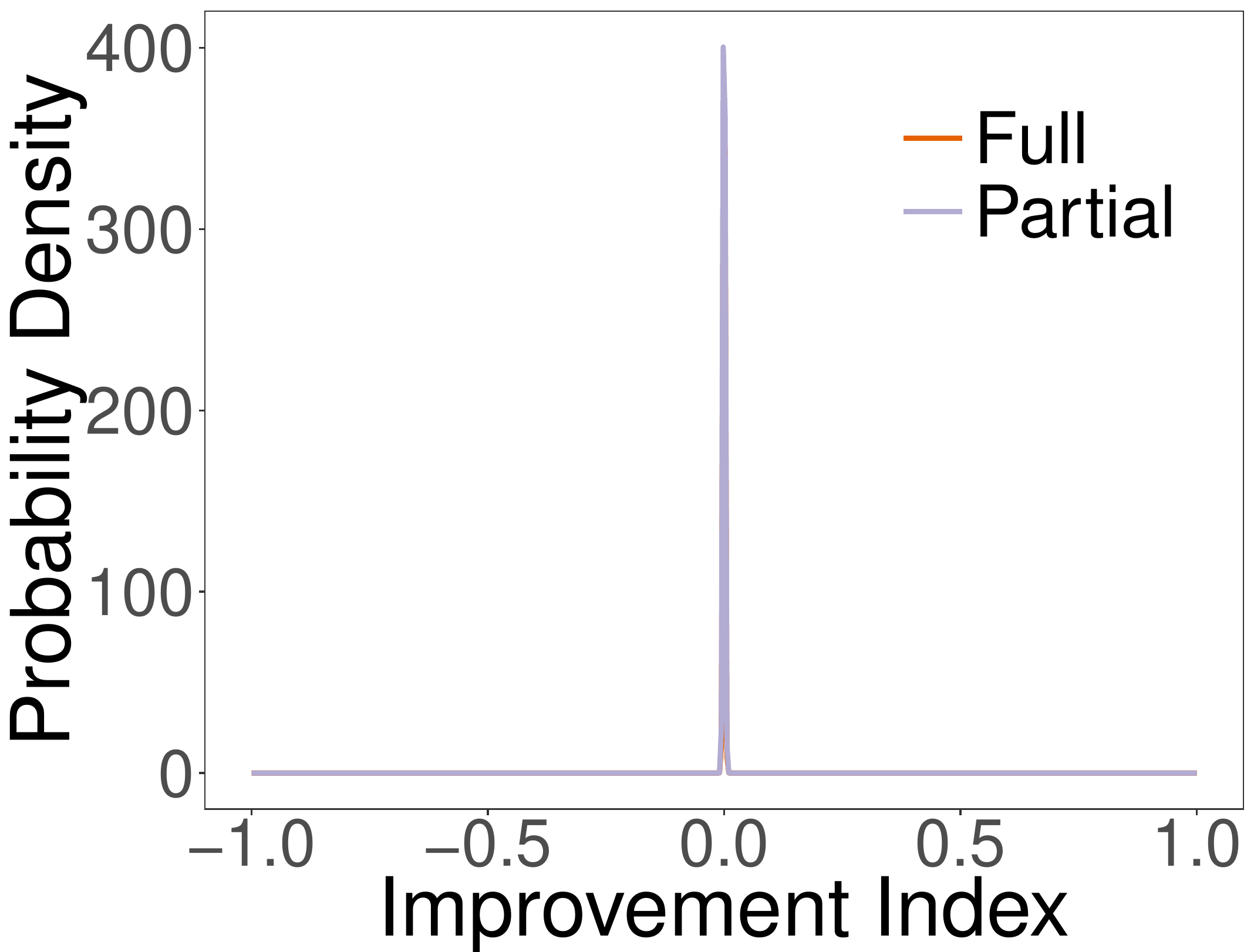}}
\subfigure[Synthetic, holarchic runtime]{\includegraphics[width=0.244\textwidth]{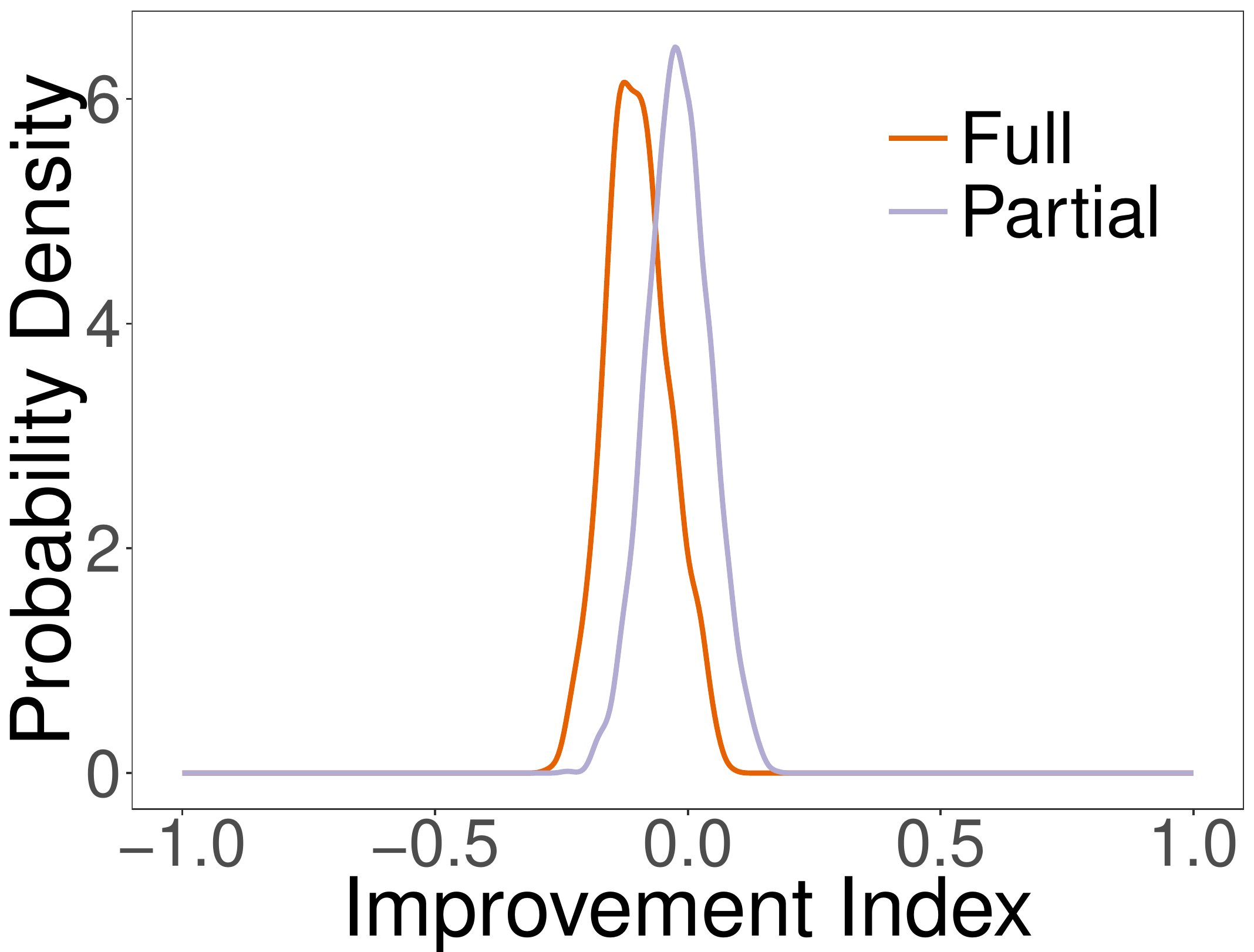}}
\subfigure[Bike sharing, holarchic runtime]{\includegraphics[width=0.244\textwidth]{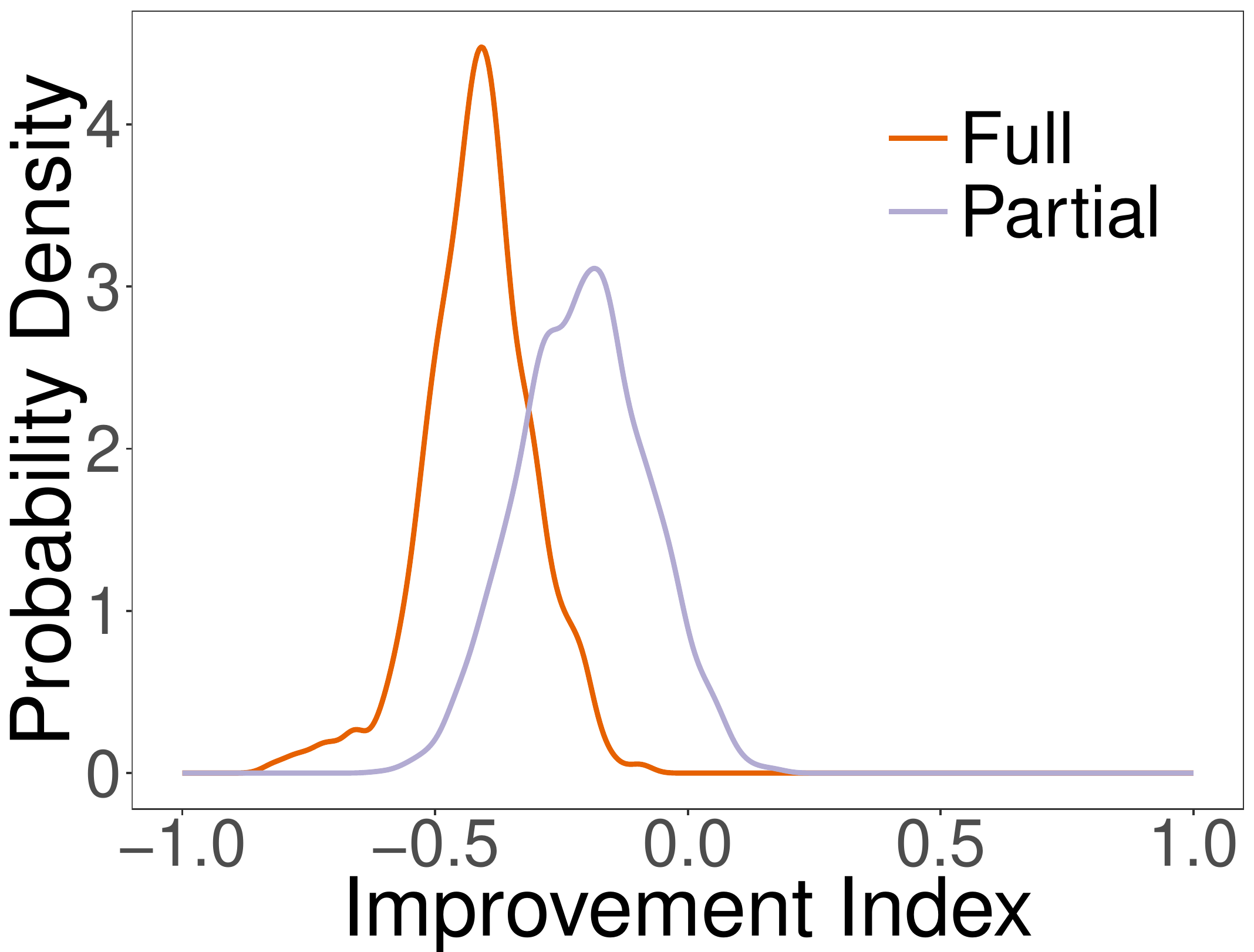}}
\subfigure[Energy demand, holarchic runtime]{\includegraphics[width=0.244\textwidth]{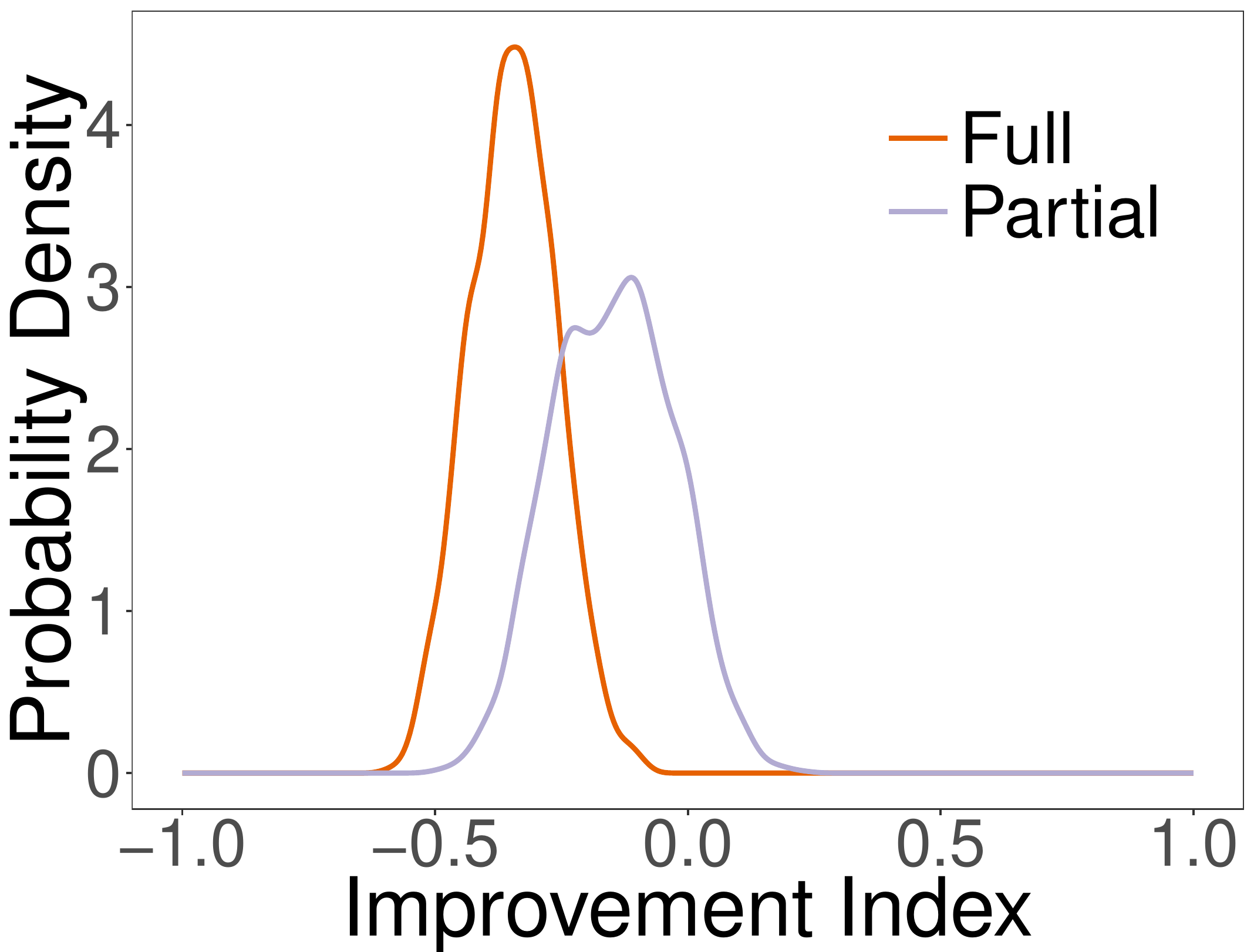}}
\subfigure[Electric vehicles, holarchic runtime]{\includegraphics[width=0.244\textwidth]{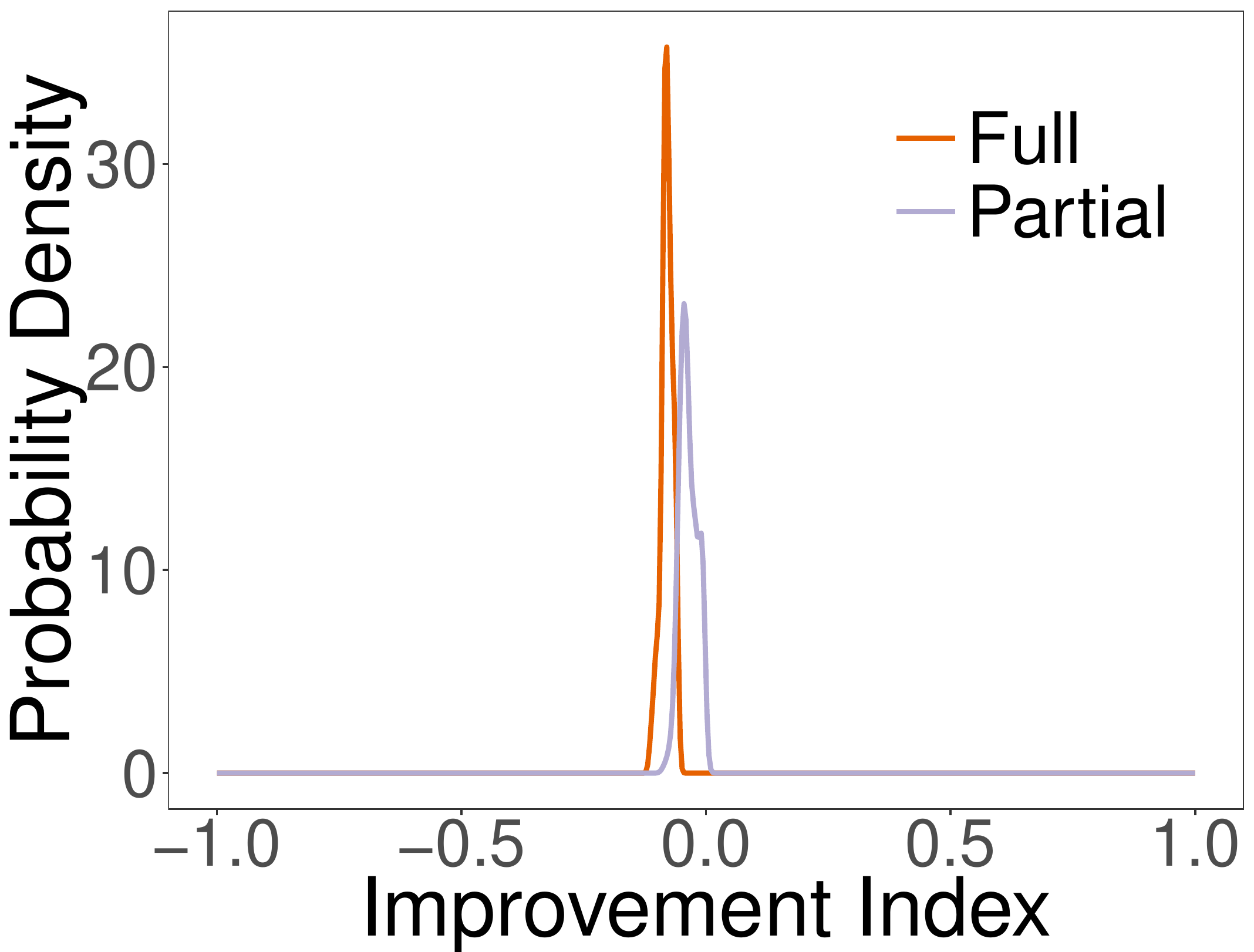}}
\subfigure[Synthetic, holarchic termination]{\includegraphics[width=0.244\textwidth]{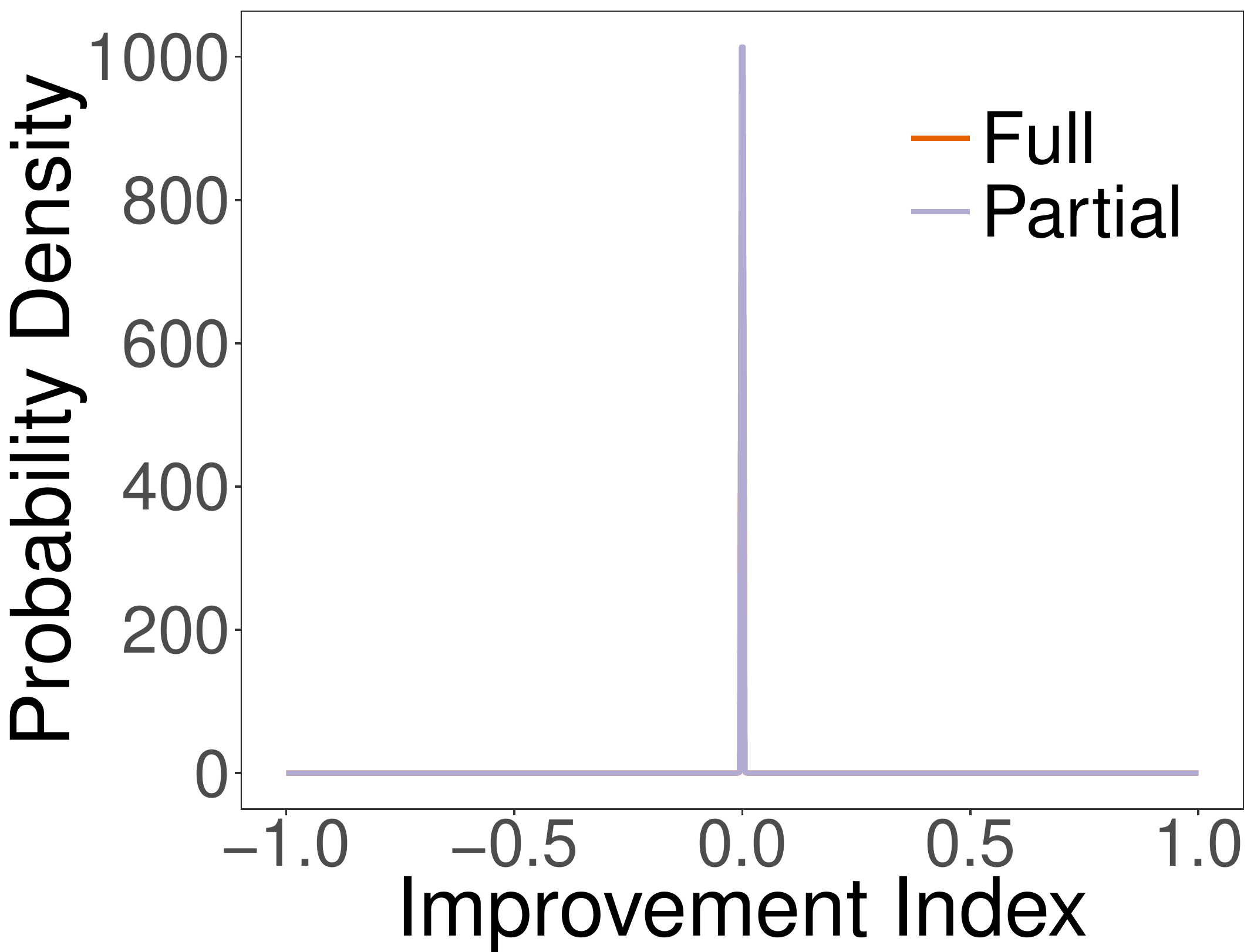}}
\subfigure[Bike sharing, holarchic termination]{\includegraphics[width=0.244\textwidth]{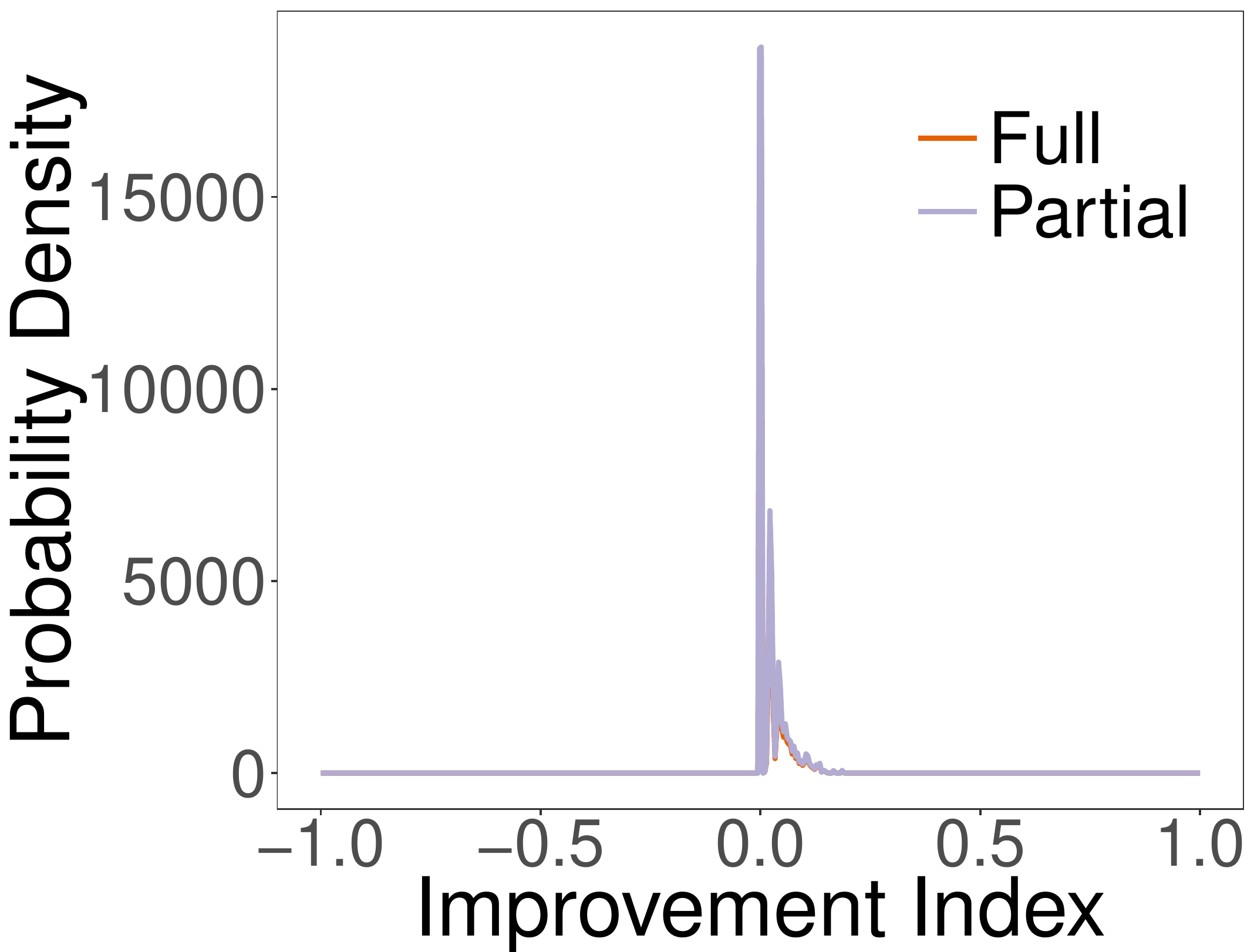}}
\subfigure[Energy demand, holarchic termination]{\includegraphics[width=0.244\textwidth]{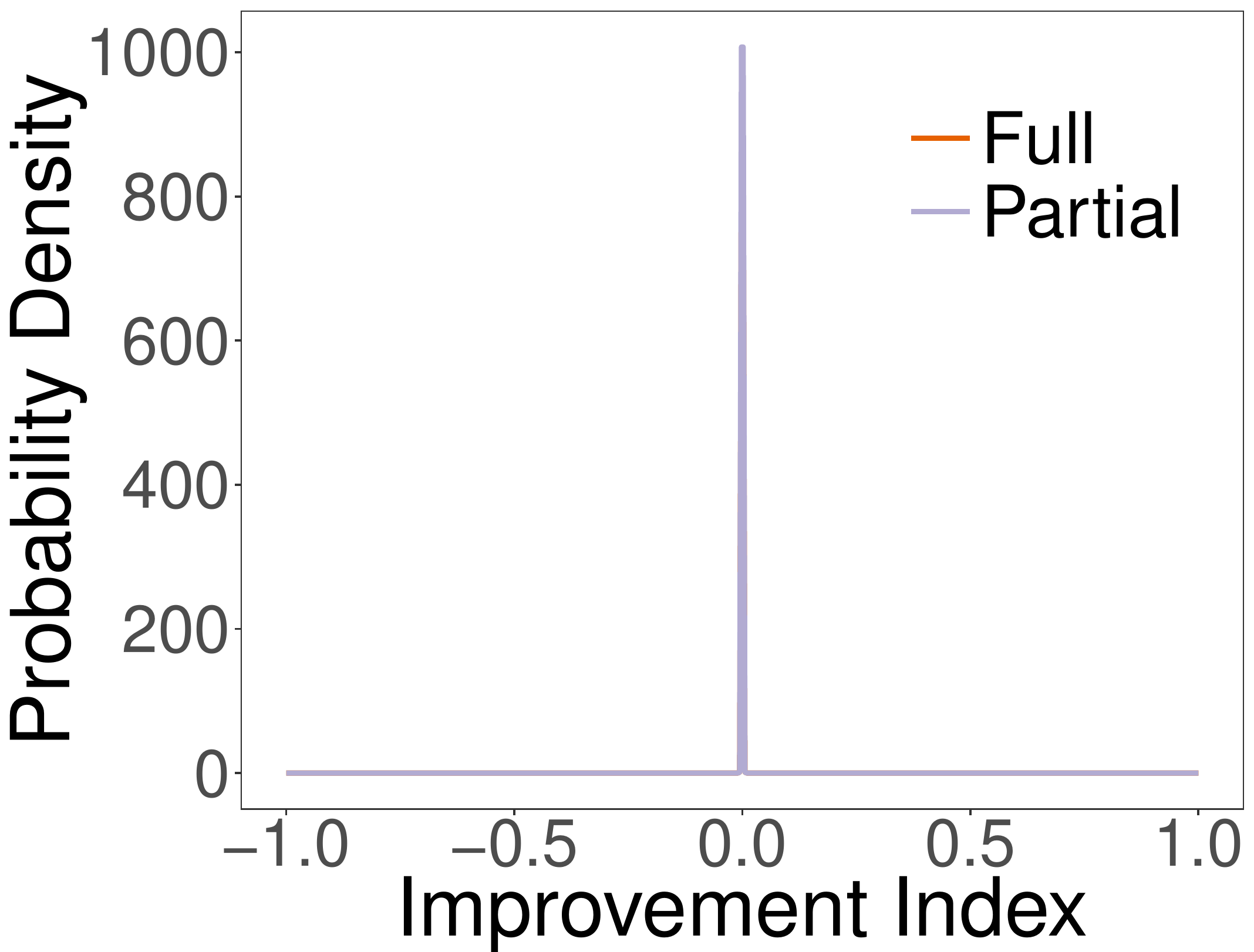}}
\subfigure[Electric vehicles, holarchic termination]{\includegraphics[width=0.244\textwidth]{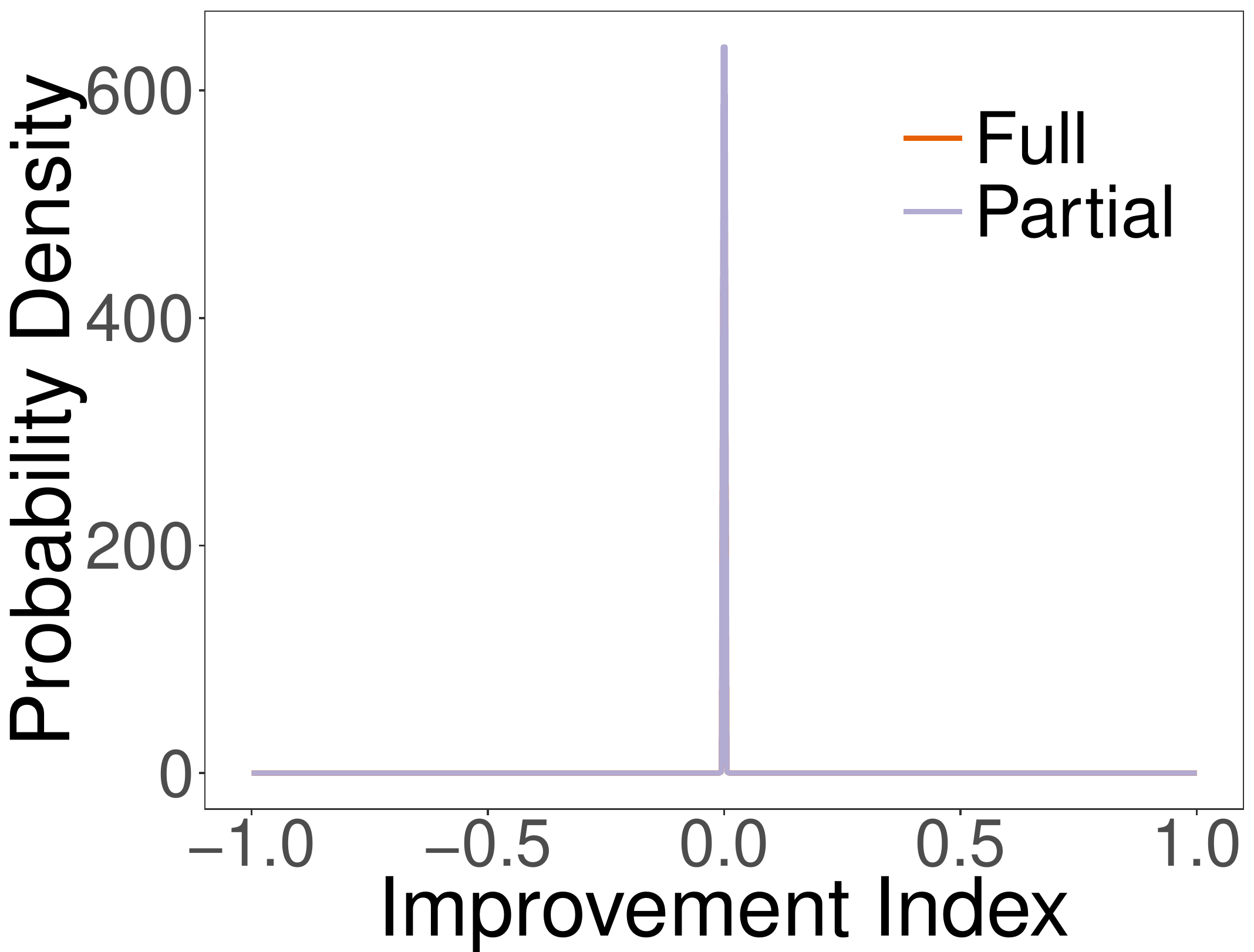}}
\caption{Probability density of the improvement index that elaborates on Figure~\ref{fig:improvement-index-scale}. \emph{Dimensions}: holarchic schemes, application scenarios, partial versus full scale. \emph{Settings}: $\lambda=0$, $c=2$.}\label{fig:density-improvement-index-scale}
\end{figure}

Figure~\ref{fig:density-improvement-index-scale} elaborates on the Figure~\ref{fig:improvement-index-scale}. It illustrates the probability density function of the improvement index by fixing $\lambda=0$, $c=2$ and varying all other dimensions. 

\begin{figure}[!htb]
\centering
\subfigure[Synthetic, total communication]{\includegraphics[width=0.244\textwidth]{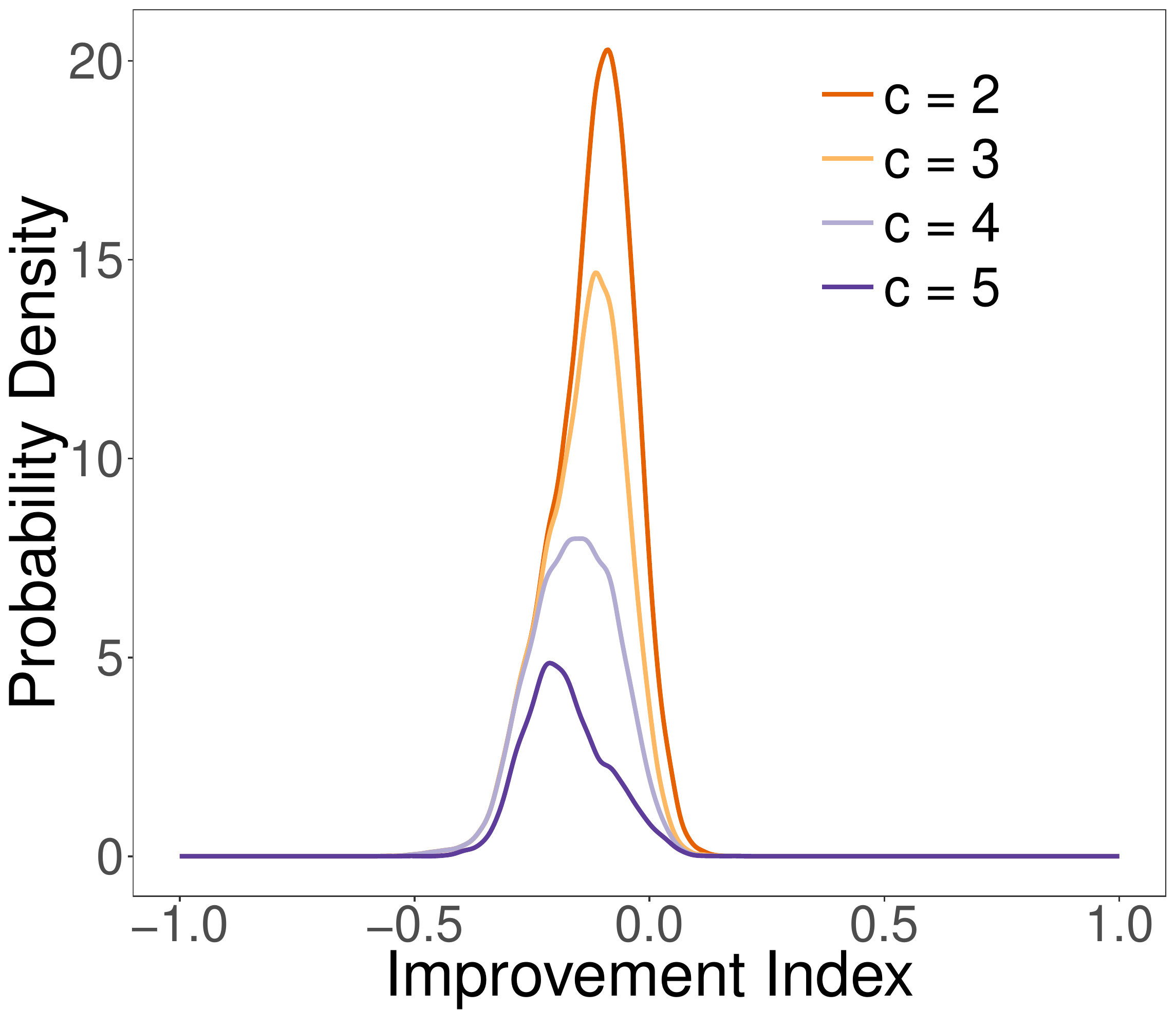}}
\subfigure[Bike sharing, total communication]{\includegraphics[width=0.244\textwidth]{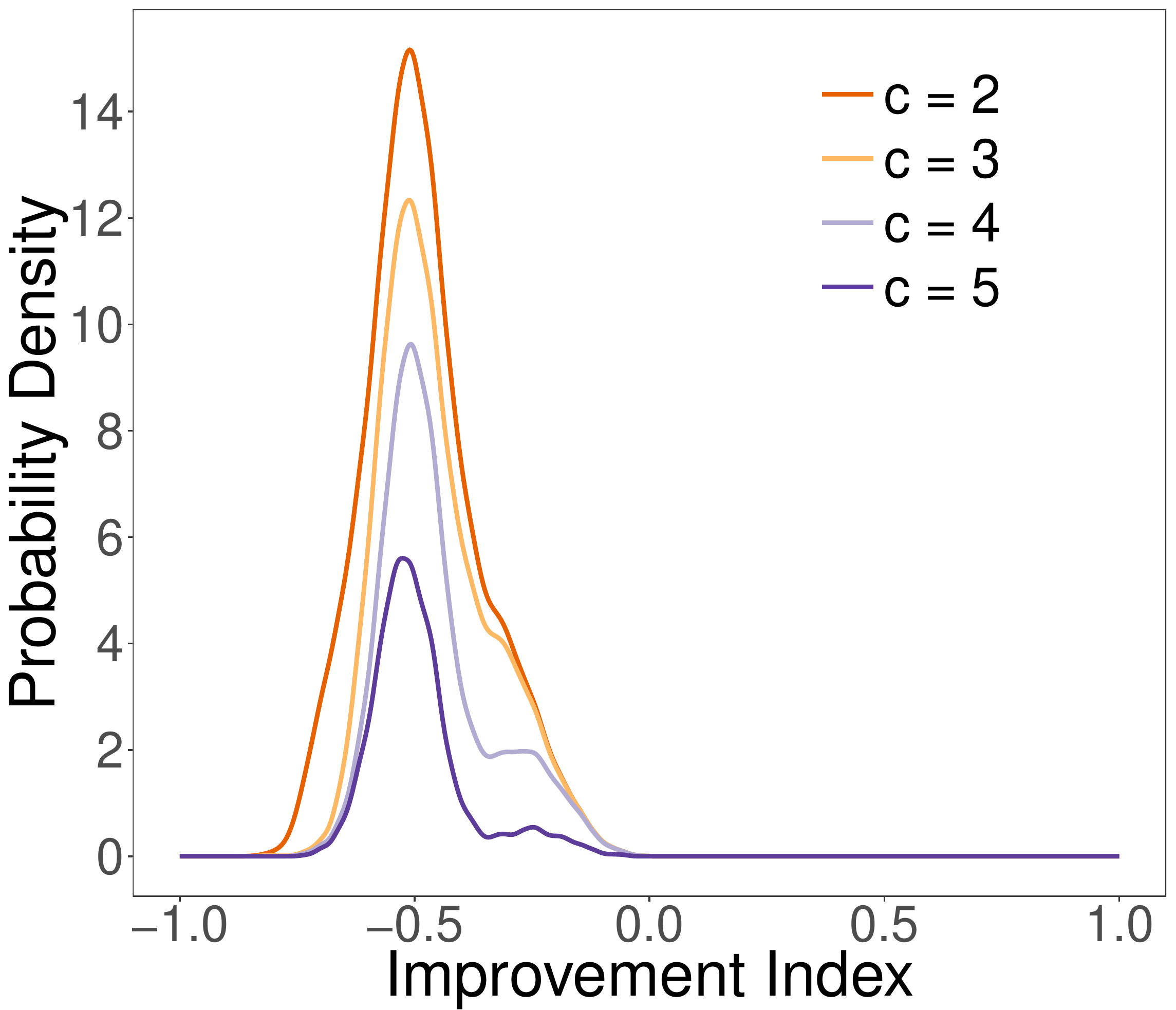}}
\subfigure[Energy demand, total communication]{\includegraphics[width=0.244\textwidth]{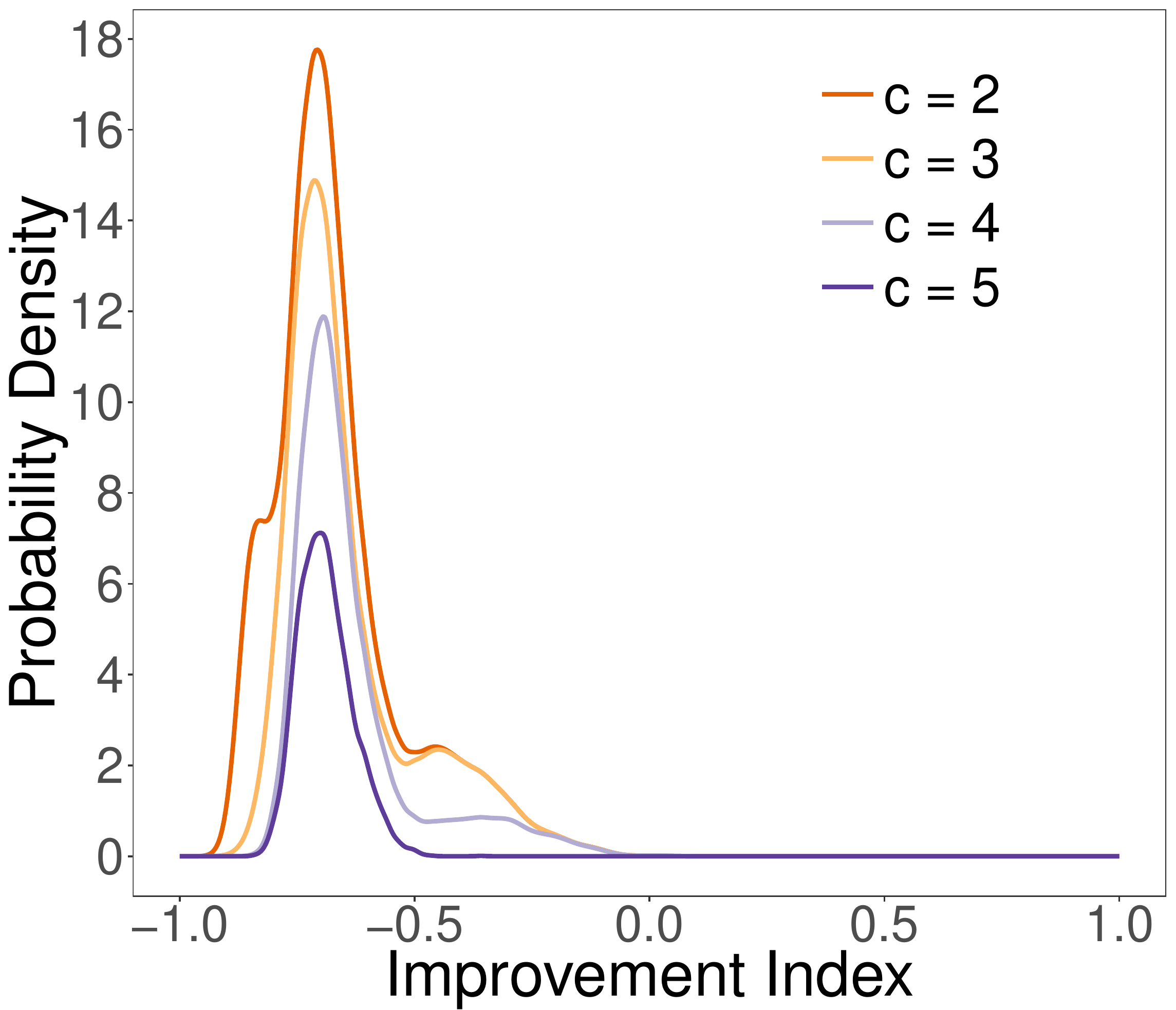}}
\subfigure[Electric vehicles, total communication]{\includegraphics[width=0.244\textwidth]{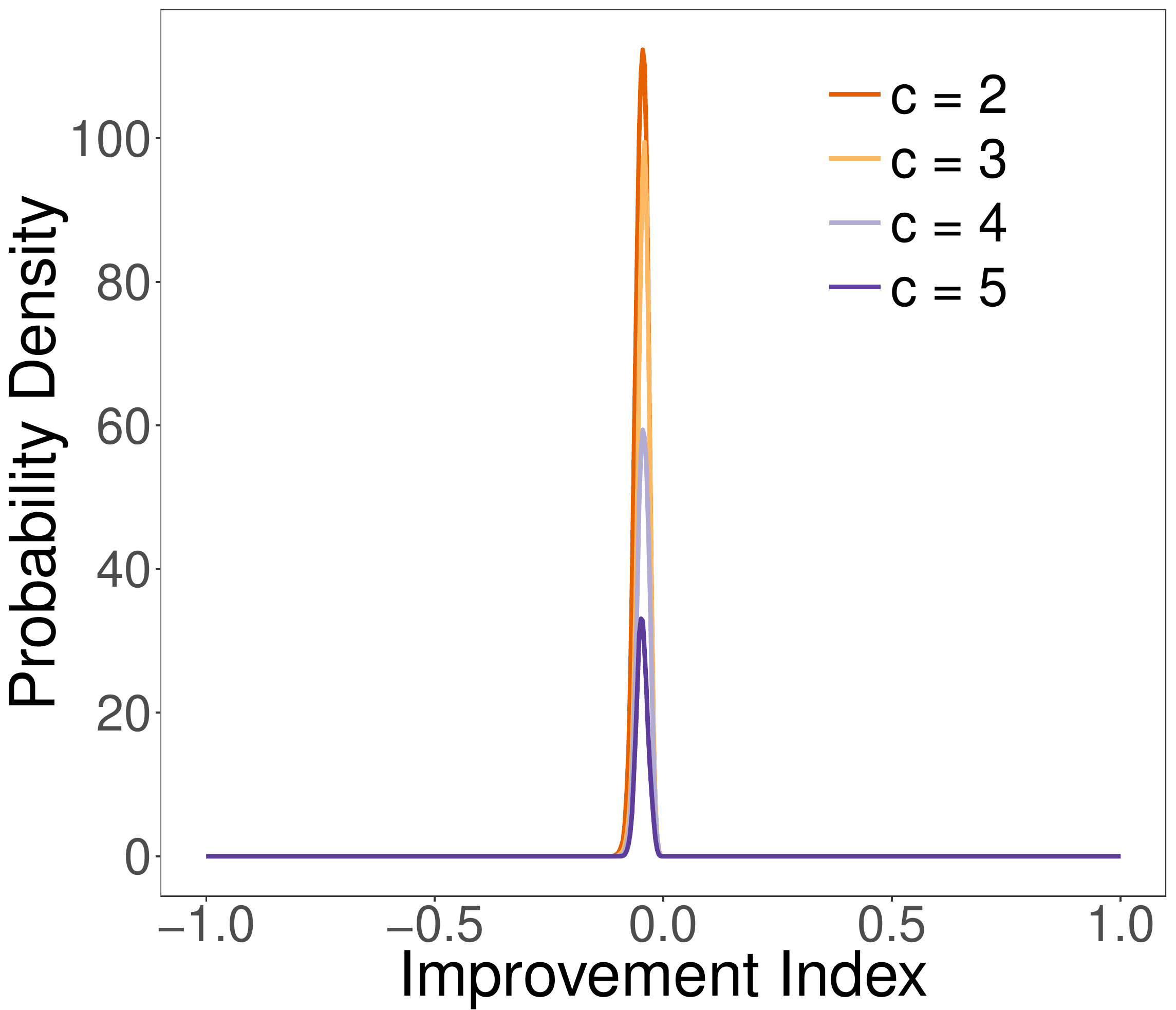}}
\subfigure[Synthetic, synchronized communication]{\includegraphics[width=0.244\textwidth]{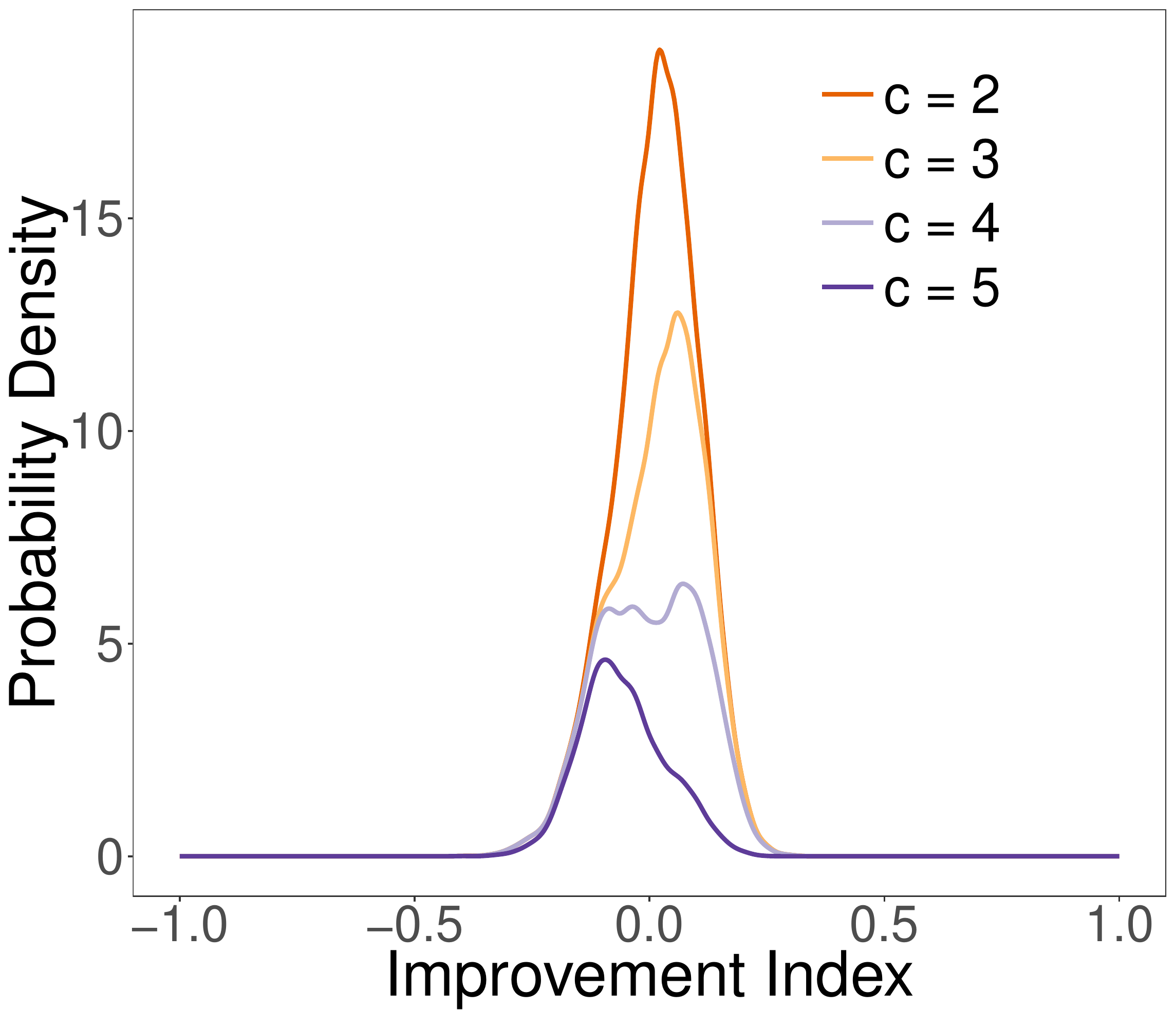}}
\subfigure[Bike sharing, synchronized communication]{\includegraphics[width=0.244\textwidth]{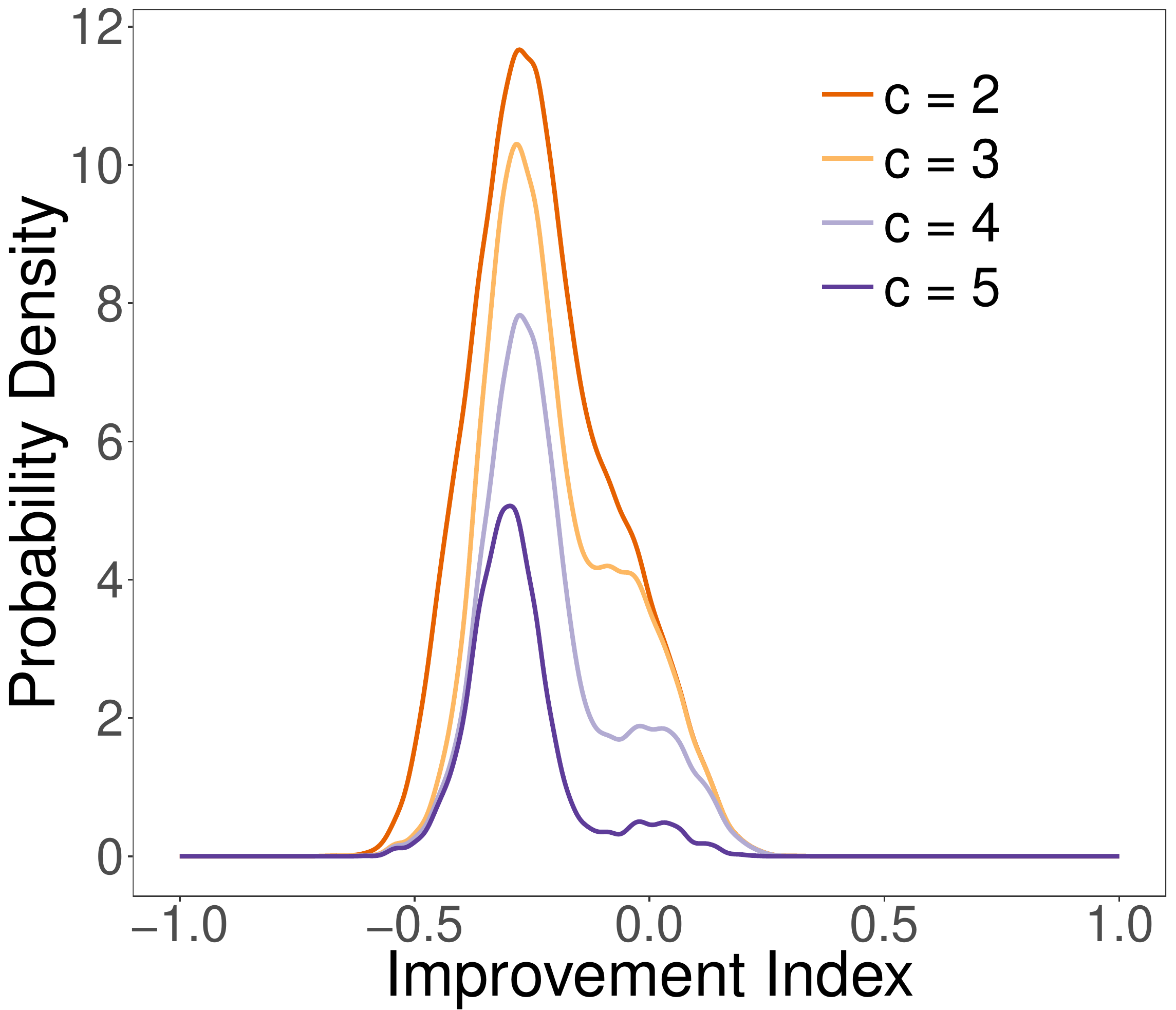}}
\subfigure[Energy demand, synchronized communication]{\includegraphics[width=0.244\textwidth]{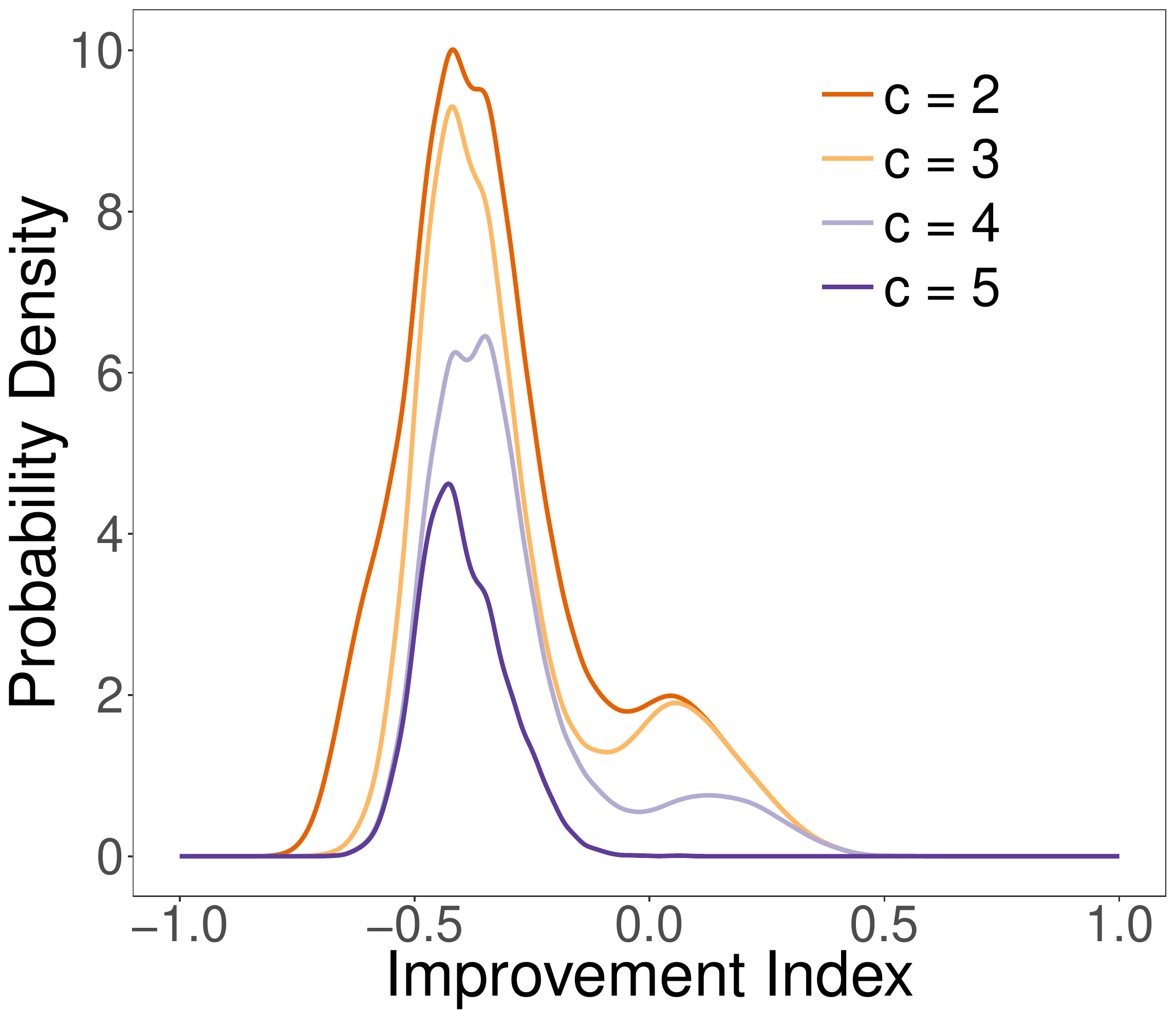}}
\subfigure[Electric vehicles, synchronized communication]{\includegraphics[width=0.244\textwidth]{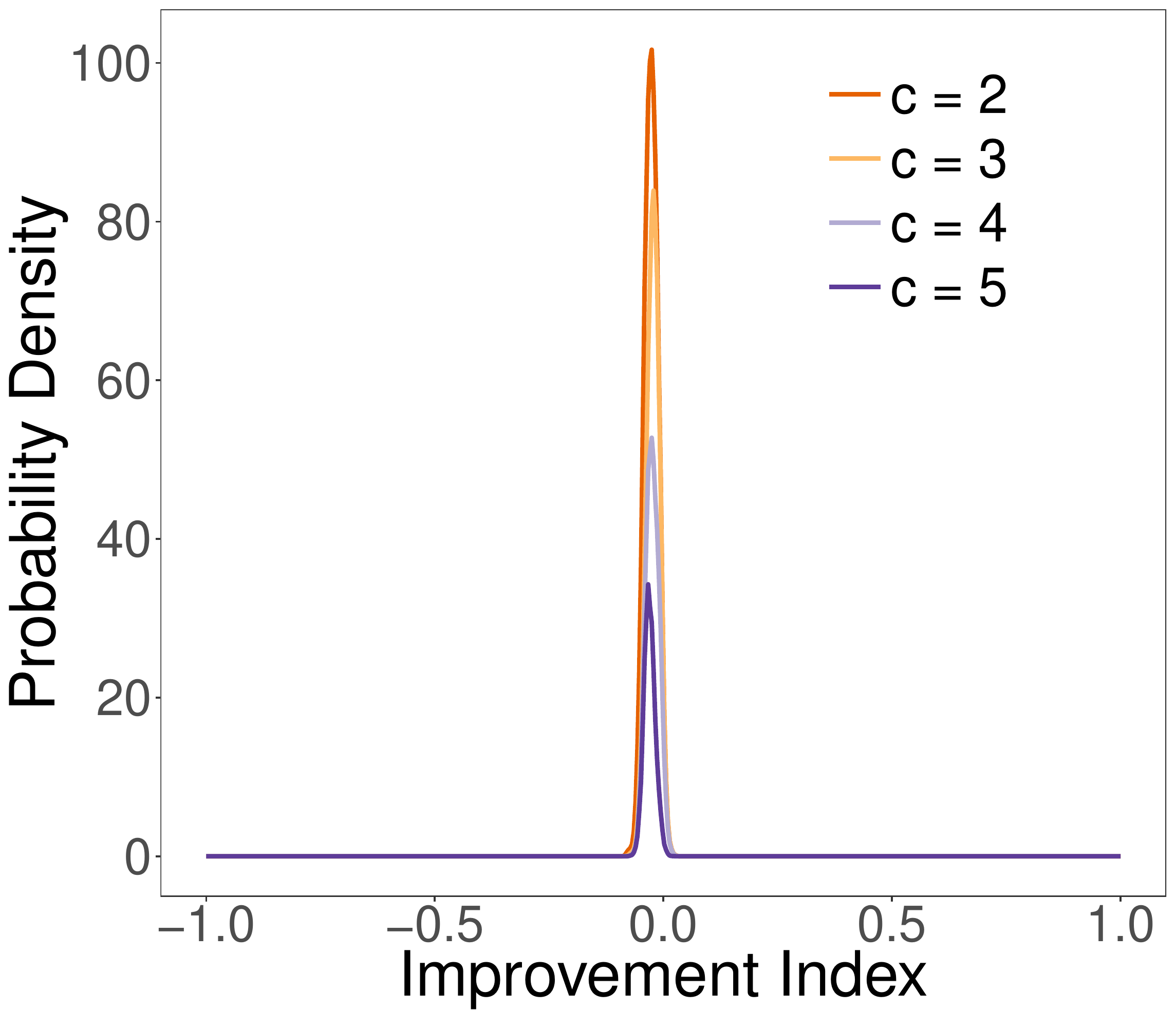}}
\caption{Probability density of the improvement index. \emph{Dimensions}: total versus synchronized communication cost, application scenarios, varying number of children. \emph{Settings}: holarchic runtime, partial scale, $\lambda=0$.}\label{fig:density-improvement-index-communication-children-partial}
\end{figure}

Figure~\ref{fig:density-improvement-index-communication-children-partial} contrasts the improvement index of the holarchic runtime on the basis of total versus synchronized communication cost. Results show an average increase of the improvement index by 9.5\%, indicated by the shift to the right for synchronized communication. 

\begin{figure}[!htb]
\centering
\subfigure[Synthetic, total communication]{\includegraphics[width=0.244\textwidth]{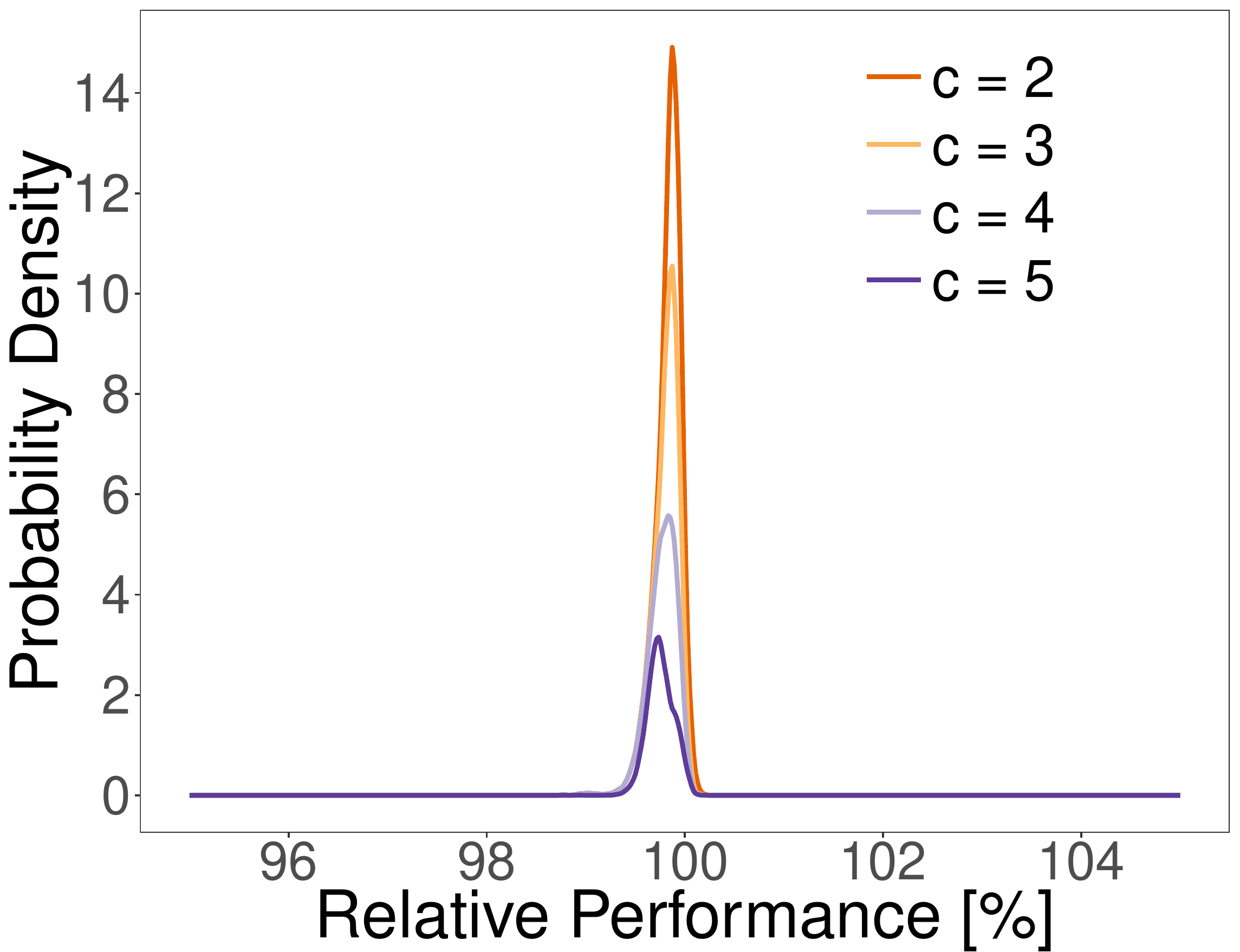}}
\subfigure[Bike sharing, total communication]{\includegraphics[width=0.244\textwidth]{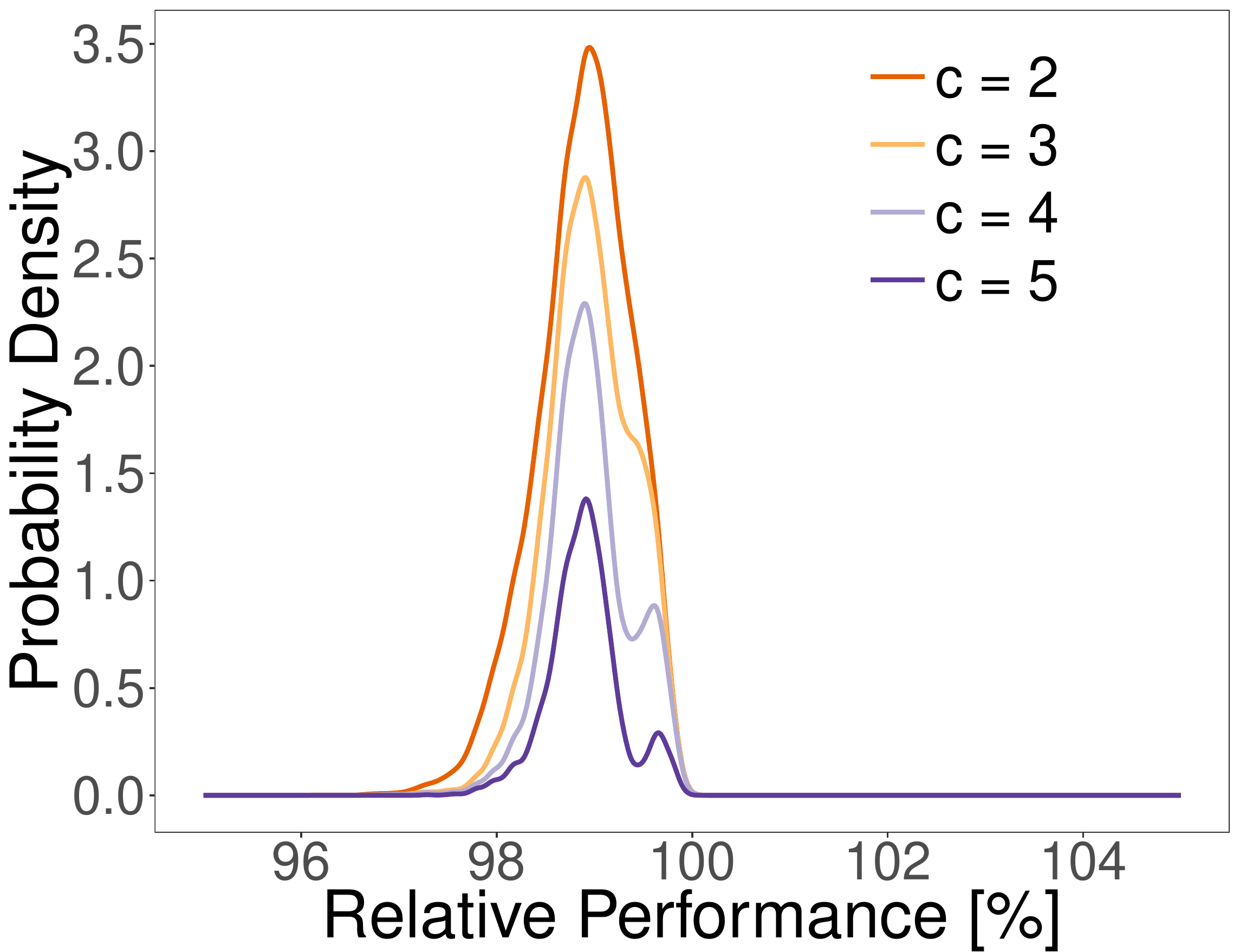}}
\subfigure[Energy demand, total communication]{\includegraphics[width=0.244\textwidth]{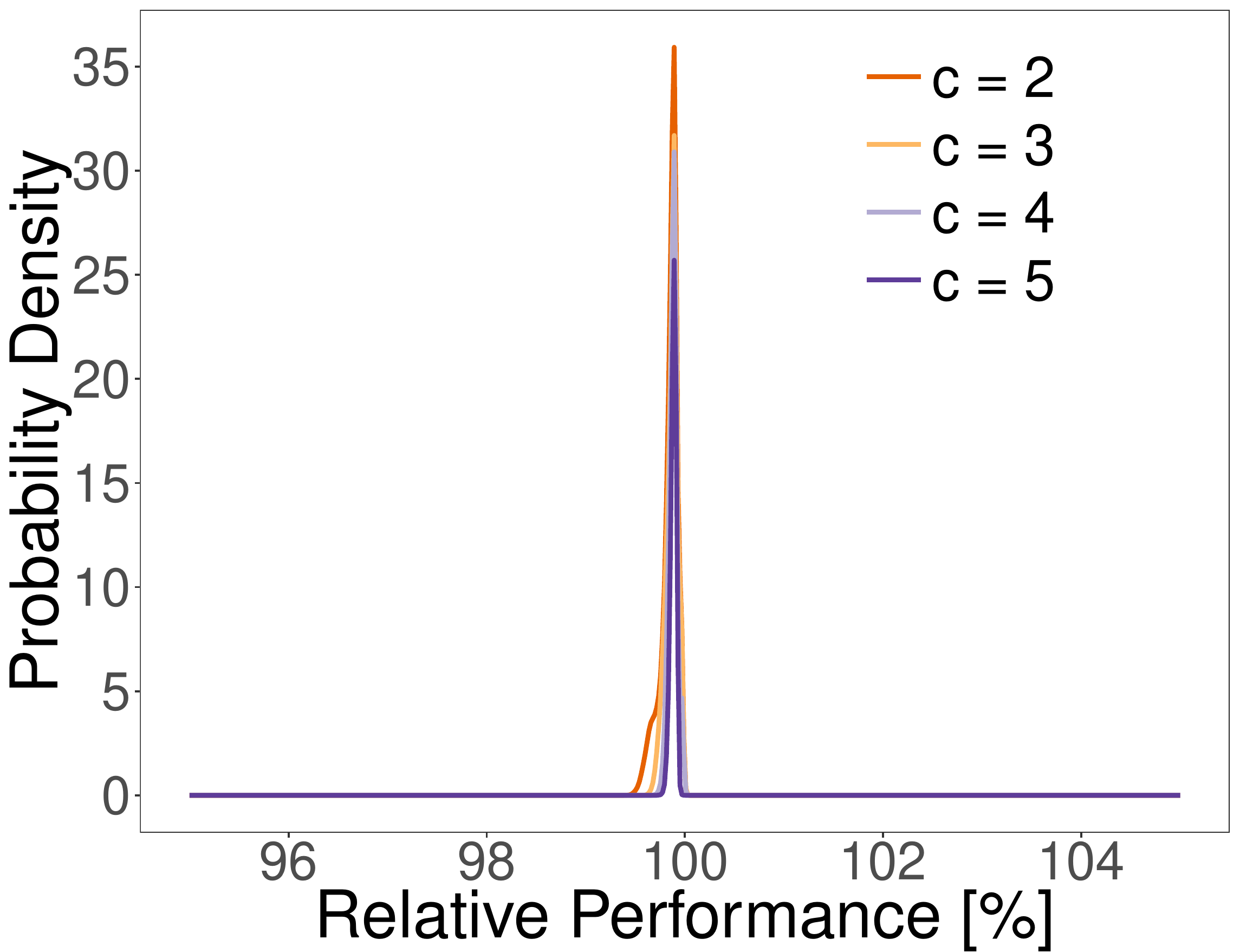}}
\subfigure[Electric vehicles, total communication]{\includegraphics[width=0.244\textwidth]{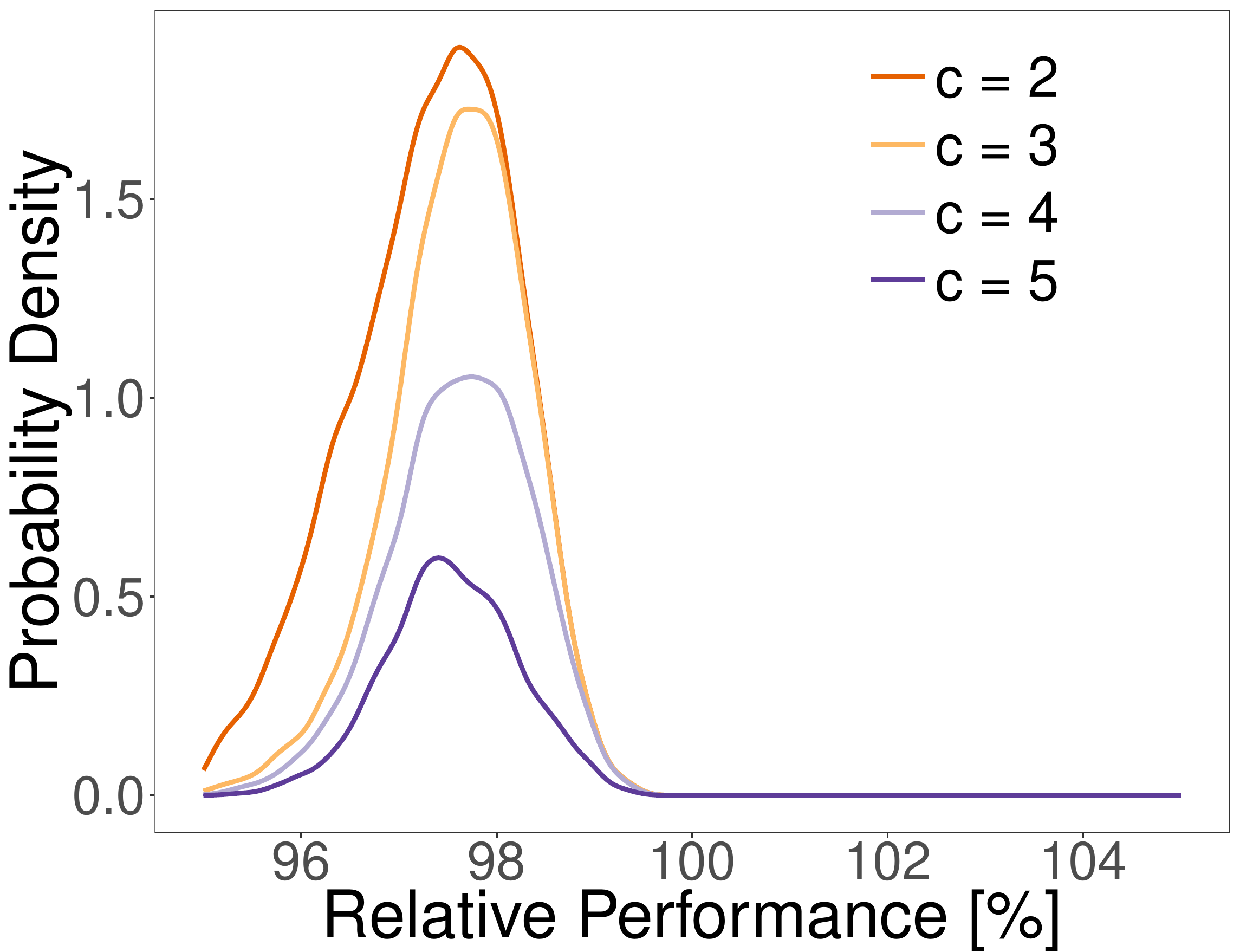}}
\subfigure[Synthetic, synchronized communication]{\includegraphics[width=0.244\textwidth]{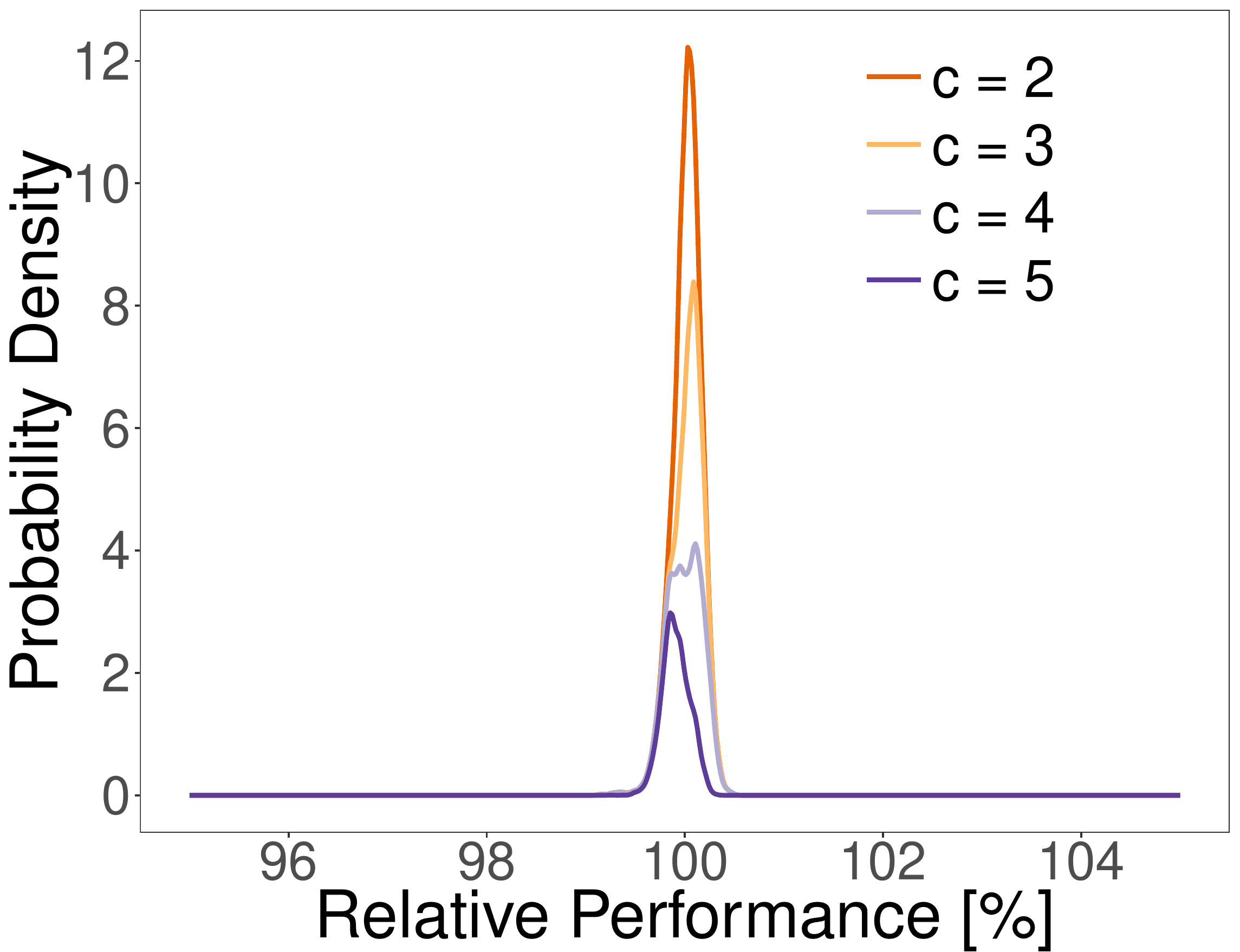}}
\subfigure[Bike sharing, synchronized communication]{\includegraphics[width=0.244\textwidth]{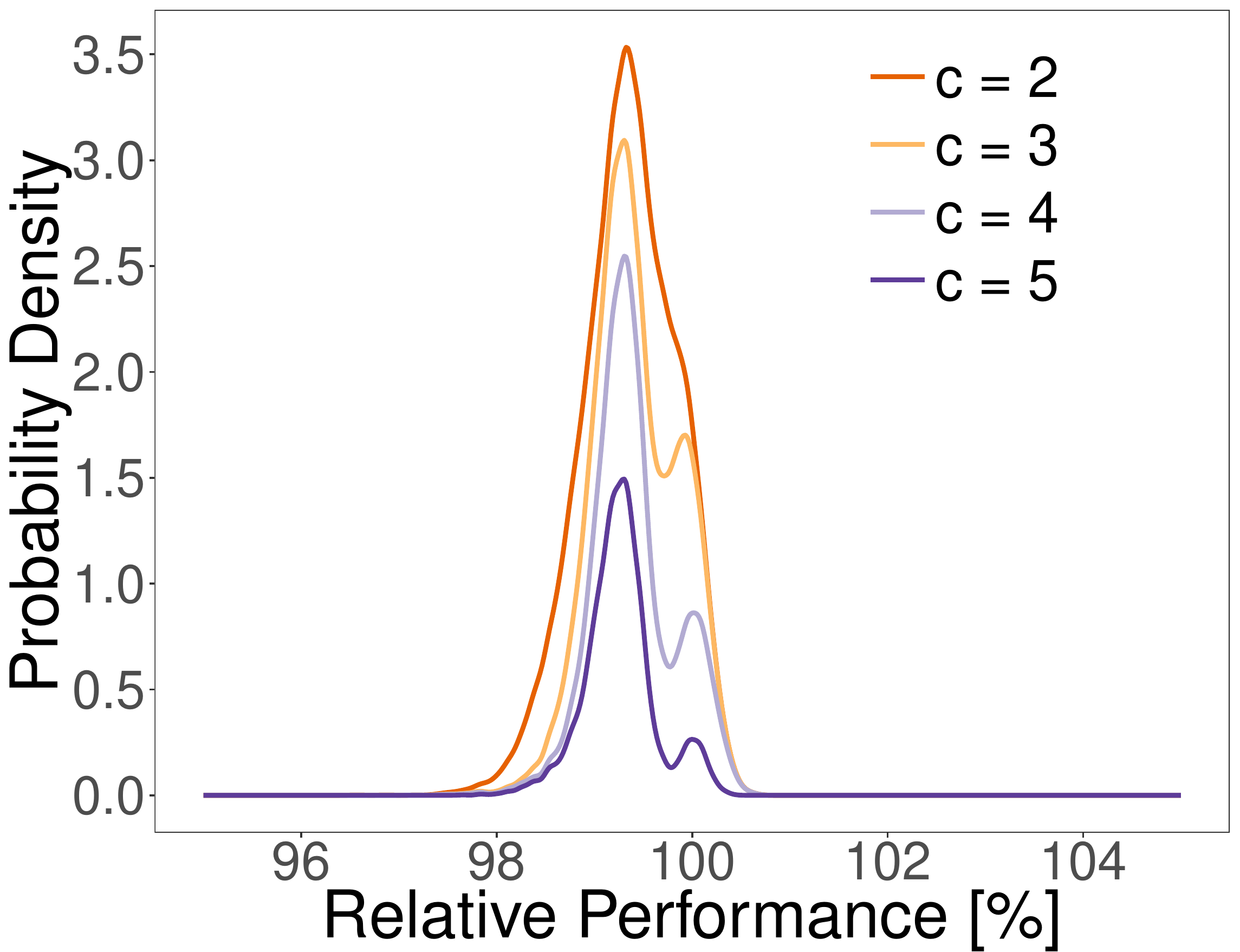}}
\subfigure[Energy demand, synchronized communication]{\includegraphics[width=0.244\textwidth]{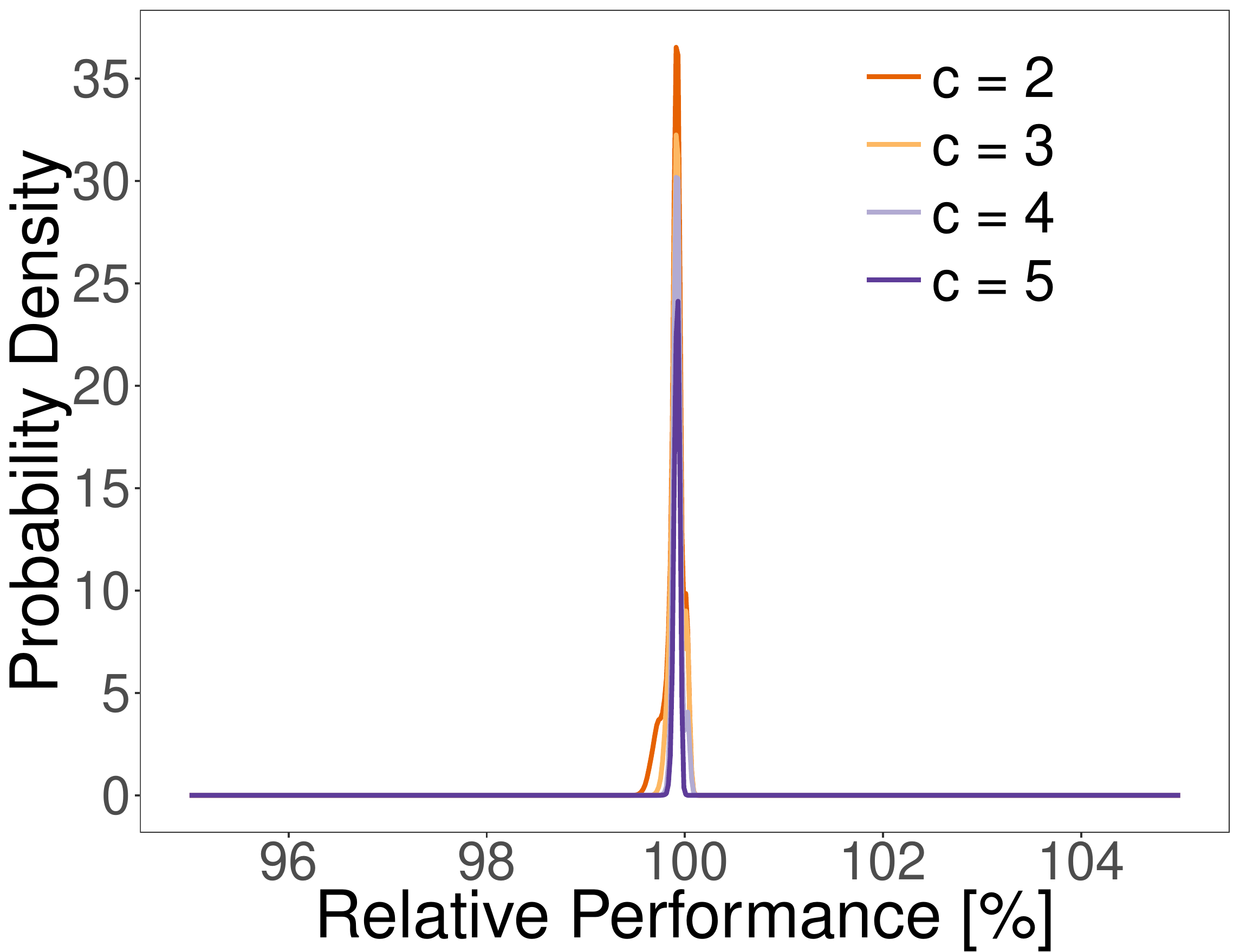}}
\subfigure[Electric vehicles, synchronized communication]{\includegraphics[width=0.244\textwidth]{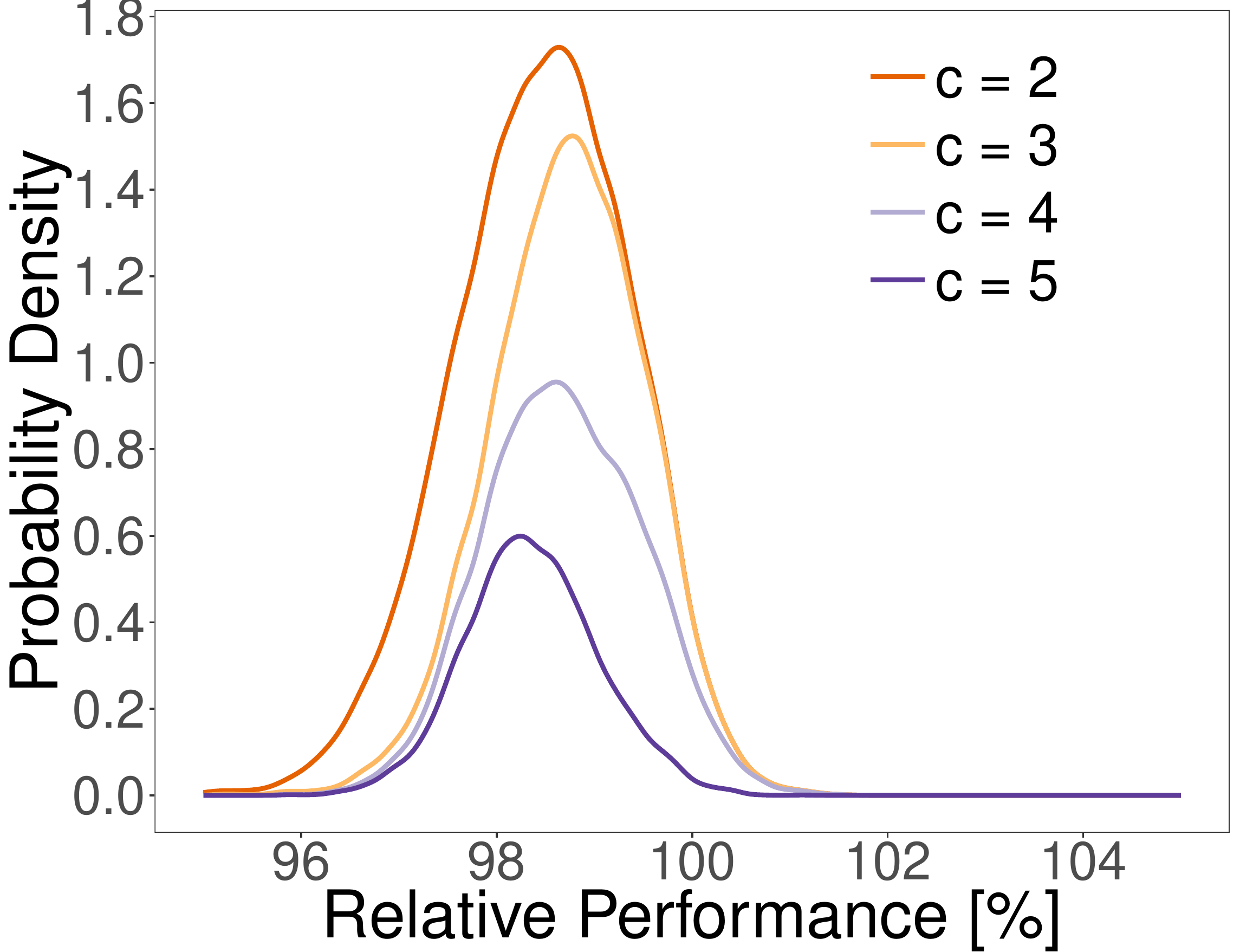}}
\caption{Probability density of the relative global cost reduction. \emph{Dimensions}: total versus synchronized communication cost, application scenarios, varying number of children. \emph{Settings}: holarchic runtime, partial scale, $\lambda=0$.}\label{fig:density-relative-global-cost-reduction-communication-children-partial}
\end{figure}

The contrast of total versus synchronized communication cost is illustrated in Figure~\ref{fig:density-relative-global-cost-reduction-communication-children-partial} via the \emph{relative performance} that is defined as follows:

\begin{align}\label{eq:relative-performance}
P=\frac{C^{(1)}_{\mathsf{h}}-C^{(T)}_{\mathsf{h}}}{C^{(1)}_{\mathsf{b}}-C^{(T)}_{\mathsf{b}}}
\end{align}

\noindent where $C^{(1)}_{\mathsf{h}}$, $C^{(1)}_{\mathsf{b}}$ is the global cost at the first iteration $t=1$ for holarchic runtime and baseline respectively, while $C^{(T)}_{\mathsf{h}}$, $C^{(T)}_{\mathsf{b}}$ is the global cost at convergence $t=T$. This metric encodes the additional information of the improvement extent during the learning iterations. For instance, while Figure~\ref{fig:density-improvement-index-communication-children-partial}f shows a density with improvement index values around -0.7, the relative performance values in Figure~\ref{fig:density-relative-global-cost-reduction-communication-children-partial}f spread very close to 100\%, which means that a performance peak is achieved. 

\begin{figure}[!htb]
\centering
\includegraphics[width=1.0\textwidth]{legend-baseline-holarchic-runtime.pdf}\\
\subfigure[Synthetic, total communication]{\includegraphics[width=0.244\textwidth]{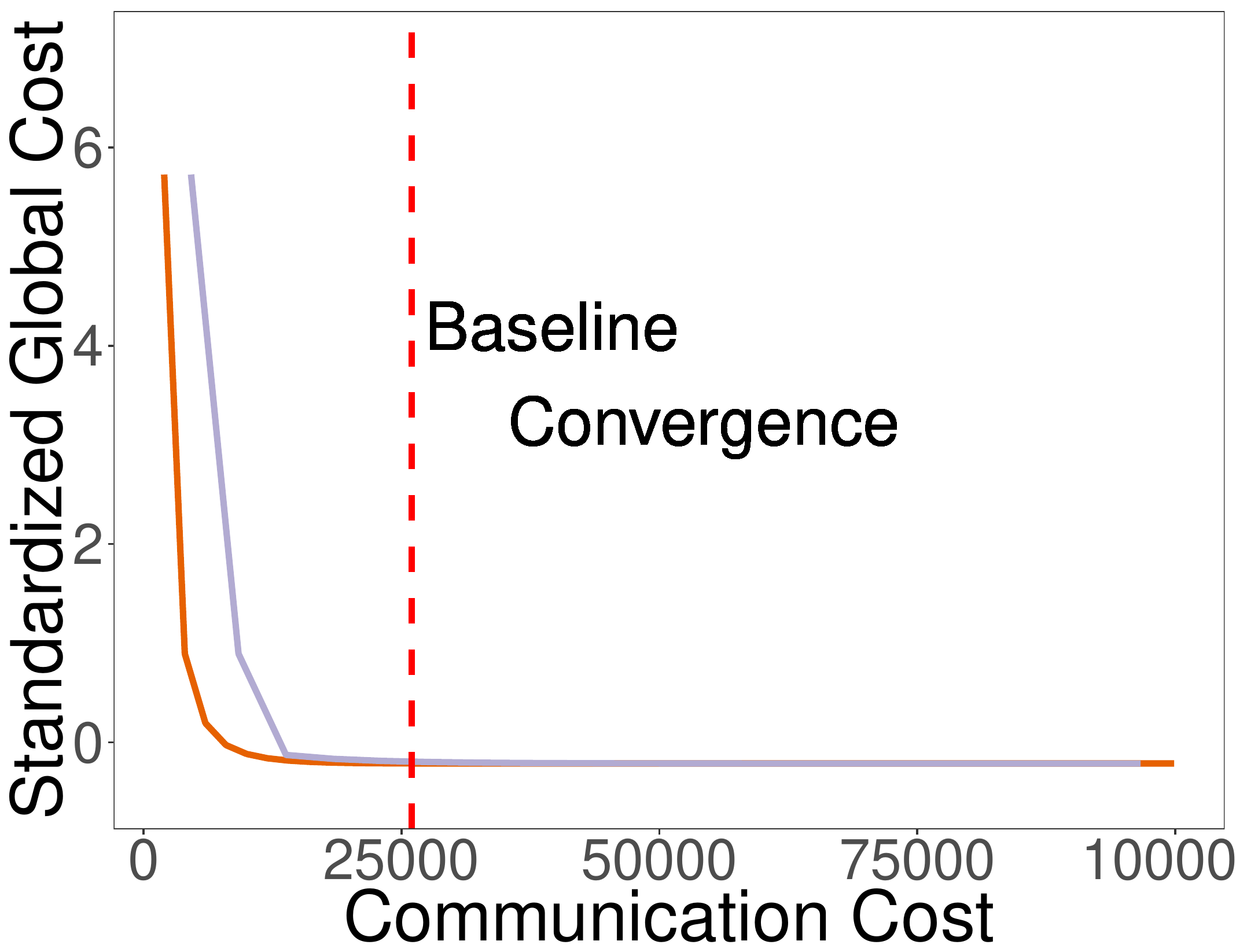}}
\subfigure[Bike sharing, total communication]{\includegraphics[width=0.244\textwidth]{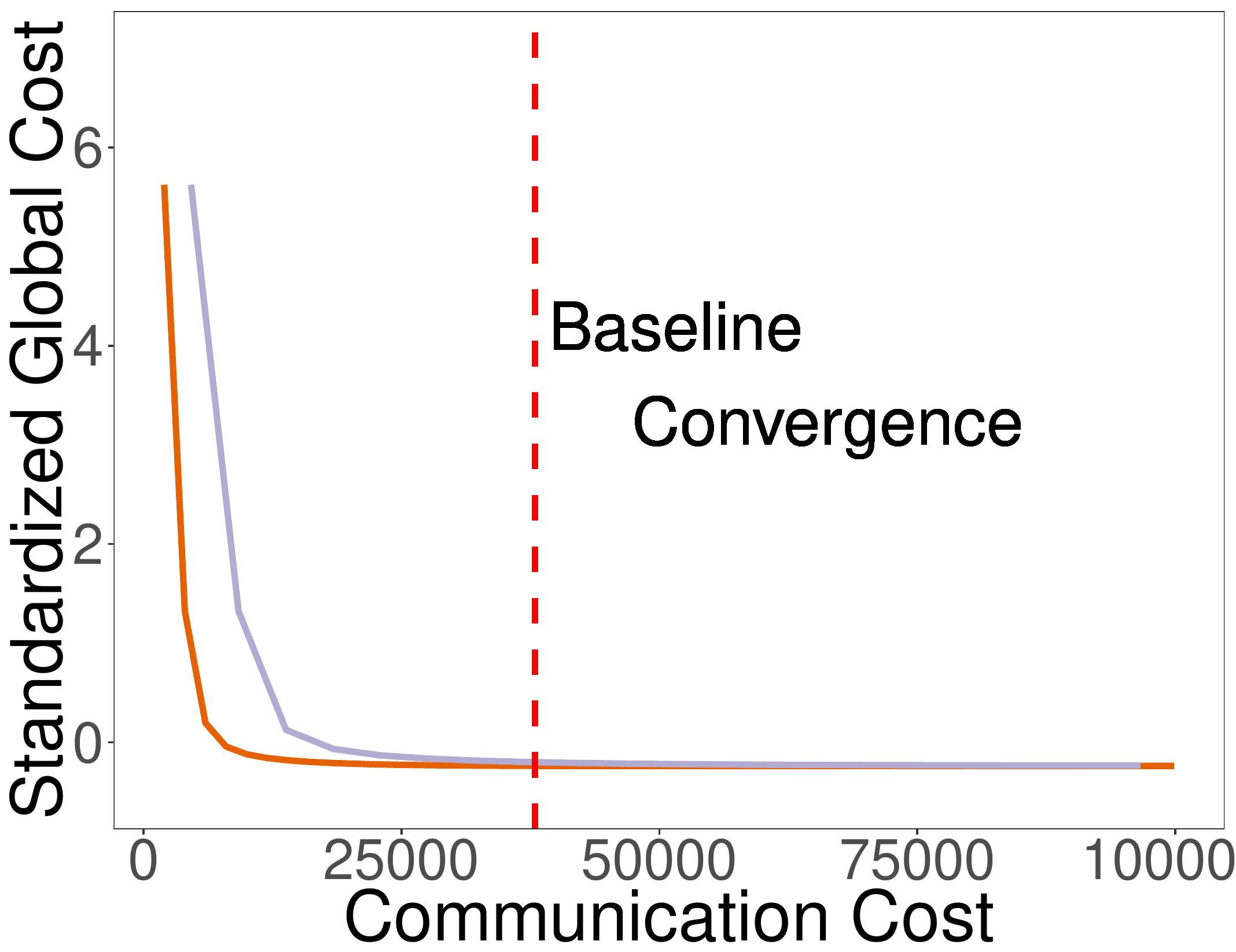}}
\subfigure[Energy demand, total communication]{\includegraphics[width=0.244\textwidth]{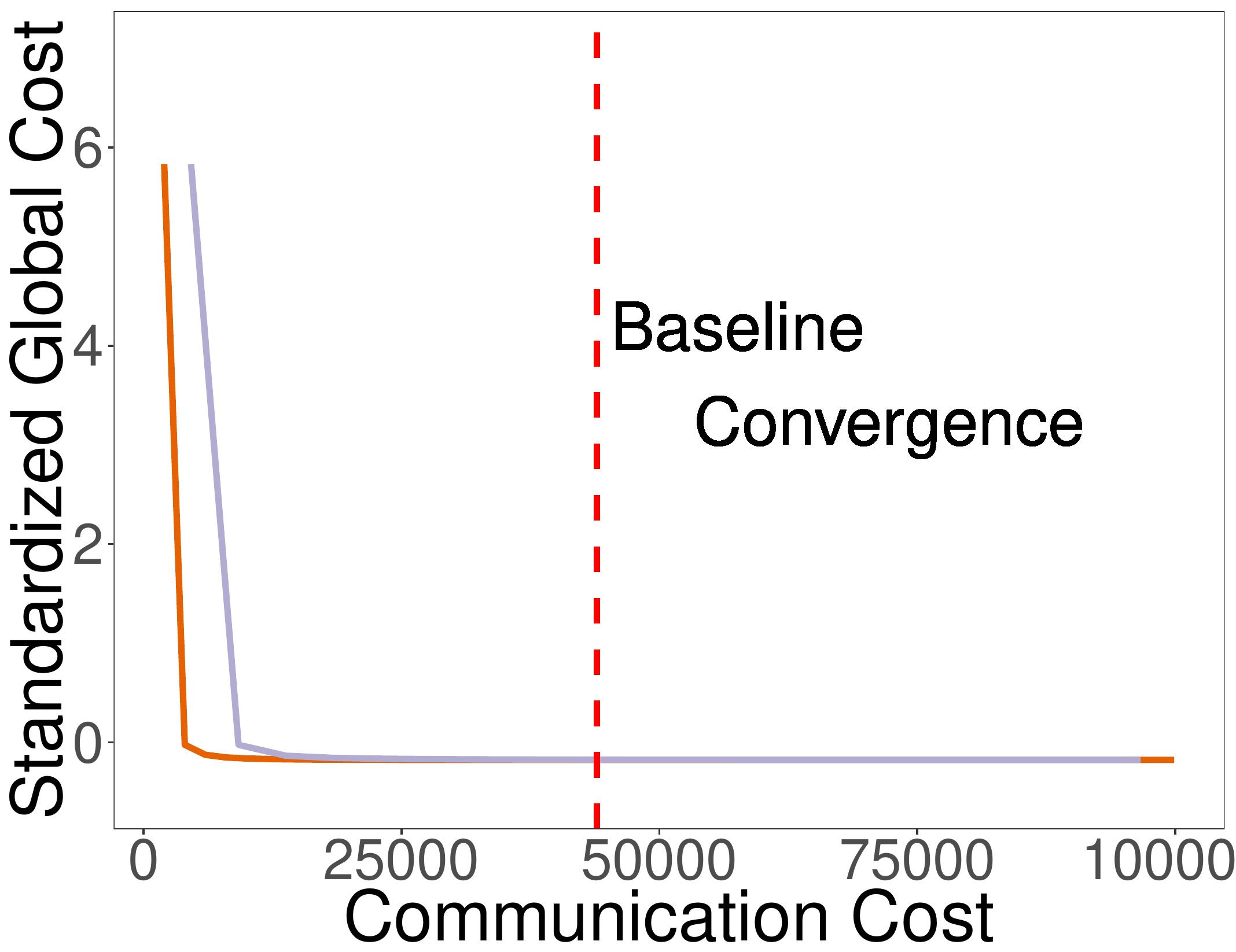}}
\subfigure[Electric vehicles, total communication]{\includegraphics[width=0.244\textwidth]{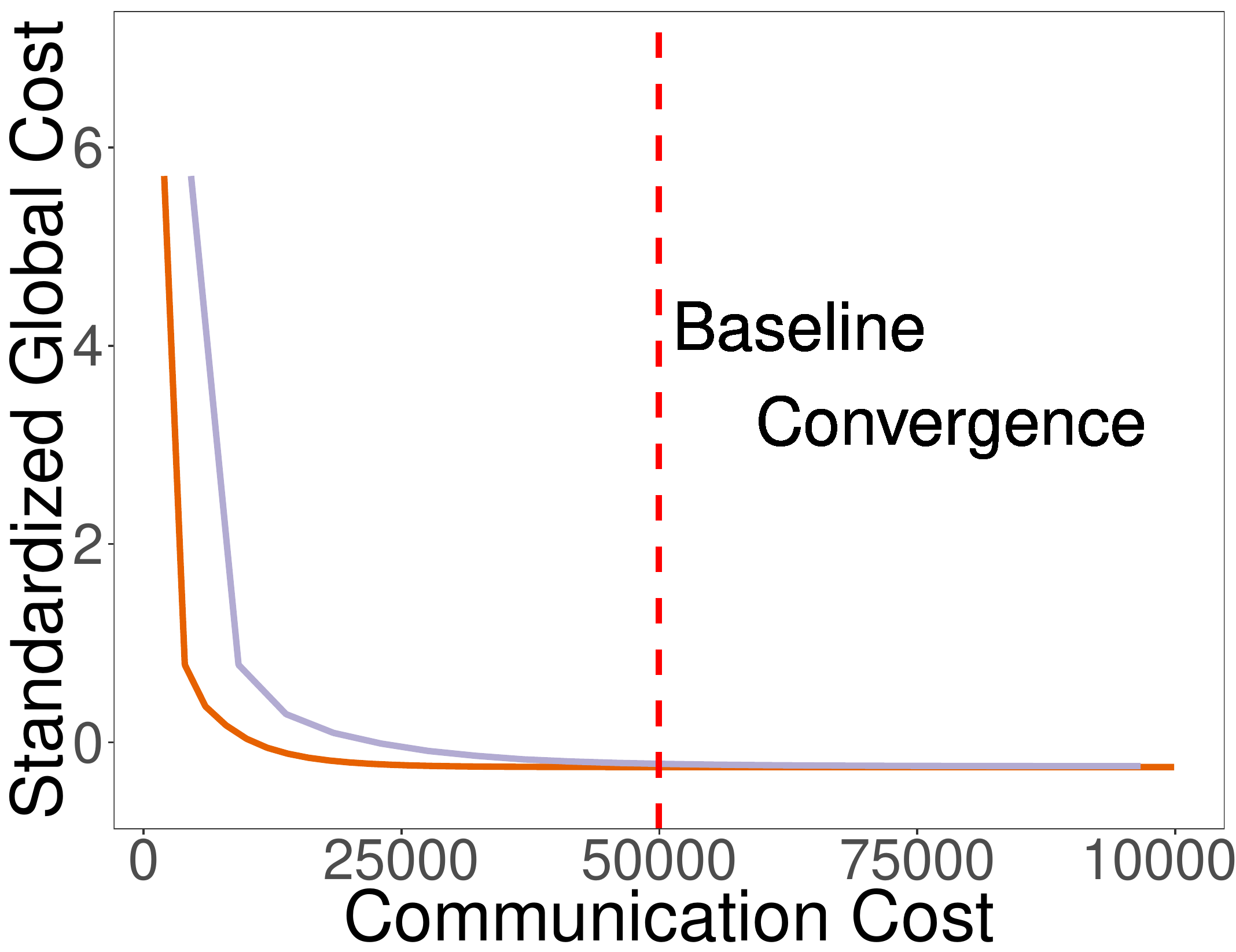}}
\subfigure[Synthetic, synchronized communication]{\includegraphics[width=0.244\textwidth]{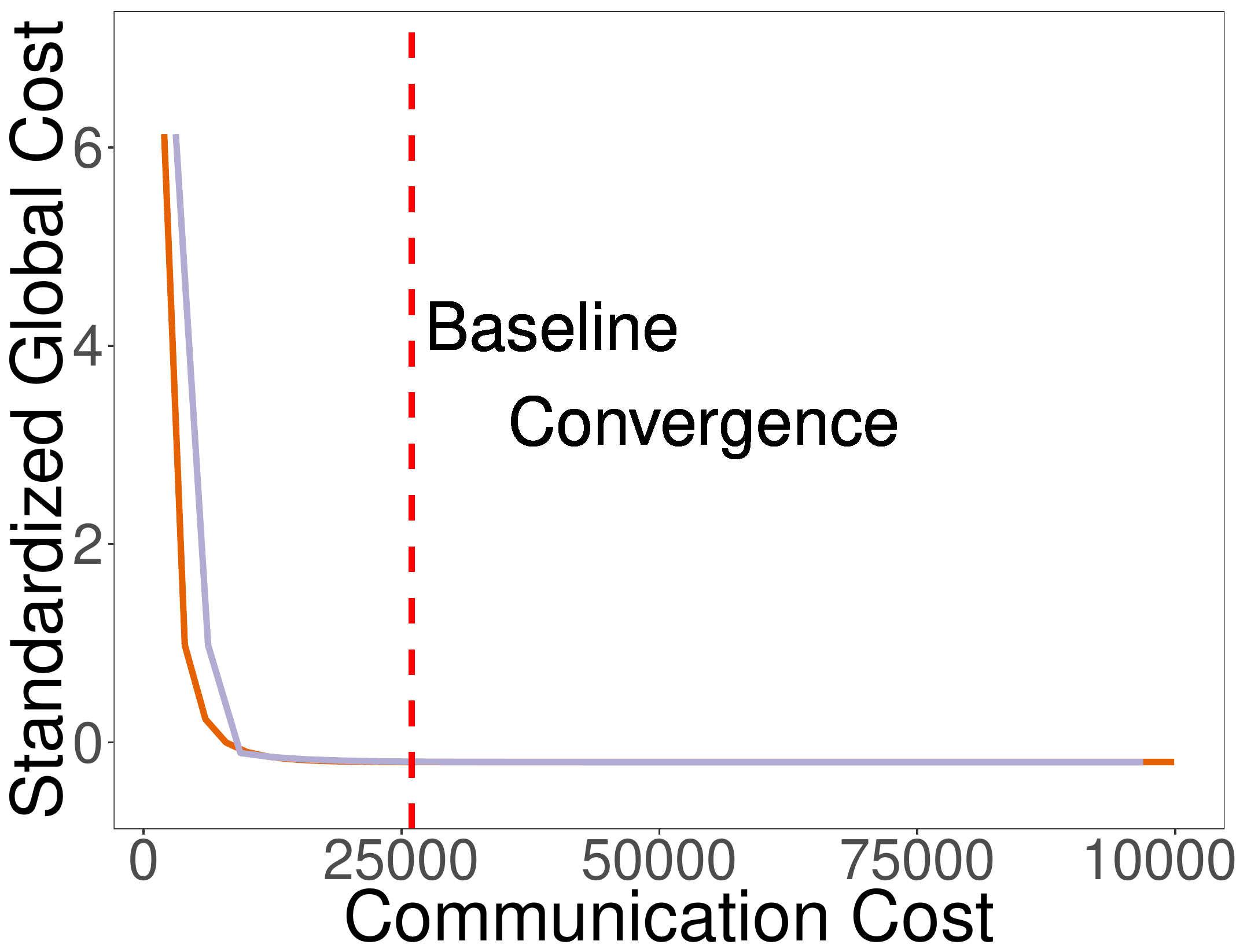}}
\subfigure[Bike sharing, synchronized communication]{\includegraphics[width=0.244\textwidth]{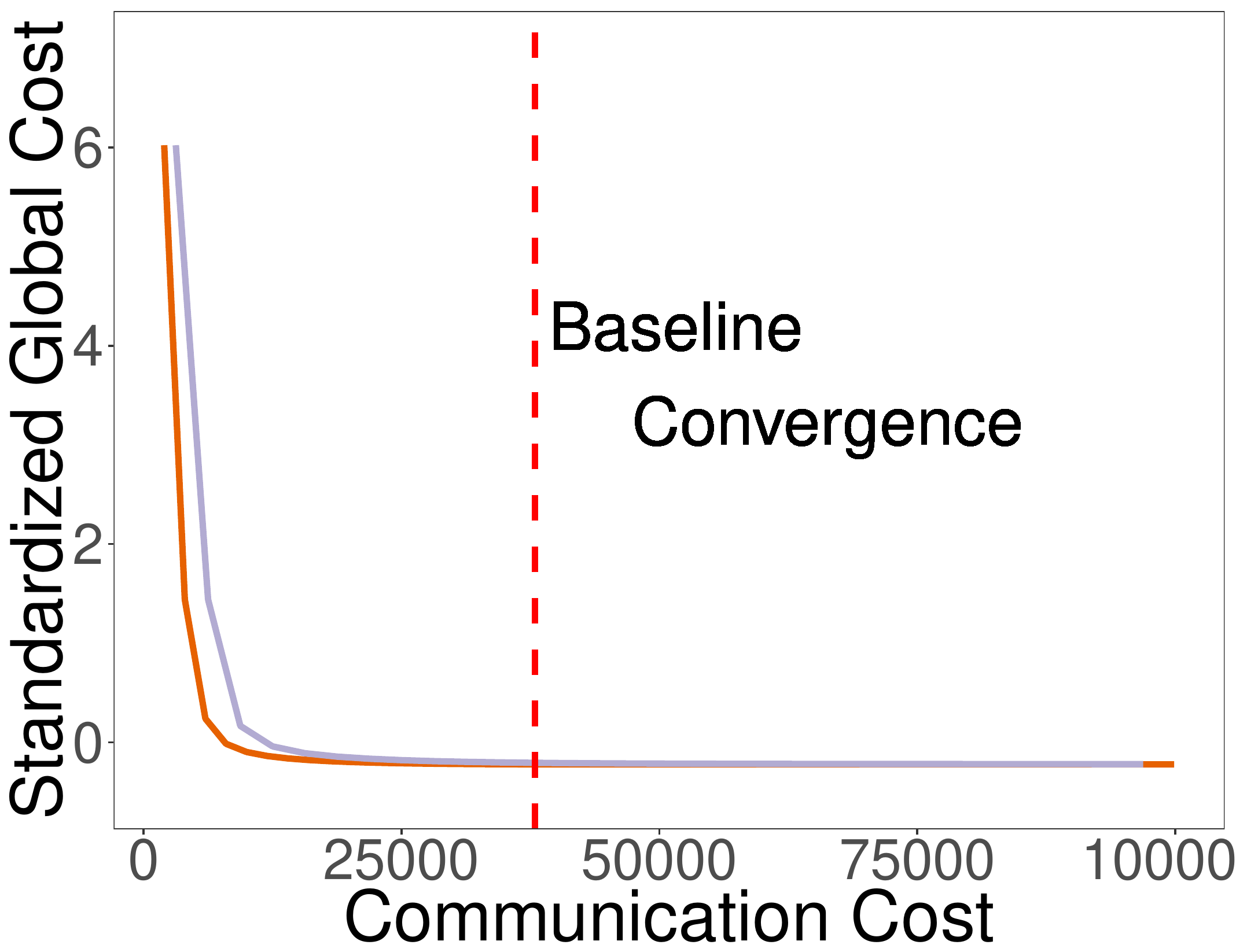}}
\subfigure[Energy demand, synchronized communication]{\includegraphics[width=0.244\textwidth]{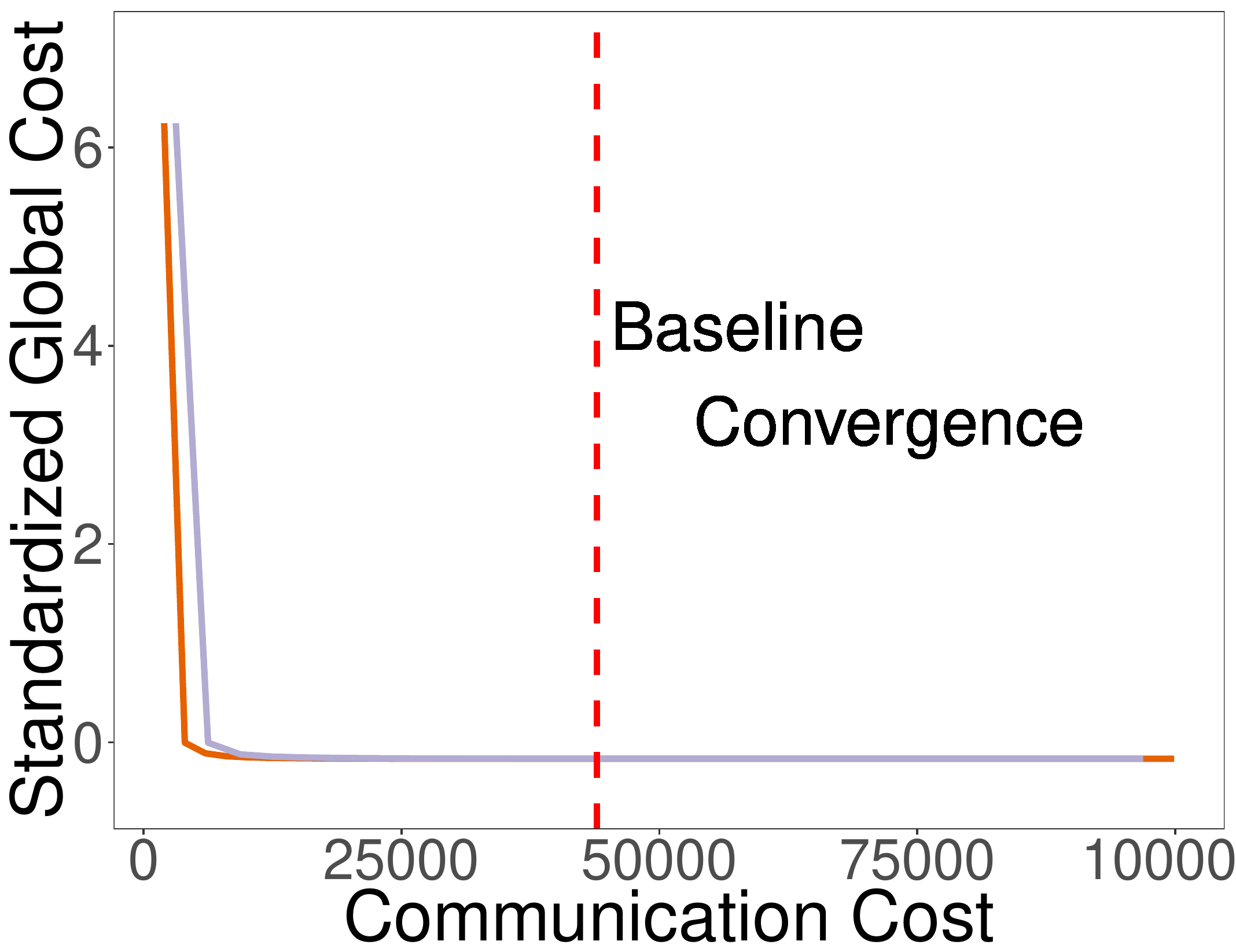}}
\subfigure[Electric vehicles, synchronized communication]{\includegraphics[width=0.244\textwidth]{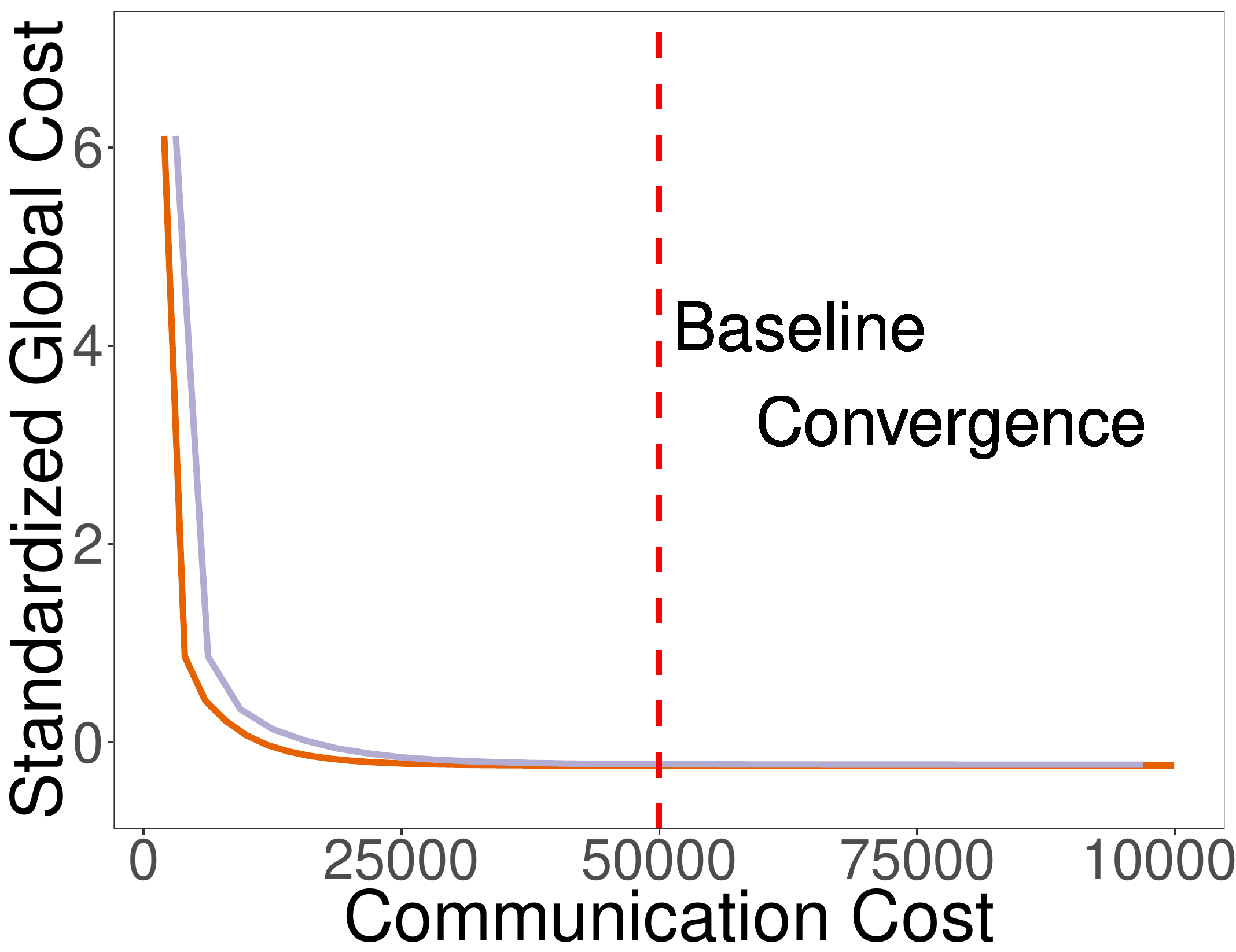}}
\caption{Cost effectiveness for comparison with Figure~\ref{fig:cost-effectiveness-two-children}. \emph{Dimensions}: baseline versus holarchic runtime, application scenarios, total versus synchronized communication cost. \emph{Settings}: partial scale, $\lambda=0$, $c=5$.}\label{fig:cost-effectiveness-five-children}
\end{figure}

Figure~\ref{fig:cost-effectiveness-five-children} compares with Figure~\ref{fig:cost-effectiveness-two-children} by varying $c=2$ to $c=5$. In this case, the global cost of the baseline and holarchic runtime is equivalent for the required communication cost to converge. The global cost of the holarchic runtime requires fewer messages to drop for $c=5$ compared to $c=2$ as indicated by the respective line shifted to the left. 

%On average X and X total and synchronized messages are required to reach a lower than 10\% and 20\% divergence in global cost, which shows a higher performance mitigation than $c=2$ with X and X respectively.

\end{document}